%% file: aWang-ML.tex
\documentclass[11pt,letterpaper]{article}
\usepackage{setspace}
\usepackage{vmargin}
\usepackage{graphicx}
\usepackage{amsmath,amsthm,amssymb}
\setpapersize{USletter}
\usepackage{amsfonts}
\usepackage{amsmath}

\usepackage{epsfig}

\usepackage[authoryear,round]{natbib}
\begin{document}
\title{%Computational and Statistical Analysis via Gradient Flow Central Limit Theorems of Gradient Descent Algorithms for Stochastic Optimization 
Asymptotic Analysis via Stochastic Differential Equations of Gradient Descent Algorithms in Statistical and Computational Paradigms
}  %for Stochastic Optimization} % for Solving Statistical and Machine Learning Problems}  
%joint asymptotic analysis in a unified framework 
\author{Yazhen Wang \\ %and Shang Wu \\
Department of Statistics, University of Wisconsin-Madison \\
Madison, WI 53706, USA. Email: yzwang@stat.wisc.edu}
\date{}
%\date{March 1, 2017}
\maketitle
\newtheorem{lem}{Lemma}[section]
\newtheorem{thm}{Theorem}[section]
\newtheorem{cor}{Corollary}[section]
\newtheorem{defn}{Definition}
\newtheorem{prop}{Proposition}[section]
\newtheorem{exam}{Example}[section]
\newtheorem{remark}{Remark}[section]
\renewcommand{\theequation} {\arabic{section}.\arabic{equation}}

\newcommand{\jp}{j^\prime}
\newcommand{\ba}{{\bf a}}
\newcommand{\bA}{{\bf A}}
\newcommand{\bB}{{\bf B}}
\newcommand{\bG}{{\bf G}}
\newcommand{\bH}{{\bf H}}
\newcommand{\bI}{{\bf I}}
\newcommand{\bM}{{\bf M}}
\newcommand{\bO}{{\bf O}}
\newcommand{\bP}{{\bf P}}
\newcommand{\bQ}{{\bf Q}}
\newcommand{\bR}{{\bf R}}
\newcommand{\bU}{{\bf U}}
\newcommand{\bV}{{\bf V}}
\newcommand{\bW}{{\bf W}}
\newcommand{\bu}{{\bf u}}
\newcommand{\bv}{{\bf v}}
\newcommand{\bw}{{\bf w}}
\newcommand{\bx}{{\bf x}}
\newcommand{\by}{{\bf y}}
\newcommand{\bz}{{\bf z}}
\newcommand{\bX}{{\bf X}}
\newcommand{\bY}{{\bf Y}}
\newcommand{\bZ}{{\bf Z}}
\newcommand{\cA}{{\cal A}}
\newcommand{\cB}{{\cal B}}
\newcommand{\cC}{{\cal C}}
\newcommand{\cF}{{\cal F}}
\newcommand{\cG}{{\cal G}}
\newcommand{\cH}{{\cal H}}
\newcommand{\cI}{{\cal I}}
\newcommand{\cL}{{\cal L}}
\newcommand{\cP}{{\cal P}}
\newcommand{\cS}{{\cal S}}
\newcommand{\cT}{{\cal T}}
\newcommand{\cY}{{\cal Y}}
\newcommand{\be}{{\bf e }}
\newcommand{\zn}{\underline{z}}
\newcommand{\hn}{\underline{h}}
\newcommand{\sigmab}{\underline{\sigma}}
\newcommand{\taub}{\bar{\tau}}
\newcommand{\boldeps}{\mbox{\boldmath $\varepsilon$}}
\newcommand{\bolddelta}{\mbox{\boldmath $\delta$}}
\newcommand{\bmu}{\mbox{\boldmath $\mu$}}
\newcommand{\bsigma}{\mbox{\boldmath $\sigma$}}
\newcommand{\bvarsigma}{\mbox{\boldmath $\varsigma$}}
\newcommand{\bSigma}{\boldsymbol{\Sigma}}
\newcommand{\bPsi}{\boldsymbol{\Psi}}
\newcommand{\bChi}{\boldsymbol{\Chi}}
\newcommand{\btau}{\mbox{\boldmath $\tau$}}
\newcommand{\boeta}{\mbox{\boldmath $\eta$}}
\newcommand{\bGamma}{\mbox{\boldmath $\Gamma$}}
\newcommand{\bgamma}{\mbox{\boldmath $\gamma$}}
\newcommand{\bLambda}{\mbox{\boldmath $\Lambda$}}
\newcommand{\bXi}{\mbox{\boldmath $\Xi$}}
\newcommand{\bxi}{\mbox{\boldmath $\xi$}}
\newcommand{\bnabla}{\mbox{\boldmath $\nabla$}}
\newcommand{\bDelta}{\mbox{\boldmath $\Delta$}}
\newcommand{\brho}{\mbox{\boldmath $\rho$}}
\newcommand{\balpha}{\mbox{\boldmath $\alpha$}}
\newcommand{\bbeta}{\mbox{\boldmath $\beta$}}
\newcommand{\R}{I\!\!R}
\newcommand{\mR}{\mathbb{R}}
\newcommand{\ot}{\frac{1}{2}}
\newcommand{\oq}{\frac{1}{4}}

%\vspace{.1in}

\begin{abstract}

This paper investigates asymptotic behaviors of gradient descent algorithms (particularly accelerated gradient descent and stochastic gradient descent) in the context of stochastic optimization arising in statistics and machine learning where objective functions are estimated from available data.  We show that these algorithms can be computationally modeled by continuous-time ordinary or stochastic differential 
equations. We establish gradient flow central limit theorems to describe the limiting dynamic behaviors of these computational algorithms %dynamic evolutions
and the large-sample performances of the related statistical procedures, %distributions
as the number of algorithm iterations and data size both go to infinity, where the gradient flow central limit theorems are governed by some linear ordinary or stochastic differential equations like time-dependent Ornstein-Uhlenbeck processes.  
We illustrate that our study can provide a novel unified framework for a joint computational and statistical asymptotic analysis, 
where the computational asymptotic analysis studies dynamic behaviors of these algorithms with the time (or the number of iterations in the algorithms), the statistical asymptotic analysis investigates large sample behaviors of the statistical procedures (like estimators and classifiers) that the algorithms are applied to compute, and in fact the statistical procedures are equal to the limits of the random sequences generated from these iterative algorithms as the  number of iterations goes to infinity. 
%on dynamic behaviors of these algorithms with the time (or the number of iterations in the algorithms) and large sample behaviors of the statistical procedures %decision rules 
%(like estimators and classifiers) that the algorithms are applied to compute, where the statistical procedures %decision rules 
%are equal to the limits of the random sequences generated from these iterative algorithms as the  number of iterations goes to infinity. 
The joint analysis results based on the obtained gradient flow central limit theorems %such as limiting behaviors of the Ornstein-Uhlenbeck processes may reveal new facts and shed light on 
can identify four factors -- learning rate, batch size, gradient covariance, and Hessian -- 
to derive new theory regarding the local minima  found by stochastic gradient descent 
%some empirically observed phenomena of stochastic gradient descent %algorithm when applied to solve 
for solving non-convex optimization problems. 
%escaping from saddle points, avoiding bad local minimizers, and converging to good local minimizers, which depends on local geometry, learning rate and batch size, when stochastic gradient descent algorithms are applied to solve non-convex optimization problems.   
%\cite{attouch2015fast}
%\cite{allen2014linear}
%\cite{arjevani2015lower}

\textbf{Key words}: %Accelerated gradient descent and stochastic Gradient descent, j
Gradient flow central limit theorem; 
joint computational and statistical asymptotic analysis,  weak convergence limit, %bootstrap, 
mini-batch, optimization, 
%(stochastic) optimization, 
ordinary or stochastic differential equation,% second order stochastic differential equation,  and stationary distribution.

\textbf{Running title}: %Gradient Flow Central Limit Theorems and  Gradient Descent Algorithms 
Computational \& Statistical Analysis of Gradient Descent
\end{abstract}

\input{bs1-ML}

\input{gbib-ML}
\newpage

\input{p1-ML}

\input{p2-ML}

\input{p4}

\input{p5}
\input{p6}

\end{document}

%% file: bs1-ML.tex
%\begin{spacing}{1.7}
%\end{spacing}
\section{Introduction}

%Stochastic optimization refers to a collection of methods for minimizing or maximizing an objective function when randomness is present. Over the last few decades these methods have become essential tools for science, engineering, business, computer science, and statistics. Specific applications are varied, but include: running simulations to refine the placement of acoustic sensors on a beam, deciding when to release water from a reservoir for hydroelectric power generation, and optimizing the parameters of a statistical model for a given data set. Randomness usually enters the problem in two ways: through the cost function or the constraint set. Although stochastic optimization refers to any optimization method that employs randomness within some communities, we only consider settings where the objective function or constraints are random. 

\subsection{Background and Motivation} 
Optimization plays an important role in scientific fields ranging from machine learning to physical sciences and statistics to engineering.  It lies at the core of data science by providing a mathematical language for handling both computational algorithms and statistical inferences in data analysis. 
Numerous algorithms and methods have been proposed to solve optimization problems. Examples include Newton's method, 
gradient and subgradient descent, conjugate gradient methods, trust region methods, and interior point methods (see Polyak, 1987; Boyd and Vandenberghe, 2004; Nocedal and Wright, 2006; Ruszczynski, 2006; Boyd et al., 2011; Shor, 2012; %Beck, 2014, 
Goodfellow et al. (2016) for expositions).
Practical problems arising in fields like statistics and machine learning usually involve optimization settings where the objective functions are empirically estimated from available data %(a training sample or a statistical sample) 
with the form of a sum of differentiable functions. 
%are risks plus some penalty functions and their corresponding empirical random counterparts that are estimated from observed data. 
We refer to such optimization problems with random objective functions as stochastic optimization. 
%For example, it is often the case that the true risk, that is used to define the optimal decision rule and classifier,  
%is unknown in machine learning and statistics, and the stochastic approximation of the risk is often to be an empirical version of the risk based on available data (a training sample or a statistical data sample).  In the context of empirical risk minimization, %%gradient descent, known as  
%stochastic gradient descent is a stochastic approximation of the gradient descent optimization method for minimizing an objective function with the form of a sum of differentiable functions. 
As data sets in practical problems grow rapidly in scale and complexity, %first-order 
methods such as stochastic gradient descent %and accelerated gradient descent 
can scale to the enormous size of big data and have been very popular. %regained popularity. 
%In a seminar work Nesterov (1983) proposed an accelerated gradient descent algorithm. Since then 
There has been recent surging interest in and 
great research work on the theory and practice of gradient descent and its extensions and variants. %various accelerated first-order schemes.
For example, a number of recent papers were devoted to investigate stochastic gradient descent and its variants for solving complex %non-convex 
 optimization problems (Ali et al. (2019), 
 Chen et al. (2016), Ge et al. (2015), Jin et al. (2017), Kawaguchi (2016),  Keskar et al. (2017), Lee et al. (2016), 
 Li et al. (2017b), Mandt et al. (2016), and Shallue et al. (2019)). In particular   
 Su et al. (2016) %showed that 
 derived the continuous-time limit of Nesterov's accelerated gradient descent as a second-order ordinary differential equation for 
  studying %understand and analyze 
 the acceleration phenomenon and generalizing Nesterov's scheme.
  %Inspired by Su et al. (2016), 
Wibisono et al. (2016) %studied acceleration from a continuous-time variational point of view and 
further developed a systematic approach based on continuous-time variations 
to understand the acceleration phenomenon and produce acceleration algorithms from continuous-time differential equations. 
%generated by a so-called Bregman Lagrangian. 
 In spite of compelling theoretical and numerical evidence on the value of the stochastic approximation idea and acceleration phenomenon, yet there remains some conceptual and theoretical mystery in the acceleration and stochastic approximation schemes.  %For non-stochastic optimization, 
  
%Optimization problems arising in fields like statistics and machine learning usually concern with data analysis, and practical data analysis algorithms involve risk minimization. Gradient descent in this context, known as  stochastic gradient descent, is a stochastic approximation of the gradient descent optimization method for minimizing an objective function with the form of a sum of differentiable functions. For example, it is often the case that the true risk is unknown in machine learning and statistics, and the stochastic approximation of the risk is often to be an empirical version of the risk based on a training sample or a statistical data sample. 

\subsection{Contributions}

This paper establishes asymptotic theory for gradient descent, stochastic gradient descent, and accelerated gradient descent in the stochastic optimization setup. We derive continuous-time ordinary or stochastic differential equations to model the
dynamic behaviors of these gradient descent algorithms and investigate their limiting algorithmic dynamics and large sample 
performances as the number of algorithm iterations and data size both go to infinity. 
Specifically for an optimization problem whose objective function is convex and deterministic, %the true risk (plus some penalty), 
we consider a matched stochastic optimization problem whose random objective function is an empirical estimator of the deterministic objective function based on available data.  
%with a corresponding empirical objective function estimated from available data. 
The solution of the stochastic optimization specifies a decision rule like an estimator or a classifier based on the sampled data in statistics and machine learning, while its corresponding deterministic optimization problem characterizes through its solution the true value of the parameter in the population model. In other words, the two connected optimization problems associate with the data sample and its corresponding population model where the data are sampled from, and the stochastic optimization is considered to be a sample version of the deterministic optimization corresponding to the population. 
These two types of optimization problems refer to the deterministic population and stochastic sample optimization problems. 
Consider random sequences that are generated from the gradient descent algorithms and their corresponding continuous-time ordinary or stochastic differential equations for the stochastic sample optimization setting. We show that the random sequences 
 converge to the ordinary differential equations for the corresponding deterministic population optimization setup,  and we derive their asymptotic distributions %governed 
 by some linear ordinary or stochastic differential equations such as time-dependent Ornstein-Uhlenbeck processes. 
%We  show that random sequences generated from these gradient descent algorithms and their continuous time ordinary or stochastic differential equations for the stochastic (sample) optimization setting converge to ordinary differential equations for the corresponding deterministic (population) optimization set-up, with asymptotic distributions governed by some linear ordinary or stochastic differential equations such as time-dependent Ornstein-Uhlenbeck processes. 
The asymptotic distributions are used to understand and quantify the limiting discrepancy between the random sequences generated from each algorithm for solving the corresponding sample and population optimization problems. 
In particular %for the (stochastic) gradient descent case, 
since the obtained asymptotic distributions characterize the limiting behavior of the normalized difference between the sample and population gradient (or Lagrangian) flows, the limiting distributions may be viewed as central limit theorems (CLT) for gradient (or Lagrangian) flows, and are then called the gradient (or Lagrangian) flow central limit theorems (GF-CLT or LF-CLT). 
Moreover, our analysis may offer a novel unified framework to carry out a joint asymptotic analysis for %computational asymptotics on 
computational algorithms and statistical decision rules that the algorithms are applied to compute. 
As iterated computational methods, these gradient descent algorithms generate sequences %of numerical values 
that converge to the exact decision rule or the true parameter value 
for the corresponding optimization problems,  when the number of the iterations goes to infinity. Thus, as time (corresponding to the number of iterations) goes to infinity, the continuous-time differential equations may have distributional limits corresponding to the large-sample distributions of statistical decision rules as the sample size goes to infinity. In other words, the asymptotic analysis can be done with both time and data size, where the time direction
corresponds to the computational asymptotics on dynamic behaviors of the algorithms, and the data size direction associates with the statistical large-sample asymptotics on the statistical behaviors of decision rules such as estimators and classifiers. 
The continuous-time modeling and the GF-CLT based joint asymptotic analysis may reveal new facts and shed some light %via %large deviation theory and 
%the limiting behaviors %the stationary distributions of the asymptotic stochastic differential equations like 
%of the time-dependent Ornstein-Uhlenbeck processes 
on the phenomenon that stochastic gradient descent algorithms can escape from saddle points 
and converge to good local minimizers for solving non-convex optimization problems in deep learning.
To the best of our knowledge, this is the first paper to establish the GF-CLT and LF-CLT,  offer the unified framework for the joint computational and statistical asymptotic analysis, and establish a novel theory to identify four factors 
for influencing the local minima found by stochastic gradient descent in non-convex optimization. 
%%provide rigorous treatments between stochastic gradient descent and stochastic differential equations 
% and discover the second order stochastic differential equations for the accelerated case. 

%continuous-time ordinary differential equations corresponding to accelerated gradient descent for empirical risk minimization and derive their 
%asymptotic behaviors as sample size goes to infinity. We show that the ordinary differential equations for the empirical risk minimization 
%case have random coefficients, and as sample size goes to infinity, the ODEs approach to the ODEs for the corresponding true risk
%minimization. Moreover, we derive the difference between the two differential equations normalized by the square root of sample size 
%converges to certain linear differential equations whose solutions are determined by a normal random variable multiplying 
%deterministic linear differential equations. 

%Some versions of stochastic differential equations (such as vague or approximate matrices for the diffusion variance) % like (\ref{GD-stoch1}) 
%are informally used in the deep learning and stochastic gradient descent literature based on some heuristic or loose reasoning without rigorous justification 
There is a large literature on stochastic approximation and recursive algorithms in particular stochastic gradient descent in deep learning (see Chen et al. (2016), Dalalyan (2017), Fan et al. (2018),  Ge et al. (2015), Jastrz\c{e}bski et al. (2018), Jin et al. (2017), Kawaguchi (2016),  Keskar et al. (2017), Kushner and Yin (2003),  Lee et al. (2016), Li et al. (2016), Li et al. (2017a),  Li et al. (2017b), Ma et al. (2019), Mandt et al. (2016), Shallue et al. (2019), Sirignano and Spiliopoulos (2017), Su et al. (2016),  Wibisono et al. (2016)).  Both continuous-time and discrete-time means are adopted by computational and statistical (as well as machine learning) communities. The work on the computational  side focuses more on the dynamics and convergence of learning algorithms, while the statistics research emphasizes more on statistical inferences  of learning rules. Our study combines both computational and statistical approaches to carry out a joint analysis on the learning algorithms and the learning rules, where the algorithms are used  to compute the rules. In a nutshell, we analyze the learning algorithms in terms of both computational 
dynamic convergence behavior and statistical large sample performance. The developed GF-CLT based theory and the created joint analysis provide 
a description of dynamic convergence behaviors of the computational algorithms as well as the statistical large sample 
performances of the learning rules computed by the algorithms. 
%In particular for the gradient descent case, the obtained asymptotic distributions characterize the asymptotic behavior of the normalized discrepancy between the sample and population gradient flows, that is, the limiting distributions may be viewed as the central limit theorem (CLT) in terms of gradient flow, and are then called the gradient flow central limit theorem (GF-CLT).  
The statistical behaviors  of the associated learning rules can be derived from the GF-CLT at an infinite time horizon. On the computational side, it has implications for optimization phenomena observed in the discrete-time case as well as in practice. For example, the obtained GF-CLT 
uncovers that the gradient flow for the stochastic sample optimization can be decomposed into the gradient flow for the corresponding deterministic population optimization plus a random fluctuation term, where the random fluctuation depends on the learning rate and batch size only through their ratio  and is governed by  
%described %determined by 
a time-dependent Ornstein-Uhlenbeck process. 
Using the joint analysis along with the algebraic Ricatti equation for characterizing the stationary covariance of the Ornstein-Uhlenbeck process, we discover a novel theory %reveal new facts and shed some light on the phenomenon 
about how the minima found by stochastic gradient descent are influenced by four factors: 
% the ratio of  learning rate to batch size, 
learning rate,  batch size, gradient covariance, and Hessian. % around the minima. 
As a case in point, our general results cover %They agree with those 
the study under a special circumstance in Jastrz\c{e}bski et al. (2018) that can identify only three of the four 
factors %-- learning rate, batch size, and gradient covariance (or Hessian) --  
to influence the local minima found by stochastic gradient descent. 
%%that the same stochastic differential equations are obtained for stochastic gradient descent, and 
%regarding the behavior of stochastic gradient descent under a special circumstance which is determined by the ratio of learning rate to batch size and the Hessian at the minima. 
 Foster et al. (2019) showed that the complexity of a stochastic optimization can be decomposed into the complexity of its corresponding deterministic population optimization and the sample complexity, where the optimization complexity represents the minimal amount of
effort required to find near-stationary points, and the sample complexity of an algorithm refers to the number of training-samples needed 
 to learn a target function sufficient well. Our results indicate that finding near-stationary points for a stochastic sample optimization can be converted into finding near-stationary points for the corresponding deterministic population optimization plus some control on the random fluctuation term. 
 As the random fluctuation has zero mean, the control can be achieved through bounding the variance of the time-dependent Ornstein-Uhlenbeck process along with selecting a sufficiently small ratio of learning rate to batch size,  which is often used to describe 
 the sample complexity of the associated statistical learning problem for the time-dependent Ornstein-Uhlenbeck process. 
This shows that our results are in agreement with Foster et al. (2019), and may point to some potential intrinsic connection between our approach and that in Foster et al. (2019).
%reveals some surprising differences between the complexity of stochastic optimization versus learning.
Furthermore, the continuous-time approach serves as a handy means as well as a beautiful framework to formulate stochastic dynamics and statistical procedures and derive their limiting behaviors and large sample performances. These are eventually used to establish limiting behaviors
of their discrete counterparts. 

%1. The results concerning the asymptotic distributions of various quantities are new, but (taking Theorem 3.1 as a prototypical example), the main object of study appears to be $V^n(t)$, measuring the discrepancy between the finite-sample continuous-time gradient flow solution $X^n(t)$ and the population-level equivalent $X(t)$. It only seems like a useful reference point when $t \to \infty$.

%2. Another argument the authors might make is that the continuous-time approach reveals behavior that has been observed in both the discrete-time case as well as in practice (this is especially brought out in Section 4.4); this is good to see, but does it merit on its own the study of continuous-time dynamics?  It would be nice if the continuous-time approach revealed something new that was not known before.  

%3. Finally, another argument that might be made is that the continuous-time approach makes all/many of the proofs snappy.

In summary, we highlight our main contributions as follows: 
\begin{itemize} 
\item We establish a new asymptotic theory for the discrepancy between the sample and population gradient (or Lagrangian) flows. In particular the new limiting distributions for the normalized discrepancy are called the gradient (or Lagrangian) flow central limit theorems (GF-CLT or LF-CLT). See Sections 3.3 and 4.1-4.2.
\item The obtained asymptotic theory provides a novel unified framework for a joint computational and statistical asymptotic analysis. Statistically the joint analysis can facilitate inferential analysis of a learning rule computed by gradient descent algorithms. Computationally the joint analysis enables us to understand and quantify a random fluctuation in and the related impact on the dynamic and convergence behavior of a gradient descent algorithm when being applied to solve a stochastic optimization problem. See Sections 3.4 and 4.3.
\item Computationally we discover a novel theory that four factors -- learning rate, batch size, gradient covariance, and Hessian -- along with the associated identities are shown to influence the local minima found by stochastic gradient descent for solving a non-convex optimization problem. 
%establish new relations among the ratio of learning rate to batch size and the local geometric characteristics (including the Hessian and gradient covariance) around the local minimum found by stochastic gradient descent for solving a non-convex optimization problem. 
It may also shed light on some intrinsic relationship among stochastic optimization, deterministic optimization, and statistical learning. See Section 4.4.
\item Statistically we illustrate implications of our results for statistical analysis of stochastic gradient descent and inference of outputs from stochastic gradient descent. See Section 4.5.
\item The continuous-time approach is employed to demonstrate that it can provide a handy means for deriving beautiful and deep 
results for stochastic dynamics of learning algorithms and statistical inference of learning rules. 
\end{itemize}

\subsection{Organization} %The paper layout}

The rest of the paper proceeds as follows. Section 2 introduces gradient descent, accelerated gradient descent, and their corresponding ordinary differential equations. Section 3 presents stochastic optimization and investigates asymptotic behaviors
of the plain and accelerated gradient descent algorithms and their associated ordinary differential equations (with random coefficients)  when the sample size 
goes to infinity. We illustrate the unified framework to carry out a joint analysis on computational and statistical asymptotics, where computational asymptotics deals with dynamic behaviors of the gradient descent algorithms with time (or iteration), 
and statistical asymptotics studies large sample behaviors of statistical decision rules that the algorithms are applied to compute. 
Section 4 considers stochastic gradient descent algorithms for large scale data and derives stochastic differential equations 
to model these algorithms. We establish asymptotic theory for these algorithms and 
their associated stochastic differential equations, % like time-dependent Ornstein-Uhlenbeck processes, 
and describe  a joint analysis on computational %asymptotics with time for the stochastic gradient descent algorithms 
and statistical asymptotics. %with sample size for statistical decision rules. 
%for risk minimization and corresponding ordinary differential equations for accelerated gradient descent. We establish their asymptotic differential equations and study the behavior of their limiting linear ordinary differential equations. 
Section 5 features an example. All technical proofs are relegated in the appendix section. 

We adopt the following notations and conventions. 
For the stochastic sample optimization problem considered in Sections 3 and 4, we add a superscript $n$ to notations for the associated processes 
and sequences in Section 3 and indices $m$ and/or $*$ to notations for the corresponding processes and sequences affiliated with 
mini-batches %(or bootstrap samples) 
in Section 4, while notations without %any 
such subscripts or superscripts are for sequences and functions corresponding to the deterministic population optimization problem given in Section 2. 
Our basic proof ideas are as follows. Each algorithm generates a sequence for computing a learning rule, a step-wise empirical process 
is formed by the generated sequence, and a continuous process is obtained from the corresponding continuous-time differential equation. 
We derive asymptotic distributions by analyzing the differential equations, and we bound the differences between the empirical processes and their corresponding continuous processes by studying the optimization problems and utilizing the empirical process theory along with the related differential equations. 

\section{Ordinary differential equations for gradient descent algorithms}
\label{section-1}

Consider the following minimization problem 
\begin{equation} \label{min-0}
 \min_{\theta \in \Theta} g(\theta), 
 \end{equation}
where the objective %target 
function $g(\theta)$ is defined on a parameter space $\Theta \subset I\!\!R^p$ and assumed to have L-Lipshitz continuous gradients. Iterative algorithms such as gradient descent methods are often employed to numerically compute the solution of the 
minimization problem. 
Starting with some initial values $x_0$, the plain gradient descent algorithm is iteratively defined by 
\begin{align}  \label{equ-GD1}
& x_k=x_{k-1}- \delta \nabla\!g(x_{k-1}), 
\end{align}
where $\nabla$ denotes gradient operator, and $\delta$ is a positive constant which is often called a step size or learning rate.  

It is easy to model $\{x_k$, $k=0, 1, \cdots \}$ by a smooth curve $X(t)$ with the Ansatz $x_k \approx X(k \delta)$ as follows. 
Define a step function $x_\delta(t) = x_k$ for $k \delta \leq t < (k+1) \delta $, %$(k-1) \delta < t \leq k \delta $
and as $\delta \rightarrow 0$, $x_\delta(t)$ approaches $X(t)$ satisfying 
\begin{equation} \label{GD-c1}
   \dot{X}(t) + \nabla g(X(t)) = 0,
\end{equation} 
where $\dot{X}(t)$ denotes the derivative of $X(t)$, and initial value $X(0)=x_0$. In fact,  $X(t)$ is a gradient flow associated with the objective function $g(\cdot)$ 
in the optimization problem \eqref{min-0}. 

Nesterov's accelerated gradient descent scheme is a well-known algorithm that is much faster than the plain gradient descent algorithm.
Starting with initial values $x_0$ and $y_0=x_0$, Nesterov's accelerated gradient descent algorithm is iteratively defined by 
\begin{align}  \label{equ-Nest1}
& x_k=y_{k-1}- \delta \nabla\!g(y_{k-1}), \qquad y_k=x_k+\frac{k-1}{k+2}(x_k-x_{k-1}), 
\end{align}
where %$\nabla$ denotes gradient operator, and 
$\delta$ is a positive constant. 
Using (\ref{equ-Nest1}) we derive a recursive relationship between consecutive increments
%Applying rescaling gives us 
\begin{equation}  \label{equ-1}
\frac{x_{k+1}-x_k}{\sqrt{\delta}}=\frac{k-1}{k+2}\frac{x_k-x_{k-1}}{\sqrt{\delta}}-\sqrt{\delta}\nabla\!g(y_k).
\end{equation}
We model $\{x_k$, $k=0, 1, \cdots \}$ by a smooth curve  in a sense that $x_k$ are its samples at discrete points, that is, 
we define a step function $x_\delta(t) = x_k$ for $k \sqrt{\delta} \leq t < (k+1) \sqrt{\delta} $, % $(k-1)\sqrt{\delta} < t \leq k \sqrt{\delta} $,
and introduce 
the Ansatz $x_\delta(k \sqrt{\delta})=x_k \approx X(k\sqrt{\delta})$ for 
some smooth function $X(t)$ defined for $t\geqslant 0$. Let $\sqrt{\delta}$ be the step size. For $t=k\sqrt{\delta}$,  as 
$\delta \rightarrow 0$, %the step size goes to 0, 
we have $x_k = x_{t/\sqrt{\delta}} = X(t)$, $x_{k+1} = x_{(t+\sqrt{\delta})/\sqrt{\delta}} = X(t + \sqrt{\delta})$, and 
\begin{eqnarray*}
&&  y_k = X(t) + \frac{t /\sqrt{\delta}-1}{t /\sqrt{\delta}+2} [X(t) - X(t-\sqrt{\delta})]  %\\
%&& = X(t) + \left(1-\frac{3\sqrt{\delta}}{t + 2 \sqrt{\delta}} \right) [X(t) - X(t-\sqrt{\delta})] \\   %= X(t) + O( X(t) - X(t - \sqrt{\delta}) ) 
%&& 
= X(t) + O(\sqrt{\delta}).
\end{eqnarray*}
Applying the Taylor expansion and using L-Lipshitz continuous gradients we obtain 
\[ \frac{x_{k+1} - x_k}{\sqrt{\delta}} = \frac{X(t+ \sqrt{\delta}) - X(t)}{\sqrt{\delta}} = \dot{X}(t) + \frac{1}{2} \ddot{X}(t) \sqrt{\delta} + o(\sqrt{\delta}), \]
\[ \frac{x_{k} - x_{k-1}}{\sqrt{\delta}} = \frac{X(t) - X(t-\sqrt{\delta}) }{\sqrt{\delta}} = \dot{X}(t) - \frac{1}{2} \ddot{X}(t) \sqrt{\delta} + o(\sqrt{\delta}), \]
\[ \sqrt{\delta} \nabla \!g(y_k) = \sqrt{\delta} \nabla \!g(X(t)) + o(\sqrt{\delta}), \]
where $\ddot{X}(t)$ denotes the second derivative of $X(t)$. Substituting above results into the equation (\ref{equ-1}) 
and letting $\delta \rightarrow 0$ we obtain  %may rewrite equation (\ref{equ-1}) as
%\begin{equation*} 
%\dot{X}(t)+\frac{1}{2}\ddot{X}(t)\sqrt{\delta}+o(\sqrt{\delta})=\left(1-\frac{3\sqrt{\delta}}{t + 2 \sqrt{\delta}} \right)(\dot{X}(t)-\frac{1}{2}\ddot{X}(t)\sqrt{\delta}+o(\sqrt{\delta}))-\sqrt{\delta}\nabla g(X(t))+o(\sqrt{\delta}).
%\end{equation*}
%Re-arranging the terms and dividing $\sqrt{\delta}$ on both sides lead to 
%\begin{equation*}
%\ddot{X}(t) +\frac{3}{t}\dot{X}(t)+\nabla \!g(X(t)) + o(1) =0.
%\end{equation*}
%Note that $X(t)$ is free of $\delta$. As $\delta \rightarrow 0$, the equation becomes 
\begin{equation} \label{equ-2}
\ddot{X}(t) +\frac{3}{t}\dot{X}(t)+\nabla \!g(X(t)) =0,
\end{equation}
with the initial conditions $X(0)=x_0$ and $\dot{X}(0)=0$. As the coefficient $3/t$  in the ordinary differential 
equation (\ref{equ-2}) is singular at $t=0$, classical ordinary differential equation theory 
is not applicable to establish the existence or uniqueness of the solution to the equation (\ref{equ-2}). The heuristic derivation of (\ref{equ-2}) is from 
Su et al. (2016) %derived (\ref{equ-2}) and proved 
who has established that the equation  (\ref{equ-2}) 
has a unique solution satisfying the initial conditions, and $x_\delta(t)$ converges to $X(t)$ uniformly on $[0, T]$ for any fixed $T>0$.
Note the step size difference between the plain and accelerated cases, where the step size is $\delta^{1/2}$ for Nesterov's accelerated gradient descent algorithm and $\delta$ for the plain gradient descent algorithm. Su et al. (2016) has shown that, because of the difference, 
the accelerated gradient descent algorithm moves much faster than the plain gradient descent algorithm along the curve $X(t)$. %See also Wibisono et al. (2016) for more elaborate explanation on the acceleration phenomenon. 
Wibisono et al. (2016) provided more elaborate explanation on the acceleration phenomenon and developed a systematic 
 continuous-time variational scheme %perspective %Bregman Lagrangian 
to generate a large class of continuous-time differential equations and produce a family of accelerated gradient algorithms. 
The variational scheme utilizes a first-order rescaled gradient flow and a second-order Lagrangian flow, which are generalizations of gradient flow. As we refer the solution $X(t)$ of the differential equation \eqref{GD-c1} to the gradient flow for the gradient descent 
algorithm \eqref{equ-GD1}, the solution $X(t)$ to the differential equation \eqref{equ-2} is called the Lagrangian flow for 
 the accelerated gradient descent algorithm \eqref{equ-Nest1}. 
 
%\section{Stochastic accelerated gradient descent and associated differential equations} 
%\label{section-2}
\section{Gradient descent for stochastic optimization} %with empirical objective functions estimated by all training data}
\label{section-2-1}

Let $\theta=(\theta_1,...,\theta_p)^{'}$ be the parameter that we are interested in, and %$\bU=(U_1,...,U_p)$ 
$U$ be a relevant random element on a probability space with a given distribution $Q$. Consider an objective function $\ell(\theta;u)$ and its corresponding expectation $E[\ell(\theta;U)]=g(\theta)$. For example, in a statistical decision problem, we may take $U$ to be a decision rule,  
$\ell(\theta;u)$ a loss function, and $g(\theta) = E[\ell(\theta;U)]$ its corresponding risk; in M-estimation, we may treat $U$ as a sample observation and $\ell(\theta, u)$ a $\rho$-function; in nonparametric function estimation and machine learning, we may choose $U$ an observation and $\ell(\theta;u)$ equal to a loss function plus some penalty. For these problems we need to consider the corresponding population minimization problem (\ref{min-0}) for characterizing the true parameter value or its function as an estimand, 
%the theoretical study such as the optimal decision rule, 
but practically, because $g(\theta)$ is usually unavailable, we have to employ its empirical version and consider a stochastic optimization problem, described as follows: 
\begin{equation} \label{min-1}
 \min_{\theta \in \Theta }  \cL^n(\theta; \bU_n), 
 \end{equation}
where $\cL^n(\theta; \bU_n)=\frac{1}{n}\sum_{i=1}^n \ell(\theta; U_{i})$,  $\bU_n = (U_1, \cdots, U_n)^\prime$ is a 
%training sample or statistical 
sample,  and we assume that $U_1, \cdots, U_n$ are i.i.d. and follow the distribution $Q$. 
%Here and after we adopt a notation convention that an extra label $Q$ is added to the objective function $\cL(\theta; \bU_n, Q)$, and we will write $\ell(\theta; U_i)$ as $\ell(\theta; U_i, Q)$ so that $Q$ is clearly specified as the distribution of $U_i$. This will be particularly useful when we handle resampling for stochastic gradient descent later.

The minimization problem (\ref{min-0}) characterizes the true value of the target estimand such as an estimation parameter %$\theta$ 
in a statistical model %the optimal decision rule  in the statistical decision framework 
%such as an underlying M-functional in the M-estimation setup or 
and a classification parameter in a machine learning task. 
As the true objective function $g(\theta)$ is usually unknown in practice, we often solve the stochastic minimization problem (\ref{min-1}) with observed data to  obtain practically useful decision rules such as an M-estimator, a smoothing function estimator, and a machine learning classifier. 
%Our goal is to minimize the risk and obtain the optimal $\theta$. However, in a lot of cases, the explicit form of $f(\theta)$ is practically unknown so we are usually dealing with $L^n(\theta):=\frac{1}{n}\sum_{i=1}^n \ell(\theta; U^{i}))$,  where $U^1, \cdots, U^n$ are training sample. 
The approach to obtaining practical procedures is based on the heuristic reasoning that as $n \rightarrow \infty$, the law of large number implies that $\cL^n(\theta; \bU_n)$ 
eventually converges to $g(\theta)$ in probability, and thus the solution of (\ref{min-1}) approaches that of the minimization problem 
(\ref{min-0}). 

\subsection{Plain gradient descent algorithm}
\label{section-2-2}
Applying the plain gradient descent scheme to the minimization problem (\ref{min-1}) 
 with initial value $x^n_0$, we obtain the following iterative algorithm to compute the solution of (\ref{min-1}), 
\begin{align} \label{min-GD2}
& x^n_k=x^n_{k-1}- \delta \nabla \! \cL^n(x^n_{k-1}; \bU_n),
\end{align}
where $\delta>0$ is a step size or learning rate, and $\cL^n$ is the objective function in the minimization problem (\ref{min-1}).

Following the continuous curve approximation described  in Section \ref{section-1} we define a 
step function $x^n_\delta(t) = x^n_k$ for $k \delta \leq t <(k+1) \delta $, and for each $n$, as $\delta \rightarrow 0$, $x^n_\delta(t)$ approaches 
a smooth curve $X^n(t)$, $t \geq 0$,  given by 
\begin{equation} \label{GD-c2}
 \dot{X}^n(t) + \nabla \! \cL^n(X^n(t); \bU_n) =  0, %\dot{X}^n(t) +  \frac{1}{n}\sum_{i=1}^n \nabla \! \ell(X^n(t); U_i)=0.
 \end{equation} 
 where $\nabla \! \cL^n(X^n(t); \bU_n)=  \frac{1}{n}\sum_{i=1}^n \nabla \! \ell(X^n(t); U_i)$, 
 gradient operator $\nabla$ here is applied to $\cL^n(\theta; \bU_n)$ and $\ell(\theta; U_i)$ with respect to $\theta$,
 and initial value $X^n(0)=x^n_0$. $X^n(t)$ is a gradient flow associated with $\cL^n$ in the optimization problem \eqref{min-1}.

 As $\bU_n$ and $X^n(t)$ are random, and our main interest is to study the distributional behaviors of the solution and algorithm, 
 we may define a solution of the equation (\ref{GD-c2}) in a weak sense that there exist a process $X^n_\dagger(t)$ and a random vector 
 $\bU_n^\dagger=(U^\dagger_1, \cdots, U^\dagger_n)^\prime$ defined on some probability space such that $\bU_n^\dagger$ is identically distributed as $\bU_n$,  $(\bU^\dagger_n, X^n_\dagger(t))$ satisfies the equation (\ref{GD-c2}), and $X^n_\dagger(t)$ is called a 
(weak) solution of the equation (\ref{GD-c2}). Note that $X^n_\dagger(t)$ is not required to be defined on a fixed probability space with given random 
variables, instead we define $X^n_\dagger(t)$ on some probability space with some associated random variables $U_i^\dagger$ whose distributions are given by $Q$. The weak solution definition, which shares the same spirit as that for stochastic differential equations (see Ikeda and Watanabe (1981) and more in Section 4), will be very handy in facilitating our asymptotic analysis in this paper. For simplicity 
we drop index $\dagger$ and `weak' when there is no confusion. 
 
\subsection{Accelerated gradient descent algorithm}
\label{section-2-3}

Nesterov's accelerated gradient descent scheme can be used to solve the minimization problem (\ref{min-1}). 
Starting with initial values $x^n_0$ and $y^n_0=x^n_0$, we obtain the following iterative algorithm to compute the 
solution of the stochastic minimization problem (\ref{min-1}), 
\begin{align} \label{min-Nest2}
& x^n_k=y^n_{k-1}-\delta \nabla \! \cL^n(y^n_{k-1}; \bU_n), \qquad  y^n_k=x^n_k+\frac{k-1}{k+2}(x^n_k-x^n_{k-1}).
\end{align}

Using the continuous curve approach described  in Section \ref{section-1} we can define a 
step function $x^n_\delta(t) = x_k$ for $k \sqrt{\delta} \leq  t < (k+1) \sqrt{\delta} $, and for every $n$, as $\delta \rightarrow 0$, we 
approximate $x^n_\delta(t)$ 
%$\{x^n_k, k=0, 1, \cdots\}$ 
by a smooth curve $X^n(t)$, $t \geq 0$, governed by 
%\[ \ddot{X}^n(t)+\frac{3}{t}\dot{X}^n(t) + \nabla \! L^n(X^n(t); \bU_n)=0, \]
%that is, 
\begin{equation} \label{equ-3}
  \ddot{X}^n(t)+\frac{3}{t}\dot{X}^n(t) + \nabla \! \cL^n(X^n(t); \bU_n)=0,
%  \ddot{X}^n(t)+\frac{3}{t}\dot{X}^n(t) +  \frac{1}{n}\sum_{i=1}^n \nabla \! \ell(X^n(t); U_i)=0,
\end{equation} 
where initial values $X^n(0) = x^n_0$ and $\dot{X}^n(0)=0$, 
$\nabla \! \cL^n(X^n(t); \bU_n)= \frac{1}{n}\sum_{i=1}^n \nabla \! \ell(X^n(t); U_i)$, and 
gradient operator $\nabla$ here is applied to $\cL^n(\theta; \bU_n)$ and $\ell(\theta; U_i)$ with respect to $\theta$. 
$X^n(t)$ is a Lagrangian flow associated with $\cL^n$ in the optimization problem \eqref{min-1}. 

Again %as $X^n(t)$ are random, and our main interests are their distribution results, 
we define a solution $X^n(t)$ of the equation (\ref{equ-3}) in the weak sense, i.e., that there exist a process $X^n(t)$ and a random vector 
$\bU_n$ %random variables $U_1, \cdots, U_n$ defined 
on some probability space so that the distribution of $\bU_n$ is specified by $Q$, and $X^n(t)$ is a 
solution of the equation (\ref{equ-3}). 
%Note that $X^n(t)$ is not required to be defined on a fixed probability space with given random 
%variables, instead we define $X^n(t)$ for some probability space with some associated random variables $U_i$ whose distributions are given by $Q$. The weak solution definition, which shares the same spirit as that for stochastic differential equation, 
%will be very handy in facilitating our asymptotic analysis in this paper.

\subsection{Asymptotic theory via ordinary differential equations}
\label{section-2-4}

To make the equations (\ref{GD-c2}) and (\ref{equ-3}) and their solutions to be well defined and study their asymptotics we need to impose the following conditions. 
\begin{enumerate}

\item[A0.] Assume initial values satisfy $x^n_0 - x_0 = o_P(n^{-1/2})$. 
\item[A1.]  $\ell(\theta;u)$ is continuously twice differentiable in $\theta$;  $\forall$ $u\in R^p$, $\exists$ $h_1(u)$, such that $\forall$ $\theta^{1}, \theta^{2} \in \Theta$, $\lVert \nabla \!\ell(\theta^{1}; u)-\nabla\! \ell(\theta^{2}; u)\rVert \leqslant h_1(u)\lVert \theta^{1}-\theta^{2}  \rVert$, where $h_1(U)$ and 
$\nabla \ell(\theta_0;U)$ for some fixed $\theta_0$ have finite fourth moments. 
% for the stochastic gradient case or moment generating functions for the accelerated case. 
%$E[h^4_1(U)] < \infty$, and 
%%$E [\nabla \ell(\theta;U)]^4  ]<\infty$ for some $\theta$.  %for each $\theta$, $[\nabla \ell(\theta;U)]^2$ 
%$\nabla \ell(\theta;U)$ has a moment generating function.
\item[A2.] $E[\ell(\theta;U)]=g(\theta)$,  $E[ \nabla\! \ell(\theta;U)]= \nabla\! g(\theta)$, $E[ \boldsymbol{I\!\! H}\! \ell(\theta;U)]= \boldsymbol{I\!\! H}\! g(\theta)$,
%$E[ \nabla^\kappa  \ell(\theta;U)]= \nabla^\kappa  g(\theta)$, $\kappa =0, 1$, $E[  \boldsymbol{I\!\! H}\! \ell(\theta;U)]=\boldsymbol{I\!\! H}\!  g(\theta)$,
%and for some $\zeta>0$, $\sup\{  |\nabla\! \ell(\vartheta;u)|:  |\vartheta - \theta| > \zeta \}  \leqslant h_2(u)$ $a.s.$, where $E[h_2(U)]<\infty$.
%Assume that 
on the parameter space $\Theta$, $g(\cdot)$ is continuously twice differentiable and strongly convex, and 
  $\nabla g(\cdot)$ is $L$-Lipschitz for some $L>0$, %with linear growth, %and $\boldsymbol{I\!\! H}\! g(\cdot)$ is bounded,
where $\nabla$ is the gradient operator (the first order partial derivatives), and $\boldsymbol{I\!\! H}\!$  %$\boldsymbol{I\!\! H}\!$ 
is the Hessian operator (the second order partial derivatives). % operator equal to the Jacobian of gradient). 
%$\boldsymbol{I\!\! H}\! = \nabla^2$ is the Laplacian operator. %We denote by FL the class of convex functions f with L-Lipschitz continuous gradients defined on $I\!\!R^p$, that is, $g$ is convex, continuously differentiable, and satisfies 
%\[  \|  \nabla g(x) - \nabla g(y) \| \leq L \| x - y\|, \]
%for any $x, y \in I\!\! R^p$, , where $\| \cdot \|$ is the standard Euclidean norm and $L > 0$ is the Lipschitz constant. 

\item[A3.]  %$E[\frac{\partial}{\partial \theta_j}\ell(\theta;U)]^2<\infty$, 
Define cross auto-covariance $\bvarsigma(\theta, \vartheta) = (\varsigma_{ij}(\theta, \vartheta))_{1\leq i,j\leq p}$,  $\theta, \vartheta \in \Theta$, 
where \\ Cov$[\frac{\partial}{\partial \theta_i}\ell(\theta;U), \frac{\partial}{\partial \vartheta_j}\ell(\vartheta;U)] =\varsigma_{ij}(\theta,\vartheta)$ 
are assumed to be continuously differentiable, and $L$-Lipschitz. %for some $L>0$.  %with linear growth in $\theta$.
Let $\sigma_{ij}(\theta) =$ Cov$[\frac{\partial}{\partial \theta_i}\ell(\theta;U), \frac{\partial}{\partial \theta_j}\ell(\theta;U)] 
  =\varsigma_{ij}(\theta, \theta)$, %$i,j=1,2,...p$, 
and $\bsigma^2(\theta) =$ Var$[\nabla \! \ell(\theta;U)] = (\sigma_{ij}(\theta))_{1\leq i,j\leq p}=\bvarsigma(\theta, \theta)$ be positive definite.  %continuously differentiable, and $L$-Lipschitz for some $L>0$.

\item[A4.] %$\sqrt{n} [ \nabla^\kappa \!\cL^n(\theta; \bU_n) - \nabla^\kappa \!g(\theta)]$ weakly converges to a normal distribution with mean zero and variance 
%$\bsigma_{\kappa}(\theta)$ uniformly over $\theta \in \Theta_X$, where $\kappa=1, 2$, $\bsigma_\kappa(\theta)$ are positive definite and continuously differentiable, and $\bsigma_\kappa(\theta)$ with $\kappa=1$ corresponds to $\bsigma^2(\theta)$ given in A3; 

$\sqrt{n} [ \nabla \!\cL^n(\theta; \bU_n) - \nabla \!g(\theta)]=\frac{1}{\sqrt{n}} \sum_{i=1}^n [\nabla \ell(\theta; U_i) - \nabla g(\theta)]$ weakly converges to $\bZ(\theta)$ %a normal distribution with mean zero and variance $\bsigma^2(\theta)$ 
uniformly over $\theta \in \Theta_X$, where $\bZ(\theta)$ is a Gaussian process  with mean zero and auto-covariance 
$\bvarsigma(\theta, \vartheta)$ %$\bsigma^2(\theta)$ is 
defined in A3, $\Theta_X$ is a bounded subset of $\Theta$, and the interior of $\Theta_X$ contains the solutions $X(t)$ of the ordinary differential equations (\ref{GD-c1}) and (\ref{equ-2}) connecting the initial value $x_0$ and the minimizer of $g(\theta)$. 

\end{enumerate} 

Conditions A1-A2 are often used to make optimization problems and differential equations to be well defined, and match the stochastic sample  optimization problem (\ref{min-1}) to the deterministic population optimization problem (\ref{min-0}). Conditions A3-A4 guarantee that the 
solution of (\ref{min-1}) and its associated differential equations provide large sample approximations of those for (\ref{min-0}).
Condition A4 %is quite reasonable, which 
can be easily justified by empirical processes with common assumptions such as that 
$\nabla \!\ell(\theta; U)$,  %$\nabla^\kappa \!\ell(\theta; U)$,  
$\theta \in \Theta_X$, form  a Donsker class (van der Vaart and Wellner (2000)), 
since the solution curves $X(t)$ of the ordinary differential equations (\ref{GD-c1}) and (\ref{equ-2}) are deterministic and bounded, and it is easy to select $\Theta_X$.  

%Let $I\!\!R_+=[0, \infty)$, denote by $C(I\!\!R_+)$  the space of all continuous functions on $I\!\!R_+$, equipped with a metric 
%$d$ for the topology of uniform convergence on compacta: 
%\[ d(h_1, h_2) = \sum_{r=1}^\infty 2^{-r} min \left\{ 1, \max_{0 \leq s \leq r} | h_1(s) - h_2(s)| \right\}. \]

For a given $T>0$, denote by $C([0, T])$ the space of all continuous functions on $[0, T]$ with the uniform metric 
%$d$ for the topology of uniform convergence on compacta: $d(h_1, h_2) = 
$ \max \{| b_1(t) - b_2(t)|:   t \in [0, T]\}$ between functions $b_1(t)$ and $b_2(t)$.
% and denote by $D([0, T])$ the Skorokhod space of all c\'adl\'ag functions on $[0, T]$, equipped with the Skorokhod metric (Billingsely (1999)).
For the solutions $X(t)$ and $X^n(t)$ of the ordinary differential equations (\ref{GD-c1}) and (\ref{GD-c2}) [or (\ref{equ-2}) and (\ref{equ-3})],  
respectively, we define $V^n(t) = \sqrt{n} [ X^n(t) - X(t) ]$.
%Then $X(t)$, $X^n(t)$ and $V^n(t)$ live on $C(I\!\!R_+)$. Treating them as random elements in $C(I\!\!R_+)$, 
Then $X(t)$, $X^n(t)$ and $V^n(t)$ live on $C([0,T])$. Treating them as random elements in $C([0,T])$, 
in the following theorem we establish a weak convergence limit of $V^n(t)$. 

%we show in the following theorem that tof $V^n(t)$ has the weak convergence limit $V(t)$ governed by the following linear differential equation,

\begin{thm} \label{thm-1}
Under conditions A0-A4, as $n \rightarrow \infty$, $V^n(t)$ weakly converges to a Gaussian process 
$V(t)$, where $V(t)$ is the unique solution of the following linear 
differential equations 
\begin{equation} \label{GD-limit-00}
  \dot{V}(t) + [\boldsymbol{I\!\! H}\! g(X(t))] V(t) + \bZ(X(t))  = 0, \;\; V(0) = 0,  
\end{equation}
for the plain gradient descent case, and 
\begin{equation} \label{limit-00}
\ddot{V}(t) + \frac{3}{t}  \dot{V}(t) + [\boldsymbol{I\!\! H}\! g(X(t))] V(t) +  \bZ(X(t))  = 0, \;\; V(0) = \dot{V}(0) = 0, 
\end{equation}
for the accelerated gradient descent case, where the deterministic functions $X(t)$ in (\ref{GD-limit-00}) and (\ref{limit-00}) are the solutions of the ordinary differential equations (\ref{GD-c1}) and (\ref{equ-2}), respectively, $\boldsymbol{I\!\! H}$ is the Hessian operator, random coefficient 
$\bZ(\cdot)$ is the Gaussian process given by  Condition A4. 

In particular if Gaussian process $\bZ(\theta) = \sigma(\theta) \bZ$, where random variable $\bZ \sim N_p(0, \bI_p)$, and $\bsigma(\theta)$ 
is defined in Condition A3, then $V(t)=\Pi(t) \bZ$ on $C([0,T])$, and the deterministic matrix $\Pi(t)$ is the unique solution of the following linear differential equations 
\begin{equation} \label{GD-limit-0}
  \dot{\Pi}(t) + [\boldsymbol{I\!\! H}\! g(X(t))] \Pi(t) + \bsigma(X(t))  = 0, \;\; \Pi(0) = 0,  
  \end{equation}
for the plain gradient descent case, and 
\begin{equation} \label{limit-0}
\ddot{\Pi}(t) + \frac{3}{t}  \dot{\Pi}(t) + [\boldsymbol{I\!\! H}\! g(X(t))] \Pi(t) + \bsigma(X(t))  = 0, \;\; \Pi(0) = \dot{\Pi}(0) = 0, 
\end{equation}
for the accelerated gradient descent case, where $X(t)$ in (\ref{GD-limit-0}) and (\ref{limit-0}) are the solutions of the ordinary differential equations (\ref{GD-c1}) and (\ref{equ-2}), respectively,  $\boldsymbol{I\!\! H}$ is the Hessian operator, and $\bsigma(\cdot)$ is defined in Condition A3. 
\end{thm}

\begin{remark} \label{remark0}
As discussed in Sections \ref{section-1} and \ref{section-2-2}, for the gradient descent case $X(t)$ and $X^n(t)$ are gradient flows 
associated with the population optimization \eqref{min-0} and the sample optimization \eqref{min-1},  respectively, and thus refer to the 
corresponding population and sample gradient flows. As a consequence,  the Gaussian limiting distribution of $V^n(t)$ describes the asymptotic distribution of the difference between the sample and population gradient flows,  with a normalization factor $\sqrt{n}$. 
%The solution of the linear differential equation \eqref{GD-limit-00} has an expression $V(t) = \Pi_0(t) \int_0^t [\Pi_0(s)]^{-1} \bZ(X(s)) ds$, where $\Pi_0(t)$ is a $p$ by $p$ deterministic matrix constructed by the Magnus expansion for solving the homogeneous linear differential equation  $ \dot{V}(t) + [\boldsymbol{I\!\! H}\! g(X(t))] V(t) =0$ (see Blanes et al. (2009)). Thus the limiting distribution of $V^n(t)$ is Gaussian. As a matter of fact, the theorem shows that in the special case of  $\bZ(\theta) = \sigma(\theta) \bZ$, we have $V(t)=\Pi(t) \bZ$, which clearly indicates the Gaussian limiting distribution, and furthermore we can easily prove $\Pi(t) = \Pi_0(t) \int_0^t [\Pi_0(s)]^{-1} \bsigma(X(s)) ds$. 
Hence, it is natural to view the Gaussian limiting distribution as the central limit theorem for the gradient flows, and we call it the gradient flow central limit theorem (GF-CLT). 
Similarly, for the accelerated case $X(t)$ and $X^n(t)$ are Lagrangian flows associated with the population optimization \eqref{min-0} and the sample optimization \eqref{min-1},  respectively, and thus refer to the corresponding population and sample Lagrangian flows. 
%We may convert the second-order homogeneous linear differential equation  $\ddot{V}(t) + \frac{3}{t}  \dot{V}(t) + [\boldsymbol{I\!\! H}\! g(X(t))] V(t) =0$ into an equivalent first-order homogeneous linear differential equation system 
%$\dot{V}(t) - \Xi(t)=0$ and $\dot{\Xi}(t) + [\boldsymbol{I\!\! H}\! g(X(t))] V(t) + \frac{3}{t}  \Xi(t)  =0$, where $\Xi(t) = \dot{V}(t)$. 
%We apply the Magnus expansion to solve the first-order homogeneous linear differential equation system and then show that the solution of the differential equation \eqref{limit-00} linearly depends on $Z(\cdot)$. 
The Gaussian limiting distribution for the normalized discrepancy between the sample and population Lagrangian 
flows can be naturally viewed as the central limit theorem for the Lagrangian flows, and we call it 
the Lagrangian flow central limit theorem (LF-CLT). 
\end{remark}

\begin{remark} \label{remark1}
As we discussed early in Section \ref{section-2-1}, as  $n \rightarrow \infty$, $\cL^n(\theta; \bU_n)=\frac{1}{n}\sum_{i=1}^n \ell(\theta; U_{i})$ converges to $g(\theta)$ in probability, and the solutions of the minimization problems (\ref{min-0}) and (\ref{min-1}) should be very close to each other. %Specifically 
We may heuristically illustrate the derivation of Theorem \ref{thm-1} as follows. 
Central limit theorem may lead us to 
%from the proof of Theorem \ref{thm-1}
see that as $n \rightarrow \infty$, $\nabla \cL^n(\theta; \bU_n)$ is asymptotically distributed as $\nabla g(\theta) + n^{-1/2} \bZ(\theta)$. 
%$\nabla g(\theta) + n^{-1/2} \bsigma(\theta) \bZ$, where random vector $\bZ \sim N(0, \bI_p)$.  
Then asymptotically the differential equations (\ref{GD-c2}) and (\ref{equ-3}) 
are, respectively, equivalent to 
\begin{align} \label{GD-c2-1}
&  \dot{X}^n(t) + \nabla g(X^n(t))  + n^{-1/2} \bZ (X^n(t))   %\bsigma (X^n(t)) \bZ
 = 0, &\\
& \ddot{X}^n(t) + \frac{3}{t} \dot{X}^n(t) + \nabla g(X^n(t))  + n^{-1/2}   \bZ (X^n(t))   %\bsigma (X^n(t)) \bZ 
  = 0. & \label{equ-3-1}
\end{align} 
Applying the perturbation method for solving ordinary differential equations, we write approximation solutions of the equations 
(\ref{GD-c2-1}) and (\ref{equ-3-1}) as $X^n(t) \approx X(t) + n^{-1/2} V(t) $ % + o(n^{-1/2})$ 
and substitute it into (\ref{GD-c2-1}) and (\ref{equ-3-1}). 
With $X(t)$ satisfying the ordinary differential equations (\ref{GD-c1}) or (\ref{equ-2}), using the Taylor expansion and ignoring higher order terms, we can easily 
obtain  the equations (\ref{GD-limit-00}) and (\ref{limit-00}) % the following equations 
for the limit $V(t)$ of $V_n(t)$ in the two cases, respectively. 
% \begin{equation} % \label{GD-limit-1}
%  \dot{V}(t) + [\boldsymbol{I\!\! H}\! g(X(t))] V(t) + \bZ(X(t))   %\bsigma(X(t)) \bZ
%     = 0, 
%\end{equation}
%\begin{equation} % \label{limit-1}
%\ddot{V}(t) + \frac{3}{t}  \dot{V}(t) + [\boldsymbol{I\!\! H}\! g(X(t))] V(t) + \bZ(X(t)) %\bsigma(X(t)) \bZ 
%= 0, 
%\end{equation}
%where %$\boldsymbol{I\!\! H}\! = \nabla^2$ is Laplacian operator, 
%$X(t)$ is a solution of the corresponding equation (\ref{GD-c1}) or (\ref{equ-2}), $\bZ(\cdot)$ is the Gaussian process defined in A4, 
% and initial conditions $V(0) = \dot{V}(0)=0$. 
 %For $\bZ(\theta) = \bsigma(\theta) \bZ$ with random variable $\bZ \sim N_p(0, \bI_p)$, 
% % random variable $\bZ$ follows the  p-variate standard normal distribution $N_p(0, \bI_p)$, 
 %using linear scaling we show that (\ref{GD-limit-1}) and (\ref{limit-1}) have unique solutions $V(t)=\Pi(t) \bZ$, where $\Pi(t)$ are unique solutions of (\ref{GD-limit-0}) and (\ref{limit-0}). 
\end{remark}

The step function $x^n_\delta(t)$ is used to model $x^n_k$ generated from the gradient descent algorithms (\ref{min-GD2}) and 
(\ref{min-Nest2}). To study their weak convergence, we need to introduce the Skorokhod space, denoted by $D([0, T])$, of all c\'adl\'ag functions on $[0, T]$, equipped with the Skorokhod metric (Billingsely (1999)). Then $x^n_\delta(t)$ lives on $D([0,T])$, and treating it 
as a random element in $D([0,T])$, we derive its weak convergence limit in the following theorem.

\begin{thm} \label{thm-1-1}
Under assumption A0-A4, as $\delta \rightarrow 0$ and $n \rightarrow \infty$, we have 
\[   \sup_{t \in [0, T]} | x_\delta^n(t) - X^n(t) | = O_P(\delta^{1/2} | \log \delta| ), \]
where $x^n_\delta(t)$ are the continuous-time step processes for discrete $x^n_k$ generated from the algorithms (\ref{min-GD2}) and (\ref{min-Nest2}), with continuous curves $X^n(t)$ defined by the ordinary differential equations (\ref{GD-c2}) and (\ref{equ-3}), 
for the cases of plain and accelerated gradient descent algorithms, respectively. 
%and continuous curves $X^n(t)$ are defined by (\ref{GD-c2}) and (\ref{equ-3}) for  the plain and accelerated cases, respectively. 
In particular, we may choose $(n,\delta)$ such that $n \delta  |\log \delta|^2 \rightarrow 0$ 
as $\delta \rightarrow 0$ and $n \rightarrow \infty$, 
and then for the chosen $(n,\delta)$, $n^{1/2} [ x^n_\delta(t) - X(t)]$ weakly converges to $V(t)$  on $D([0, T])$, 
%if we take $(n,\delta)$ such that as $\delta \rightarrow 0$ and $n \rightarrow \infty$, $n \delta  |\log \delta|^2 \rightarrow 0$, 
%then on $D([0, T])$, $n^{1/2} [ x^n_\delta(t) - X(t)]$ weakly converges to $V(t)$, %$\Pi(t) \bZ$, 
where $X(t)$ is the solution of the ordinary differential equations (\ref{GD-c1}) or (\ref{equ-2}), and $V(t)$ %$\Pi(t)$  and $\bZ$ are 
is given by Theorem \ref{thm-1}. That is, $\sqrt{n} [ x^n_\delta(t) - X(t)]$ and $\sqrt{n} [ X^n(t) - X(t)] $ share the same weak convergence limit. 
\end{thm}

\begin{remark} \label{remark2}
There are two types of asymptotic analyses in the set up. One type is to employ continuous differential equations to model discrete sequences generated from the gradient descent algorithms, which is associated with $\delta$ treated as the step size between consecutive sequence points. Another type involves the use of random %empirical estimators of true objective functions in minimization problems, 
objective functions in stochastic optimization, which are estimated from sample data of size $n$.
%is associated with the size $n$ of sampled data used to estimate the objective functions. 
We refer the first and second types as computational %(modeling) 
and statistical asymptotics, respectively. 
The computational asymptotic analysis is that for each $n$, the ordinary differential equations (\ref{GD-c2}) and (\ref{equ-3})[or (\ref{GD-c2-1}) and (\ref{equ-3-1})] provide continuous solutions as the limits of discrete sequences generated from the algorithms (\ref{min-GD2}) and (\ref{min-Nest2}), respectively, 
when $\delta$ is allowed to go to zero.  Theorem \ref{thm-1} provides the statistical asymptotic analysis to describe the behavior difference between the %data based solution 
sample gradient flow $X^n(t)$ and the %true solution 
population gradient flow $X(t)$, as the sample size $n$ goes to infinity. 
%As studied in Su et al. (2016, Theorem 2) for (\ref{equ-Nest1}) and (\ref{equ-2}), we has shown in  the proof of Theorem \ref{thm-1} that 
Theorem \ref{thm-1-1} involves both types of asymptotics and shows that %uniformly over $n$, 
as $\delta \rightarrow 0$ and $n \rightarrow \infty$, 
$x^n_\delta(t) - X^n(t)$ is of order $\delta^{1/2}$. %go to zero as fast as some power of $\delta$. 
%As a computational modeling parameter, $\delta$ can be any arbitrary sequence approaching zero, and we may let 
%it depend on $n$ and choose $\delta=\delta_n$ that goes to zero fast enough 
It is easy to choose $(n, \delta)$ so that $x^n_{\delta_n}(t) - X^n(t)$ is of order smaller than $n^{-1/2}$. Then $x^n_{\delta_n}(t)$ has the same asymptotic distribution $V(t)$ as $X^n(t)$. 
%As $\delta \rightarrow 0$, 
%\[  \max_{0 \leq t \leq T}  | x^n(t) - X^n(t)| = O(\delta^{1/2}). \]
%If $\delta = o(n^{-1})$, then $\sqrt{n} [ x^n(t) - X(t)]$ has the same weak limit as $V^n(t)$ given in Theorem 2.

\end{remark}

\subsection{A framework to unify computational and statistical asymptotic analysis}
\label{section-2-5}

%\begin{remark} \label{remark3}
The two types of asymptotics associated with $\delta$ and $n$ seem to be  quite different, with one for computational algorithms and one for 
statistical procedures. This section will elaborate further about these analyses and provide a framework to unify both viewpoints.   
%We have seen the situation with $\delta\rightarrow 0$ %first and then $n \rightarrow \infty$ 
%in Remark \ref{remark2}. We may describe the scenario analysis below with $n \rightarrow \infty$. %when we first let $n$ goes to infinity. 
Denote the solutions of the optimization problems (\ref{min-0}) and (\ref{min-1}) by $\check{\theta}$ and $\hat{\theta}_n$, respectively. In the 
statistical set-up, %like M-estimation, 
$\check{\theta}$ and $\hat{\theta}_n$ represent the true estimand %true parameter value and 
and its associated estimator, %of the parameter $\theta$, 
respectively. 
Using the definitions of $\check{\theta}$ and $\hat{\theta}_n$ and the Taylor expansion, we have $\nabla g(\check{\theta}) =0$, 
\[   0 = \nabla \cL^n(\hat{\theta}_n; \bU_n) = \nabla \cL^n(\check{\theta}; \bU_n) + \boldsymbol{I\!\! H}\! \cL^n(\check{\theta}; \bU_n) (\hat{\theta}_n - \check{\theta}) + \mbox{reminder},  \]
 the law of large number implies that $\boldsymbol{I\!\! H}\! \cL^n(\check{\theta}; \bU_n)$ converges in probability to $\boldsymbol{I\!\! H}\! g(\check{\theta})$ as $n \rightarrow \infty$, and 
 Condition A4 indicates that 
\[ \nabla \cL^n(\check{\theta}; \bU_n) = \nabla g(\check{\theta}) + n^{-1/2} \bZ(\check{\theta}) %\bsigma(\check{\theta}) \bZ 
+ \mbox{reminder}
= n^{-1/2} \bsigma(\check{\theta}) \bZ + \mbox{reminder}, \]
where $\bZ$ stands for a standard normal random vector. 
Thus, $n^{1/2} (\hat{\theta}_n - \check{\theta})$ is asymptotically distributed as $[ \boldsymbol{I\!\! H}\! g(\check{\theta}) ]^{-1} 
     \bsigma(\check{\theta}) \bZ$. 
%which leads to the asymptotic distribution of $n^{1/2} (\hat{\theta}_n - \check{\theta})$. The asymptotic derivation can be illustrated from 
%the optimization point of view. We apply Taylor expansion to obtain 
%\[  \cL^n(\theta, \bU_n) = \cL^n(\check{\theta}, \bU_n) + \nabla \cL^n(\check{\theta}, \bU_n) (\theta - \check{\theta})
%+ \boldsymbol{I\!\! H}\! \cL^n(\check{\theta}, \bU_n) (\theta - \check{\theta})^2/2 + \mbox{reminder}. \] 
%Define $\vartheta = n^{1/2} (\theta - \check{\theta})$. Substitute $\vartheta$ into above equation to turn $\cL^n(\theta, \bU_n)$ into 
%\[ \cL^n(\check{\theta}, \bU_n) + n^{-1} \bsigma(\check{\theta} ) \bZ \vartheta + n^{-1}  \boldsymbol{I\!\! H}\!  g(\check{\theta}) \vartheta^2/2 + \mbox{reminder}. \] 
%The minimizer $\hat{\theta}_n$ of $\cL^n(\theta, \bU_n)$ is immediately converted into the minimizer $\hat{\vartheta}_n = 
%n^{1/2} (\hat{\theta}_n- \check{\theta})$, which asymptotically minimizes $\bsigma(\check{\theta} ) \bZ \vartheta +  
%\boldsymbol{I\!\! H}\!  g(\check{\theta})  \vartheta^2/2$, and is asymptotically distributed as $[ \boldsymbol{I\!\! H}\!  g(\check{\theta})]^{-1} 
%\bsigma(\check{\theta} ) \bZ$. Such asymptotic analysis is often rigorously carried out by empirical process arguments.
On the other hand,  the gradient descent algorithms generate sequences corresponding to $X(t)$ and $X^n(t)$, which are expected to approach the solutions of the two optimization problems  (\ref{min-0}) and (\ref{min-1}), respectively. Hence $X(t)$ and $X^n(t)$ must move towards $\check{\theta}$ and $\hat{\theta}_n$, respectively, and $V_n(t)$ and $V(t)$ are reaching their corresponding targets $n^{1/2} (\hat{\theta}_n- \check{\theta})$ and $[ \boldsymbol{I\!\! H}\! g( \check{\theta})]^{-1} \bsigma(\check{\theta} ) \bZ$. Below we will provide a framework to connect 
$(X^n(t), X(t))$ with $(\hat{\theta}_n, \check{\theta})$ and $(V^n(t), V(t))$ with $\left( n^{1/2}(\hat{\theta}_n - \check{\theta}), 
[ \boldsymbol{I\!\! H}\! g( \check{\theta})]^{-1} \bsigma(\check{\theta} ) \bZ \right)$.
%\end{remark}

Since the time interval considered so far is $[0, T]$ for any arbitrary $T>0$, we may extend the time interval to $I\!\!R_+=[0, +\infty)$, and consider $C(I\!\!R_+)$, the space of all continuous functions on $I\!\!R_+$, equipped with a metric $d$ for the topology of uniform convergence on compacta: 
\[ d(b_1, b_2) = \sum_{r=1}^\infty 2^{-r} min \left\{ 1, \max_{0 \leq s \leq r} | b_1(s) - b_2(s)| \right\}. \]
The solutions $X(t)$, $X^n(t)$, $V(t)$ and $V^n(t)$ of the ordinary differential equations (\ref{GD-c1}), (\ref{equ-2}), (\ref{GD-c2}),   
(\ref{equ-3})-(\ref{equ-3-1}) 
%(\ref{limit-1}) 
%(\ref{GD-c1}) \&(\ref{GD-c2}),  (\ref{equ-2}), (\ref{equ-3}), and (\ref{GD-limit-0})-(\ref{limit-1}) 
all live on $C(I\!\!R_+)$, and we can study their weak convergence on $C(I\!\!R_+)$. Similarly we may adopt the Skorokhod space $D(I\!\!R_+)$ equipped with the Skorokhod metric for the weak convergence study of $x^n_\delta(t)$ (see Billingsely (1999)).
%$C(I\!\!R_+)$ is a subspace of $D(I\!\!R_+)$, and 
%%because of their metrics used, %defined in $C(I\!\!R_+)$ and $D(I\!\!R_+)$, 
%the weak convergence of these process on $D(I\!\!R_+ )$ is determined by their weak convergence on $D([0, T])$ for all integers $T$ only (see Billingsely (1999)).  
The following theorem establishes the weak convergence of these processes on $D(I\!\!R_+ )$ and studies their 
asymptotic behaviors as $t \rightarrow \infty$.

%Treating them as random elements in $D(I\!\!R_+)$, 
%since Theorems \ref{thm-1} and \ref{thm-1-1} establish the weak convergence of $X^n(t)$ and $x^n_\delta(t)$ to $V(t)=U(t) \bZ$ on $[0, T]$ for any $T>0$, we may conclude from these weak convergence results that $V^n(t) = \sqrt{n} [X^n(t) - X(t)]$ and $\sqrt{n} [x^n_\delta(t) - X(t)]$, $t \in [0, +\infty)$, weakly converge to $V(t)$, $t \in [0, +\infty)$.

\begin{thm} \label{thm-1-2}
Suppose that the assumption A0-A4 are met, $\boldsymbol{I\!\! H}\! g(\check{\theta})$ is positive definite, % = \boldsymbol{I\!\! H}\! g(X(\infty)) > 0$,
all eigenvalues of $\int_0^t \boldsymbol{I\!\! H}\! g(X(s)) ds $ diverge as $t \rightarrow \infty$, %(in terms of matrix spectral norm),  
$\boldsymbol{I\!\! H}\! g(\theta_1)$ and $\boldsymbol{I\!\! H}\! g(\theta_2)$ commute for any $\theta_1 \neq \theta_2$, 
and $n \delta |\log \delta|^2 \rightarrow 0$ as $\delta \rightarrow 0$ and $n \rightarrow \infty$. 
 Then on  $D(I\!\!R_+ )$,  as $\delta \rightarrow 0$ and $n \rightarrow \infty$, 
 $V^n(t) = \sqrt{n} [X^n(t) - X(t)]$ and $\sqrt{n} [x^n_\delta(t) - X(t)]$ weakly converge to 
 $V(t)$, $t \in [0, +\infty)$. 
 
 Furthermore,  for the plain gradient descent case we have as $t  \rightarrow \infty$ and $k \rightarrow \infty$, 
 \begin{itemize}
 \item[(1)]  $x_k$, $x_\delta(t)$ and $X(t)$ converge to $\check{\theta}$, where $x_k$, $x_\delta(t)$ and $X(t)$ are defined 
 in Section \ref{section-1}  (see the ordinary differential equations  (\ref{equ-GD1})-(\ref{equ-Nest1}) and (\ref{equ-2})). 
    
 \item[(2)]  $x^n_k$, $x^n_\delta(t)$ and $X^n(t)$ converge to $\hat{\theta}_n$ in probability, and thus $V^n(t)$ converges to $\sqrt{n} (\hat{\theta}_n - \check{\theta})$ in probability, where $x^n_k$, $x^n_\delta(t)$ and $X^n(t)$ are defined in the algorithms and equations 
 (\ref{min-GD2})-(\ref{equ-3}). 

 \item[(3)] 
 The limiting distributions of $V(t)$ as $t \rightarrow \infty$ and $\sqrt{n} (\hat{\theta}_n - \check{\theta})$ as $n \rightarrow \infty$ are identical and given by a normal distribution with mean zero and variance 
$[ \boldsymbol{I\!\! H}\! g(\check{\theta})]^{-1} \bsigma^2(\check{\theta})  [ \boldsymbol{I\!\! H}\! g(\check{\theta})]^{-1}$, where $V(t)$, defined in the ordinary differential equations (\ref{GD-limit-00}) and (\ref{limit-00}),  is the weak convergence limit of $V^n(t)$ as $n \rightarrow \infty$. 
\end{itemize}
\end{thm}

\begin{remark} \label{remark3-3}
Denote the limits of the processes in Theorem \ref{thm-1-2} as $t,k \rightarrow \infty$ by the corresponding processes with $t$ and $k$ 
replacing by $\infty$. Then  Theorem \ref{thm-1-2} shows that for the plain gradient descent case, 
$x_\infty = x_\delta(\infty) = X(\infty) = \check{\theta}$,  $x^n_\infty = x^n_\delta(\infty) = X^n(\infty) = \hat{\theta}_n$,  $V^n(\infty)=\sqrt{n} [X^n(\infty)  - X(\infty)] = \sqrt{n} [x^n_\delta(\infty)  - X(\infty)] = \sqrt{n} ( \hat{\theta}_n - \check{\theta})$, 
 $V(\infty)= [\boldsymbol{I\!\! H}\! g(X(\infty)) ]^{-1} \bsigma(X(\infty)) \bZ = [\boldsymbol{I\!\! H}\! g(\check{\theta}) ]^{-1} \bsigma(\check{\theta}) \bZ$,  
 $V(t)$ weakly converges to $V(\infty)$ as $t \rightarrow \infty$, and $V^n(\infty)$ weakly converges to $V(\infty)$ 
 as $n \rightarrow \infty$.  In particular, as the process $V^n(t)$ is indexed by $n$ and $t$, its limits are the same regardless the order 
 of $n \rightarrow \infty$ and $t \rightarrow \infty$. Also as $\check{\theta}=X(\infty)$ is the minimizer of the convex function $g(\cdot)$, the positive definite assumption 
 $\boldsymbol{I\!\! H}\! g(\check{\theta}) = \boldsymbol{I\!\! H}\! g(X(\infty))>0$ is very reasonable; since the limit $\boldsymbol{I\!\! H}\! g(X(\infty))$ of $\boldsymbol{I\!\! H}\! g(X(t))$ as $t \rightarrow \infty$ has all 
 positive eigenvalues, it is natural to expect that $\int_0^\infty \boldsymbol{I\!\! H}\! g(X(s)) ds$ has diverging eigenvalues. 
 We conjecture that for the accelerated gradient descent case, similar asymptotic results might hold as $k, t \rightarrow \infty$.
 \end{remark}

 With the augmentation of $t=\infty$, we extend $[0, +\infty)$ further to $[0, +\infty]$, consider $X(t)$, $x_\delta(t)$, 
$X^n(t)$, $x^n_\delta(t)$, $V(t)$, and $V^n(t)$ on $t \in [0, \infty]$ and derive the limits of $V^n(t)$ and $\sqrt{n} [x^n_\delta(t) - X(t)]$ 
on $[0, \infty]$ by Theorem \ref{thm-1-2}. 
As $\delta \rightarrow 0$ and $n \rightarrow \infty$,
the limiting distributions of $V^n(t)=\sqrt{n}[ X^n(t) - X(t)]$ and $\sqrt{n} [ x^n_\delta(t) - X(t)]$ are $V(t)$ for $t \in [0, \infty]$, 
where $(V^n(t), V(t))$ describe the dynamic evolution of the gradient descent algorithms  for $ t \in [0, \infty)$ and the 
statistical distribution of $\sqrt{n}( \hat{\theta}_n -\check{\theta})$ for $t =\infty$.

The joint asymptotic analysis provides a unified framework to describe distribution limits of $X^n(t)$ and $x^n_\delta(t)$ 
from both computation and statistical viewpoints as follows. For $t \in [0, \infty)$, $X(t)$ and $V(t)$ give the limiting behaviors 
of $X^n(t)$ and $x^n_\delta(t)$ corresponding to the computational algorithms, and $X(\infty)$ and $V(\infty)$ illustrate their limiting 
behaviors of the corresponding statistical decision rule  $\hat{\theta}_n$ (or the exact solutions of the corresponding optimization problems 
(\ref{min-0}) and (\ref{min-1}) that the algorithms are designed to compute).  We use the following simple example to explicitly illustrate 
the joint asymptotic analysis. \\

\textbf{Example 1}. Suppose that $U_i = (U_{1i}, U_{2i})^\prime$, $i=1, \cdots, n$, are iid random vectors,  %and follow from a distribution $Q$, 
where $U_{1i}$ and $U_{2i}$ are independent, and follow a normal distribution $N(\theta_1, \tau^2)$ 
 %a distribution $F_0( \frac{u - \mu}{\sigma})$ with symmetric $F_0$ 
with mean $\theta_1$ and variance $\tau^2$ 
and an exponential distribution with mean $\theta_2$, respectively, and $\theta=(\theta_1, \theta_2)^\prime$.  Define 
$\ell(\theta; U_i) = (U_i - \theta)^\prime (U_i - \theta)/2$, 
%$\ell(\theta; U_i) =[ (U_i - \mu)/\sigma]^\prime [(U_i - \mu)/\sigma]$, 
and denote by $\check{\theta}$ the true value of the parameter $\theta$ in the model. Then $\cL(\theta; \bU_n) = \frac{1}{n} \sum_{i=1}^n (U_i - \theta)^\prime (U_i - \theta)/2$, %is the mean square error, 
$g(\theta) = E[\ell(\theta; U_i) ] =  [(\theta - \check{\theta})^\prime (\theta - \check{\theta}) + \tau^2 + \check{\theta}_2^2]/2$,
$\nabla g(\theta) = \theta - \check{\theta}$, $\nabla \ell(\theta; U_i) = \theta - U_i$, $\nabla \cL(\theta; \bU_n) = \theta - \bar{U}_n$,
and $\bsigma^2(\theta) =Var(U_1 - \theta)= \mbox{diag}( \tau^2, \check{\theta}_2^2)$, 
where $\bar{U}_n=(\bar{U}_{1n}, \bar{U}_{2n})^\prime$  is the sample mean.
It is easy to see that the corresponding minimization problems 
(\ref{min-0}) and (\ref{min-1}) have explicit solutions: $g(\theta)$ has the minimizer $\check{\theta}$,  and 
$\cL(\theta; \bU_n) $ has the minimizer $\hat{\theta}_n= \bar{U}_n$. For this example, the algorithms (\ref{equ-GD1}), 
(\ref{min-GD2}),  (\ref{equ-Nest1})  and (\ref{min-Nest2}) yield recursive formulas $x_k = x_{k-1} + \delta (\check{\theta} - x_{k-1})$, and 
$x^n_k = x^n_{k-1} + \delta (\bar{U}_n  - x^n_{k-1})$ for the plain gradient descent case; and 
$x_k = x_{k-1} + \delta (\check{\theta} - y_{k-1})$, $y_k = x_k + \frac{k-1}{k+2} (x_k - x_{k-1})$, 
$x^n_k = x^n_{k-1} + \delta (\bar{U}_n  - y^n_{k-1})$, $y^n_k = x^n_k + \frac{k-1}{k+2} (x^n_k - x^n_{k-1})$
for the accelerated gradient descent case. While it may not be so obvious to explicitly describe 
the dynamic behaviors of these algorithms %in particular 
for the accelerated case, 
%(\ref{equ-GD1}) and (\ref{min-GD2}) as well as (\ref{equ-Nest1})  and (\ref{min-Nest2}) (particularly for the accelerated case), 
below we will clearly illustrate 
the behaviors of their corresponding ordinary differential equations through closed form expressions. First we consider the plain gradient descent case where 
closed form expressions are very simple.  The ordinary differential equations (\ref{GD-c1}) and (\ref{GD-c2}) admit simple solutions 
\[ X(t) = (X_1(t), X_2(t))^\prime = \check{\theta} + (x_{0} - \check{\theta}) e^{-t}, \;\;  X^n(t) = (X^n_1(t), X^n_2(t) )^\prime = \bar{U}_n  +  (x^n_{0} - \bar{U}_n )  e^{-t},  \]
\[ V^n(t) =  (V^n_1(t), V^n_2(t) )^\prime = \sqrt{n} ( \bar{U}_n  - \check{\theta})  ( 1 - e^{-t} ) + \sqrt{n} 
 (x^n_0 - x_0)  e^{-t} . \]
Note that $Z_1 = \sqrt{n} (\bar{U}_{1n} - \check{\theta}_1)/\tau \sim N(0,1)$, $\sqrt{n} (\bar{U}_{2n} /\check{\theta}_2 - 1)$ converges in distribution to a standard normal random variable $Z_2$, and $Z_1$ and $Z_2$ are independent.  As in Theorem \ref{thm-1}, let 
$\bZ=(Z_1, Z_2)^\prime$, $V(t) = \Pi(t) \bZ$, where $\Pi(t)=- ( 1 - e^{-t} ) \mbox{diag}(\tau, \check{\theta}_2)$ is the matrix solution of the linear differential equation (\ref{GD-limit-0}) in this case. Then for  
$t \in [0, \infty)$,
\[ V^n(t) = \left( \begin{array}{l}  \tau Z_1 \\ \check{\theta}_2 Z_2 \end{array} \right) (1 - e^{-t} ) + o_P(1)
     = V(t) + o_P(1), \]
which confirms that $V^n(t)$ converges to $V(t)$,  as shown in Theorem \ref{thm-1}.  Furthermore, as $t \rightarrow \infty$,  $X(t) \rightarrow \check{\theta} = X(\infty)$,  $X^n(t) \rightarrow \hat{\theta}_n = 
\bar{U}_n = X^n(\infty)$, and $V^n(t) \rightarrow V^n(\infty) = \sqrt{n} (\bar{U}_n  - \check{\theta})$; as $n \rightarrow \infty$, 
$V^n(\infty) \rightarrow V(\infty) =\Pi(\infty) \bZ = -(\tau Z_1, \check{\theta}_2 Z_2)^\prime$,
which gives the asymptotic distribution of the estimator $\hat{\theta}_n = X^n(\infty)$. In summary, the behaviors of 
$X(t)$, $X^n(t)$,  $V^n(t)$, and $V(t)$ over $[0, \infty]$ provide a complete description on the dynamic evolution of the gradient descent algorithms when applied to solve the stochastic sample optimization problem. %statistical and machine learning problems  
For example, as functions of $t$, $X(t)$ and $X^n(t)$ can be used to describe how the sequences generated from the algorithms evolve 
along iterations;  we may use the convergence of $V^n(t)$ to $V^n(\infty)$ and $V(t)$ to $V(\infty)$, as $t \rightarrow \infty$, to 
illustrate how the generated sequences converge to the target optimization solutions (estimators); the convergence of $V^n(\infty)$ to $V(\infty)$ as $n \rightarrow \infty$ may be employed to characterize the asymptotic distributions of the target optimization solutions;
and their relationship with $n$ and $t$ can be used to investigate the joint dynamic effect of data size and algorithm iterations on 
the computational and statistical errors in the sequences generated by the algorithms. The key signature in this case is the 
exponential decay factor $e^{-t}$ that appears in all relationships. 
The joint asymptotic analysis with both $n$ and $t$ provides a unified picture for the statistical asymptotic analysis with 
 $n \rightarrow \infty$ and the computational asymptotic analysis with $t \rightarrow \infty$. 
 %It illustrates the dynamic structure  as $t$ evolves. 
  
  For the accelerated case, the solution $X(t)$ of the ordinary differential equation (\ref{equ-2}) admits 
an expression via the Bessel function, 
\[ X(t)   =  \check{\theta} + \frac{ 2 (x_{0} - \check{\theta}) }{t} J_1(t),    \]
where $x_0=(x_{0,1}, x_{0, 2})^\prime$ is an initial value of $X(t)=(X_1(t), X_2(t))^\prime$, and $J_1(u)$ is the Bessel function of the first kind of order one, 
\[ J_1(u) = \sum_{j=0}^\infty \frac{ (-1)^j} { (2j)!! (2j+2)!!} u^{2j+2}, \]
with the following symptotic behaviors as $u \rightarrow 0$ and $u \rightarrow \infty$
\[ J_1(u) \sim \frac{u}{2} \mbox{ as } u \rightarrow 0, \mbox{ and } J_1(u) \sim \sqrt{ \frac{2}{ \pi u} } cos\left( u - \frac{3 \pi}{4} \right)  \mbox{ as } u \rightarrow \infty. \]
The ordinary differential equation (\ref{equ-3}) has the solution 
\[ X^n(t)   =  \bar{U}_n  + \frac{ 2 (x^n_{0} - \bar{U}_n) }{t} J_1(t),   \;\;  V^n(t) = \sqrt{n}  (\bar{U}_n - \check{\theta}) \left[ 1 - \frac{ 2 }{t} J_1(t) \right]
+ \sqrt{n} (x_0^n - x_0)  \frac{ 2 }{t} J_1(t) ,   \]
As in Theorem \ref{thm-1}, let $V(t) = \Pi(t) \bZ$, where it is relatively simple to use the properties of the Bessel function $J_1(u)$ to 
verify that $\Pi(t)=-[1 - 2 J_1(t)/t ] \mbox{diag}(\tau, \check{\theta}_2)$ is the matrix solution of the linear differential equation 
(\ref{limit-0}) in this case. Then for  $t \in [0,\infty)$, 
\[ V^n(t) = \left( \begin{array}{l}  \tau Z_1 \\ \check{\theta}_2 Z_2 \end{array} \right) \left[ 1 - \frac{ 2 }{t} J_1(t) \right]+ o_P(1)
     = V(t) + o_P(1). \]
The result matches the weak convergence of $V^n(t)$ to $V(t)$ shown in Theorem \ref{thm-1}, and as $t \rightarrow \infty$,  
$X(t) \rightarrow \check{\theta} = X(\infty)$,  $X^n(t) \rightarrow \hat{\theta}_n = 
\bar{U}_n = X^n(\infty)$, and $V^n(t) \rightarrow V^n(\infty) = \sqrt{n}  (\bar{U}_n - \check{\theta}) $; as $n \rightarrow \infty$, 
$V^n(\infty) \rightarrow V(\infty) =\Pi(\infty) \bZ = - (\tau Z_1, \check{\theta}_2 Z_2)^\prime$,
which indicates the asymptotic distribution of the estimator $\hat{\theta}_n = X^n(\infty)$. Again the behaviors of 
$X(t)$, $X^n(t)$,  $V^n(t)$, and $V(t)$ over $[0, \infty]$ describe the dynamic evolution of the accelerated gradient descent algorithm, 
such as how the sequences generated from the algorithm evolve along iterations (via $X(t)$ and $X^n(t)$ as functions of $t$), and converge to the target optimization solutions (via the convergence of $V^n(t)$ to $V^n(\infty)$ and $V(t)$ to $V(\infty)$
as $t \rightarrow \infty$), as well as connect to the asymptotic distributions of the target optimization solutions (via the convergence of $V^n(\infty)$ to $V(\infty)$ as $n \rightarrow \infty$). 
We find that the polynomial decay factor $\frac{2}{t} J_1(t)$ appears in all relationships for the accelerated case, and  the major difference for the two cases is exponential decay $1 - e^{-t}$ for the plain 
case vs polynomial decay $1 - \frac{2}{t} J_1(t)$ for the accelerated case.

%It describes how the generated sequences from the algorithms evolves along iterations  (via $X(t)$ and $X^n(t)$ as functions of $t$), and 
%converge to the target optimization solutions (estimators) (via the convergence of $V^n(t)$ to $V^n(\infty)$ and $V(t)$ to $V(\infty)$
%as $t \rightarrow \infty$), as well as connect to asymptotic distributions of the target solutions 
%(via the convergence of $V^n(\infty)$ to $V(\infty)$ as $n \rightarrow \infty$).

%\begin{remark} 
%The rapid growth in the size and scope of datasets in science and technology has created a need for novel foundational perspectives on data analysis that blend computer science and statistics. 
%Modern massive datasets create a fundamental problem at the intersection of the computational and statistical sciences
%Computational and statistical tradeoffs
%Modern statistics and machine learning 

\begin{remark}
Solving problems with large-scale data often require some tradeoffs between statistical efficiency and computational efficiency,
and thus we need to handle both statistical errors and computational errors. We illustrate the potential of the joint asymptotic analysis framework 
for the study of the two types of errors. 
%relationship between the derived asymptotic differential equations and the two types of errors as follows. 
%computation accuracy and statistical uncertainty, which can be illustrated as follows. 
%In terms of notations used in Remarks \ref{remark2} and \ref{remark3}, 
Note that
\[  x^n_\delta(t) - \check{\theta} = x^n_\delta(t) - \hat{\theta}_n + \hat{\theta}_n - \check{\theta}, \]
where $x^n_\delta(t)$ (or $x^n_k$) are the values computed by the gradient descent algorithms for solving the stochastic sample 
optimization problem 
(\ref{min-1}) based on sampled data, and $\check{\theta}$ is the exact solution of the deterministic population optimization problem (\ref{min-0}) corresponding to the true value of $\theta$, with $\hat{\theta}_n$ the exact solution of the optimization problem 
(\ref{min-1}) corresponding to the estimator of $\theta$. The total error $x^n_\delta(t) - \check{\theta}$ consists of computational error $x^n_\delta(t) - \hat{\theta}_n$ (of order $t^{-1}$
or $t^{-2}$) and statistical error $\hat{\theta}_n- \check{\theta}$ (of order usually $n^{-1/2}$). Since $X(t)$ approaches the solution $\check{\theta}$
of the optimization problem (\ref{min-0}), and  in fact numerically $\check{\theta}$ can be only evaluated by $X(t)$ and its 
corresponding algorithms, using $X(t)$ as a proxy of $\check{\theta}$ we may treat $x^n_\delta(t) - X(t)$ as a surrogate of the total error, and find its asymptotic distribution $V(t)$ via differential equation useful for the analysis of  the total error.
%surrogate total error $x^n_\delta(t) - X(t)$ (or $X^n(t) - X(t)$). 
\end{remark}

\section{Stochastic gradient descent via stochastic differential equations for stochastic optimization}

%With all training sample, 
Solving the stochastic sample optimization problem (\ref{min-1}) by the algorithms (\ref{min-GD2}) and (\ref{min-Nest2}) requires evaluating the sum-gradient for all data, 
that is, it requires expensive evaluations of the gradients $\nabla\! \ell(\theta; U_{i})$
from summand functions $\ell(\theta; U_{i})$ with all data $U_i$, $i=1, \cdots, n$.
For big data problems, data are enormous, and such evaluation of  the sums of 
gradients for all data becomes prohibitively  expensive. To overcome the computational burden, stochastic gradient descent uses 
a so-called mini-batch of data to evaluate % a 
the corresponding subset of summand functions at each iteration. Each mini-batch is a relatively small data set 
that is sampled from (i) the large training data set $\bU_n$ or (ii) the underlying population distribution $Q$.  
%samples 
%%a single example or 
%a minibatch of examples from the training set %(or bootstraps a subsample from the data) 
%and evaluates a corresponding subset of summand functions at each iteration. %every step. 
For the case of subsampling from the original data set $\bU_n$, it turns out that mini-batch subsampling 
in the stochastic gradient descent scheme is similar to the $m$ out of $n$ (with or without replacement) 
bootstraps for gradients (Bickel et al. (1997)). While bootstrap resampling %from the original data set %$\bU_n$ 
is widely used to draw inferences in statistics, resampling used here in learning community is motivated purely from the computational purpose. 
Specifically, %denote by $\bU_m^*=(U_1^*, \cdots, U_m^*)^\prime$ a minibatch. 
%the $m$ out $n$ bootstrap sample. %(with or without replacement). 
assume integer $m$ is much smaller than $n$, and denote by $\bU_m^*=(U_1^*, \cdots, U_m^*)^\prime$ a mini-batch. For the case (ii),
$\bU_m^*=(U_1^*, \cdots, U_m^*)^\prime$ is an i.i.d. sample taken from distribution $Q$. 
For case (i),  $\bU_m^*=(U_1^*, \cdots, U_m^*)^\prime$ is a subsample taken from $\bU=(U_1, \cdots, U_n)^\prime$, 
where %for the case of without replacement, 
$U_1^*, \cdots, U_m^*$ %form a subsample that 
are randomly drawn with or without replacement from $U_1, \cdots, U_n$. For the case of with replacement, 
$U_1^*, \cdots, U_m^*$ are an i.i.d. sample taken from $\hat{Q}_n$, and $\hat{Q}_n$ is the empirical distribution of 
$U_1, \cdots, U_n$. 
%In machine learning, $\bU_m^*$ is often referred to as a mini-batch, as the mini-batch size %(or the bootstrap sample size) 
%$m$ is much smaller than the size $n$ of the whole training sample. 
%Denote by $\hat{Q}_n$ the empirical distribution of $U_1, \cdots, U_n$, and 
In this paper we consider the case that mini-batches are sampled from the underlying distribution $Q$. Since mini-batch size $m$ is negligible in 
comparison with data size $n$, %and $\nabla\! \ell(\theta; U)$ has a moment generating function, 
the bootstrap sampling case (ii) can be handled via strong approximation (Cs\"{o}rg\"{o} and Mason (1989), Cs\"{o}rg\"{o} et al. (1999), Massart (1989), Rio (1993a, b)) by converting case (ii)  into the essentially proven scenario of case (i) where mini-bataches are sampled from the underlying distribution $Q$. 

The main computational idea in the stochastic gradient descent algorithm is to replace $\cL^n(\theta; \bU_n)$ in the algorithms 
(\ref{min-GD2}) and (\ref{min-Nest2}) by a smaller sample version 
$\hat{\cL}^m(\theta; \bU^*_m)$ at  each iteration, where 
\[ \hat{\cL}^m(\theta; \bU^*_m)=\frac{1}{m} \sum_{i=1}^m \ell(\theta; U^*_{i}). \]
%to be the bootstrap version of $L^n(\theta; \bU_n)$. 

\subsection{Stochastic gradient descent}
\label{section-3-1}
The stochastic gradient descent scheme replaces $\nabla\! \cL^n(x^n_{k-1}; \bU_n)$ in the algorithm (\ref{min-GD2}) by a smaller sample version at each iteration to obtain the following recursive algorithm, 
\begin{align} \label{min-GD3}
& x^m_k=x^m_{k-1}-\delta \nabla \!\hat{\cL}^m(x^m_{k-1}; \bU_{mk}^*), 
\end{align}
where $\bU_{mk}^* = (U_{1k}^*, \cdots, U_{mk}^*)^\prime$, $k=1, 2, \cdots$, are independent mini-batches. 
%independent given $\bU_n$, and for each $k$,  $U_{1k}^*, \cdots, U_{mk}^*$ form a $m$ out of $n$ bootstrap sample. 

%For the accelerated case, the stochastic gradient descent scheme replaces $\nabla\!L^n(\theta; \bU_n)$ in (\ref{min-Nest2}) by a bootstrap version at each iteration to obtain 
%\begin{align} \label{min-Nest3}
%& x^m_k=y^m_{k-1}-\delta \nabla \!\hat{L}^m(y^m_{k-1}; \bU_{mk}^*), \qquad  y^m_k=x^m_k+\frac{k-1}{k+2}(x^m_k-x^m_{k-1}),
%\end{align}
%where $\bU_{mk}^* = (U_{1k}^*, \cdots, U_{mk}^*)^\prime$, $k=1, 2, \cdots$, are independent given $\bU_n$, and for each $k$,
% $U_{1k}^*, \cdots, U_{mk}^*$ are a $m$ out of $n$ bootstrap. 

We may naively follow the continuous curve approach described  in Section \ref{section-1} to approximate $\{x^m_k, k=0, 1, \cdots\}$ 
by a smooth curve similar to the case in Section \ref{section-2-1}. However, unlike the scenario in Section \ref{section-2-1}, the algorithms (\ref{min-GD3}) [and (\ref{min-Nest3}) for the accelerated case in Section \ref{section-accelerate} later] are designed for the 
computational purpose, they do not correspond to any optimization problem with a well-defined objective function like $g(\theta)$ in 
the optimization problem (\ref{min-0}) or $\cL^n(\theta; \bU_n)$ in the optimization problem
 (\ref{min-1}), since samples $\bU^*_{mk}$ used in $\hat{\cL}^m(x^m_{k-1}; \bU_{mk}^*)$ change with iteration $k$. The analysis for stochastic gradient descent will be quite different from those studied in Section \ref{section-2-1}.  Below we consider the stochastic gradient descent case, and may define a `pseudo objective function'. 

Define  a mini-batch process $\bU^*_m(t) = (U_1^*(t), \cdots, U_m^*(t))^\prime$ and 
%$U_i^*(t)= U_{ik}^*$, 
%$\bU_m^*(t) = \bU_{km}^*$ for $(k-1) \delta < t \leq k \delta $,  and 
a step process $x^m_\delta(t)$, $t \geq 0$, for $x^m_k$ in (\ref{min-GD3}) as follows, 
\begin{equation} \label{GD-xt} 
\bU_m^*(t) = \bU_{mk}^* \mbox{ and } x^m_\delta(t) = x^m_k \mbox{ for } k \delta \leq t <(k+1) \delta. 
 \end{equation}
To facilitate the analysis we adopt a convention $x^m_\delta(t) = x^m_0$ for $t<0$. Then $\hat{\cL}^m(x^m_\delta(t-\delta); \bU_m^*(t))$ $ = \hat{\cL}^m(x^m_{k-1}; \bU_{mk}) $ for $k \delta \leq t < (k+1) \delta$. 
$\hat{\cL}^m(\theta; \bU_m^*(t))$ may be treated as a counterpart of $\cL^n(\theta; \bU_n)$.
 As $m \rightarrow \infty$, $\hat{\cL}^m(\theta; \bU_m^*(t))$ approaches $g(\theta)$ for each $\delta$, and the stochastic 
 gradient descent algorithm (\ref{min-GD3}) can still solve the optimization problem (\ref{min-1}) numerically. But as $t$ evolves, 
$\hat{\cL}^m(\theta; \bU_m^*(t))$ changes from iteration to iteration, and depends on $\delta$ as well as $m$, since mini-batches change as the algorithm iterates, and the number of the mini-batches involved is determined by the time $t$ and the step size $\delta$.
There is no single bona fide objective function here, and the `pseudo objective function' $\cL^m(\theta; \bU^*_m(t))$ can't serve the role of genuine objective functions like $g(\theta)$ and $\cL^n(\theta;\bU_n)$.  The approach in Sections \ref{section-1}
and \ref{section-2-1} can not be directly applied to obtain an ordinary differential equation like the equation (\ref{GD-c2}). In fact as we will see below, for this case there exists no such analog ordinary differential equation. Instead we will derive asymptotic stochastic differential equations for the algorithm (\ref{min-GD3}). The new asymptotic stochastic differential equations may be considered as a counterpart of the ordinary differential equation (\ref{GD-c2-1}), which is an asymptotic version of the ordinary differential equation 
(\ref{GD-c2}), but the key difference is that the asymptotic stochastic differential equations must depend on the step size $\delta$ as well 
as $m$ to account for the mini-batch effect (see more details later after the stochastic differential equations (\ref{GD-stoch1}) and (\ref{GD-stoch2}) regarding the associated random variability). 
Our derivation and stochastic differential equations rely on the asymptotic behavior of $\nabla \hat{\cL}^m(\theta; \bU_m^*(t)) - \nabla g(\theta)$ as  $\delta \rightarrow 0$ and $m\rightarrow \infty$.

We need the following initial %usual sample size 
condition to guarantee the validity of  our asymptotic analysis. %mini-batch subsampling.

\begin{enumerate}

\item[A5.] Assume initial values satisfy $x^m_0 - x_0 = o_P((\delta /m)^{1/2})$. 
%For the case that mini-batches are sampled from training data $\bU_n$, %mini-batch %(or bootstrap sample) %size $m$ satisfies that 
%as $n \rightarrow \infty$, we choose mini-batch size $m \rightarrow \infty$ and $m/n \rightarrow 0$, and add to Condition A1 further 
%requirement that %for the stochastic gradient case or moment generating functions for the accelerated case. 
%$E[h^4_1(U)] < \infty$, and 
%%$E [\nabla \ell(\theta;U)]^4  ]<\infty$ for some $\theta$.  %for each $\theta$, $[\nabla \ell(\theta;U)]^2$ 
%$\nabla \ell(\theta;U)$ has a moment generating function.

%\item[A5.] Assume $\delta$ to obey that as $m,n \rightarrow \infty$, $\delta = O( m^{-c})$ for certain positive constant $c$.

\end{enumerate}

We describe the asymptotic behavior of $\nabla \hat{\cL}^m(\theta; \bU_m^*(t) ) $ in the following theorem.
\begin{thm} \label{thm2}
%For $ t \in [0, T]$, 
Define a partial sum process
\begin{equation} \label{equ-H1}
  H^m_\delta(t) = % (m \delta/ T )^{1/2} 
(m \delta )^{1/2}  \sum_{t_k \leq t} \left[ \nabla \hat{\cL}^m(x^m_\delta(t_{k-1}); \bU_m^*(t_k)) - \nabla g(x^m_\delta(t_{k-1})) %\nabla g(X(t_k))    
\right], \; t \geq 0, 
\end{equation}
where $t_k = k \delta$, $k=0, 1, 2, \cdots $. %[T/\delta]$.
Under Conditions A1-A5,
as  $\delta \rightarrow 0$ and $m\rightarrow \infty$, we have that on $D([0, T])$, $H^m_\delta(t)$
weakly converges to $H(t) = \int_0^t \bsigma ((X(u)) d\bB(u)$,  $t \in [0, T]$, where $\bB$ is a $p$-dimensional standard Brownian motion, $\bsigma(\theta)$ is defined in Condition A3, and $X(t)$ is the solution of the ordinary differential equation (\ref{GD-c1}).
\end{thm}

\begin{remark} \label{remark-4-1}
As we have discussed early, due to mini-batches %or bootstrap samples 
used in the algorithm (\ref{min-GD3}), there is no corresponding optimization problem with a well-defined objective function. As a result, we do not have any $\delta$-free differential equation analog to the ordinary differential equation 
(\ref{GD-c2-1}). In other words, here there is no analog continuous modeling to derive differential equations free of $\delta$, obtained by letting $\delta \rightarrow 0$.
This may be explained from Theorem \ref{thm2} as follows. It is easy to see that $H^m_\delta(t)$ is a normalized partial sum process for 
$[T/\delta]$ random variables $\nabla \hat{\cL}^m(x^m_\delta(t_{k-1}); \bU_m^*(t_k))$ whose variances are of order $m^{-1}$, 
and the weak convergence theory for partial sum processes indicates that a normalized factor $(m \delta)^{1/2}$ in the definition (\ref{equ-H1}) is needed to obtain a weak convergence limit for $H^m_\delta(t)$. On the 
other hand, to obtain an analog to the $\bZ$ term in the equation (\ref{GD-c2-1}) we need to find  some kind of continuous-time limit for $\nabla \hat{\cL}^m(\theta; \bU^*_m(t) ) - \nabla g(\theta)$. As $\bU^*_m(t)$ is an empirical process for  %(conditionally) 
independent %bootstrap 
subsamples $\bU^*_{mk}$, $\nabla \cL^m(\theta; \bU^*_m(t)) - \nabla g(\theta)$ may behave like a sort of discrete-time weighted white noise (in fact a martingale difference sequence). Therefore, a possible continuous-time limit for $\nabla \cL^m(\theta; \bU^*_m(t) ) - \nabla g(\theta)$ is related to a continuous-time white noise, which is defined 
as the derivative $\dot{\bB}(t)$ of Brownian motion $\bB(t)$ in the sense of the Dirac delta function (a generalized function). In the notation of Theorem \ref{thm2}, we may informally write $H(t) =\int_0^t \bsigma(X(u)) \dot{\bB}(u) du$ in terms of white noise 
$\dot{\bB}(t)$, and $\nabla \hat{\cL}^m(x^m_\delta(t-\delta); \bU_m^*(t) ) - \nabla g(X(t))$ corresponds to the derivative 
$\dot{H}(t) = \bsigma(X(t)) \dot{\bB}(t)$ of $H(t)$. While the factor $\delta^{1/2}$ on the right hand side of the definition (\ref{equ-H1}) is needed to normalize a partial sum process with $[T/\delta]$ random variables for obtaining a weak convergence limit, from the white noise point of view, here we need a normalized factor $\delta^{1/2}$ to move from a discrete-time white noise to a continuous-time white noise. As a matter of fact, the weak convergence is very natural from the viewpoint of limit theorems for stochastic processes (Jacod and Shiryaev (2003), He et al. (1992)). 
Because of the white noise type stochastic variation due to different mini-batches %(or bootstrap samples) 
used from iteration to iteration in the algorithm (\ref{min-GD3}),  the continuous modeling for stochastic gradient descent will be $\delta$-dependent, which will be given below. 
\end{remark} 

Using the definitions of $x^m_\delta(t)$ in \eqref{GD-xt} and $H^m_\delta(t)$ in \eqref{equ-H1}, 
we recast algorithm (\ref{min-GD3}) as %indicate that with time increment $\delta t$ at $t$, 
%\[  \frac{ x^m_\delta(t + \delta ) - x^m_\delta(t) } {\delta } = - \nabla g(x^m_\delta(t)) - (\delta /m)^{1/2} \,\frac{H^m_\delta(t + \delta ) - H^m_\delta(t)}{\delta}.\]
\[ x^m_\delta(t + \delta ) - x^m_\delta(t) = - \nabla g(x^m_\delta(t))  \delta - (\delta /m)^{1/2} \,[ H^m_\delta(t + \delta ) - H^m_\delta(t)].\]
Theorem \ref{thm2} suggests an approximation of the step process $H^m_\delta(t)$ by the continuous process $H(t)$, and 
we approximate the step process $x^m_\delta(t)$ by a continuous process $X^m_\delta(t)$. 
Taking the step size $\delta$ as $dt$,  $H^m_\delta(t + \delta ) - H^m_\delta(t)$ as $d H(t) = \bsigma(X(t)) d\bB(t)$, and 
$x^m_\delta(t + \delta ) - x^m_\delta(t)$ as $d X^m_\delta(t)$, we tranform the above difference equation into 
%and above difference equation becomes 
%\[  \frac{ d X^m_\delta(t)}{ dt } = - \nabla g(X^m_\delta(t)) - (\delta /m)^{1/2}\, \frac{d H(t) } {dt }, \]
%which leads to 
the following stochastic differential equation 
\begin{equation} \label{GD-stoch1}
 dX^m_\delta(t) = - \nabla g(X^m_\delta(t)) dt - (\delta /m)^{1/2}  \bsigma (X(t))  d\bB(t),  %(\delta T/m)^{1/2}
  \end{equation}
 where  $X(t)$ is the solution of the ordinary differential equation (\ref{GD-c1}), and $\bB(t)$ is a $p$-dimensional standard Brownian motion.  The solution 
 $X^m_\delta(t)$ of the stochastic differential equation (\ref{GD-stoch1}) may be considered as a continuous approximation of  
 $x^m_k$ [or $x^m_\delta(t)$] generated from the stochastic gradient descent algorithm (\ref{min-GD3}) [or (\ref{GD-xt})]. 
 Since $X^m_\delta(t)$ is expected to be close to $X(t)$, and the Brownian term in (\ref{GD-stoch1}) is of higher order, 
 we may replace $X(t)$ in (\ref{GD-stoch1}) by $X^m_\delta(t)$ to better mimic the recursive relationship in (\ref{min-GD3}). That is, 
 we also consider the following stochastic differential equation 
\begin{equation} \label{GD-stoch2}
  d\check{X}^m_\delta(t) = - \nabla g(\check{X}^m_\delta (t)) dt - ( \delta /m)^{1/2} \bsigma(\check{X}^m_\delta(t)) d\bB(t). 
\end{equation}
As our interest is on their distributional behaviors, we consider solutions of the stochastic differential equations 
(\ref{GD-stoch1}) and  (\ref{GD-stoch2}) in the weak sense 
that for each fixed $\delta$ and $m$, there exist versions of the continuous process $X^m_\delta(t)$ (or $\check{X}^m_\delta(t)$)  and Brownian motion $\bB(t)$ on some probability space
to satisfy the equation (\ref{GD-stoch1}) (or (\ref{GD-stoch2})) (see Ikeda and Watanabe (1981)).

%Some versions of stochastic differential equations (such as vague or approximate matrices for the diffusion variance) % like (\ref{GD-stoch1}) 
%are informally used in the deep learning and stochastic gradient descent literature based on some heuristic or loose reasoning without rigorous justification 

%There is a large literature on stochastic differential equations for recursive algorithms in particular stochastic gradient descent in deep learning (see Chen et al. (2016), Fan et al. (2018), Kushner and Yin (2003), Li et al. (2016), Li et al. (2017a),  Li et al. (2017b), Mandt et al. (2016), Sirignano and Spiliopoulos (2017)). 

%As we will see, this is the first paper to provide explicit stochastic differential equations and establish rigorous weak convergence for stochastic gradient descent algorithms. 

The stochastic Brownian terms in (\ref{GD-stoch1}) and (\ref{GD-stoch2}) are employed to account for the random fluctuations due to 
the use of min-batches %(or bootstrap samples) 
for gradient estimation from iteration to iteration in the stochastic gradient descent algorithm (\ref{min-GD3}),
where $m^{-1/2}$ and $\delta^{1/2}$ are statistical normalization factors with $m$ for mini-batch size and $[T/\delta]$ for the total number of iterations considered in $[0,T]$ (as $\delta$ for the step size). At each iteration we resort to a mini-batch 
 for gradient estimation, so the number of iterations in $[0,T]$ is equal to the number of mini-batches %(or bootstrap samples) 
 used in $[0, T]$, and the factor $\delta^{1/2}$ accounts for the effect due to the total number of mini-batches used in $[0, T]$, 
while $m^{-1/2}$ accounts for the effect of $m$ observations in each mini-batch. % (bootstrap sample). 

%Although the models are related to $m$ as in the empirical, their dependence on
%$\delta$ is due to the number of iterations (or steps) that we do bootstrap ??  ( \delta T /m)^{-1/2}
The theorem below derives the asymptotic distribution of $X^m_\delta(t)$ and $\check{X}^m_\delta(t)$. Let $V^m_\delta(t) = ( 
m/ \delta)^{1/2} [ X^m_\delta (t) - X(t)]$ and $\check{V}^m_\delta(t) = ( m/\delta )^{1/2} [ \check{X}^m_\delta (t) - X(t)]$. Treating them as 
random elements in $C([0, T])$, we derive their weak convergence limit in the following theorem. 

\begin{thm} \label{thm3}
Under Conditions A1-A5, %and the same initial value for $X^m_\delta(t)$ and $\check{X}^m_\delta(t)$, 
as  $\delta \rightarrow 0$ and $m\rightarrow \infty$, we have 
\begin{equation} \label{equ-XX}
  \sup_{0 \leq t \leq T} |X^m_\delta(t) - \check{X}^m_\delta(t)| = O_P(m^{-1} \delta), 
\end{equation}
and both $V^m_\delta(t) $ %= [ m/(\delta T) ]^{1/2} [ X^m_\delta (t) - X(t)]$ 
and $\check{V}^m_\delta(t)$, % = [ m/(\delta T) ]^{1/2} [ \check{X}^m_\delta (t) - X(t)]$
 $t \in [0, T]$,  weakly converge to $V(t)$ which is a time-dependent Ornstein-Uhlenbeck process satisfying 
\begin{equation} \label{GD-v0}
 dV(t)  = - [\boldsymbol{I\!\! H}\! g(X(t))] V(t) dt - \bsigma (X(t))  d\bB(t), \;\; V(0)=0, 
 \end{equation} 
where $\boldsymbol{I\!\! H}$ is the Hessian operator, 
$\bB$ is a $p$-dimensional standard Brownian motion, $\bsigma(\theta)$ is defined in Condition A3, and $X(t)$ is the solution of the ordinary differential equation (\ref{GD-c1}).
\end{thm}

\begin{remark}
As  $X(t)$ and $X^n(t)$ in the gradient descent case are viewed as the population and sample gradient flows in Remark \ref{remark0} respectively, we may treat $X^m_\delta(t)$ and $\check{X}^m_\delta(t)$ as stochastic gradient flows in the stochastic gradient descent case, and regard  
the Gaussian limiting distribution of $V^m_\delta(t)$ and $\check{V}^m_\delta(t)$ as the central limit theorem for the stochastic gradient flows, 
which simply refers to the gradient flow central limit theorem (GF-CLT). 
\end{remark}

\begin{remark}
Theorem \ref{thm3} shows that while $X^m_\delta(t)$ and $\check{X}^m_\delta(t)$ have the same weak convergence limit, they are an order of  magnitude closer to each other than to $X(t)$. This may also be seen from the fact that the difference between the stochastic differential equations 
(\ref{GD-stoch1}) and (\ref{GD-stoch2}) is at the high order Brownian term with $X^m_\delta(t)$ replaced by its limit $X(t)$. 
 The linear stochastic differential equation (\ref{GD-v0}) indicates that $V(t)$ 
%the limiting process $V(t)$ is a time-dependent Ornstein-Uhlenbeck process, which 
has the following explicit expression for $t \in [0, T]$ under the 
condition that $\boldsymbol{I\!\! H}\! g(X(u))$ and $\boldsymbol{I\!\! H}\! g(X(v))$ commute for all $u \neq v$, 
\begin{equation} \label{GD-v1}
%V(t) = - \exp\left [- \int_0^t \boldsymbol{I\!\! H}\! g(X(v)) dv \right] \int_0^t \exp \left[  \int_0^u \boldsymbol{I\!\! H}\! g(X(v))  dv \right] \bsigma(X(u)) d\bB(u). 
V(t) = -  \int_0^t \exp \left[  - \int_u^t \boldsymbol{I\!\! H}\! g(X(v))  dv \right] \bsigma(X(u)) d\bB(u). 
\end{equation}
\end{remark}

%What happen for $V(t)$ as $t \rightarrow \infty$, not well defined, as it is defined only on $[0, T]$.
%$t \rightarrow \infty$ or $k \rightarrow \infty$, points to $x^m_k$ and $x^m_\delta(t)$ approach to the true optimizer, 

The step process $x^m_\delta(t)$ defined in (\ref{GD-xt}) is the empirical process for $x^m_k$ generated from the stochastic gradient descent 
algorithm (\ref{min-GD3}). Treating $x^m_\delta(t)$ as a random element in $D([0,T])$ we consider its asymptotic distribution in the follow theorem.

\begin{thm}  \label{thm4}
Under assumption A1-A5, as $\delta \rightarrow 0$ and $m \rightarrow \infty$, we have  
%\[    \max_{k \leq T/\delta}  |x^m_k - \check{X}^m_\delta(k \delta) | = o_P( m^{-1/2}  \delta^{1/2}) + O_P( n^{-1/2}  + \delta + \delta m^{-1/2} |\log \delta|^{1/2}), \]
\[   \sup_{t \leq T} |x^m_\delta(t) - X^m_\delta(t)| = o_P( m^{-1/2}  \delta^{1/2}) + O_P(  \delta |\log \delta|^{1/2}),\]
where $x^m_\delta(t)$ and $X^m_\delta(t)$ are defined by the algorithm (\ref{GD-xt}) and the stochastic differential equation  (\ref{GD-stoch1}), respectively. 
%\[ % \max_{0 \leq t \leq T} |x^m_\delta(t) - X^m_\delta(t)| = , \max_{0 \leq t \leq T} |X^m_\delta(t) - \check{X}^m_\delta(t)| = O_P(m^{-1} \delta), \]
In particular if we choose $(\delta, m)$ such that %$m /(n \delta) \rightarrow 0$, 
$m \delta |\log \delta| \rightarrow 0$ as $\delta \rightarrow 0$ and $m \rightarrow \infty$, 
then for the chosen $(\delta, m)$, 
$ (m/\delta )^{1/2} [ x^m_\delta (t) - X(t)] $ weakly converges to $V(t)$, where $V(t)$ is governed by the stochastic differential equation (\ref{GD-v0}).
\end{thm}

\begin{remark} Theorem \ref{thm4} indicates that sequences $x^m_k$ generated from the stochastic gradient descent algorithm 
(\ref{min-GD3}) can be very close to the continuous curves $X^m_\delta(t)$ and $\check{X}^m_\delta (t)$
governed by the stochastic differential equations (\ref{GD-stoch1}) and (\ref{GD-stoch2}), respectively, and with proper 
choices of $(\delta, m)$ we can make the empirical process $x^m_\delta(t)$ for $x^m_k$ to share the same weak convergence limit as 
the continuous curves $X^m_\delta(t)$ and $\check{X}^m_\delta (t)$. The results allow us to study discrete algorithms by analyzing their 
corresponding continuous stochastic differential equations and their relatively simple weak limit. 
\end{remark}

\begin{remark} We may consider stochastic gradient descent with momentum and/or diminishing learn rate and obtain the corresponding stochastic differential equations. 
For example, $\delta$ in (\ref{min-GD3}) can be replaced by 
diminishing learning rate $\delta_k = \eta k^{-\alpha}$ for some $\alpha \in (0, 1)$ and constant $\eta >0$, and the same arguments lead us to stochastic differential equations like (\ref{GD-stoch1}) and  (\ref{GD-stoch2}) with extra factor $(t+1)^{-\alpha}$. %in the Brownian terms: 
%\[  dX^m_\delta(t) = - \nabla g(X^m_\delta(t)) (t+1)^{-\alpha} dt - (\eta /m)^{1/2}  \bsigma (X(t)) (t+1)^{-\alpha} d\bB(t),  \]
%\[   d\check{X}^m_\delta(t) = - \nabla g(\check{X}^m_\eta (t)) (t+1)^{-\alpha} dt - ( \delta /m)^{1/2} \bsigma(\check{X}^m_\delta(t)) (t+1)^{-\alpha}  d\bB(t). \]
For the momentum case, we need to add an extra linear term in the drifts of $X^m_\delta(t)$ (or $\check{X}^m_\delta(t)$).  For example, 
we consider stochastic gradient descent with momentum, 
\[ x^m_k=\gamma x^m_{k-1}-\delta \nabla \!\hat{\cL}^m(x^m_{k-1}; \bU_{mk}^*), \qquad \delta = \eta k^{-\alpha}, \;\; \gamma = 1 - \beta \eta, 
 \]
and obtain the following stochastic differential equation, %for $X^m_\delta(t)$, 
\[  dX^m_\delta(t) = -[  \nabla g(X^m_\delta(t)) (t+1)^{-\alpha}  + \beta X^m_\delta(t) ] dt - (\eta /m)^{1/2}  \bsigma (X(t)) (t+1)^{-\alpha} d\bB(t).   \]

%For example, if we take $m  \sim \delta^{-\alpha_1}$, and $n \sim m^{\alpha_2}$, the requirement is $0<\alpha_1<2$ %$\alpha_2>1$ and $\alpha_1 (\alpha_2-2)>1$.
\end{remark}

\subsection{Accelerated stochastic gradient descent} 
\label{section-accelerate}

We apply Nesterov's acceleration scheme to stochastic gradient descent by replacing $\nabla\! \cL^n(y^n_{k-1}; \bU_n)$ in the algorithm (\ref{min-Nest2}) with a subsampled version at each iteration 
as follows,  
\begin{align} \label{min-Nest3}
& x^m_k=y^m_{k-1}-\delta \nabla \!\hat{\cL}^m(y^m_{k-1}; \bU_{mk}^* ), \qquad  y^m_k=x^m_k+\frac{k-1}{k+2}(x^m_k-x^m_{k-1}),
\end{align}
where we use initial values $x^m_0$ and $y^m_0=x^m_0$, and 
$\bU_{mk}^* = (U_{1k}^*, \cdots, U_{mk}^*)^\prime$, $k=1, 2, \cdots$, are independent mini-batches. 
%independent given $\bU_n$, and for each $k$,  $U_{1k}^*, \cdots, U_{mk}^*$ form a $m$ out of $n$ bootstrap sample. 
 
The continuous modeling for the algorithm (\ref{min-Nest3}) is conceptually in parallel with the case for the stochastic gradient descent 
algorithm (\ref{min-GD3}) in Section \ref{section-3-1}, but the tricky part is on the technical side that we face many mathematical 
difficulties in multiple steps related to singularity in the second order stochastic differential equations involved.

As we have illustrated the continuous modeling of $x^m_k$ generated from the algorithm (\ref{min-GD3}) in Section \ref{section-3-1},  it is easy to see that our derivation of  stochastic differential equations relies on the asymptotic behavior of $\nabla \hat{\cL}^m(\theta; \bU_m^*(t) ) - \nabla g(\theta)$ as  $\delta \rightarrow 0$ and $m\rightarrow \infty$.
Similar to the cases in Sections \ref{section-1} and  \ref{section-2-3}, we define step processes 
\begin{equation} \label{Nest-xyt}
x^m_\delta(t) = x^m_k, \;\; y^m_\delta(t)=y^m_k, \;\; \bU_m^*(t) = \bU^*_{mk}, 
\;\mbox{  for } \; k \sqrt{\delta} \leq t < (k+1) \sqrt{\delta} ,
\end{equation} 
and  approximate $x^m_\delta(t)$
%$\{x^n_k, k=0, 1, \cdots\}$
by a smooth curve $X^m_\delta(t)$ given by (\ref{Nest-stoch1}) below.  Note the difference between the step sizes 
$\delta$ and $\delta^{1/2}$ for the plain and accelerated cases, respectively, as pointed out at the end of Section \ref{section-1}.

\begin{thm} \label{thm5} 
Define a partial sum process 
\begin{equation} \label{equ-H2} 
   H^m_\delta(t) = (m^2 \delta) ^{1/4}  \sum_{t_k \leq t} \left[ \nabla \hat{L}^m(y^m_\delta(t_{k-1}); \bU_m^*(t_k)) - 
     \nabla g(y^m_\delta(t_{k-1}) ) \right], \; t \geq 0, 
 \end{equation}
 where $t_k=k \delta^{1/2}$, $k=0, 1, 2, \cdots$. %[ T/\delta^{1/2} ]$.  
 Under Conditions A1-A5,
as  $\delta \rightarrow 0$ and $m\rightarrow \infty$, we have that on $D([0, T])$, $H^m_\delta(t)$
weakly converges to $H(t) = \int_0^t \bsigma ((X(u)) d\bB(u)$, $t \in [0, T]$, where $\bB$ is a $p$-dimensional standard Brownian motion, $\bsigma(\theta)$ is defined in Condition A3, and $X(t)$ is the solution of the ordinary differential equation (\ref{equ-2}). %(\ref{GD-c1}).
\end{thm}

%Theorem ?? is similar to Theorem \ref{thm2}, but the step size and normalization factor are related to $\delta^{1/2}$ instead of $\delta$. 
%This is due to the accelerated gradient descent algorithm ??

%It is not obvious that we can directly adopt the simple arguments %between (\ref{equ-1}) and (\ref{equ-2}) 
%in Sections \ref{section-1} and \ref{section-3-1} 
Now we are ready to derive the second order stochastic differential equation corresponding to the algorithm (\ref{min-Nest3}). 
%We illustrate an alternative approach instead. 
First, note that the second order ordinary differential 
equation (\ref{equ-2}) can be equivalently written as  
\begin{equation} \label{Nest-2-1}
 \left\{ \begin{array}{l}
       dX(t) = Z(t) dt, \\
       dZ(t) = -\left[ \frac{3}{t} Z(t) + \nabla g(X(t))\right] dt, \end{array} \right. 
\end{equation}
where $Z(t) = \dot{X}(t)$; 
and the algorithm  \eqref{equ-Nest1} %(\ref{equ-1}) 
is equivalent to 
\begin{equation} \label{Nest-1-1}
 \left\{ \begin{array}{l}
    x_{k+1} =  x_k  + \sqrt{\delta} \,z_k, \\     
    z_{k+1}  = \left[ 1 - \frac{3}{k+3} \right]  z_k - \sqrt{\delta} \,\nabla g\left( x_k + \frac{2k+3}{k+3} \sqrt{\delta} \,z_k \right),
     \end{array} \right. 
\end{equation}
where $z_k = (x_{k+1} - x_k)/\sqrt{\delta}$, which can be recasted as 
\begin{equation} \label{Nest-1-2}
 \left\{ \begin{array}{l}
    \frac{ x_{k+1} - x_k }{ \sqrt{\delta}}  = z_k, \\
    \frac{ z_{k+1}  -z_k }{\sqrt{\delta} }  = - \frac{3}{t_k + 3 \sqrt{\delta}  }  z_k - \nabla g\left( x_k + \frac{2k+3}{k+3} \sqrt{\delta} \,z_k \right),
     \end{array} \right. 
\end{equation}
where we take $t_k = k \sqrt{\delta}$.
We approximate $(x_k, z_k)$ by continuous curves $(X(t), Z(t))$. Noting that as $\delta \rightarrow 0$, 
$3 \sqrt{\delta} \rightarrow 0$  and $\frac{2k+3}{k+3} \sqrt{\delta} \,z_k \rightarrow 0$ in (\ref{Nest-1-2}), which are negligible relative to $t_k$ and $x_k$.
 We take step size $\sqrt{\delta}$ as $dt$ and turn the discrete difference 
equation system (\ref{Nest-1-2}) into the continuous differential equation system (\ref{Nest-2-1}). 

Second, we replace $(x_k, z_k)$ in (\ref{Nest-1-1}) by $(x^m_k, z^m_k)$, where $ z^m_k = (x^m_{k+1} - x^m_k)/\sqrt{\delta}$,  and write the algorithm (\ref{min-Nest3}) in the following equivalent forms 
 \begin{equation} \label{min-Nest3-1}
 \left\{ \begin{array}{l}
    x^m_{k+1} =  x^m_k  + \sqrt{\delta} \,z^m_k, \\     
    z^m_{k+1}  = \left[ 1 - \frac{3}{k+3} \right]  z^m_k - \sqrt{\delta} \,\nabla g\left( x^m_k + \frac{2k+3}{k+3} \sqrt{\delta} \,z^m_k \right) 
      - \frac{ \delta^{1/4}  } { \sqrt{m} } [H^m_\delta(t_{k+1}) - H^m_\delta(t_k) ], 
      \end{array} \right. 
\end{equation}
or equivalently, 
\begin{equation} \label{min-Nest3-2}
 \left\{ \begin{array}{l}
    \frac{ x^m_{k+1} - x^m_k }{ \sqrt{\delta}}  = z^m_k, \\
    \frac{ z^m_{k+1}  -z^m_k }{\sqrt{\delta} }  = - \frac{3}{t_k + 3 \sqrt{\delta}  }  z^m_k - \nabla g\left( x^m_k + \frac{2k+3}{k+3} \sqrt{\delta} \,z^m_k \right) - \frac{ \delta^{1/4}  } { \sqrt{m} } \frac{H^m_\delta(t_{k+1}) - H^m_\delta(t_k) }{\sqrt{\delta} },      \end{array} \right. 
\end{equation}
where again $t_k = k \sqrt{\delta}$. 
Third, we approximate $(x^m_k, z^m_k)$ by some continuous process $(X^m_\delta(t), Z^m_\delta(t))$. As Theorem \ref{thm5} suggests to 
substitute $H^m_\delta(t)$  by $H(t)$, with $dH(t) = \bsigma(X(t)) d\bB(t)$, dropping the negligible terms $3 \sqrt{\delta} $ and $\frac{2k+3}{k+3} \sqrt{\delta} \,z_k^m$, and taking the step size $\sqrt{\delta}$ as $dt$ we move from discrete difference equations (\ref{min-Nest3-2}) to the following stochastic differential equation system,
\begin{equation} \label{Nest-stoch1-1}
 \left\{ \begin{array}{l}
       dX^m_\delta(t) = Z^m_\delta(t) dt, \\
       dZ^m_\delta(t) = -\left[ \frac{3}{t} Z^m_\delta(t) + \nabla g(X^m_\delta(t))\right] dt - \frac{ \delta^{1/4}  } { \sqrt{m} } \bsigma(X(t)) d\bB(t), 
       \end{array} \right. 
\end{equation}
which together with $\dot{X}^m_\delta(t) = Z^m_\delta(t)$ is equivalent to the following second order stochastic differential equation,
\begin{equation}\label{Nest-stoch1}
  \ddot{X}^m_\delta(t)+\frac{3}{t}\dot{X}^m_\delta(t) +  \nabla \!g(X^m_\delta(t)) + (\delta /m^2)^{1/4} \bsigma(X(t)) \dot{\bB}(t)  =0,
\end{equation}
where initial conditions $X^m_\delta(0)=x^m_0$ and $\dot{X}^m_\delta(0)=0$, $X(t)$ is defined by the ordinary differential equation 
(\ref{equ-2}), $\bB(t)$ is a $p$-dimensional Brownian motion, and $\dot{\bB}(t)$ is a white noise defined as the time derivative of $\bB(t)$ in the sense of generalized functions (Hida and Si (2008)). 
%The main reason is that the {B?(t),t ? R} forms a system of i.i.d. (independent identically distributed) random variables. The i.i.d. property of the variables makes the analysis simpler and efficient.

As we have discussed and demonstrated for the stochastic gradient descent case in Section \ref{section-3-1}, 
similar to the stochastic differential equations (\ref{GD-stoch1}) and (\ref{GD-stoch2}) for the %plain 
stochastic gradient descent algorithm, 
the second order stochastic differential equations (\ref{Nest-stoch1-1}) and (\ref{Nest-stoch1}) depend on $\delta$ and $m$ through 
the stochastic Brownian terms. %in (\ref{Nest-stoch1-1}) and (\ref{Nest-stoch1}) 
They are used to account for the random fluctuation due to the use of min-batches for gradient estimation from iteration to iteration in %the accelerated stochastic gradient descent 
the algorithm (\ref{min-Nest3}),
where $m^{-1/2}$  and $\delta^{1/4}$ are statistical normalization factors with $m$ for the mini-batch size and $[T /\delta^{1/2}]$ for the total number of iterations considered in $[0, T ]$ (as $\delta^{1/2}$  for the step size), or equivalently, the total number 
of mini-batches used in $[0, T ]$. 

The theorem below will show that the second order stochastic differential equation (\ref{Nest-stoch1}) has a unique solution. Here again 
we consider the solution in the weak sense that for each fixed $\delta$ and $m$, there exist continuous process $X^m_\delta(t)$ and Brownian motion $\bB(t)$ on some probability space to satisfy the equation (\ref{Nest-stoch1}).
As in Section \ref{section-3-1}, the process $X^m_\delta(t)$ provides a continuous approximation of $x^m_k$ given by the algorithm (\ref{min-Nest3}). As $\delta \rightarrow 0$ and $m \rightarrow \infty$, the Brownian term in the equation (\ref{Nest-stoch1}) disappears, and $X^m_\delta (t)$ approaches $X(t)$ defined by  the ordinary differential equation (\ref{equ-2}). 
%For solutions $X(t)$ and $X^m_\delta(t)$ of and (\ref{Nest-stoch1}), respectively, we 
Define $V^m_\delta(t) = (m^2/ \delta) ^{1/4} [ X^m_\delta(t) - X(t) ]$.
Then $X(t)$, $X^m_\delta(t)$ and $V^m_\delta(t)$ live on $C([0,T])$.  %$C(I\!\!R_+)$. 
Treating them as random elements in $C([0,T])$, in the following theorem 
we derive a weak convergence limit of $V^m_\delta(t)$. 
%we show in the following theorem that tof $V^n(t)$ has the weak convergence limit $V(t)$ governed by the following linear differential equation,

\begin{thm} \label{thm6}
Under conditions A1-A5, the second order stochastic differential equation (\ref{Nest-stoch1}) has a unique solution in the weak sense,  
and as $\delta \rightarrow 0$, $m \rightarrow \infty$, $V^m_\delta(t)$ weakly converges to a Gaussian process $V(t)$ on $C([0,T])$, where $V(t)$ is the unique solution of the following linear second order stochastic differential equation,
\begin{equation} \label{Nest-limit-0}
\ddot{V}(t) + \frac{3}{t}  \dot{V}(t) + [\boldsymbol{I\!\! H}\! g(X(t))] V(t) + \bsigma(X(t)) \dot{\bB}(t) = 0, 
\end{equation}
where $\boldsymbol{I\!\! H}\!$ is the Hessian %= \nabla^2$ is Laplacian 
operator, $X(t)$ is the solution of the ordinary differential equation (\ref{equ-2}), $\bsigma(\theta)$ is defined in Condition A3, 
$\bB(t)$ is a $p$-dimensional standard Brownian motion, and initial conditions $V(0) = \dot{V}(0)=0$.
%Then as $n \rightarrow \infty$,  $V^n(t)$ weakly converges to $V(t)$, where $V(t)$ satisfies the linear differential equation (\ref{limit-0}).
\end{thm}

\begin{remark}
As  $X(t)$ and $X^n(t)$ in the accelerated case are viewed as the population and sample Lagrangian flows in Remark \ref{remark0} respectively, we may treat $X^m_\delta(t)$ as a stochastic Lagrangian flow in the accelerated stochastic gradient descent case, and regard 
%Similarly we can show that the solution of the second-order stochastic differential equation \eqref{Nest-limit-0} linearly depends on $\bB(t)$, and thus the limiting distribution of $V^m_\delta(t)$ is Gaussian. 
 the Gaussian limiting distribution of $V^m_\delta(t)$ as the central limit theorem for the 
stochastic Lagrangian flows, which we simply call the Lagrangian flow central limit theorem (LF-CLT). 
\end{remark}

%Step process $x^m_\delta(t)$ in (\ref{Nest-xyt}) is the empirical process for $x^m_k$ generated from algorithm (\ref{min-Nest3}). 
%Treating $x^m_\delta(t)$ as a random element in $D([0,T])$ we %conjecture 
%consider its asymptotic distribution in the follow theorem and show that $x^m_\delta(t)$ and $X^m_\delta(t)$ share the same weak convergence limit $V(t)$.
The step process $x^m_\delta(t)$ in the definition (\ref{Nest-xyt}) is the empirical process for $x^m_k$ generated from the 
algorithm (\ref{min-Nest3}). Treating $x^m_\delta(t)$ as a random element in $D([0,T])$ we consider its asymptotic distribution in the follow theorem.

\begin{thm}  \label{thm7} (Due to Shang Wu)
Assume that there exists $a \in (0, 1/2)$ such that %as $\delta \rightarrow 0$ and $m \rightarrow \infty$, 
$\delta m^{2/(1-2a)}$ is bounded below from zero. Then under assumption A1-A5, as $\delta \rightarrow 0$ and $m \rightarrow \infty$, we have  
%and there exists $p, C$, such that $\delta\geq Cm^{-p}$, then as $\delta\rightarrow 0$, $m\rightarrow \infty$, 
$$\sup_{t\leq T}|x_\delta^m(t)-X_\delta^m(t)|=o_p(m^{-1/2}\delta^{1/4})+O_p( \delta^{1/2}|\log\delta|),$$
%\[   \max_{t \leq T} |x^m_\delta(t) - X^m_\delta(t)| = o_P( m^{-1/2}  \delta^{1/4}) + O_P( m^{-1/2} \delta^{1/4+a/2}  |\log \delta | + \delta^{1/2} |\log \delta|^{1/2}),\]
where $x^m_\delta(t)$ and $X^m_\delta(t)$ are given by the definition (\ref{Nest-xyt}) and the stochastic differential equation (\ref{Nest-stoch1}), respectively. 
In particular if we choose $(\delta, m)$ to further satisfy that $m^{1/2} \delta^{1/4} |\log \delta| \rightarrow 0$ 
as $\delta \rightarrow 0$ and $m \rightarrow \infty$,  then for the chosen $(\delta, m)$, %on $D([0,T])$, 
$ (m^2/ \delta)^{1/4} [ x^m_\delta (t) - X(t)] $ weakly converges to $V(t)$, $t \in [0, T]$, where $V(t)$ is governed by the stochastic differential equation (\ref{Nest-limit-0}).
\end{thm}

\begin{remark} 
As we mentioned before, similar to the stochastic gradient descent case, the continuous modeling depends on both $\delta$ and $m$, 
and Theorems \ref{thm5}-\ref{thm7} are in parallel with Theorems \ref{thm2}-\ref{thm4}. 
However, for the accelerated case, the challenges are largely on the technical proofs.  For example, we need to handle 
second order stochastic differential equations like (\ref{Nest-stoch1}) with singularity (similar to the singularity case for the ordinary differential equations  (\ref{equ-2}) and (\ref{limit-0})); 
%the lack of adequate theory and technical tools for handling well-behaved second order stochastic  differential equations, let alone the singularity difficulty; 
it is hard to analyze the complex recursive relationship in the accelerated stochastic gradient descent algorithm (\ref{min-Nest3}). 
%There are issues for future study. For example, as $t$ and $k$ increase, $x^m_k$, $x^m_\delta(t)$ and $X^m_\delta(t)$ approach to the true solution of optimization (\ref{min-1}), with $X(t)$ to that of optimization (\ref{min-0}). We may consider the difference among them and study their asymptotic behaviors as $t \rightarrow \infty$.  
%Fixed $\delta$ and $k$ steps, one moves $k\delta$ and another $k \sqrt{\delta}$, and algorithm speed $1/k$ and $1/k^2$.
Theorems \ref{thm6}-\ref{thm7} show that sequences $x^m_k$ generated from the accelerated stochastic gradient descent algorithm 
(\ref{min-Nest3}) may be very close to the continuous curve $X^m_\delta(t)$ governed by the stochastic differential equation 
(\ref{Nest-stoch1}), and proper 
choices of $(\delta, m)$ enable the empirical process $x^m_\delta(t)$ for $x^m_k$ to have the same weak convergence limit as 
the continuous curve $X^m_\delta(t)$.  
\end{remark}

\begin{remark} 
The two conditions on $(\delta, m)$ are compatible. The bound condition indicates that for some generic constant $C$, $\delta^{a/2-1/4}m^{-1/2}<C$ or $\delta>C m^{-2/(1-2a)}$, and the condition $m^{1/2}\delta^{1/4}|\log\delta|\rightarrow 0$ requires that $\delta$ should decrease faster than $m^{-2}$.  For example, if we take $\delta=m^{-b}$ for any $b>2$, then $\delta^{a/2-1/4}m^{-1/2}\leq 1$  
holds for $1/2>a>1/2-1/b$, and $m^{1/2}\delta^{1/4}|\log\delta|=b  m^{1/2-p/4} \log m\rightarrow 0$. 

 %$m \delta^{1/2} |\log \delta|^2 \rightarrow 0$ and $m \delta^{1/4} /|\log \delta|^2 \rightarrow \infty$ are satisfied for any $2<b<4$
\end{remark}

%Consider Example 1 in this case give expression for $X^m_\delta(t)$ Joint asymptotic analysis??? Diverge as $t \rightarrow \infty$ since Brownian motion diverges like $\sqrt{t}$. 
Below we study the example considered in Section \ref{section-2-5} under the stochastic gradient descent case. 

\textbf{Example 1}(continue). We have already calculated $\nabla \! g(\theta) =\theta - \check{\theta}$, $\boldsymbol{I\!\! H}\! g(\theta) =I$, $\bsigma(\theta) = \mbox{diag}(\tau, \check{\theta}_2)$, 
$X(t) = \check{\theta} + (x_0 - \check{\theta}) e^{-t}$. %$= \check{\theta} (1 - e^{-t}) + x_0 e^{-t}$,
For the stochastic gradient descent case, solving the stochastic differential equation (\ref{GD-stoch1})  we obtain 
%\[ X^m_\delta(t) = \check{\theta} + e^{-t} \left[ x^m_0 - \check{\theta} - \sqrt{ \frac{\delta}{ m}}  \int_0^t e^u \bsigma(X(u)) d\bB(u) \right] \]
\begin{align} \label{GD-recursive3-0}
& X^m_\delta(t) 
%= \check{\theta} + e^{-t} \left[ x^m_0 - \check{\theta} - \sqrt{ \frac{\delta}{ m}}  \int_0^t e^u \bsigma(X(u)) d\bB(u) \right]  & \nonumber \\
 = x^m_0 e^{-t}  + \check{\theta} (1 - e^{-t})  - \sqrt{ \frac{\delta}{ m}}  \int_0^t e^{u-t} \bsigma(X(u)) d\bB(u)   & \nonumber \\
%&  = \check{\theta} + e^{-t} \left[ x^m_0 - \check{\theta} - \sqrt{ \frac{\delta}{ m}}  \int_0^t e^u %\bsigma(X(u)) 
%   \mbox{diag}(\tau, \check{\theta}_2)  d\bB(u) \right]   &\\
&  = \check{\theta} +  (x^m_0 - \check{\theta}) e^{-t}
   - \sqrt{ \frac{\delta}{ m}}  \left(  \tau   \int_0^t e^{u-t} dB_1(u),   \;  \check{\theta}_2   \int_0^t e^{u-t} dB_2(u)  \right)^\prime & \nonumber \\
%& = x^m_0 e^{-t}  + \check{\theta} (1 - e^{-t})  - \sqrt{ \frac{\delta}{ m}}  \left(   \tau   \int_0^t e^{u-t} dB_1(u) ,  \check{\theta}_2
%      \int_0^t  e^{u-t}  dB_2(u)  +  (x_{0,2} - \check{\theta}_2) e^{-t} B_2(t)  \right)^\prime & \nonumber \\
& = X(t) + (x^m_0 - x_0) e^{-t}    %x^m_0 e^{-t}  + \check{\theta} (1 - e^{-t})  
+  \sqrt{ \frac{\delta}{ m}}   \mbox{diag}(\tau, \check{\theta}_2)  \Lambda(t) 
  %- \sqrt{ \frac{\delta}{ m}} \left( \begin{array}{c}  0 \\  (x_{0,2} - \check{\theta}_2) e^{-t} B_2(t)  \end{array} \right), 
  & \nonumber \\
 & = X(t) + (x^m_0 - x_0) e^{-t} + \sqrt{ \frac{\delta}{ m}} V(t), & 
\end{align}
%\[  \left(  \tau  \int_0^t e^{u-t} dB_1(u),   \int_0^t \left[ (x_{0,2} - \check{\theta}_2) e^{-t} + \check{\theta}_2
%     e^{u-t} \right] dB_2(u)  \right)^\prime,  \] 
where $\Lambda(t)=-( \int_0^t e^{u-t} dB_1(u), \int_0^t e^{u-t} dB_2(u) )$ is an Ornstein-Uhlenbeck process whose stationary distribution 
is a bivariate normal distribution with mean zero and variance equal to the half of identity matrix, and 
$V(t)= \mbox{diag}(\tau, \check{\theta}_2)  \Lambda(t)$ is the solution of the stochastic differential equation (\ref{GD-v0}).
%\begin{align}
%& X^m_\delta(t) = \check{\theta} + 
%e^{-t} \left[ x^m_0 - \check{\theta} - \sqrt{ \frac{\delta}{ m}}  \int_0^t e^u %\bsigma(X(u)) 
%   \mbox{diag}(\tau, \check{\theta}_2)  d\bB(u) \right]   &\\
%&  = \check{\theta} +  (x^m_0 - \check{\theta}) e^{-t}
%   - \sqrt{ \frac{\delta}{ m}}  \left(  \tau   \int_0^t e^{u-t} dB_1(u),   \;  \check{\theta}_2   \int_0^t e^{u-t} dB_2(u)  \right)^\prime &\\
%& = X(t) + (x^m_0 - x_0) e^{-t} + \sqrt{ \frac{\delta}{ m}} V(t), &
%\end{align}
%where $V(t)$ is the Ornstein-Uhlenbeck process as the solution of the stochastic differential equation (\ref{GD-v0}). 
It is easy to see that the weak convergence of $V^m_\delta(t)= (m/\delta)^{1/2} [ X^m_\delta(t) - X(t)]$ to $V(t)$.
%\[ \left.   x^m_0  e^{-t}-  \sqrt{ \frac{\delta}{ m}} \left[ (x_{0,2} - \check{\theta}_2) e^{-t} B_2(t) + \check{\theta}_2
%    \int_0^t e^{u-t} dB_2(u) \right]   \right)^\prime,  \]
%\[ X^m_\delta(t) = \left(  X_1(t) -  \tau \sqrt{ \frac{ \delta  }{ m} }   B_1(t), 
%   X_2(t) - \sqrt{ \frac{ \delta  }{ m } } \int_0^t  X_2(u) d\bB(u)       \right)^\prime,  \]
%\[ = \left( \check{\theta} _1 + (x_{0,1} - \check{\theta}_1) e^{-t}  -  \tau \sqrt{ \frac{ \delta  }{ m} }   B_1(t), 
%  \check{\theta} _2 + (x_{0,2} - \check{\theta}_2) e^{-t}  - \sqrt{ \frac{ \delta  }{ m } } \left[ \check{\theta} _2 B_2(t) +
%       (x_{0,2} - \check{\theta}_2)  \int_0^t  e^{-u} d\bB(u)   \right]    \right)^\prime, \]
For the accelerated case, as we have seen, the solution of the ordinary differential equation (\ref{equ-2}) has the form  
 \[ X(t) = \check{\theta} + \frac{ 2 (x_{0} - \check{\theta} ) }{t } J_1\left( t \right). \]
 Below we will give solutions of the stochastic differential equations (\ref{Nest-stoch1}) and (\ref{Nest-limit-0}) in this case.  First we consider the solution $V(t)$ of the stochastic differential equation (\ref{Nest-limit-0}). It is easy to check that $t V(t)$ satisfies the inhomogeneous Bessel equation of the first order with constant 
 term $t^3 \mbox{diag}(\tau, \check{\theta}_2) \dot{\bB}(t)$, and its solution can be expressed as follows, 
%  \[ Y(t) = t X(t), \;\; \]
%\[ \ddot{X}(t) + \frac{3}{t} \dot{X}(t) + X(t) + G(t) =0, \;\; t^2 \ddot{Y}(t) + t \dot{Y}(t) + (t^2-1)Y(t) + t^3 G(t) =0  \]
%$X(0)=x_{0}$. 
%\[ X(t) = \frac{\pi}{2} \frac{J_1(t)}{t} \int_0^t Y_1(u) u^2 G(u) du - \frac{\pi}{2} \frac{Y_1(t)}{t} \int_0^t  J_1(u) u^2 G(u) du \]
%where $J_1(t)$ and $Y_1(t)$ are the Bessel functions of the first and second kind of order one, respectively. 
\[ V(t) = \frac{\pi}{2} \frac{J_1(t)}{t} \int_0^t \check{J}_1(u) u^2 \mbox{diag}(\tau, \check{\theta}_2) d \bB(u) - 
    \frac{\pi}{2} \frac{\check{J}_1(t)}{t} \int_0^t  J_1(u) u^2 \mbox{diag}(\tau, \check{\theta}_2) d \bB(u), \]
 where $J_1(t)$ and $\check{J}_1(t)$ are the Bessel functions of the first and second kind of order one, respectively. 
 Since in this case, $\nabla g$ is linear, $\boldsymbol{I\!\! H}\! g=1$, and the 
 stochastic differential equations (\ref{Nest-stoch1}) and (\ref{Nest-limit-0}) differ by a shift $\check{\theta}$ and a scale $m^{-1/2} \delta^{1/4}$,  
 we can easily find 
\begin{align*}
& X^m_\delta(t) = \check{\theta} + \frac{ 2 (x^m_{0} - \check{\theta} ) }{t } J_1\left( t \right)  + m^{-1/2} \delta^{1/4}  V(t) &\\
& =  X(t) + \frac{ 2 (x^m_{0} - x_0 ) }{t } J_1\left( t \right)  + m^{-1/2} \delta^{1/4}  V(t). &
\end{align*}
      %In particular for the same initial value $x^m_0=x_0$, $X^m_\delta(t) = X(t) + m^{-1/2} \delta^{1/4}  V(t)$.

 %\[ X^m_\delta(t) =  \check{\theta} + \frac{ 2 (x^m_{0} - \check{\theta} ) }{t } J_1\left( t \right) - \frac{ \delta^{1/4}  }{ \sqrt{m} } \int_0^t \left[  \int_0^v \mbox{diag}(\tau, \check{\theta}_2)  d\bB(u)  \right] dv, \]
%\[  = \check{\theta} + \frac{ 2 (x^m_{0} - \check{\theta} ) }{t } J_1\left( t \right) - \frac{ \delta^{1/4}  }{ \sqrt{m} }  \left(    \tau   \int_0^t B_1(s) ds, \; \check{\theta}_2  \int_0^t B_2(s) ds    \right)^\prime.  \]
With the initial condition in A5, it is clear that $V^m_\delta(t)= (m^2/\delta)^{1/4} [ X^m_\delta(t) - X(t)]$ weakly converges to $V(t)$.

\subsection{Joint computational and statistical asymptotic analysis for stochastic gradient descent}
%\textbf{Example 1}(Continue). 
As we advocate a joint asymptotic analysis framework in Section \ref{section-2-5}, here $X(t)$, $X^m_\delta(t)$, $V^m_\delta(t)$ and $V(t)$ 
provide a joint asymptotic analysis for the dynamic behaviors of %stochastic gradient descent 
the algorithms  (\ref{min-GD3}) and (\ref{min-Nest3}), and the weak convergence results established in Theorems \ref{thm2}-\ref{thm7} 
can be used to demonstrate the corresponding weak convergence results in $C(I\!\!R_+ )$ and $D(I\!\!R_+ )$. It is more complicated to consider the asymptotic analysis with $t \rightarrow \infty$ for the 
stochastic gradient descent case and extend the convergence results further from $[0, \infty)$ to $[0, \infty]$.
%Unfortunately, here we can not extend the results further from $[0, \infty)$ to $[0, \infty]$, 
As $t \rightarrow \infty$, Brownian motion $\bB(t)$ behaves like $(2 t \log \log t)^{1/2}$, and process $H(t)$ often 
diverges, however, there may exist meaningful distributional limits for processes $X^m_\delta(t)$, $x^m_\delta(t)$, $V^m_\delta(t)$ and $V(t)$. %as $t \rightarrow \infty$.
For the stochastic gradient descent case  we establish the weak convergence of $V^m_\delta(t)$ to $V(t)$ 
on $D(I\!\!R_+ )$ and study their asymptotic behaviors as $t \rightarrow \infty$ in the following theorem. 

\begin{thm} \label{thm8}
Suppose that the assumptions A1-A5 are met, %$\int_0^\infty \boldsymbol{I\!\! H}\! g(X(t)) dt = \infty$ in terms of matrix spectral norm, 
$\boldsymbol{I\!\! H}\! g(\check{\theta})$ is positive definite, %= \boldsymbol{I\!\! H}\! g(X(\infty))  > 0$, 
all eigenvalues of  $\int_0^t \boldsymbol{I\!\! H}\! g(X(s)) ds$ diverge as $t \rightarrow \infty$, $\boldsymbol{I\!\! H}\! g(\theta_1)$ and $\boldsymbol{I\!\! H}\! g(\theta_2)$ commute for any $\theta_1 \neq \theta_2$,  and 
assume %$m(n\delta)^{-1/2} \rightarrow 0$ and 
$m^{1/2} \delta |\log \delta|^{1/2} \rightarrow 0$, as $\delta \rightarrow 0$ and $m \rightarrow \infty$.
We obtain the following results. 

(i) As $\delta \rightarrow 0$ and $m \rightarrow \infty$, $V^m_\delta(t)=(m/\delta)^{1/2} [ X^m_\delta(t) - X(t)]$ and $(m/\delta)^{1/2} [ x^m_\delta(t) - X(t)]$ 
weakly converge to $V(t)$ on  $D(I\!\!R_+ )$. 

(ii) The stochastic differential equation (\ref{GD-v0}) %has the explicit solution (\ref{GD-v1}) of SDE (\ref{GD-v0}) we find that $V(t)$ follows a normal distribution with mean zero and variance 
admits a unique stationary distribution denoted by $V(\infty)$, where $V(\infty)$ follows a normal distribution with mean zero and covariance matrix $\Gamma(\infty)$ satisfying the algebraic Ricatti equation, 
\begin{equation} \label{stationary-xx} 
 \Gamma(\infty) \boldsymbol{I\!\! H}\! g(X(\infty)) +  \boldsymbol{I\!\! H}\! g(X(\infty)) \Gamma(\infty) = \bsigma^2(X(\infty)) . 
    %\bsigma(X(\infty))^{\prime}. 
\end{equation}
%$[\boldsymbol{I\!\! H}\! g(X(\infty))]^{-1} \bsigma(X(\infty)) \bsigma(X(\infty))^{\prime} /2$. 

(iii) Further assume that there exists a unique stationary distribution, denoted by $X^m_\delta(\infty)$, for the stochastic differential equation (\ref{GD-stoch1}). Then as $\delta \rightarrow 0$ and $m \rightarrow \infty$, 
$V^m_\delta(\infty)=(m/\delta)^{1/2} [ X^m_\delta(\infty) - X(\infty)]$ converges in distribution to $V(\infty)$.
%In particular the limiting distribution of $V^m_\delta(t)=(m/\delta)^{1/2} [ X^m_\delta(t) - X(t)]$ %and $(m/\delta)^{1/2} [ x^m_\delta(t) - X(t)]$ are 
%is $V(t)$ for $t \in [0, \infty]$. 
%where $V(\infty)$ follows a bivariate normal distribution with mean zero and variance $[\boldsymbol{I\!\! H}\! g(X(\infty))]^{-1} \bsigma(X(\infty)) \bsigma(X(\infty))^{\prime} /2$, 
%and $(V^m_\delta(t), V(t))$ describe the dynamic evolution of gradient descent algorithms  for $ t \in [0, \infty)$ and the 
%statistical distribution?? of $\sqrt{n}( \hat{\theta}_n -\check{\theta})$ for $t =\infty$. 
\end{thm}

\begin{remark} Similar to Theorem \ref{thm-1-2} and Remark  \ref{remark3-3}, %as $t,k \rightarrow \infty$ 
Theorem \ref{thm8} indicates that for the stochastic gradient descent 
case, as $\delta \rightarrow 0$ and $m \rightarrow \infty$, $X^m_\delta(\infty)$ approaches $X(\infty)=\check{\theta}$, 
%$x_\infty = x_\delta(\infty) = X(\infty) = \check{\theta}$,  $x^n_\infty = x^n_\delta(\infty) = X^n(\infty) = \hat{\theta}_n$,  
$V^m_\delta(t)=\sqrt{m/\delta} [X^m_\delta (t)  - X(t)] $ converges to $V(t)$, $t \in [0, \infty]$,
%$V^m_\delta(\infty)=\sqrt{m/\delta} [X^m_\delta(\infty)  - X(\infty)] = \sqrt{m/\delta} [x^m_\delta(\infty)  - X(\infty)]$ %$ = \sqrt{m/\delta} ( \hat{\theta}_n - \check{\theta})$, 
 %$V(\infty)= [\boldsymbol{I\!\! H}\! g(X(\infty)) ]^{-1} \bsigma(X(\infty)) \bZ = [\boldsymbol{I\!\! H}\! g(\check{\theta}) ]^{-1} \bsigma(\check{\theta}) \bZ$,  
 and $V(t)$ weakly converges to $V(\infty)$ as $t \rightarrow \infty$. 
  %In particular, as process $V^m_\delta(t)$ is indexed by $(\delta, m)$ and $t$, its limits are the same regardless the order of $m \rightarrow \infty$ and $t \rightarrow \infty$. 
 Intuitively, %the SDE and solution expression for $V(t)$ indicate that 
 $V(t)$ is a time-dependent Ornstein-Uhlenbeck process with stationary distribution $V(\infty)$ as its limit when $t \rightarrow \infty$, and similarly the solution $X^m_\delta(t)$ of the stochastic differential equation (\ref{GD-stoch1}) may admit a stationary distribution $X^m_\delta(\infty)$ as the limiting distribution of $X^m_\delta(t)$ when $t \rightarrow \infty$ (see Da Prato and Zabczyk (1996)  and Gardiner (2009) for the existence of stationary distributions). 
 Naturally $X^m_\delta(\infty)$ corresponds to $V(\infty)$. Mandt et al. (2017) essentially takes these results as its major model assumptions to establish that stochastic gradient  descent can be treated as a statistical estimation procedure in the Bayesian framework. 
\end{remark}

%Treating them as random elements in $D(I\!\!R_+)$, 
%since Theorems \ref{thm-1} and \ref{thm-1-1} establish the weak convergence of $X^n(t)$ and $x^n_\delta(t)$ to $V(t)=U(t) \bZ$ on $[0, T]$ for any $T>0$, we may conclude from these weak convergence results that $V^n(t) = \sqrt{n} [X^n(t) - X(t)]$ and $\sqrt{n} [x^n_\delta(t) - X(t)]$, $t \in [0, +\infty)$, weakly converge to $V(t)$, $t \in [0, +\infty)$.

%\begin{remark}
% $(V^m_\delta(t), V(t))$ describe the dynamic evolution of gradient descent algorithms  for $ t \in [0, \infty)$ and the 
%statistical distribution?? of $\sqrt{n}( \hat{\theta}_n -\check{\theta})$ for $t =\infty$. 
%fixed m, n $X^m_\delta(\infty) \neq \hat{\theta}_n$, varying $\delta$??
%\end{remark}

Note that stochastic gradient descent is designed for the pure computational purpose, and 
there is no corresponding objective function nor analog of minimizer $\hat{\theta}_n$ for the stochastic gradient descent algorithm, as
mini-batches (and their corresponding gradient estimators) %$\hat{\theta}^*_{mk}$)
change along iterations.  It is not clear whether we 
have known statistical estimation methods corresponding to the limits of $x^m_\delta(t)$ and $X^m_\delta(t)$ as $t \rightarrow \infty$. 
%With $V^m_\delta(\infty)=\sqrt{m/\delta} [X^m_\delta(\infty) - X(\infty)]$, $X(\infty)=\check{\theta}$, and $V(\infty)$ follows a normal
%distribution with zero mean, it is expected to show that $V^m_\delta(\infty)$ converges in distribution to $V(\infty)$. 
 Below we provide an explicit illustration of the point  through Example 1 considered in Sections \ref{section-2-5} and \ref{section-accelerate}.  
 
 \textbf{Example 1}(continue). First we evaluate 
 \[ H(t) = \int_0^t \bsigma(X(u)) d\bB(u) = \left(  \tau  B_1(t),  \check{\theta}_2 B_2(t) %+ (x_{0,2} - \check{\theta}_2) \int_0^t   e^{-u} dB_2(u)  
 \right)^\prime,  \] 
 % \left(  \tau  \int_0^t dB_1(u),  \int_0^t  \left[   \check{\theta}_2 + (x_{0,2} - \check{\theta}_2) e^{-u}   \right] dB_2(u)  \right)^\prime,  \] 
 where $\bsigma(X(u)) = \mbox{diag}(\tau, \check{\theta}_2)$, and $X(u) =   \check{\theta} + (x_{0} - \check{\theta}) e^{-u}$. 
 By the law of the iterated logarithm for Brownian motion, $H(t)$ diverges like $(t \log \log t)^{1/2}$ as $t \rightarrow \infty$.
%and thus the exploding effect is due to the factor 
%$\delta^{-1/2}$, where $\delta^{-1}$ proportional to the number of bootstrap samples (mini-batches) used. Also $\delta^{-1/2}$ is the 
%factor to fill the gap between the discrete and continuous white noise, as we have discussed in Remark \ref{remark-4-1}. 
%Solving the stochastic differential equation (\ref{GD-stoch1})  
From \eqref{GD-recursive3-0} we have 
%\[ X^m_\delta(t) = \check{\theta} + e^{-t} \left[ x^m_0 - \check{\theta} - \sqrt{ \frac{\delta}{ m}}  \int_0^t e^u \bsigma(X(u)) d\bB(u) \right] \]
\begin{eqnarray} \label{GD-recursive3}
 X^m_\delta(t) 
%%= \check{\theta} + e^{-t} \left[ x^m_0 - \check{\theta} - \sqrt{ \frac{\delta}{ m}}  \int_0^t e^u \bsigma(X(u)) d\bB(u) \right]  & \nonumber \\
% = x^m_0 e^{-t}  + \check{\theta} (1 - e^{-t})  - \sqrt{ \frac{\delta}{ m}}  \int_0^t e^{u-t} \bsigma(X(u)) d\bB(u)   & \nonumber \\
%%& = x^m_0 e^{-t}  + \check{\theta} (1 - e^{-t})  - \sqrt{ \frac{\delta}{ m}}  \left(   \tau   \int_0^t e^{u-t} dB_1(u) ,  \check{\theta}_2
%%      \int_0^t  e^{u-t}  dB_2(u)  +  (x_{0,2} - \check{\theta}_2) e^{-t} B_2(t)  \right)^\prime & \nonumber \\
 &=& X(t) + (x^m_0 - x_0) e^{-t}    %x^m_0 e^{-t}  + \check{\theta} (1 - e^{-t})  
+  \sqrt{ \frac{\delta}{ m}}   \mbox{diag}(\tau, \check{\theta}_2)  \Lambda(t)  \nonumber  \\
  %- \sqrt{ \frac{\delta}{ m}} \left( \begin{array}{c}  0 \\  (x_{0,2} - \check{\theta}_2) e^{-t} B_2(t)  \end{array} \right), 
 &=& X(t) + (x^m_0 - x_0) e^{-t}  +  \sqrt{ \frac{\delta}{ m}}  V(t), 
\end{eqnarray}
%\[  \left(  \tau  \int_0^t e^{u-t} dB_1(u),   \int_0^t \left[ (x_{0,2} - \check{\theta}_2) e^{-t} + \check{\theta}_2
%     e^{u-t} \right] dB_2(u)  \right)^\prime,  \] 
$\Lambda(t)=-( \int_0^t e^{u-t} dB_1(u), \int_0^t e^{u-t} dB_2(u) )$ is an Ornstein-Uhlenbeck process whose stationary distribution 
is a bivariate normal distribution with mean zero and variance equal to the half of identity matrix, 
$V(t) = \mbox{diag}(\tau, \check{\theta}_2)  \Lambda(t)$, and $V^m_\delta(t) = (m/\delta)^{1/2} (x^m_0 - x_0) e^{-t}  + V(t)$. 
%\[%% \mbox{diag}\left(\tau^2 \int_0^t e^{2 u - 2t} du,  \int_0^t \left[ (x_{0,2} - \check{\theta}_2) e^{-t} + \check{\theta}_2 e^{u-t} \right] ^2 du\right)
%\mbox{diag}\left(\tau^2 \int_0^t e^{2 u - 2t} du, \check{\theta}_2 \int_0^t  e^{2 u- 2t} du\right) = \mbox{diag}(\tau^2, \check{\theta}^2_2)/2. 
%% \rightarrow \mbox{diag}(\tau^2, \check{\theta}^2_2)/2, \;\;\mbox{ as } t \rightarrow \infty. 
 %\]
As $t \rightarrow \infty$, $\Lambda(t)$ approaches its stationary distribution given by 
$%\mbox{diag}(\tau, \check{\theta}_2) 
\bZ/\sqrt{2}$, where $\bZ=(Z_1, Z_2)^\prime$, and $Z_1$ and $Z_2$ are independent standard 
normal random variables. Using the expression (\ref{GD-recursive3}) we conclude that as 
$t \rightarrow \infty$, $X^m_\delta(t)$ converges in distribution to 
$X^m_\delta(\infty)=\check{\theta} + (\delta/m)^{1/2} \mbox{diag}(\tau, \check{\theta}_2) \bZ/\sqrt{2}$. 
For the initial values satisfying Condition A5, 
$x^m_0 - x_0 =o( (\delta/m)^{1/2} )$, $V^m_\delta(t)$ weakly converges to $V(t)$, and $V^m_\delta(\infty) $ weakly converges to $V(\infty) = (\tau Z_1, \check{\theta}_2 Z_2)/\sqrt{2}$. 

On the other hand,  the algorithm (\ref{min-GD3}) gives 
\[ x^m_k = x^m_{k-1} + \delta (\bar{U}^*_{mk} - x^m_{k-1}), \;\; k=1, 2, \cdots, \]
where %we consider the case that 
$\bU^*_{mk}=(U^*_{1k}, \cdots, U^*_{mk})$, $k\geq 1$, are mini-batches, 
%are sampled from large training data set $\bU_n$, 
and $\bar{U}^*_{mk}$ is the sample mean of $U^*_{1k}, \cdots, U^*_{mk}$. In comparison with the recursive 
relationship $x^n_k = x^n_{k-1} + \delta (\bar{U}_{n} - x^n_{k-1})$ for the stochastic sample optimization (\ref{min-1}) based on all data, 
and $x_k = x_{k-1} + \delta ( \check{\theta} - x_{k-1})$ for the deterministic population optimization (\ref{min-0}), the differences are $\delta (\bar{U}^*_{mk} - \bar{U}_n)$ and 
$\delta (\bar{U}^*_{mk} - \check{\theta})$, respectively. In fact, for the stochastic gradient descent case, we rewrite the recursive relationship as 
$x^m_k = (1- \delta) x^m_{k-1} + \delta \bar{U}^*_{mk} $, and obtain 
\begin{equation}  \label{GD-recursive4}
 x^m_\delta(t) = x^m_0 (1-\delta)^{[t/\delta]} + \delta  \sum_{k \delta \leq t} (1- \delta)^{[t/\delta]-k} \bar{U}^*_{mk}. 
\end{equation}
Similarly, we have 
\[ x^n_\delta(t ) = x^n_0  (1-\delta)^{[t/\delta]}  + \bar{U}_n \delta  \sum_{k \delta \leq t} (1- \delta)^{[t/\delta]-k} , \;\;
   x_\delta(t ) = x_0  (1-\delta)^{[t/\delta]}  + \check{\theta} \delta  \sum_{k \delta \leq t} (1- \delta)^{[t\delta]-k} . \]
Letting $t\rightarrow \infty$, we get 
\[ x^n_\delta(\infty) =  \bar{U}_n \delta \sum_{k =1}^\infty (1- \delta)^{k-1} = \bar{U}_n,  \;\; x_\delta(\infty) = \check{\theta}, 
\]
\[   x^m_\delta(\infty) =  \delta \lim_{\ell \rightarrow \infty} \sum_{j=0}^\ell (1- \delta)^{j} \bar{U}^{*}_{m, \ell-j} = 
\delta \sum_{k=1}^\infty (1- \delta)^{k-1} \bar{U}^{**}_{mk}, \;\; \]
where sequence $\{\bar{U}^{**}_{mk}\}_k$ is defined as the reverse sequence of $\{\bar{U}^{*}_{mk}\}_k$. 
We can clearly see that $X(\infty) = x_\delta(\infty)=\check{\theta}$, $X^n(\infty) = x_\delta(\infty) = \hat{\theta}_n$, and 
$X^m_\delta(\infty)$ and $x^m_\delta(\infty)$ approach $\check{\theta}$ but do not correspond to any statistical estimation procedures like $\hat{\theta}_n$. 
For $t \in [0, \infty)$, when $\delta$ is small, and $m$ is relatively large, 
%with enormous $n$ and relatively large $m$, 
$x^m_\delta(t)$ can be naturally approximated by 
its `limit' $x^m_0 e^{-t} + \check{\theta} (1 - e^{-t}) - (\delta/m)^{1/2} \int_0^t e^{u-t} \bsigma(X(u)) d\bB(u)$, which is equal to 
$X^m_\delta(t)$, where the last term on the right hand side of (\ref{GD-recursive4}) after centered at $\check{\theta}$ and normalized 
by $\delta^{1/2}$ weakly converges to $\int_0^t e^{u-t} \bsigma(X(u)) d\bB(u)$. 
To compare these processes, we assume initial values $x^m_0=x^n_0=x_0$ for simplicity. Then
\begin{eqnarray} 
&& x^n_\delta(t) = x_\delta(t)  + (\bar{U}_n - \check{\theta})   \left[ 1 -  (1- \delta)^{[t/\delta]  %+1
     } \right], \label{GD-recursive1} \\
 && x^m_\delta(t) = x^n_\delta(t) + \delta  \sum_{k \delta \leq t} (1- \delta)^{[t/\delta]-k} (\bar{U}^*_{mk} - \bar{U}_n) \nonumber \\
 && =  x_\delta(t) + (\bar{U}_n - \check{\theta})   \left[ 1 -  (1- \delta)^{[t/\delta]  %+1
     } \right] + 
  \delta  \sum_{k \delta \leq t} (1- \delta)^{[t/\delta]-k} (\bar{U}^*_{mk} - \bar{U}_n).  \label{GD-recursive2}
 \end{eqnarray}
 The second and third terms on the right hand side of (\ref{GD-recursive2}) account for, respectively,  the variability due to statistical estimation and the random fluctuation due to the use of min-batches %(or bootstrap samples) 
 for gradient estimation from iteration to iteration in the stochastic gradient descent algorithm. 
%Note that $\bar{U}_{n} - \check{\theta} $ is of order $n^{-1/2}$, and $\bar{U}^*_{mk} - \bar{U}_n$ is of order $m^{-1/2}$ (as $m/n \rightarrow 0$).  
Note that $\bar{U}_{n} - \check{\theta} $ and $\bar{U}^*_{mk} - \bar{U}_n$ are of orders $n^{-1/2}$ and $m^{-1/2}$, respectively. 
%This is true even for the case that mini-batches $\bU^*_{mk}=(U^*_{1k}, \cdots, U^*_{mk})$, $k\geq 1$, are sampled from large data set $\bU_n$. In fact we may resort  the strong approximation (Koml\'os et al. (1975, 1976), %) and its related bootstrap strong approximation 
% Cs\"{o}rg\"{o} et al. (1999), and Cs\"{o}rg\"{o} and Mason (1989)) to obtain %lead to 
%\begin{equation} \label{KMT-1}
% \bar{U}^*_{mk} - \bar{U}_n = m^{-1/2} A_{mk}  + O_P(m^{-1} \log m), \;\; \bar{U}_{n} - \check{\theta}  = n^{-1/2} D_n + O_P(n^{-1} \log n),
%\end{equation}
%where $A_{mk}$, $k=1, 2, \cdots, $ are nearly i.i.d. random variables defined by a sequence of independent Brownian bridges on some probability spaces, with random variables $D_n$ defined by another sequence of independent Brownian bridges on the probability spaces. 
%From (\ref{KMT-1}) and $m/n \rightarrow 0$, we 
As $m/n \rightarrow 0$, we easily conclude that the second term 
on the right hand side of (\ref{GD-recursive2}) is of order higher than the third term, where the the third term represents the cumulative 
min-batch-subsampling %(or bootstrapping) 
effect up to the $k=[t/\delta]$-th iteration, with the second term for the statistical estimation error. 
%Stochastic gradient descent algorithm suffers from the cumulative effect of random fluctuations due to different min-batches (bootstrap samples) employed from iterations to iterations. 
(\ref{GD-recursive1}) and (\ref{GD-recursive2}) show that as $m,n \rightarrow \infty$, $x^n_\delta(t)$ and $x^m_\delta(t)$ approach 
$\check{\theta}$, and on average both gradient descent and stochastic gradient descent algorithms stay on target, the difference is
their random variabilities. Theorems \ref{thm-1} and \ref{thm-1-1} establish an order of $n^{-1/2} \bZ$ for the random variability of the 
gradient descent algorithm using all data, while Theorems \ref{thm3} and \ref{thm4} indicate that for the stochastic gradient descent algorithm, the cumulative random fluctuation up to the $[t/\delta]$-iteration can be modeled by process $(\delta/m)^{1/2} V(t)$, where 
$V(t)$  given by the stochastic differential equation (\ref{GD-v0}) (or its expression (\ref{GD-v1})) is a time-dependent Ornstein-Uhlenbeck process that may admit a 
stationary distribution with mean zero and variance $\bsigma^2(X(\infty))/[2 \boldsymbol{I\!\! H}\! g(X(\infty))]$, factor $m^{-1/2}$ accounts for the effect of each mini-batch %(or bootstrap sample) 
of size $m$, and factor $\delta^{1/2}$ represents the effect of the total number of mini-batches %(or bootstrap samples) 
 that is proportional to $1/\delta$. The normalized factor $(\delta/m)^{1/2}$ means that while each mini-batch %(or bootstrap sample) 
of size $m$ is not as efficient as full data sample of size $n$, but repetitive use of min-batch subsampling %(or bootstrapping) 
in stochastic gradient descent utilizes more data and improves its efficiency, with the improvement represented by $\delta^{1/2}$, 
where $1/\delta$ is proportional to the total number of min-batches %(or bootstrap samples) 
up to the time $t$ (or the $t/\delta$-th iteration). In other words, repeatedly subsampling compensates the efficiency loss due to a mini-batch %(or bootstrap sample) 
of small size at each iteration. Intuitively, it means that the stochastic gradient descent algorithm invokes different min-batches %(or bootstrap samples)  
 resulted some random fluctuation  when moving from one iteration to another, and as the number of iterations increases,  
subsampling  improves efficiency with factor $(\delta/m)^{1/2}$ instead of $m^{-1/2}$, to make up loss from $n^{-1/2}$ to $m^{-1/2}$, that is, updating with the use of many mini-batches %(or bootstrap samples) 
can improve accuracy for the stochastic gradient descent algorithm.

\subsection{Convergence analysis of  stochastic gradient descent for non-convex optimization} %Saddle points and local minimizers}

Our asymptotic results may have implications for stochastic gradient descent used in non-convex optimization particularly in deep learning. Recent studies often suggest that stochastic gradient descent algorithms can escape from saddle points and find good local 
minima (Jastrz\c{e}bski et al. (2018), Jin et al. (2017), Keskar et al. (2017), Lee et al. (2016), Shallue et al. (2019)). 
%As our analysis for the stochastic gradient descent case may extend to cover non-convex optimizations, 
We will provide new rigorous analysis and heuristic intuition to shed some light on the phenomenon. First note that we can relax the convexity assumption on the objective function $g(\theta)$ for the deterministic population optimization (\ref{min-0}) in Theorems \ref{thm2}-\ref{thm4}, and thus Theorem \ref{thm8} can be easily adopted to non-convex optimization with $\check{\theta}$ being a critical point of $g(\theta)$. Suppose that stochastic gradient descent processes converge to the critical point $\check{\theta}$. 
Applying large deviation theory to the stochastic differential equations (\ref{GD-stoch1}) and (\ref{GD-stoch2}) corresponding to the gradient descent algorithm, we obtain that as $\delta/m$ goes to zero, if the critical point is a saddle point of $g(\theta)$, the continuous processes generated from the stochastic differential equations can escape from the saddle point in a polynomial time (proportional to $ (m/\delta)^{1/2} \log (m/\delta)$)
(see Kifer (1981, theorems 2.1-2.3) and Li et al. (2017b, theorem 3.3)); while, if the critical point is a local minimizer of $g(\theta)$, the continuous processes will take an exponential time 
(proportional to $\exp \{c (m/\delta)^{1/2} \}$ for some generic constant $c$) to get out a neighborhood of the local minimizer (see Dembo and Zeitouni (2010, chapter 5) and Li et al. (2017b, theorem 3.2)). We may also explain the phenomenon 
from the limiting distribution point of view. Theorem \ref{thm3} indicates that the continuous processes $X^m_\delta(t)$ and $\check{X}^m_\delta(t)$ 
generated from the stochastic differential equations (\ref{GD-stoch1}) and (\ref{GD-stoch2}) are asymptotically the same as the deterministic solution $X(t)$ of the ordinary differential equation (\ref{equ-2}) plus $(\delta/m)^{1/2} V(t)$, where $V(t)$ is the solution of the stochastic differential equation (\ref{GD-v0}). The limiting process $V(t)$ is a time-dependent Ornstein-Uhlenbeck process %with a specialized explicit expression 
given by the expression (\ref{GD-v1}). 
%Again suppose that the stochastic gradient descent algorithm converges to some critical point $\check{\theta}$, and its corresponding continuous processes should approach the critical point $\check{\theta}$. 
We have the following theorem for the behaviors of $g(X^m_\delta(t))$ and $g(\check{X}^m_\delta(t))$ around the critical point $\check{\theta}$. 
\begin{thm} \label{thm9}
Suppose that the assumptions A1-A5 (except for the convexity of $g(\cdot)$) are met, and the gradient descent process $X(t)$ given by the ordinary differential equation (\ref{GD-c1}) converges to a critical point 
$\check{\theta}$ of $g(\cdot)$. %,  and assume $m^{1/2} \delta |\log \delta|^{1/2} \rightarrow 0$,  as $\delta \rightarrow 0$ and $m \rightarrow \infty$. 
Then we have the following results, 
%\[  \cL^n( \theta; \bU_n) = g(\theta) + O_P(n^{-1/2}), \qquad \nabla \cL(\theta; \bU_n) = \nabla g(\theta) + n^{-1/2} \bZ(\theta) + o_P(n^{-1/2}), \]
%\[  \hat{\cL}^m( \theta; \bU_m^*) = g(\theta) + O_P(m^{-1/2}), \qquad \]
%\[ \nabla \hat{\cL}^m(X^m_\delta(t)); \bU_m^*(t)) = \nabla g(X^m_\delta(t)) + (\delta/m)^{1/2} \bZ(\theta) + o_P(m^{-1/2}),\]
\begin{align}
& g(X^m_\delta(t)) = g(X(t)) + (\delta /m)^{1/2} \nabla g(X(t)) V^m_\delta(t) + \frac{\delta}{2 m} [V^m_\delta(t)]^\prime \boldsymbol{I\!\! H}\! g(X(t)) V^m_\delta(t)  +  o_P(\delta/m) &  \nonumber \\
& =  g(\check{\theta}) +  \frac{1}{2} 
 [X(t) - \check{\theta} + (\delta/ m)^{1/2}  V^m_\delta(t)]^\prime \boldsymbol{I\!\! H}\! g(\check{\theta}) [X(t) - \check{\theta} + (\delta/ m)^{1/2}  V^m_\delta(t)]  \nonumber \\
 % \left\{  [X(t) - \check{\theta}]^2 + 2 (\delta/ m)^{1/2}  [X(t) - \check{\theta}] V(t) + \delta [V(t)]^2/m  \right\} 
& \qquad + o_P\left(\delta/m + [X(t) - \check{\theta}]^2  \right) , \label{optimization-x1} \\
%\end{align*}
%\begin{align*}
& \nabla g(X^m_\delta(t)) = \nabla g(X(t)) + (\delta /m)^{1/2} \boldsymbol{I\!\! H}\! g(X(t)) V^m_\delta(t) +  o_P((\delta/m)^{1/2}) & \nonumber \\
& =  \boldsymbol{I\!\! H}\! g(\check{\theta}) [X(t) - \check{\theta} + (\delta/ m)^{1/2}  V^m_\delta(t)] + o_P\left( (\delta/m)^{1/2}  + | X(t) - \check{\theta} |   \right) , \label{optimization-x2}
\end{align}

and the same equalities hold with $X^m_\delta$ replaced by $\check{X}^m_\delta$,  %and $x^m_\delta$.
where $X(t)$, $X^m_\delta(t)$ and $\check{X}^m_\delta(t)$ are the solutions of the differential equations (\ref{GD-c1}), (\ref{GD-stoch1}) and (\ref{GD-stoch2}),
respectively, $V^m_\delta(t) = (m/\delta)^{1/2} [X^m_\delta(t) - X(t)]$, and the equalities hold in the weak sense that we may consider $X^m_\delta(t)$, %(or $\check{X}^m_\delta(t)$), 
$V^m_\delta(t)$, and $V(t)$ on some common probability spaces through Skorokhod's representation. 

If $\check{\theta}=X(\infty)$ is a local minimizer with positive definite $\boldsymbol{I\!\! H}\! g(\check{\theta})$, then as $t \rightarrow \infty$, $V(t)$ has a limiting stationary distribution with mean zero and covariance matrix $\Gamma(\infty)$ satisfying the algebraic Ricatti equation
(\ref{stationary-xx}), % $X(\infty)=\check{\theta}$, 
and
\begin{eqnarray} 
 && E[ g(X^m_\delta(t))] = g(X(t)) + \frac{\delta}{4 m} tr[\bsigma^2(X(\infty))] + o(\delta /m),  \label{optimization-x3} \\
 && E[ |\nabla\!g(X^m_\delta(t))|^2] = |\nabla \! g(X(t))|^2 + \frac{\delta}{2 m} tr[  \bsigma^2(X(\infty)) \boldsymbol{I\!\! H}\! g(X(\infty)) ] + o(\delta /m). \qquad  \label{optimization-x4} 
   \end{eqnarray}
If $\check{\theta}$ is a saddle point, 
%$V(t)$ has covariance function 
%\[ [Cov(V(t), V(s))]_{ii} = \frac{ \bsigma_{ii}(X(t)) }{2 \lambda_i} \left[  e^{ - \lambda_i |t + s| } - e^{ -\lambda_i |t-s| } \right], \]
%where $\lambda_i$ are eigenvalues of $\boldsymbol{I\!\! H}\! g(\check{\theta})$. Thus, 
$V(t)$ diverges and thus does not have any limiting distribution.
\end{thm}

Theorem \ref{thm9} shows that as $X(t)$ gets close to the critical point $\check{\theta}$ within the range of order $(\delta /m)^{1/2}$, 
$g(X^m_\delta(t))$ and $g(\check{X}^m_\delta(t))$ are approximately quadratic. As Theorem \ref{thm3} indicates that 
$V(t)$ is the limit of $V^m_\delta(t)$, we may replace $V^m_\delta(t)$ by $V(t)$ in the expansions of $g(X^m_\delta(t))$ and 
$\nabla \!g(X^m_\delta(t))$ and find that $V(t)$ plays a key role in determining the behavior of the stochastic gradient descent algorithm. 
If the critical point $\check{\theta}$ is a saddle point of $g(\theta)$, $\boldsymbol{I\!\! H}\! g(\cdot)$ is non-positive definite around the saddle point, and the time-dependent Ornstein-Uhlenbeck process $V(t)$ does not have any stationary distribution, and in fact, it diverges. Thus processes 
$X^m_\delta(t)$ and $\check{X}^m_\delta(t)$ have unstable behaviors around the saddle point and can make big moves, which leads them to escape from the saddle point. On the other hand, if the critical 
point $\check{\theta}$ is a local minimizer of $g(\theta)$, then $g(\cdot)$ may be approximately quadratic with positive definite 
$\boldsymbol{I\!\! H}\! g(\cdot)$ around the local minimizer. 
Also $V(t)$ has a stationary distribution, and all the processes maintain stable stochastic behaviors. 
%and tend to stay around the local minimizer. 
%Moreover,  after further analyzing the behaviors of $X^m_\delta$ and $\check{X}^m_\delta$ around a local minimizer $\check{\theta}$, we find that stochastic gradient descent behaves quite differently depending on factor $(\delta /m)^{1/2}$ and the local geometry of $g(\cdot)$ around $\check{\theta}$. 
%Keskar et al. (2017) considers two kinds of local minimizers: sharp and flat local minimizers. 
%We characterize the sharpness of a local minimizer $\check{\theta}$ by the Hessian matrix $\boldsymbol{I\!\! H}\! g(\check{\theta})$ and range index $\rho$ such that the whole local minimizer well falls inside $\{\theta: |\theta - \check{\theta}| <  \rho\}$, or equivalently,
%gradient descent process $X(t)$ will move away from the local minimizer $\check{\theta}$ if $X(t)$ starts outside $\{\theta: |\theta - \check{\theta}| <  \rho\}$. 
%We can easily see that the smaller $\rho$ and larger $\boldsymbol{I\!\! H}\! g(\check{\theta})$ are, the bigger $\nabla g(\cdot)$ and steeper $g(\cdot)$ are around 
%$\check{\theta}$, while the larger $\rho$ and smaller $\boldsymbol{I\!\! H}\! g(\check{\theta})$ are, the smaller $\nabla g(\cdot)$ and flatter $g(\cdot)$ are around $\check{\theta}$. From 
Moreover, %Theorem \ref{thm9} indicates that  
the stochastic component $(\delta/ m)^{1/2} V(t)$ plays a dominant role in determining the 
behaviors of $g(X^m_\delta(t))$ and $g(\check{X}^m_\delta(t))$ around the local minimizer. 
In fact, %Theorems \ref{thm3} and  \ref{thm9} 
the equations  \eqref{optimization-x1}-\eqref{optimization-x4} imply that $g(X^m_\delta(t))$ and $g(\check{X}^m_\delta(t))$ behave like 
  \begin{align*}
&  g(X(t)) + (\delta /m)^{1/2} \nabla g(X(t)) V(t) + \frac{\delta}{2 m} [V(t)]^\prime \boldsymbol{I\!\! H}\! g(X(t)) V(t), &
\end{align*}
whose mean is asymptotically equal to 
\[ g(X(t)) + \frac{\delta}{4 m} tr [\bsigma^2(X(\infty))],  \]
and $ \nabla \!g(X^m_\delta(t))$ and $ \nabla \!g(\check{X}^m_\delta(t))$ act similar to  
 \begin{align*}
& \nabla \! g(X(t)) + (\delta /m)^{1/2} \boldsymbol{I\!\! H}\! g(X(t)) V(t), &
\end{align*}
which has mean $\nabla \! g(X(t))$ and  variance asymptotically equal to 
\[ \frac{\delta}{2 m} tr [\bsigma^2(X(\infty)) \boldsymbol{I\!\! H}\! g(X(\infty))  ].   \]
First note that the stochastic components in the equations  \eqref{optimization-x1}-\eqref{optimization-x4} 
depend on the learning rate $\delta$ and the batch size $m$ only through their ratio $\delta/m$. 
 Second, they are characterized by the local geometry of the objective function around the local minimizer, where the local geometric 
 characteristics include the Hessian $\boldsymbol{I\!\! H}\! g( X(t))$ and the gradient covariance $\bsigma^2(X(t))$.  
In particular utilizing the joint analysis along with the algebraic Ricatti equation \eqref{stationary-xx} for the stationary covariance 
of the Ornstein-Uhlenbeck process, we establish the equations  \eqref{optimization-x3}-\eqref{optimization-x4} to specify 
%the relationship between minima fund by 
 how the minima found by % the behavior of 
 stochastic gradient descent %around a local minimum is 
 are influenced by four factors: 
 %the ratio $\delta/m$ of learning rate to batch size, 
 the learning rate $\delta$, batch size $m$,  gradient covariance $\bsigma^2(\check{\theta})$,  and Hessian $\boldsymbol{I\!\! H}\! g(\check{\theta})$.  
 These may have implications regarding the behavior of stochastic gradient descent for non-convex optimization. 
  %what kind of  minima stochastic gradient descent can find.  
 For example, 
 %(\ref{stationary-xx}) shows that the stationary variance $\Gamma(\infty)$ varies positively with $\bsigma^2(\check{\theta})$ but inversely with 
 %$\boldsymbol{I\!\! H}\! g( \check{\theta})$. That is, for smaller (or larger) $\bsigma^2(\check{\theta})$, or 
 %for a more (or less) sharp local minimizer $\check{\theta}$ with a larger (or smaller) $\boldsymbol{I\!\! H}\! g(\check{\theta})$, 
% (\ref{stationary-xx}) implies that the corresponding stationary distribution of $V(t)$ may have a smaller (or larger) variance  
 %$\Gamma(\infty)$, and thus $V(t)$ tend to produce values of smaller (or larger) magnitude.  
 \eqref{optimization-x3}-\eqref{optimization-x4} indicate that the ratio $\delta/m$ of learning rate to batch size is inversely proportional to $tr[ \bsigma^2(\check{\theta})]$ for a given level of the expected loss at $\check{\theta}$ and $tr[ \bsigma^2(\check{\theta}) \boldsymbol{I\!\! H}\! g(\check{\theta})]$ for a specific level of the expected loss gradient at $\check{\theta}$. That is, for a larger $\delta/m$, stochastic gradient descent tends to find a local minimum with smaller $tr[ \bsigma^2(\check{\theta})]$ and $tr[ \bsigma^2(\check{\theta}) \boldsymbol{I\!\! H}\! g(\check{\theta})]$. 
 For a more sharp (or wide) local minimizer $\check{\theta}$, we have larger (or smaller) $\boldsymbol{I\!\! H}\! g(\check{\theta})$ 
 as well as faster (or slower) changing gradient around $\check{\theta}$, which points to a tendency of larger (or smaller) 
 %gradient covariance trace 
 $tr[\bsigma^2(\check{\theta})]$ and $tr[ \bsigma^2(\check{\theta}) \boldsymbol{I\!\! H}\! g(\check{\theta})]$. 
 The scenario analysis seems to indicate that for a higher ratio of learning rate to batch size, stochastic gradient descent tends to avoid 
 %escape from 
 sharp minima and lead towards more wide minima that typically generalize well. 
  
%three factors - learning rate, batch size and gradient covariance - influence the minima found by SGD. In particular we find that
%the ratio of learning rate to batch size is a key determinant of SGD dynamics and of the width of the final minima, and that higher values of the ratio lead to wider minima and often better generalization
A case in point is %the equation (\ref{optimization-x3}) which was studied under a special case in Jastrz\c{e}bski et al. (2018) 
%about a relation between the ratio of learning rate to batch size and the width of the minimum found by stochastic gradient descent. 
a special case studied in Jastrz\c{e}bski et al. (2018) that identified three factors to influence the minimum found by stochastic gradient descent. We describe the special case as follows. 
Suppose that $U$ has a pdf $f(u;\theta)$, and the loss function  $\ell(\theta; u) = - \log f(u; \theta)$. 
 %and the gradient descent process $X(t)$ given by  (\ref{GD-c1}) converges to a local minimizer $\check{\theta}$ of $g(\cdot)$, then  
 %we have the following theorem. 
 %\begin{thm} \label{thm10}
 %Suppose that the assumptions A1-A5 (except for the convexity of $g(\cdot)$) are met, and the gradient descent process $X(t)$ given by  (\ref{GD-c1}) converges to a local minimizer $\check{\theta}$ of $g(\cdot)$. 
 %Assume that $U$ has a pdf $f(u;\theta)$, and the loss function  $\ell(\theta; u) = - \log f(u; \theta)$. Then 
 %\[   E[ g(X^m_\delta(t))] = g(X(t)) + \frac{\delta}{4 m} tr[\boldsymbol{I\!\! H}\! g(X(t))] + o(\delta /m). \] 
 % \end{thm} 
 %Indeed, 
 Since we take the loss as a negative log likelihood, this is the MLE case, and the gradient covariance $\bsigma^2(\theta)$ corresponds to 
 the negative Fisher information, which in turn is equal to  $E [\boldsymbol{I\!\! H}\! \ell(\theta; U)]  =  \boldsymbol{I\!\! H}\! g(\theta)$. 
 In this case, because the stochastic differential equation  
 \eqref{GD-v0}  has commutable diffusion coefficient $\bsigma(X(t))$ and drift $\boldsymbol{I\!\! H}\! g(X(t)) = \bsigma^2(X(t)) $, we have an explicit expression \eqref{GD-v1}  
 for the time-dependent Ornstein-Uhlenbeck process $V(t)$, with the simple stationary distribution $N(0, \Gamma(\infty)) = N(0, \bI)/2$.  With these explicit forms and $\boldsymbol{I\!\! H}\! g(X(t)) = \bsigma^2(X(t))$,  Jastrz\c{e}bski et al. (2018, equation (9)) employed direct calculations for this specific example to essentially 
 establish a special form of (\ref{optimization-x3}) with only three of the four factors about a relation between the ratio of learning rate to batch size and the width of the minimum found by stochastic gradient descent. However, their corresponding formula no longer holds for the general case.  In fact, for this case, given $\boldsymbol{I\!\! H}\! g(X(t)) = \bsigma^2(X(t))$ and the explicit expressions of $V(t)$ and its stationary distribution, our general results can 
 easily recover the relation in Jastrz\c{e}bski et al. (2018, equation (9)). Moreover, we conclude that 
 smaller values for both $tr[\bsigma^2(\check{\theta})] = tr[ \boldsymbol{I\!\! H}\! g(\check{\theta})]$ and $tr[ \bsigma^2(\check{\theta}) \boldsymbol{I\!\! H}\! g(\check{\theta})] = tr[ \{ \boldsymbol{I\!\! H}\! g(\check{\theta}) \}^2 ]$ point to a wider minimum.

 %The findings seem to be in consistent with some empirical studies (e.g. Jastrz\c{e}bski et al. (2018)) regarding the behavior of stochastic gradient descent involving the learning rate, batch size, and local geometry.  
 
% Suppose that $\boldsymbol{I\!\! H}\! g(X(t))$ has an eigen-decomposition $ \boldsymbol{I\!\! H}\! g(X(t)) =  O_t \Lambda_t O_t^\prime$, 
% where $O_t$ and $\Lambda_t$ are orthogonal and diagonal matrices, respectively. Then we have 
% $\bsigma(X(t)) = O_t \Lambda^{1/2}_t O_t^\prime$, and $\check{V}_t = O^\prime_t V(t)$ obeys 
% \[ d \check{V}_t = - \Lambda_t  \check{V}_t dt + \Lambda^{1/2}_t d \bB(t), \]
%  which has a stationary distribution $N(0, \bI/2)$. 
%  \[   E [ ( V(t)) ^\prime \boldsymbol{I\!\! H}\! g(X(t)) V(t) ] = E[(  \check{V}_t)^\prime   \Lambda_t \check{V}_t ] = tr(  \Lambda_t )/2
%    = tr[\boldsymbol{I\!\! H}\! g(X(t))]/2.   \]

In Section 1.2 we have discussed Foster et al. (2019) regarding %decomposing the complexity of a stochastic optimization into the complexity of its corresponding deterministic optimization and the sample complexity. 
transforming the problem of finding near-stationary points for a stochastic optimization into finding near-stationary points for the corresponding deterministic optimization plus the related sample complexity. The equation 
\eqref{optimization-x2} indicates that finding near-stationary points of $g(X^m_\delta(t))$ can be converted into making $| \nabla \! g(X(t)) |$ small and controlling $(\delta /m)^{1/2} \boldsymbol{I\!\! H}\! g(X(t)) V(t)$. Making $| \nabla \! g(X(t)) |$ small 
means finding a near-stationary point for the corresponding deterministic optimization. The equation 
 \eqref{optimization-x4} implies that the control of $(\delta /m)^{1/2} \boldsymbol{I\!\! H}\! g(X(t)) V(t)$ can be achieved through bounding its 
 variance, namely, imposing a bound on $tr [\bsigma^2(X(\infty))$ $\boldsymbol{I\!\! H}\! g(X(\infty))  ]$ along with selecting a sufficiently small 
 ratio $\delta/m$ of learning rate to batch size, where 
%which is related to the sample complexity of the associated learning algorithm.
%We utilize the variance $tr [\bsigma^2(X(t)) \boldsymbol{I\!\! H}\! g(X(t))  ]$ along with the ratio $\delta/m$ to control $(\delta /m)^{1/2} \boldsymbol{I\!\! H}\! g(X(t)) V(t)$.
the variance $tr [\bsigma^2(X(\infty)) \boldsymbol{I\!\! H}\! g(X(\infty))  ]$ is used to describe the sample complexity of the associated statistical learning problem for the time-dependent Ornstein-Uhlenbeck process.  
This indicates that our results are in line with Foster et al. (2019), and future study may reveal further intrinsic connection between our results and those in Foster et al. (2019). 
Also we again find that the behavior of stochastic gradient descent depends on the gradient covariance $\bsigma^2(X(t))$ and the Hessian $ \boldsymbol{I\!\! H}\! g(X(t)) $ along with the ratio $\delta/m$ of learning rate to batch size.

\textbf{Example 2}. Consider the problem of orthogonal tensor decomposition (Ge et al. (2015) and Li et al. (2016)). 
%%A $4$-th order
%Tensors are $d$-dimensional arrays, and we consider $4$-th order tensors for simplicity. 
%Denote by $I\!\!R^{d^4}$ the set of all $4$-th order tensors, and set $[d]=\{1, \cdots, d\}$. 
%For a $4$-th order tensor $\Upsilon \in I\!\!R^{d^4}$, denote by $\Upsilon_{i_1 i_2 i_3 i_4}$ its $(i_1,i_2,i_3, i_4)$-th entry, where $i_1, i_2, i_3, i_4 \in [d]$.  
%%Tensors can be constructed from tensor products, 
%For a given vector $\balpha =(\alpha_i) \in I\!\!R^{d}$, we denote by $\balpha^{\otimes4}$ its $4$-th order tensor product with 
%\[  \left[ \balpha^{\otimes4} \right]_{i_1 i_2 i_3 i_4} = \alpha_{i_1} \alpha_{i_2} \alpha_{i_3} \alpha_{i_4}.   \]
%We say that a 
A $4$-th order tensor $\Upsilon \in I\!\!R^{d^4}$ has an orthogonal tenor decomposition if it can be written as 
\[ \Upsilon = \sum_{j=1}^d \balpha_j^{\otimes 4}, \]
where $\balpha_j$'s are orthonormal vectors in $I\!\!R^{d}$ satisfying $\|\balpha_j\|=1$ and $\balpha_j^{\dagger} \balpha_k=0$ for $j \neq k$, and the problem is to find tensor components $\balpha_j$'s given such a tensor $\Upsilon$. 
Since the tensor decomposition problem has inherent symmetry, that is, a tensor decomposition is unique only up to component permutation and sign-flips, the symmetry property makes the corresponding %natural 
optimization problem multiple local minima and thus 
non-convex.

A formulation of orthogonal tensor decomposition as an optimization problem to find one component was proposed in Frieze et al. (1996) %and show that this new problem satisfies the strict saddle property.  solves the problem of finding one component, 
with the following objective function
\[ \max_{ \| \bbeta \|_2=1} \Upsilon( \bbeta, \bbeta, \bbeta, \bbeta). \]
Take $\Upsilon = E [U^{\otimes 4}]$ to be the 4-th order tensor whose $(i_1, i_2, i_3, i_4)$-th entry is $E(U_{i_1} U_{i_2} U_{i_3} U_{i_4})$, 
%We consider the problem of independent component analysis (ICA) and its associated stochastic gradient descent algorithm. 
where $U$ is  a $d$-dimensional random vector with distribution $Q$. Assume $U = \bA W$, where $W$ is bounded, and has symmetric and i.i.d. components with unit variance, 
% and identical moments up to the $8$-th order, 
and $\bA$ is an orthonormal matrix whose column vectors $\balpha_1, \cdots, \balpha_d$ form an orthonormal basis. 
%Our goal is to find orthonormal basis $\balpha_j$ based on i.i.d. data $U_1, \cdots, U_m$ sampled from $Q$ with a tensor decomposition approach. 
%Note that we assume $U = \bA W$ and
Let $\psi_k$ be the $k$-th moment of i.i.d. components of $W$, with $\psi_1=0$, $\psi_2=1$, and $\psi_4$ equal to its kurtosis. 
The optimization problem can be equivalently casted as 
%It is easy to check 
%\[ \Upsilon (\bbeta, \bbeta,\bbeta,\bbeta) = E[ (\bbeta^\dagger U)^4] = 3 + (\psi_4 - 3) \sum_{j=1}^d (\balpha_j^\dagger \bbeta)^4, \]
%and thus we cast 
the problem of finding components $\balpha_j$'s into the solution to the following population optimization problem
\[  \min - sign(\psi_4-3) E[ (\bbeta^\dagger U)^4] = \min  \sum_{j=1}^d  - (\balpha_j^\dagger \bbeta)^4
     \;\;\; \mbox{subject to } \|\bbeta \|=1. \] 
It is well known that there is an unidentifiable tensor structure for $\psi_4= 3$. 
%which corresponds to the normal case where $U$ and $W$ have identical normal distribution. 
%%similar to % call the value $|\psi - 3|$ the tensor gap. The reader will see later that, 
% %to eigengap for principal component analysis. 
 %We assume $\psi_4 \neq 3$, 
 For $\psi_4 \neq 3$,  we may consider empirical objective function 
 $\sum_{i=1}^n  -sign(\psi_4-3) (\bbeta^\dagger U_i)^4$ based on available data $U_1, \cdots, U_n$, and study the corresponding 
 stochastic optimization. 
 %with the corresponding stochastic optimization problem
%\[  \min  \sum_{i=1}^n  -sign(\psi_4-3) (\bbeta^\dagger U_i)^4 \;\;\; \mbox{subject to } \|\bbeta \|=1. \] 

%It is straightforward to conclude that all stable equilibria of (2.2) are $\pm a_i$ whose number linearly
%grows with d. Meanwhile, by analyzing the Hessian matrices the set of unstable equilibria of (2.2)
%includes (but not limited to) all $v^*= d^{-1/2}(\pm1, · · · ,\pm)$, whose number grows exponentially as d increases [18, 44].
The objective function of the population optimization has the gradient and Hessian in the tangent space 
\[ sign(\psi_4-3)  \nabla \Upsilon(\bbeta, \bbeta, \bbeta, \bbeta ) = 4 \left( [ \beta_1^2 - \| \bbeta \|_4^4 ] \beta_1, \cdots, [ \beta_d^2 - \| \bbeta \|_4^4 ] \beta_d \right) , \]
\[ sign(\psi_4-3)  \boldsymbol{I\!\! H}\! \Upsilon(\bbeta, \bbeta, \bbeta, \bbeta ) = -12 \mbox{diag}(\beta_1^2, \cdots, \beta_d^2) + 4 \|\bbeta \|_4^4 \bI_d. \]
%Gradient:  the gradient in the tangent space is equal to [Tensor-Ge (equ 158-159)]:
%\[ 4 \left( [ x_1^2 - \| x \|_4^4 ] x_1, \cdots, [ x_d^2 - \| x \|_4^4 ] x_d \right) \]
%Hessian:  the second-order partial derivative of Lagrangian is equal to
%\[ -12 \mbox{diag}(x_1^2, \cdots, x_d^2) + 4 \|x \|_4^4 I_d \]
Applying gradient descent and stochastic gradient descent algorithms for solving the population and stochastic optimization problems we obtain sequence $x_k$ and $x^m_k$.
%We have the following stochastic gradient descent algorithm for solving the stochastic optimization problem. Denote by 
%$\cS^{d-1} = \{u :\| u \| = 1\}$  the unit sphere in $R^d$, and $\Pi u = u/\|u\|$ for $ u \neq 0$ the projection operator onto $\cS^{d-1}$. With appropriate initialization, the stochastic gradient descent for tensor method has the following on-line iterative update scheme
%\[ u_k = \Pi \left\{ u_{k-1} + %sign( \psi_4 - 3)  \cdot 
%  4  \delta \left[ (u_{k-1} )^\dagger U^{(k)} \right]^3 U^{(k)} \right\}.  \]
%%By symmetry, our algorithm in Eq. (2.3) converges to a uniformly random tensor component from d
%%components. In order to solve the problem completely, one can repeatedly run the algorithm using
%%different set of online samples until all tensor components are found. In the case where d is high, the
%%well-known coupon collector problem [16] implies that it takes $\approx d \log d$ runs of SGD algorithm to obtain all d tensor components.
%For simplicity we consider the transformed iteration $x_k \equiv \bA^\dagger u_k$ and obtain the following equivalent update iteration,
%\[ x_k= A^\dagger u_k = \Pi \left\{ A^\dagger u_{k-1} + 4 \delta \left[ (u_{k-1})^\dagger A A^\dagger U^{(k)} \right]^3 A^\dagger U^{(k)} \right\} \]
%\[ = \Pi \left\{ x_{k-1} + 4 \delta \left[ (x_{k-1})^\dagger W^{(k)} \right]^3  W^{(k)} \right\} . \]
%%where $\pm \beta$ has the same sign as $\psi_4 - 3 $. 
As learning rate $\delta \rightarrow 0$, $x_{[t/\delta]}$ converges in probability to gradient flow $X(t)$ satisfying 
\[ \frac{d X_i}{dt} =  4 X_i \left( X_i^2 - \sum_{\ell=1}^d X_\ell^4 \right), \;\; i=1, \cdots, d, \]
and $( m/\delta)^{1/2} [x^m_{[t/\delta]} - X(t)]$ has a weak convergence limit $V(t)$ satisfying,
\[  dV(t) = - \bmu(X(t)) V(t)dt - \bsigma(X(t)) d\bB(t), \]
where 
\[ \bmu(\bbeta)  = -12 \mbox{diag}(\beta_1^2, \cdots, \beta_d^2) + 4 \|\bbeta \|_4^4 \bI_d, 
\;\;  \bsigma^2(\bbeta)  = 16 Cov( [\bbeta^\dagger \bW]^3 \bW). \]
To better understand the complex gradient flow system and time-dependent Ornstein-Uhlenbeck process limit, we derive explicit expressions for the case of 
$d = 2$ where $X(t)=(X_1(t), X_2(t))^\prime$ has a closed-form solution 
 \[ X_1^2(t)  = 0.5 +  0.5[ 1 + c  \exp (- 4 t )]^{-0.5}, \;\; X_2^2(t) = 1 - X_1^2(t), \]
with constant $c$ depending on the initial value. In particular, if the initial vector$([X_1(0)]^2 < [X_2(0)]^2$ (resp. $([X_1(0)]^2 > [X_2(0)]^2$),
then $X_1(t)$  approaches $1$ (resp. $0$) as $t \rightarrow \infty$. %Second we consider 
Direct calculations lead to 
\[ \bsigma^2(u) /16= E( [u_1 W_1 + u_2 W_2 ]^3  \bW \bW^\dagger ) - E ( [u_1 W_1 + u_2 W_2 ]^3 \bW)  [E ( [u_1 W_1 + u_2 W_2 ]^3 \bW)]^\dagger, \]%= Cov( [u_1^3 Y_1^3 + u_2^3 Y_2^3 + 3 u_1^2 u_2 Y_1^2 Y_2 + 3 u_1 u_2^2 Y_1 Y_2^2] Y) \]
where $ E ( [u_1 W_1 + u_2 W_2 ]^3 \bW) = (u_1^3 \psi_4 + 3 u_1 u_2^2 , u_2^3 \psi_4 + 3 u_1^2 u_2), $
\begin{align*}
%& E ( [u_1 W_1 + u_2 W_2 ]^3 \bW) = (u_1^3 \psi_4 + 3 u_1 u_2^2 , u_2^3 \psi_4 + 3 u_1^2 u_2) , &\\
& E ( [u_1 W_1 + u_2 W_2 ]^6 W_1^2)  =\sum_{\ell=0}^6 C^6_\ell u_1^\ell u_2^{6-\ell} \psi_{\ell+2} \psi_{6-\ell} , &\\
& E ( [u_1 W_1 + u_2 W_2 ]^6 W_2^2)  =\sum_{\ell=0}^6 C^6_\ell u_2^\ell u_1^{6-\ell} \psi_{\ell+2} \psi_{6-\ell} , & \\
& E ( [u_1 W_1 + u_2 W_2 ]^6 W_1 W_2)  =\sum_{\ell=0}^6 C^6_\ell u_1^\ell u_2^{6-\ell} \psi_{\ell+1} \psi_{6-\ell+1} . & 
\end{align*}
We may simplify $\bmu(X(t))$ and $\bsigma(X(t))$ further by approximating $X(t)$ with its limit $w_*$(some critical point).  For example, if $X(t)$ approaches 
critical point $w_*=(1,0)$ (saddle point), we may approximate $\bmu(X(t)) $ and $ \bsigma^2(X(t))$ by 
$\bmu (w_*) $ and $\bsigma^2(w_*)$, where 
\[ \bmu (w_*) = - 12 \mbox{diag}(w_{*1}^2, w_{*2}^2) + 4\bI  = \mbox{diag}(-8, 4 ),  \] % \mbox{or diag}(4, -8) \]
\[ \bsigma^2(w_*) =\mbox{diag}( \psi_8 - \psi_4^2, \psi_6) , \] %or  \mbox{diag}( \psi_6 , \psi_8 - \psi_4^2) \]
%\[ E ( [u_1 Y_1 + u_2 Y_2 ]^3 Y) = \psi_4 u_* \]
%\[ E ( [u_1 Y_1 + u_2 Y_2 ]^6 Y_1 Y_2) =0\]
%\[ E ( [u_1 Y_1 + u_2 Y_2 ]^6 Y_1^2) = u_1^6 \psi_8 + u_2^6 \psi_6 \]
%at $u_*=(1,0)$,
and obtain an approximate stochastic differential equation for the weak convergence limit $V(t)$, 
\[ dV(t) =  4 \left[ - \mbox{diag}(-2, 1) V(t) dt -  [\mbox{diag}(\psi_8 - \psi_4^2, \psi_6)]^{1/2} d\bB_t  \right]  . \]
On the other hand, if $X(t)$ approaches critical point $w_*= %2^{-1/2}(-1,1) or 
2^{-1/2}(1,-1)$ (local minimizer), we have %may approximate $\boldsymbol{I\!\! H}\! g(X(t)) $ and $ \bsigma^2(X(t))$ by $\boldsymbol{I\!\! H}\! g(u_*) $ and $\bsigma^2(u_*)$, 
\[ \bmu(w_*) = - 12 \mbox{diag}(w_{*1}^2, w_{*2}^2)  + 4 \bI/d  = -8 I/d=-4 \bI, \]
\[  \bsigma^2(w_*) = \frac{1}{8} \left( \begin{matrix}  \psi_8 + 16 \psi_6 + 15 \psi_4^2 - 26 \psi_3 \psi_5 - (\psi_4+3)^2 & 
  30 \psi_3 \psi_5 -12 \psi_6 - 20 \psi_4^2 + (\psi_4+3)^2 \\
  30 \psi_3 \psi_5 -12 \psi_6 - 20 \psi_4^2 + (\psi_4+3)^2 &    \psi_8 + 16 \psi_6 + 15 \psi_4^2 - 26 \psi_3 \psi_5 - (\psi_4+3)^2
    \end{matrix} \right),  \]
\[ dV(t) =  - 4 V(t) dt -   \bsigma(u_*) d\bB_t.  \]
It is easy to see from the stochastic differential equations that $V(t)$ has a stationary distribution for the local minimizer case, while $V(t)$ 
diverges for the saddle point case (in fact the first component of $V(t)$ has variance with exponential growth in $t$).

\subsection{Statistical analysis of stochastic gradient descent for output inference}
%\subsection{Implication for statistical inference based on outputs from stochastic gradient descent}
There is a great current interest in statistical analysis of stochastic gradient descent. Examples include statistical variability analysis 
and Bayesian inference (Chen et al. (2018), Li et al. (2018), Mandt et al. (2017), Toulis and Airoldi (2017)). Our results may have important %fundamental 
implications on statistical analysis of stochastic gradient descent. For the case of stochastic gradient descent, 
Theorems \ref{thm3}-\ref{thm4} show that output sequence $x^m_k$ generated from the stochastic gradient descent algorithm 
(\ref{min-GD3}) is asymptotically equivalent to the continuous processes $X^m_\delta(t)$ and $\check{X}^m_\delta(t)$ generated from 
the stochastic differential equations (\ref{GD-stoch1}) and (\ref{GD-stoch2}), respectively, and they both in turn are asymptotically the 
same as $(\delta/m)^{1/2} V(t)$ plus the deterministic solution $X(t)$ of the ordinary differential equation (\ref{GD-c1}), 
where $V(t)$ is the solution of the stochastic differential equation (\ref{GD-v0}). The limiting process $V(t)$ is 
a time-dependent Ornstein-Uhlenbeck process, %with a specialized explicit expression given by (\ref{GD-v1}).  
and its stationary distribution is a normal distribution given by Theorem \ref{thm8} with mean zero and covariance $\bGamma(\infty)$ specified in the algebraic Ricatti equation (\ref{stationary-xx}).
This suggests that statistical inference based on $x^m_k$ can be asymptotically equivalent to statistical inference based on 
discrete samples from $X(t) + (\delta/m)^{1/2} V(t)$. As $t \rightarrow \infty$,  $X(t)$ converges to the true 
minimizer $X(\infty)=\check{\theta}$, $V(t)$ converges in distribution to $V(\infty)$, which follows its stationary distribution 
$N(0, \bGamma(\infty))$. Thus inferences based on $x^m_k$ can be asymptotically equivalent to inferences based on 
discrete samples from the Ornstein-Uhlenbeck process with stationary distribution $N(\check{\theta}, \delta \bGamma(\infty)/m)$. 
Below we discuss two specific cases. 

Consider the Bayesian treatment of stochastic gradient descent in Mandt et al. (2017).  As described above, Theorems \ref{thm3}-\ref{thm4} and Theorem \ref{thm8} imply that outputs from the stochastic gradient descent algorithm (\ref{min-GD3}) are asymptotically 
equivalent to discrete samples from 
%$\check{\theta}+ (\delta/m)^{1/2} V(t)$, where 
%%deterministic  $X(t)$ is the solution of (\ref{equ-2}) with limit $\lim_{t \rightarrow \infty} X(t) = \check{\theta}$, 
%$V(t)$ is a time-dependent Ornstein-Uhlenbeck process with stationary distribution $N(0, \bGamma(\infty))$, and 
the Ornstein-Uhlenbeck process with stationary distribution $N(\check{\theta}, \delta \bGamma(\infty)/m)$, where $\bGamma(\infty)$ 
is given by the algebraic Ricatti equation (\ref{stationary-xx}). 
Applying the Bernstein-von Mises theorem to discrete samples from the Ornstein-Uhlenbeck process, 
%$\check{\theta} + (\delta/m)^{1/2} V(t)$ with stationary distribution $N(\check{\theta}, \delta \bGamma(\infty)/m)$, 
we obtain that the posterior distribution is 
asymptotically equal to a normal distribution with mean and covariance equal to the MLE of $\check{\theta}$ and the Fisher information evaluated at the MLE, respectively.  
Since the stochastic gradient descent outputs are asymptotically equivalent to discrete samples from the Ornstein-Uhlenbeck process, 
%$\check{\theta} + (\delta/m)^{1/2} V(t)$, 
the posterior distribution based on outputs from stochastic gradient descent is asymptotically the same as the posterior distribution for 
the Ornstein-Uhlenbeck model, and thus is asymptotically equal to the normal distribution. 
The obtained results can be employed to justify the essential inference assumptions in Mandt et al. (2017) and Li et al. (2018) that stochastic gradient descent is a stationary Ornstein-Uhlenbeck process, 
%with stationary distribution $N(\check{\theta},  \delta \bGamma(\infty)/m)$, 
and the corresponding posterior distribution is Gaussian. 
%This suggests that statistical inference based on $x^m_k$ can be asymptotically equivalent to statistical inference based on 
%discrete sample from $X(t) + (\delta/m)^{1/2} V(t)$. As $t \rightarrow \infty$, deterministic $X(t)$ converges to true 
%minimizer $X(\infty)=\check{\theta}$, and $V(t)$ converges in distribution to its stationary distribution 
%$V(\infty) \sim N(0, \bGamma(\infty))$, thus inference based on $x^m_k$ can be asymptotically equivalent to inference based on 
%a sample from an Ornstein-Uhlenbeck process with stationary distribution $N(\check{\theta}, \delta \bGamma(\infty)/m)$.
%Hellinger distribution between two normal distributions 
% \[ H^2\left( N(\mu_1,\Sigma_1), N(\mu_2, \Sigma_2)\right) = 1 -  \left(\frac{ 4 |\Sigma_1 | | \Sigma_2|}{ |\Sigma_1 +\Sigma_2|^2 } \right)^{1/4} \exp\left\{ - \frac{1}{4} (\mu_1-\mu_2)^\prime \left( \Sigma_1 + \Sigma_2\right)^{-1} (\mu_1 - \mu_2)     \right\}  \]

Another case is the average output from stochastic gradient descent. Denote by $\bar{x}^m_\delta$ the average of $N$ outputs $x^m_{k_i}=x^m_\delta(t_{k_i})$, $i=1, \cdots, N$, from the stochastic gradient descent algorithm 
(\ref{min-GD3}), where $N$ may depend on $m$ and $\delta$, and $N \rightarrow \infty$ as $\delta \rightarrow 0$ and 
$m \rightarrow \infty$. 
By Skorokhod's representation theorem and Theorems \ref{thm3}-\ref{thm4} we have that as $\delta \rightarrow 0$ and 
$m \rightarrow \infty$, the average of $x^m_\delta(t)$ has the same asymptotic distribution as the average of $X^m_\delta(t)$, and the difference between 
$$(m/\delta)^{1/2}  \left[ N^{-1} \sum_{i=1}^N X^m_\delta(t_{k_i})  - N^{-1} \sum_{i=1}^N X(t_{k_i}) \right]  \;\mbox{ and } \; N^{-1} \sum_{i=1}^N V(t_{k_i}) $$  is negligible. Note that deterministic $N^{-1} \sum_{i=1}^N X(t_{k_i}) $ 
converges to $X(\infty)=\check{\theta}$, and %Theorem \ref{thm8} indicates 
that for large $N$ the distribution of $N^{-1/2} \sum_{i=1}^N V(t_{k_i})$ can be approximated by 
%the stationary distribution $N(0, \bGamma(\infty))$ of $V(t)$. 
a normal distribution with mean zero and covariance $A^{-1} S A^{-1}$, where with notations in Theorem \ref{thm8} we set 
\begin{equation} \label{notationA-S}
 A = \boldsymbol{I\!\! H}\! g(\check{\theta}) = \boldsymbol{I\!\! H}\! g(X(\infty)), \;\;
S = \bsigma^2(\check{\theta}) %\bsigma(\check{\theta} )^{\prime} 
= \bsigma^2(X(\infty)). % \bsigma(X(\infty))^{\prime} . 
\end{equation}
This suggests that $(m/\delta)^{1/2} ( \bar{x}^m_\delta - \check{\theta} )$ has an asymptotic normal distribution with mean zero 
and covariance $A^{-1} S A^{-1}$, and we may use outputs from stochastic gradient descent to estimate $A^{-1} S A^{-1}$ and 
employ the associated Ornstein-Uhlenbeck process to justify 
the estimation approaches. See Chen et al. (2018) and Li et al. (2018) for the covariance estimation study of stochastic gradient descent.  

Note the asymptotic covariance difference between  $\bGamma(\infty)$ and $A^{-1} S A^{-1}$
for stochastic gradient descent above and in the literature. For example, in  Chen et al. (2018), Kushner and Yin (2003), Li et al. (2018), and Polyak and Juditsky (1992), the average output from stochastic gradient descent has asymptotic covariance  $A^{-1} S A^{-1}$,
while Mandt et al. (2017) and Theorem \ref{thm8} indicate that the asymptotic covariance of the corresponding outputs is equal to the stationary covariance $\bGamma(\infty)$ [defined in (\ref{stationary-xx})] of the associated Ornstein-Uhlenbeck process. We explain and reconcile 
the difference between the covariances $A^{-1} S A^{-1}$ and $\bGamma(\infty)$ as follows. 
%$$[\boldsymbol{I\!\! H}\! g(X(\infty))]^{-1} \bsigma(X(\infty)) \bsigma(X(\infty))^{\prime} [ \boldsymbol{I\!\! H}\! g(X(\infty)) ]^{-1} = 
% [\boldsymbol{I\!\! H}\! g(\check{\theta})]^{-1} \bsigma(\check{\theta}) \bsigma(\check{\theta} )^{\prime} [ \boldsymbol{I\!\! H}\! 
% g(\check{\theta} ) ]^{-1}, $$
On the one hand, although the Ornstein-Uhlenbeck process $V(t)$ approaches its normal stationary distribution with mean zero 
and covariance $\bGamma(\infty)$, its 
re-scaled average $ \frac{1}{\sqrt{N}} \sum_{i=1}^N V(t_{k_i})  \approx \frac{1}{\sqrt{N}} \int_{0}^N V(s) ds$ has asymptotic covariance 
$A^{-1} S A^{-1}$. Indeed, without confusion we denote by $V(t)$ the stationary solution of the Ornstein-Uhlenbeck model
$dV(t) = - A  V(t) dt +  S^{1/2} d\bB_t$, and define its auto-covariance function $\zeta(s_1-s_2)=E[V(s_1)( V(s_2))^\prime ]$. Then 
the variance of $\frac{1}{\sqrt{N}} \int_{0}^N V(s) ds$ is equal to 
\begin{eqnarray*}
&& \frac{1}{N} \int_{0}^N \int_0^N E[V(s_1) (V(s_2))^\prime] ds_1 ds_2 = \frac{1}{N} \int_{0}^N \int_0^N \zeta(s_1 - s_2) ds_1 
ds_2  \\
%&& = \frac{1}{N} \int_{0}^N  ds_2 \int_{-s_2}^{N-s_2} \zeta(u) du  
&& = \int_{-\infty}^\infty \zeta(u) du + O(N^{-1})  =  A^{-1} S A^{-1} + O(N^{-1}),   \mbox{ as } N \rightarrow \infty, 
\end{eqnarray*}
where $A$ and $S$ are given by the expressions (\ref{notationA-S}), and the last equality is due to the facts that 
\begin{eqnarray*}
&& \zeta(0) = V\!ar(V(s))= \int_0^\infty e^{-A s} S e^{-A s} ds = \Gamma(\infty) \mbox{ satisfying } \zeta(0) A + A \zeta(0) = S, \\
&&  \zeta(s) = e^{- A s} \zeta(0), \;\; \zeta(-s) = \zeta(0) e^{- A s}, \;\;  s\geq 0, \\
&& \int_{-\infty}^\infty \zeta(u) du = A^{-1} \zeta(0) + \zeta(0) A^{-1} = A^{-1} [ \zeta(0) A + A \zeta(0) ] A^{-1} = 
 A^{-1} S A^{-1} .
 \end{eqnarray*}
On the other hand, the stationary covariance $\bGamma(\infty)$ is derived by treating the 
stochastic gradient descent recursive equation as an approximate VAR(1) %(vector autoregression of order $1$) 
model (see  Polyak and Juditsky (1992)), where the VAR(1) model can be expressed as $V_k =  \Psi V_{k-1} + e_k$, with $\Psi = I - \delta A$, 
and random errors $e_k$ has covariance $V\!ar(e_k) = \delta S$. The VAR(1) model can be approximated by an 
Ornstein-Uhlenbeck model $dV(t) = - A  V(t) dt +  S^{1/2} d\bB_t$. 
%$dV_t = - \boldsymbol{I\!\! H}\! g(\check{\theta}) V_t dt + \bsigma(\check{\theta}) d\bB_t$ can be treated as a continuous 
%approximation of the VAR(1) model with $\Psi = I - \delta A$, and $\tau^2 = \delta S$. 
From the VAR(1) equation and stationarity we obtain 
\begin{eqnarray*}
&& V\!ar(V_k) =  \Psi V\!ar(V_{k-1})  \Psi  + \delta S = 
 (I - \delta A ) V\!ar(V_{k})  (I - \delta A ) + \delta S \\
 && = V\!ar(V_k) - \delta A V\!ar(V_{k}) - \delta V\!ar(V_{k})  A + \delta^2 A V\!ar(V_{k})  A + \delta S,  
 \end{eqnarray*}
 Canceling out $V\!ar(V_k)$, dividing by $\delta$ on both sides, and then letting $\delta \rightarrow 0$ we obtain 
  \[  A \, V\!ar(V_k) + V\!ar(V_k) \,A = S , \] 
 which recovers the algebraic Ricatti equation (\ref{stationary-xx}) for the stationary covariance $\bGamma(\infty)$ of the Ornstein-Uhlenbeck process $V(t)$.  
%$V\!ar(V(\infty))$ of the Ornstein-Uhlenbeck process $V(t)$  \[  A \, V\!ar(V(\infty)) + V\!ar(V(\infty)) \,A = S , \]
 In particular, for the one-dimensional case, the AR(1) variance has an expression $Var(e_k)/(1 - \Psi^2)$. 
 Plugging $\Psi = I - \delta A$ and $Var(e_k) = \delta S$ into the variance formula we obtain 
 \[ \frac{ Var(e_k)}{ 1 - \Psi^2}  = \frac{ \delta S}{ 1 - (1 - \delta A)^2} = \frac{S}{2 A} + O(\delta),  \]
 where the leading term $\frac{S}{2 A} $ is the exact stationary variance of the Ornstein-Uhlenbeck process. 

\section{A numerical example}

This section considers a simple example with some numerical study to illustrate the approximation of gradient descent algorithms by ordinary or stochastic differential equations.

\textbf{Example 3}. Consider the following simple linear regression model 
\begin{equation} \label{regression1}
   U_{1i} = U_{2i} ^\prime \theta +\varepsilon_i, \;\; i=1, \cdots, n, 
\end{equation}
where $U_{1i}$ and $U_{2i}$ are response and covariate, respectively,  $\theta =(\theta_1, \theta_2)^\prime$ is the parameter, and random errors $\varepsilon_i$ are i.i.d. normal random variables with mean zero and variance $\tau^2$. We consider both fixed and random designs. For the random design case, we assume that 
%\[   U_1 =  U_2 \theta + \varepsilon, \qquad E(\varepsilon_i)=0,\;\;  Cov(\varepsilon_1, \cdots, \varepsilon_n) = \Sigma, \]
%\[ \ell(\theta; U_{1i}, U_{2i}) = ( U_{1i} - U_{2i} \theta)^2, \qquad g(\theta) = \frac{1}{n} \sum_{i=1}^n E[( U_{1i} - U_{2i} \theta)^2] = Var(\varepsilon_1) +   (\theta - \check{\theta})^\prime Var(U_{2i}) (\theta - \check{\theta}), \]
%Select $\check{\theta}=0$, $Var(U_2) = H/2 = \mbox{diag}(h_1, \cdots, h_p)/2$. Take in ODE (\ref{equ-2}) 
%\[ g(\theta) = \frac{1}{2} \langle \theta, H \theta \rangle +c, \;\; \nabla g(\theta) = H \theta, \;\; H = \mbox{diag}(h_1, \cdots, h_p) \]
%Suppose that $U_i = (U_{1i}, U_{2i}^\prime)^\prime$, $i=1, \cdots, n$, are sampled from the regression model
%where $\theta =(\theta_1, \theta_2)^\prime$, 
$U_{2i}$ and $\varepsilon_i$ are independent, %$\varepsilon_i$ are i.i.d. normal random variables with mean zero and variance $\tau^2$, 
and $U_{2i}$ are i.i.d. mean zero bivariate normal random vectors. %with mean zero and covariance matrix $\balpha=E[ U_{2i} U_{2i}^\prime] = (\alpha_{ij})_{i,j=1,2}$. %$\mbox{diag}(a_{11}, a_{22})/2$. 
For the fixed design case,  we set $U_{2i}$ to be deterministic instead of bivariate 
normal random variables, where observations $U_{1i}$ are not i.i.d, and we need to make some obvious modification. 

%\textbf{Example 3}. 
%We may also consider regression with fixed design. Model (\ref{regression1}) is the same except for that now we set $U_{2i}$ to be deterministic instead of bivariate 
%normal random variables. Note that observations $U_{1i}$ are not i.i.d, and we need to make some obvious modification. 
First consider the fixed design case. Denote by $\check{\theta}$ the true value of the parameter $\theta$ in the regression model. 
Let $U_1=(U_{11}, \cdots, U_{1n})^\prime$,  $U_2 = (U_{21}, \cdots, U_{2n})^\prime$, and $\varepsilon=(\varepsilon_1, \cdots, \varepsilon_n)$. Assume that we have an orthogonal design so that $U_2^\prime U_2/n$ is equal to the identity matrix. 
Then we may write regression model in a matrix form $U_1 = U_2 \theta + \varepsilon$,  %$\bU_n = (U_1, U_2)$. 
and define $\cL^n (\theta; \bU_n ) = (U_1 - U_2 \theta)^\prime  (U_1 - U_2 \theta)/(2n)$, and 
$g(\theta) = E[ \cL^n (\theta; \bU_n ) ]$. Simple calculations show $g(\theta) = [(\theta - \check{\theta})^\prime (\theta - \check{\theta})  + \tau^2]/2$, 
$\nabla \! g(\theta) = \theta - \check{\theta}$, 
$\nabla \! \cL^n (\theta; \bU_n ) =- U_2^\prime  (U_1 - U_2 \theta)/n = \nabla \! g(\theta) - n^{-1/2} \tau \bZ$, where $\bZ=n^{-1/2} U_2^\prime \varepsilon/\tau$ follows a standard bivariate normal distribution. 

It is easy to see that the deterministic population minimization problem (\ref{min-0}) and 
the stochastic sample minimization problem (\ref{min-1}) have explicit solutions in this case: $g(\theta)$ has the minimizer being the true 
parameter value $\check{\theta}$,  and $\cL^n(\theta; \bU_n ) $ has the minimizer  equal to the least squares estimator $\hat{\theta}_n$. 

%We consider the accelerated case, as the plain gradient descent case is similar but much simpler. 
The ordinary differential equations  (\ref{GD-c2}) and (\ref{GD-c2-1}) [or (\ref{equ-3}) and (\ref{equ-3-1})] are identical in this case, 
%exactly the same,  and with linear $\nabla \!g(\theta)= \theta - \check{\theta}$, 
and as $n \rightarrow \infty$, their limits are given by the ordinary differential equations 
(\ref{GD-c1}) and (\ref{equ-2}). Specially, the differential equations admit solutions
with the following expressions 
\[ X^n(t) = (X^n_1(t), X^n_2(t) )^\prime = \frac{U_2^\prime U_1}{n}  + \left (x^n_{0} - \frac{U_2^\prime U_1}{n} \right)  e^{-t},  X(t) = (X_1(t), X_2(t))^\prime = \check{\theta} + (x_{0} - \check{\theta}) e^{-t} \]
for the plain gradient descent case. For the accelerated case, 
\[ X^n(t) = \left( \begin{array}{l} X^n_1(t) \\ X^n_2(t) \end{array} \right)  = \left( \check{\theta} + n^{-1/2}\tau \bZ 
  + \frac{ 2 ( x_0^n  - \check{\theta} - n^{-1/2} \tau \bZ) }{t  } J_1(t) \right)_{i=1,2},  \] 
  \[ X(t) = \left( \begin{array}{l} X_1(t) \\ X_2(t) \end{array} \right)  = \left( \check{\theta}  
  + \frac{ 2 ( x_0  - \check{\theta} ) }{t  } J_1(t) \right)_{i=1,2},  \]  
where $x_0^n$ and $x_0$ are initial values of $X^n(t)$ and $X(t)$, respectively, and $J_1(u)$ is the Bessel function of the first kind of order one. %As $n \rightarrow \infty$, the solution has limit $X(t)$ as the solution of (\ref{equ-2}), with expression 
The results are identical to those for Example 1 given in Section \ref{section-2-5}.  For both stochastic gradient descent and accelerated stochastic gradient descent cases,  the situation is also the same as the part of Example 1 considered in Section \ref{section-accelerate} 
with explicit forms. 
%Numerical results were computed from these explicit expressions to illustrate gradient descent algorithms and the corresponding differential equations as illustrated in Figure \ref{Figure1}.
%check the good approximation of finite sample distributions by the asymptotic theory. 

Now consider the random design case.  %Assume $U_{2i}$ are i.i.d. bivariate normal random vectors with mean zero and covariance matrix $\balpha=E[ U_{2i} U_{2i}^\prime] = (\alpha_{ij})_{i,j=1,2}$.
Denote the covariance matrix of $U_{2i}$ by $\balpha=E[ U_{2i} U_{2i}^\prime] = (\alpha_{ij})_{i,j=1,2}$, let $\check{\theta}$ be the true value of the parameter $\theta$ in the regression model, and define 
$\ell(\theta; U_i ) = (U_{1i} - U_{2i}^\prime \theta)^2/2$. Then $\cL^n(\theta; \bU_n ) = \frac{1}{n} \sum_{i=1}^n (U_{1i} - U_{2i}^\prime \theta)^2/2$ 
%\[ (U_{1i} - U_{2i}^\prime \theta)^\prime (U_{1i} - U_{2i}^\prime \theta) =    [\varepsilon_i - (\theta-\check{\theta})^\prime U_{2i}][ \varepsilon_i - U_{2i}^\prime (\theta - \check{\theta}) ] =   \]
is the half mean residual square error, $g(\theta) = E[\ell(\theta; U_i ) ] = \tau^2/2 + (\theta - \check{\theta})^\prime \balpha  (\theta - \check{\theta})^2/2$,
$\nabla g(\theta) = \balpha %\mbox{diag}(a_1, a_2) 
(\theta - \check{\theta})$, $\nabla \ell(\theta; U_i) = U_{2i} ( U_{2i}^\prime \theta - U_{1i})$, and 
$\nabla \! \cL^n(\theta; \bU_n) = \frac{1}{n} \sum_{i=1}^n U_{2i} ( U_{2i}^\prime \theta - U_{1i})$. 
Also from the regression model (\ref{regression1}) we have 
\[ \nabla \ell(\theta; U_i ) = U_{2i} U_{2i}^\prime ( \theta - \check{\theta}) - U_{2i} \,\varepsilon_i, \;\; 
 E[ \nabla \ell(\theta; U_i )] = \balpha (\theta - \check{\theta})=\nabla g(\theta), \]
 \[ \sigma(\theta)= Var [\nabla \ell(\theta; U_i )] =  
    E[ U_{21} U_{21}^\prime ( \theta - \check{\theta})  ( \theta - \check{\theta})^\prime  U_{21} U_{21}^\prime ]
  + \tau^2  \balpha - \balpha (\theta - \check{\theta})(\theta - \check{\theta})^\prime \balpha^\prime, 
    \]
 where we set $U_{21}=(H_1, H_2)^\prime$ and $\beta=(\beta_1, \beta_2)^\prime = \theta - \check{\theta}$, and compute 
$E(H_1^4)=3\alpha_{11}^2$, $E(H_1^3 H_2)=3\alpha_{11}\alpha_{12}$, $E(H_1^2 H_2^2)=\alpha_{11}\alpha_{22}+2\alpha_{12}^2$, $E(H_1 H_2^3)=3\alpha_{22}\alpha_{21}$, $E(H_2^4)=3\alpha_{22}^2$, and 
%$\frac{\partial \ell}{\partial \theta}=-2H^\mathsf{T}H(\beta-\theta)-2H^\mathsf{T}\varepsilon$.\\
$ U_{21} U_{21}^\prime ( \theta - \check{\theta})  ( \theta - \check{\theta})^\prime  U_{21} U_{21}^\prime =$
\[ 
\begin{pmatrix}
H_1^4\beta_1^2+2H_1^3 H_2\beta_1\beta_2+ H_1^2 H_2^2\beta_2^2 & 
H_1^3 H_2\beta_1^2+2 H_1^2 H_2^2\beta_1\beta_2+ H_1 H_2^3\beta_2^2 \\
H_1^3 H_2\beta_1^2+2 H_1^2 H_2^2\beta_1\beta_2+ H_1 H_2^3\beta_2^2 & 
H_1^2 H_2^2\beta_1^2+2 H_1 H_2^3\beta_1\beta_2+ H_2^4\beta_2^2
\end{pmatrix}.
\]
Again it is easy to see that the minimization problems corresponding to
(\ref{min-0}) and (\ref{min-1}) have explicit solutions in this case: $g(\theta)$ has the minimizer being the true 
parameter value $\check{\theta}$,  and $\cL^n(\theta; \bU_n ) $ has the minimizer  equal to the least squares estimator $\hat{\theta}_n$. 

Take $\balpha=
\begin{pmatrix}
0.02 & 0\\
0 & 0.005
\end{pmatrix}$, $\tau=0.1$, and $\check{\theta}=(0,0)^\prime$. Then $g(\theta)=(0.02\theta_1^2+0.005\theta_2^2+0.1)/2$, and 
\[ \bsigma^2(\theta)=
\begin{pmatrix}
2 \alpha_{11}^2 \theta_1^2+ \alpha_{11} \alpha_{22} \theta_2^2 + \tau^2 \alpha_{11} & 
  \alpha_{11} \alpha_{22} \theta_1\theta_2  \\
 \alpha_{11} \alpha_{22} \theta_1\theta_2 & 
\alpha_{11} \alpha_{22} \theta_1^2 + 2 \alpha_{22} \theta_2^2 + \tau^2 \alpha_{22}
\end{pmatrix} . \]
Unlike the fixed design case, the random design case lacks explicit expressions for the processes corresponding to the 
 accelerated (or stochastic) gradient descent algorithms and the ordinary (or stochastic) differential equations. 
%Starting from initial value $x_0=(0.1,0.1)^\prime$, 
We applied the gradient descent, accelerated gradient descent, and stochastic gradient descent algorithms and 
solved the corresponding %differential equation (\ref{equ-3}) and stochastic differential equation (\ref{GD-stoch1}) 
%\[  \ddot{X}^n(t) + \frac{3}{t} \dot{X}^n(t) + \nabla g(X^n(t))  + n^{-1/2} \bsigma (X^n(t)) \bZ = 0, \]
ordinary or stochastic differential equations by the Euler scheme for various initial values and $(m, n, \delta)$. We conducted simulations to check the finite sample performance of the obtained gradient descent central limit theorem. 
Figure 1 illustrates sample paths of sequences generated from the accelerated gradient descent (based on all data) and stochastic gradient 
descent algorithms and their corresponding ordinary or stochastic differential equations. It also displays the comparison of 
the limiting distribution with the corresponding finite sample distribution of the scaled discrepancy between the sample and 
population gradient flows. 
As analytically demonstrated in the fixed design case,  the numerical results show that the algorithms and the ordinary or stochastic differential equations can successfully find the solutions of the corresponding minimization problems. The sample paths generated by the algorithms and differential equations for solving  the stochastic sample optimization problem are random sequences or curves distributed around the true targets for the corresponding deterministic population optimization, 
 and the sample path for each algorithm is closely tracked by that for its corresponding differential equation. We found that the finite sample distributions of $V^n(t)$ and $V^m_\delta(t)$ can be well approximated by their corresponding limiting distributions provided by the obtained gradient descent central limit theorem. 
%Numerical results were computed from these explicit expressions to illustrate gradient descent algorithms and the corresponding differential equations as illustrated in Figure \ref{Figure1}.
%check the good approximation of finite sample distributions by the asymptotic theory. 

\begin{figure}[!htpb]
\vspace{-0.2in}
\includegraphics[width=0.5\textwidth]{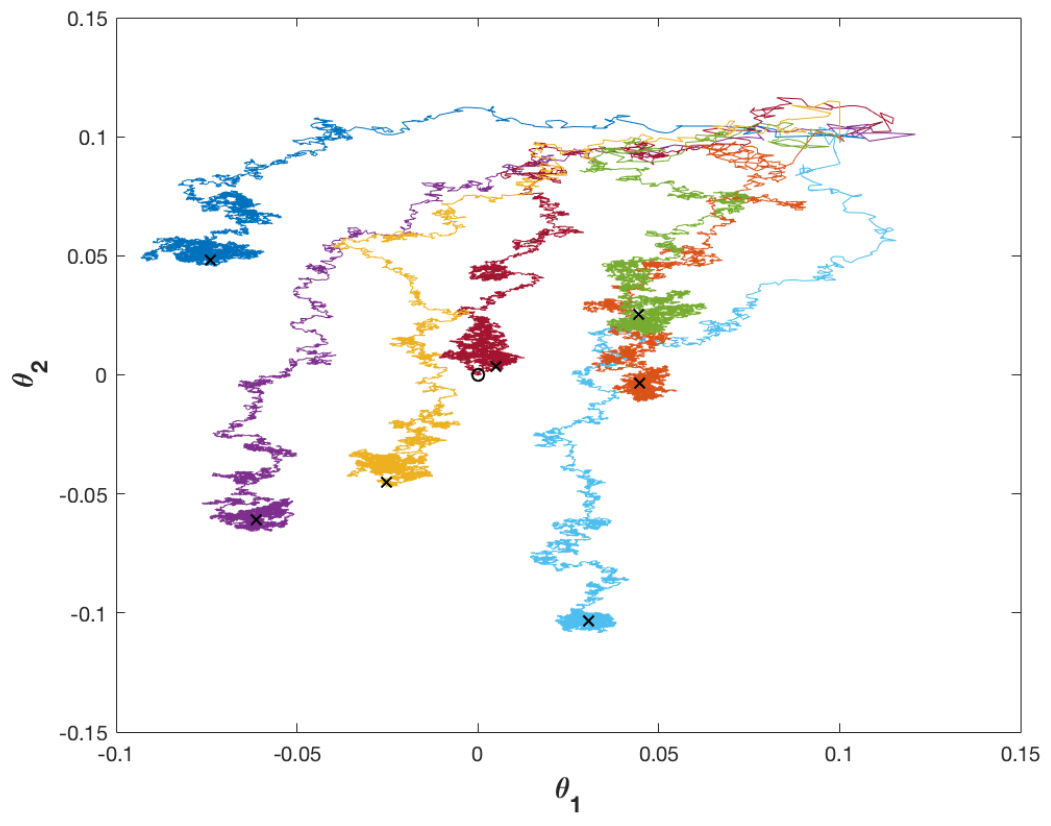} 
\includegraphics[width=0.5\textwidth]{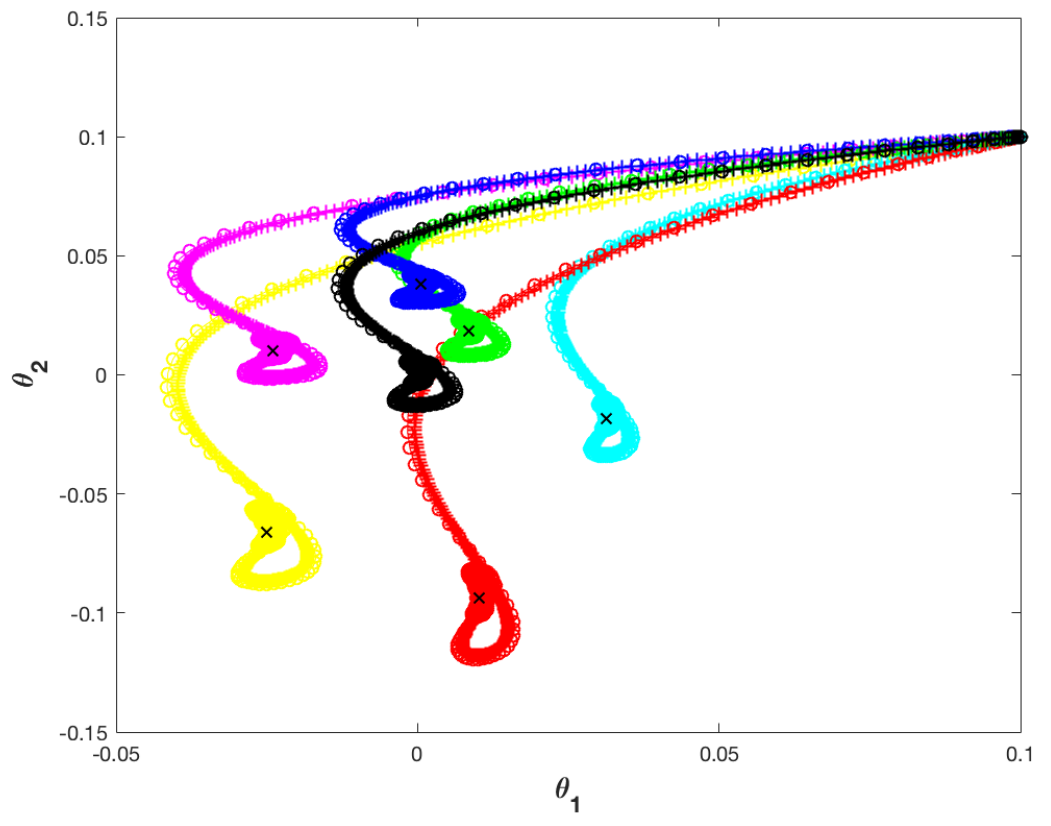}

\includegraphics[width=0.50\textwidth]{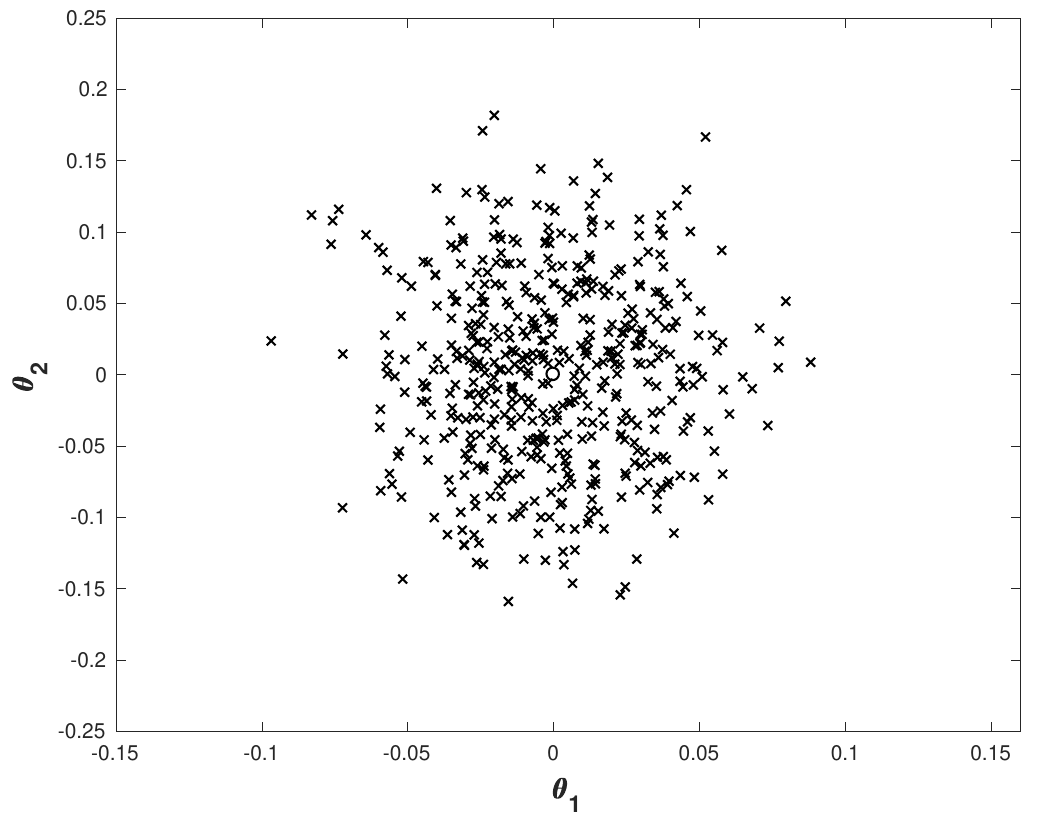}  %0.52
\includegraphics[width=0.485\textwidth]{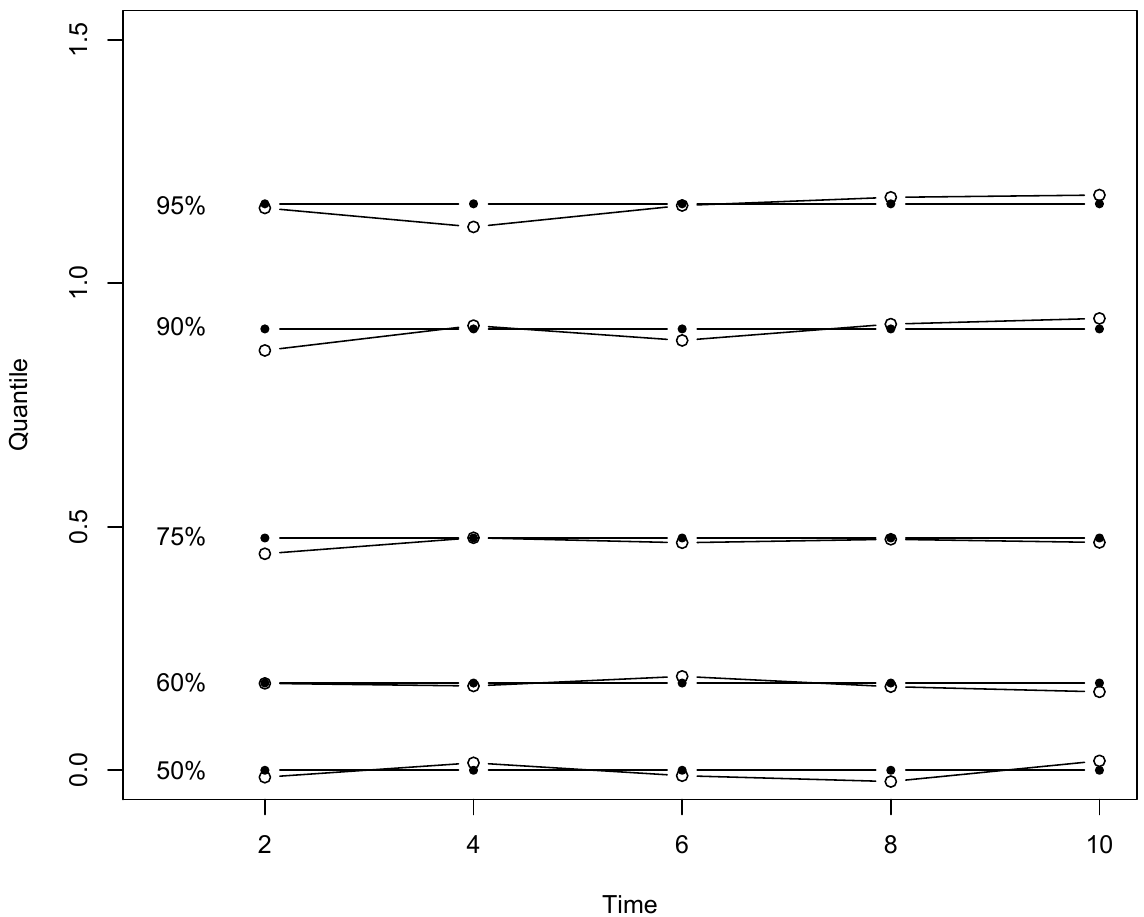}  %0.46 0.44

 \begin{singlespace}
\caption {The scatter plot of the estimator $\hat{\theta}_n$ and the plots of sample paths for accelerated and stochastic gradient descent algorithms and their corresponding ordinary or stochastic differential equations, where $\delta=0.05$, $n=1000$, $m=200$, initial values $x_0=x^n_0=x^m_0=(0.1, 0.1)^\prime$, and $\tau=1$. The top panels are sample paths of the gradient descent algorithms to compute 
$\hat{\theta}_n$, with the top left and right panels for the stochastic gradient descent case, and the accelerated gradient descent case based on all data, respectively. In both top panels, $\times$ denotes the estimator $\hat{\theta}_n$ which is the solution of the stochastic sample optimization (\ref{min-1}), with $\circ$ for the true parameter value $\check{\theta}=(0,0)$ which is the solution of the deterministic population  optimization (\ref{min-0}); 
in the top left panel, color curves~are different sample paths of stochastic gradient descent; and in the top right panel, color curves with $+$ and $\circ$ correspond to sample paths of the algorithm (\ref{min-Nest2}) and their corresponding differential 
equation (\ref{equ-3}), respectively, for accelerated gradient descent based on all data, with the black curve for sample paths of the algorithm (\ref{equ-Nest1}) and the ordinary differential equation (\ref{equ-2}). 
The bottom left panel is the scatter plot of $500$ simulated $\hat{\theta}_n$, and the bottom right panel is the marginal quantiles for $V(t)$ (the curve with $\bullet$) and the corresponding estimated quantiles for $V^m_\delta(t)$ (the curve with $\circ$)  based on $500$ repetitions.  }
\label{Figure1}
\end{singlespace}
\end{figure}

\section*{Acknowledgements}
The research of Yazhen Wang was supported in part by NSF grants DMS-1528375,  DMS-1707605, and DMS-1913149. 
The authors thank action editor Ryan Tibshirani, an associate editor and three referees for helpful comments and suggestions which led to 
improvements of the paper.

%% file: gbib-ML.tex
%\vspace{0.2in}
%
%{\bf Acknowledgements}: 
%The research of Yazhen Wang was supported in part by NSF grants DMS-105635, DMS-1265203 and DMS-1528375. 
%The authors thank the editor, associate
%editor, and two anonymous referees for comments and suggestions, which led to improvements of the paper.  
%
\newcommand{\refitem}{\vskip 1pt \par\sloppy\hangindent=1pc \hangafter=1  \noindent}
\renewcommand\baselinestretch{1.3}
%\tiny
\normalsize
\begin{center}
\section*{REFERENCES}
\end{center}

%\refitem
%Zeyuan Allen-Zhu and Lorenzo Orecchia. Linear coupling: An ultimate unification of gradient
%and mirror descent. ArXiv preprint arXiv:1407.1537, 2014.

\refitem
A. Ali, Z. Kolter, and R. Tibshirani. A continuous-time view of early stopping for least squares regression. International Conference on Artificial Intelligence and Statistics, 2019.

%\refitem
% Felipe Alvarez, J\'er\^ome Bolte, and Olivier Brahic. Hessian Riemannian gradient flows in convex
%programming. SIAM Journal on Control and Optimization, 43(2):477-501, 2004.

\refitem
L. Ambrosio, N. Gigli, and G. Savar\'e.  Gradient Flows: In Metric Spaces and in the Space of Probability Measures. 
Birkh\"auser. Second edition. 2008. 

\refitem
 Y. Arjevani, S. Shalev-Shwartz, and O. Shamir. On lower and upper bounds for
smooth and strongly convex optimization problems. ArXiv preprint arXiv:1503.06833, 2015.

%\refitem
% Hedy Attouch and Zaki Chbani. Fast inertial dynamics and FISTA algorithms in convex
%optimization: Perturbation aspects. ArXiv preprint arXiv:1507.01367, 2015.

\refitem
 H. Attouch, J. Peypouquet, and P. Redont. On the fast convergence of an inertial
gradient-like system with vanishing viscosity. ArXiv preprint arXiv:1507.04782, 2015.

%\refitem
 %Michel Baes. Estimate sequence methods: Extensions and approximations. Manuscript, available
%at http://www.optimization-online.org/DB_FILE/2009/08/2372.pdf, August 2009.

% \refitem 
%Arindam Banerjee, Srujana Merugu, Inderjit S. Dhillon, and Joydeep Ghosh. Clustering with Bregman
%divergences. J. Mach. Learn. Res., 6:1705-1749, December 2005.
 
% \refitem  
%Amir Beck and Marc Teboulle. Mirror descent and nonlinear projected subgradient methods for convex
%optimization. Oper. Res. Lett., 31(3):167-175, May 2003.

%\refitem
% Amir Beck and Marc Teboulle. A fast iterative shrinkage-thresholding algorithm for linear
%inverse problems. SIAM Journal on Imaging Sciences, 2(1):183-202, March 2009.
 
%\refitem  
%A. Ben-Tal and A. Nemirovski. Lectures on Modern Convex Optimization. SIAM, 2001.
 
%\refitem 
%Aharon Ben-Tal, Tamar Margalit, and Arkadi Nemirovski. The ordered subsets mirror descent optimization
%method with applications to tomography. SIAM J. on Optimization, 12(1):79-108, January 2001.
 
 \refitem
 P. Billingsley.  Convergence of Probability Measures. Wiley, 2nd Edition. 1999.
 
%\refitem  
%Anthony Bloch. Hamiltonian and Gradient Flows, Algorithms, and Control. American Mathematical
%Society, 1994.

%\refitem
%S. Becker, J. Bobin, and E. J. Cand\'es. NESTA: A fast and accurate first-order method for
%sparse recovery. SIAM Journal on Imaging Sciences, 4(1):1-39, 2011.

\refitem
P. J. Bickel, F. G\"otze,  and W. R. van Zwet. Resampling fewer than n observations: Gains, loses, and remedies for loses. 
Statistica Sinica 7, 1-31. 1997.

\refitem
S. Blanes, F.  Casas, J. A. Oteo, and J. Ros. The Magnus expansion and some of its applications. 
Physics Reports 470, 151-238, 2009.

 \refitem 
M. Bogdan, E. V. D. Berg, C. Sabatti, W. Su, and E. J. Cand\'es. SLOPE adaptive variable
selection via convex optimization. Annals of Applied Statistics, 9(3):1103-1140, 2015.

\refitem 
L. Bottou. Online learning and stochastic approximations. On-line learning in neural networks, 17(9):142, 1998.

 \refitem 
S. Boyd and L. Vandenberghe. Convex Optimization. Cambridge University Press, 2004.

 \refitem 
S. Boyd, N. Parikh, E. Chu, B. Peleato, and J. Eckstein. Distributed optimization and
statistical learning via the alternating direction method of multipliers. Foundations and
Trends in Machine Learning, 3(1):1-122, 2011.

 \refitem 
A. A. Brown and M. C. Bartholomew-Biggs. Some effective methods for unconstrained optimization
based on the solution of systems of ordinary differential equations. Journal of Optimization Theory and
Applications, 62(2):211-224, 1989.

\refitem
 S. Bubeck, Y. T. Lee, and M. Singh. A geometric alternative to Nesterov's
accelerated gradient descent. ArXiv preprint arXiv:1506.08187, 2015.

\refitem
 S. Bubeck and N. Cesa-Bianchi. Regret analysis of stochastic and nonstochastic multi-armed
bandit problems. Foundations and Trends in Machine Learning, 5(1):1-122, 2012.
 
 \refitem 
 J. C. Butcher. Numerical Methods for Ordinary Differential Equations. John Wiley. %\& Sons, Ltd, 
 2008.
 
% \refitem 
%Nicol\'o Cesa-Bianchi and G\'abor Lugosi. Prediction, Learning, and Games. Cambridge, 2006.

\refitem
V. Chandrasekarana and M. I. Jordan. 
Computational and statistical tradeoffs via convex relaxation.  PNAS 110, no. 13, E1181-E1190. 2012. 
%doi: 10.1073/pnas.1302293110

\refitem
C. Chen, D. Carlson, Z. Gan, C. Li, and L. Carin. 
Bridging the gap between stochastic gradient MCMC and stochastic optimization. arXiv:1512.07962v3, 2016.

 \refitem
C. Chen, N. Ding,  and L. Carin. 
On the convergence of stochastic gradient MCMC algorithms with high-order integrators.  arXiv:1610.06665v1, 2016.

 \refitem
 X. Chen, J. D. Lee, X. T. Tong, Y. Zhang.  Statistical inference for model parameters in stochastic gradient 
 descent.   arXiv:1610.08637, 2018.  
 
 \refitem
Y. S. Chow  and H. Teicher.   Probability Theory. Independence, Interchangeability, Martingales.   
  Third edition.   New York, Springer. 1997.
%\refitem
% Jorge Cort\'es. Finite-time convergent gradient flows with applications to network consensus.
%Automatica, 42(11):1993-2000, 2006.

\refitem
M.  Cs\"{o}rg\"{o}, L.  Horv\'{a}th, and P. Kokoszka.  Approximation for bootstrapped empirical processes. 
Proceedings of the American Mathematical Society 128, 2457-2464. 1999.

\refitem
S.  Cs\"{o}rg\"{o}, and D. M. Mason. Bootstrapping empirical functions. 
Ann. Statist. 17, 1447-1471. 1989.

\refitem
G. Da Prato and J. Zabczyk. Ergodicity for Infinite Dimensional Systems. Cambridge University Press, Cambridge; 1996.

\refitem
A. S. Dalalyan. Further and stronger analogy between sampling and optimization: Langevin Monte Carlo and gradient descent. 
Proceedings of Machine Learning Research,  65:1-12, 2017. 
%\refitem 
%Ofer Dekel, Ran Gilad-Bachrach, Ohad Shamir, and Lin Xiao. Optimal distributed online prediction. In
%Proceedings of the 28th International Conference on Machine Learning (ICML), June 2011.

%\refitem
%H.-B. Durr and C. Ebenbauer. On a class of smooth optimization algorithms with applications
%in control. Nonlinear Model Predictive Control, 4(1):291-298, 2012.

% \refitem 
%H.-B. Durr, E. Saka, and C. Ebenbauer. A smooth vector field for quadratic programming.
%In 51st IEEE Conference on Decision and Control, pages 2515-2520, 2012.

% \refitem 
%S. Fiori. Quasi-geodesic neural learning algorithms over the orthogonal group: A tutorial.
%Journal of Machine Learning Research, 6:743-781, 2005.

\refitem
J. Fan, W. Gong, C. J. Li, and Q. Sun. 
Statistical sparse online regression: A diffusion approximation perspective. Proceedings of the Twenty-First International 
Conference on Artificial Intelligence and Statistics, PMLR 84, 1017-1026 (2018). 

\refitem
 N. Flammarion and F. R. Bach. From averaging to acceleration, there is only a
step-size. In Proceedings of the 28th Conference on Learning Theory (COLT), 2015.

\refitem
D. J. Foster, A. Sekhari, O. Shamir, N. Srebro, K. Sridharan, B. Woodworth. 
The complexity of making the gradient small in stochastic convex optimization. 
Proceedings of Machine Learning Research 99, 1-27 (2019). 
%2019 32nd Annual Conference on Learning Theory

\refitem %Alan, Mark, Ravi
A. Frieze, M. Jerrum,  and R. Kannan. Learning linear transformations. FOCS 1996.

\refitem
%Gardiner, C.W. et al. (1985). Handbook of stochastic methods, volume 4. Springer Berlin. %Crispin 
C. W. Gardiner. Stochastic Methods: A Handbook for the Natural and Social Sciences. Springer, 4th edition. 2009.

 \refitem
R. Ge, F. Huang, C. Jin, and Y. Yuan. Escaping from saddle points - online stochastic gradient for tensor decomposition. In COLT, 2015.

\refitem
 S. Ghadimi and G. Lan. Accelerated gradient methods for nonconvex nonlinear
and stochastic programming. Mathematical Programming, 156(1):59-99, 2015.

\refitem
I. Goodfellow, Y. Bengio, and A. Courville.  Deep Learning. MIT press. 2016.

\refitem
%T. Hida, H. H. Kuo, J. Potthoff, and L. Streit. Whiote noise, An Infinite Dimensional Calculus. Kluwer Academic Publication. 1993.
T. Hida and S. Si. Lectures on White Noise Functionals. World Scientific. 2008.

\refitem
S. W.  He, J. G. Wang,   and J. A. Yan. Semimartingale Theory and Stochastic Calculus. Science Press and CRC Press. 1992.
%Inc., Beijing. %MR1219534

% \refitem 
%U. Helmke and J.B. Moore. Optimization and dynamical systems. Communications and control engineering series. Springer-Verlag, 1994.

% \refitem 
%U. Helmke and J. Moore. Optimization and dynamical systems. Proceedings of the IEEE, 84(6):907, 1996.

% \refitem 
%D. Hinton. Sturm's 1836 oscillation results evolution of the theory. In Sturm-Liouville theory, pages 1-27. Birkhauser, Basel, 2005.

\refitem
 C. Hu, J. T. Kwok, and W. Pan. Accelerated gradient methods for stochastic
optimization and online learning. In Advances in Neural Information Processing Systems %(NIPS) 
22, 2009.

 \refitem
 N. Ikeda and S. Watanabe. Stochastic Differential Equations and Diffusion Processes, Volume 24 (North-Holland Mathematical Library). 1981.
 
\refitem
J. Jacod  and A. Shiryaev. Limit Theorems for Stochastic Processes. Springer, 2nd edition, 2003. 
%JACOD, J. and SHIRYAEV, A. N. (2003). Limit Theorems for Stochastic Processes, 2nd ed. Springer, New York. MR1943877

%\refitem
% Shuiwang Ji, Liang Sun, Rong Jin, and Jieping Ye. Multi-label multiple kernel learning. In Advances in Neural Information Processing Systems (NIPS) 21, 2009.

\refitem
S. Jastrz\c{e}bski, Z. Kenton, D. Arpit, N. Ballas, A. Fischer, Y. Bengio, and A. Storkey. 
Three factors influencing minima in SGD.  arXiv:1711.04623v3. 2018. 

\refitem
 S. Ji and J.Ye. An accelerated gradient method for trace norm minimization. In
Proceedings of the 26th International Conference on Machine Learning (ICML), 2009.

\refitem
C. Jin, R. Ge, P. Netrapalli, S. M. Kakade, M. I. Jordan. How to escape saddle points efficiently.
 arXiv:1703.00887v1, 2017.

\refitem
 V. Jojic, S. Gould, and D. Koller. Accelerated dual decomposition for MAP
inference. In Proceedings of the 27th International Conference on Machine Learning (ICML),
2010.

\refitem
 A. Juditsky. Convex Optimization II: Algorithms. Lecture Notes, 2013.
 
 \refitem 
A. Juditsky, A. Nemirovski, and C. Tauvel. Solving variational inequalities with stochastic
mirror-prox algorithm. Stoch. Syst., 1(1):17-58, 2011.
 
 \refitem
K. Kawaguchi. Deep learning without poor local minima. In Advances In Neural Information Processing Systems, pages 586-594, 2016.
 
% \refitem   H.K. Khalil. Nonlinear Systems. Macmillan, 1992. % Pub. Co., 1992.

\refitem
N. S. Keskar, D. Mudigere, J. Nocedal, M. Smelyanskiy, and P. T. P. Tang.   
On Large-Batch Training for Deep Learning: Generalization Gap and Sharp Minima.  arXiv:1609.04836v2, 2017.   

\refitem
Y. Kifer. The exit problem for small random perturbations of dynamical systems with a hyperbolic fixed point. 
Israel Journal of Mathematics 40, 74-96 (1981). 

 \refitem 
 J. Kim and D. Pollard (1990). Cube root asymptotics. Ann. Statist.18, 191-219. 
  
\refitem
J. Koml\'{o}s, P. Major, and G. Tusn\'{a}dy.  An approximation of partial sums of independent
R.V.'s and the sample DF. I. Z. Wahrschein. Verw. Gebiete 32, 111-131. 1975. %MR 51:11605b

\refitem
J. Koml\'{o}s, P. Major, and G. Tusn\'{a}dy.  An approximation of partial sums of independent
R.V.'s and the sample DF. II. Z. Wahrschein. Verw. Gebiete 34, 33-58. 1976. %MR 53:6697
 
\refitem
 W. Krichene, A. Bayen, and P. Bartlett. Accelerated mirror descent in continuous
and discrete time. In Advances in Neural Information Processing Systems (NIPS) 29,
2015.
%W. Krichene,  A. M. Bayen, P. L. Bartlett (2015). Accelerated mirror descent in continuous and discrete time. Proceeding NIPS'15 Proceedings of the 28th International Conference on Neural Information Processing Systems Pages 2845-2853. 

 \refitem 
W. Krichene, S. Krichene, and A. Bayen. Efficient Bregman projections onto the simplex.
In 54th IEEE Conference on Decision and Control, 2015.

 \refitem 
H. Kushner and G. Yin. Stochastic Approximation and Recursive Algorithms and Applications. 
Springer, 2003. %www.springer.com. ISBN 9780387008943. Retrieved 2016-05-16.

\refitem
 G. Lan. An optimal method for stochastic composite optimization. Mathematical
Programming, 133(1-2):365-397, 2012.

\refitem
 G. Lan, Z. Lu, and R. Monteiro. Primal-dual first-order methods with
$O(1/\epsilon)$ iteration-complexity for cone programming. Mathematical Programming, 126(1):1-29,
2011.

%\refitem  J. J. Leader. Numerical Analysis and Scientific Computation. Pearson Addison Wesley, 2004.

 \refitem
J. D. Lee, M. Simchowitz, M. I Jordan, and B. Recht. Gradient descent only converges to minimizers. In Conference on Learning Theory, pages 1246-1257, 2016.

\refitem
 L. Lessard, B. Recht, and A. Packard. Analysis and design of optimization
algorithms via integral quadratic constraints. SIAM Journal on Optimization, 26(1):57-95, 2016.

\refitem
T. Li, L. Liu,  A. Kyrillidis, and C. Caramanis. Statistical Inference Using SGD. The Thirty-Second AAAI Conference on Artificial Intelligence (AAAI-18), 3571-3578, 2018. 

\refitem 
C. J. Li, Z. Wang, H. Liu. Online ICA: Understanding global dynamics of nonconvex optimization via 
diffusion processes. 30th Conference on Neural Information Processing Systems (NIPS 2016), Barcelona, Spain. 

\refitem 
C. J. Li, M. Wang, H. Liu, and T. Zhang. Diffusion approximations for online principal component estimation and global convergence.
31th Conference on Neural Information Processing Systems (NIPS 2017), Long Beach, CA, USA.  2017a. 

\refitem 
C. J. Li, L. Li, J. Qian, J. Liu. 
Batch Size Matters: A Diffusion Approximation Framework on Nonconvex Stochastic Gradient Descent. arXiv:1705.07562v2, 2017b. 

\refitem
 H.  Li and Z. Lin. Accelerated proximal gradient methods for nonconvex programming.
In Advances in Neural Information Processing Systems (NIPS) 28, 2015.

\refitem 
Q. Li, C. Tai, W. E.
Stochastic modified equations and adaptive stochastic gradient algorithms. arXiv:1511.06251v3,  2015. 

% \refitem A.M. Lyapunov. General Problem of the Stability Of Motion. Control Theory and Applications Series. Taylor \& Francis, 1992.

\refitem 
Y. A. Ma,  N. S. Chatterji, X. Cheng, N. Flammarion, P. L. Bartlett, c, and M. I. Jordan
Is there an analog of Nesterov acceleration for MCMC?  arXiv:1902.00996v1. 2019.

 \refitem
S. Mandt, M. D. Hoffman, and D. M. Blei. 
A variational analysis of stochastic gradient algorithms. arXiv:1602.02666v1,  2016.

\refitem
S. Mandt, M. D. Hoffman, and D. M. Blei. Stochastic gradient descent as approximate Bayesian inference. 
Journal of Machine Learning Research 18, 1-35, 2017. %arXiv:1704.04289v1,  2017. 

\refitem
P. Massart. 
Strong Approximation for Multivariate Empirical and Related Processes, Via KMT Constructions. Ann. Probab. 17, 266-291, 1989.

 \refitem
Q. Meng,  W. Chen, J. Yu,T. Wang, Z. Ma, T. Liu. 
Asynchronous Accelerated Stochastic Gradient Descent. 
Proceedings of the Twenty-Fifth International Joint Conference on Artificial Intelligence ((IJCAI-16). 2016.

%\refitem
% Indraneel Mukherjee, Kevin Canini, Rafael Frongillo, and Yoram Singer. Parallel boosting with momentum. In Machine Learning and Knowledge Discovery in Databases. Springer, 2013.

\refitem
R. Monteiro, C. Ortiz, and B. Svaiter. An adaptive accelerated first-order method for convex optimization. Technical report, ISyE, Gatech, 2012.  
 
\refitem  A. Nemirovskii and D. Yudin. Problem Complexity and Method Efficiency in  Optimization. John Wiley \& Sons, 1983.

\refitem
 Y. Nesterov. A method of solving a convex programming problem with convergence rate
$O(1/k^2)$. Soviet Mathematics Doklady, 27(2):372-376, 1983.

\refitem
 Y. Nesterov. Introductory Lectures on Convex Optimization: A Basic Course. Applied
Optimization. Kluwer, 2004. %Boston, 2004.  % volume 87. Springer Science \& Business Media, 2004.

\refitem
 Y. Nesterov. Smooth minimization of non-smooth functions. Mathematical Programming, 103(1):127-152, 2005.

\refitem
 Y. Nesterov. Accelerating the cubic regularization of Newton's method on convex problems.
Mathematical Programming, 112(1):159-181, 2008.

\refitem
 Y. Nesterov. Gradient methods for minimizing composite functions. Mathematical Programming, 140(1):125-161, 2013.

\refitem
 Y. Nesterov and Boris T. Polyak. Cubic regularization of Newton's method and its global
performance. Mathematical Programming, 108(1):177-205, 2006.

\refitem
J. Nocedal and S. Wright. Numerical Optimization. Springer Science \& Business Media,
2006.

\refitem
 B. O'Donoghue and E. Cand\'es. Adaptive restart for accelerated gradient
schemes. Foundations of Computational Mathematics, 15(3):715-732, 2015.

\refitem
B. O'Donoghue and E. J. Cand\'es. Adaptive restart for accelerated gradient schemes. Found.
Comput. Math., 2013.
 
% \refitem 
%S. Osher, F. Ruan, J. Xiong, Y. Yao, and W. Yin. Sparse recovery via differential inclusions. arXiv preprint arXiv:1406.7728, 2014.

 \refitem 
 D. Pollard. Empirical Processes: Theory and Applications. CMS. 1988. 
 
 \refitem 
B. T. Polyak. Introduction to optimization. Optimization Software New York, 1987.

 \refitem 
B. T. Polyak and A. B. Juditsky. Acceleration of stochastic approximation by averaging.  SIAM Journal on Control and Optimization 
30 (4): 838, 1992. % doi:10.1137/0330046.

% \refitem 
%Z. Qin and D. Goldfarb. Structured sparsity via alternating direction methods. Journal of Machine Learning Research, 13(1):1435-1468, 2012.

  \refitem 
M. Raginsky and J. Bouvrie. Continuous-time stochastic mirror descent on a network: Variance reduction,
consensus, convergence. In CDC 2012, pages 6793-6800, 2012.

%\refitem
% Garvesh Raskutti and Sayan Mukherjee. The information geometry of mirror descent. IEEE Transactions on Information Theory, 61(3):1451-1457, 2015.

\refitem 
A. Rakhlin, O. Shamir, and K. Sridharan. 
Making Gradient Descent Optimal for Strongly Convex Stochastic Optimization. arXiv:1109.5647v7, 2012. 

\refitem
E. Rio.
Strong Approximation for Set-Indexed Partial Sum Processes Via KMT Constructions I. Ann. Probab. 21, 759-790 (1993a).

\refitem
E. Rio. Strong Approximation for Set-Indexed Partial-Sum Processes, Via KMT Constructions II.  Ann. Probab.  21, 1706-1727 (1993b).  

\refitem
R. T. Rockafellar. Convex Analysis. Princeton Landmarks in Mathematics. Princeton
University Press, 1997. Reprint of the 1970 original.

\refitem
S. Ruder. An overview of gradient descent optimization algorithms.  arXiv:1609.04747v1, 2016.

 \refitem 
D. Ruppert. Efficient estimators from a slowly converging Robbins-Monro process. Technical report, 1988.

 \refitem 
A. P. Ruszczy\'nski. Nonlinear Optimization. Princeton University Press, 2006.

% \refitem 
%J. Schropp and I. Singer. A dynamical systems approach to constrained minimization. Numerical functional analysis and optimization, 21(3-4):537-551, 2000.

 \refitem 
 C. J. Shallue, J. Lee, J. Antognini, J. Sohl-Dickstein, R. Frostig, and G. E. Dahl. 
Measuring the effects of Data Parallelism on Neural Network Training. Journal of Machine Learning Research 20,  1-49 (2019). 

 \refitem 
N. Z. Shor. Minimization Methods for Non-Differentiable Functions. Springer Science \&
Business Media, 2012.

\refitem
J. Sirignano and K. Spiliopoulos. Stochastic gradient descent in continuous time.   
arXiv:1611.05545v3, 2017.

\refitem 
J. Sirignano and K. Spiliopoulos. Stochastic gradient descent in continuous time: A central limit theorem. 
arXiv:1710.04273v2, 2017.

\refitem 
I. Sutskever, J. Martens, G. Dahl, and G. Hinton. On the importance of initialization
and momentum in deep learning. In Proceedings of the 30th International Conference on
Machine Learning, pages 1139-1147, 2013.

\refitem
 W. Su, S. Boyd, and E. J. Cand\'es. A differential equation for modeling Nesterov's
accelerated gradient method: Theory and insights. In Advances in Neural Information
Processing Systems (NIPS) 27, 2014.

\refitem
%Su, W., Boyd, S., and Cand\'es, E. J. 
Weijie Su, Stephen Boyd, and Emmanuel J. Cand\'es. 
A differential equation for modeling Nesterov's accelerated gradient method: Theory and Insights.
Journal of Machine Learning Research 17, 1-43. 2016.

% \refitem 
%Gerald Teschl. Ordinary differential equations and dynamical systems, volume 140. American Mathematical Soc., 2012.

\refitem
 P. Tseng. On accelerated proximal gradient methods for convex-concave optimization. SIAM
Journal on Optimization, 2008.

 \refitem 
P. Tseng. Approximation accuracy, gradient methods, and error bound for structured
convex optimization. Mathematical Programming, 125(2):263-295, 2010.

\refitem
P. Toulis and E. M. Airoldi. Asymptotic and finite-sample properties of estimators based on stochastic gradients. Ann. Statist. 45(4), 1694-1727 (2017). 

\refitem
A. W. van der Vaart and J. Wellner.   Weak Convergence and Empirical Processes
With Applications to Statistics. Springer. 2000.

%\refitem  Cedric Villani. Optimal Transport, Old and New. Springer, 2008.

 \refitem 
G. N. Watson. A Treatise on the Theory of Bessel Functions. Cambridge Mathematical
Library. Cambridge University Press, 1995. Reprint of the second (1944) edition.

 \refitem
A. Wibisono, A. Wilson, and M. Jordan. A variational perspective on accelerated methods in optimization. 
Proc Natl Acad Sci  113,  E7351-E7358. 2016.  %Published online 2016 Nov 9. doi:  10.1073/pnas.1614734113

 \refitem % Zeyuan Allen-Zhu
 Z. A. Zhu. Katyusha: The first direct acceleration of stochastic gradient methods. Journal of Machine Learning Research 18,
 1-51 (2018). 
 
 \refitem 
 Z. Zhu, J. Wu, B. Yu, L. Wu, and J. Ma  The anisotropic noise in stochastic gradient descent: Its behavior of escaping from sharp minima and regularization effects. Xiv:1803.00195v5, 2019. 

%% file: p1-ML.tex
\section{Appendix: Proofs}
Denote by $C$ generic constant free of $(\delta, m,n)$ whose value may change from appearance to appearance. For simplicity we take initial values $x^n_0=x^m_0=x_0$. 
In appendix sections of theorem proofs, lemmas are established under the conditions and assumptions in corresponding theorems, and we often do not repeatedly list these 
conditions and assumptions in the lemmas.  
%In this appendix section we adopt notation conventions to facilitate our proofs and improve the presentation. 
To track processes under different circumstances and facilitate long technical arguments we adopt the following notations and conventions.

It is often necessary to put processes and random variables on some common probability spaces. At such occasions, we often automatically change probability spaces and consider versions of the processes and the random variables on new probability spaces, without altering notations. Because of this convention and Skorokhod's representation theorem, we often switch between ``convergence in probability'' and ``convergence in distribution.'' Also because of the convention, when no confusion occurs, we try to use the same notation for random variables or processes with identical distribution.

\textbf{Convention 1}. We reserve $x$'s and $y$'s for sequences generated from gradient descent algorithms and the corresponding empirical processes, $X$'s for solutions of 
ordinary differential equations (ODEs) and stochastic differential equations (SDEs). 

\textbf{Convention 2}. As described at the end of Section 1, for gradient descent algorithms to solve optimization (\ref{min-1}), we add 
superscripts 
$n$ and $m$ to notations for the associated processes and sequences based on all data in Section 3 and based on mini-batches %(or bootstrap samples) 
in Section 4, respectively, 
while notations without any superscript are for sequences and functions corresponding to the optimization (\ref{min-0}).

\textbf{Convention 3}. We reserve $V$'s for normalized solutions difference between differential equations associated with the optimization (\ref{min-0}) and the optimization (\ref{min-1}) under the cases for all data and mini-batches, %(bootstrap samples), 
while we reserve $V$ without any superscript as their corresponding weak convergence limits. 

\textbf{Convention 4}. As described at the end of Section 1, we add a superscript $*$ to notations $U$'s associated with mini-batches 
%(or bootstrap samples), 
and  as in Convention 2,  their corresponding process notations have a superscript $m$.

\textbf{Convention 5}. We denote by $| \Psi |$ the absolute value of scalar $\Psi$, the Euclidean norm of vector $\Psi$, or the spectral norm of matrix $\Psi$.

%\textbf{Convention 6}. If needed we may add an extra label $Q$ to the loss and objective functions $\ell(\theta; U_i)$ and $\cL(\theta; \bU_n)$, and write them as $\ell(\theta; U_i, Q)$ and $\cL(\theta; \bU_n, Q)$ so that $Q$ is clearly specified as the distribution of $U_i$. 
%%This will be particularly useful when we need to distinguish the cases with all data and with subsamples for stochastic gradient descent. % later.

%Our basic proof ideas are as follows. Each algorithm generates a sequence for computing a learning rule, a step-wise empirical process 
%is formed from the generated sequence, and a continuous process is obtained from the corresponding continuous-time differential equation. 
%We derive asymptotic distributions by analyzing the differential equations and bound the differences between the empirical processes and the corresponding continuous processes by studying the optimization problems and the empirical process theory along with the related differential equations. 

\subsection{Proofs of Theorem \ref{thm-1} }
We show that the solution of the linear differential equations \eqref{GD-limit-00} and \eqref{limit-00}  are Gaussian, 
assuming the existence and uniqueness. 
For the equation \eqref{GD-limit-00} its solution has an expression $V(t) = \Pi_0(t) \int_0^t [\Pi_0(s)]^{-1} \bZ(X(s)) ds$, where $\Pi_0(t)$ is a $p$ by $p$ deterministic matrix constructed by the Magnus expansion for solving the homogeneous linear differential equation  $ \dot{V}(t) + [\boldsymbol{I\!\! H}\! g(X(t))] V(t) =0$ (see Blanes et al. (2009)). Thus the limiting distribution of $V^n(t)$ is Gaussian. 
For the equation  \eqref{limit-00} in the accelerated case we may convert the second-order homogeneous linear differential equation  
$\ddot{V}(t) + \frac{3}{t}  \dot{V}(t) + [\boldsymbol{I\!\! H}\! g(X(t))] V(t) =0$ into an equivalent first-order homogeneous linear differential equation system 
\[  \left( \begin{array}{c} \dot{V}(t) \\ \dot{\Xi}(t) \end{array} \right) = \left[ \begin{matrix} 0 & 1 \\ - \boldsymbol{I\!\! H}\! g(X(t)) 
    &- \frac{3}{t}  \end{matrix} \right]    \left( \begin{array}{c} V(t) \\ \Xi(t) \end{array} \right),  \]
%$\dot{V}(t) - \Xi(t)=0$ and $\dot{\Xi}(t) + [\boldsymbol{I\!\! H}\! g(X(t))] V(t) + \frac{3}{t}  \Xi(t)  =0$, 
where $\Xi(t) = \dot{V}(t)$. 
Similar to the first order case, 
we apply the Magnus expansion to solve the first-order homogeneous linear differential equation system and then show that the solution of the differential equation \eqref{limit-00} linearly depends on $Z(\cdot)$. Therefore, the limiting 
distribution of $V^n(t)$ is also Gaussian. 
As a matter of fact, the theorem shows that in the special case of  $\bZ(\theta) = \sigma(\theta) \bZ$, we have $V(t)=\Pi(t) \bZ$ to clearly indicates the Gaussian limiting distribution. 
%and furthermore we can easily prove $\Pi(t) = \Pi_0(t) \int_0^t [\Pi_0(s)]^{-1} \bsigma(X(s)) ds$ in the first order case. 

%We provide arguments for the accelerated case only, as results for the plain case are relatively much easier to establish. 
Now we are ready to provide detailed arguments for the accelerated case, as results for the plain case are relatively easier to show and will be established later. From now on for simplicity we provide proof arguments only for the case of $\bZ(\theta) = \bsigma(\theta) \bZ$, as the proof for general $\bZ(\theta)$ is essentially the same. 

\subsubsection{Differential equation derivation} 
\label{section-Xconvergence}

With $\bU_n =(U_1,\cdots, U_n)^\tau$, let $R^n(\theta; \bU_n) = (R^n_1(\theta;\bU), \cdots, R^n_p(\theta;\bU_n))^\tau$, where 
\begin{equation*}
R_{j}^{n}(\theta;\bU_n)=\sqrt{n}\left[ \frac{1}{n}\sum_{i=1}^n  \frac{\partial}{\partial \theta_j} \ell(\theta; U_{i})-\frac{\partial}{\partial \theta_j}g(\theta)\right], \qquad j =1, \cdots, p.
\end{equation*}
Then 
\[ R^n(\theta; \bU_n) =\sqrt{n}\left[ \frac{1}{n}\sum_{i=1}^n  \nabla\! \ell(\theta; U_{i}) - \nabla\!g(\theta)\right]. \]

For the accelerated case, we can re-express ODE (\ref{equ-3}) as 
\begin{equation} \label{equ-4}
 \ddot{X}^n(t)+\frac{3}{t}\dot{X}^n(t) +  \nabla \!g(X^n(t)) + \frac{1}{\sqrt{n}} R^n(X^n(t);\bU_n) =0.
\end{equation} 

%As $\nabla \ell(\theta; u)$ form a Donsker class, we have as $n \rightarrow \infty$, 
%$\sup_{\theta} |n^{-1/2} R^n(\theta; \bU_n)| \rightarrow 0$, 
%and Donsker theorem shows that $R^{n}(\theta; \bU_n) \rightarrow_d \bsigma(\theta) \bZ$ uniformly over $\theta$, 

By Lemma \ref{lem2} below we obtain that  $X^n(t)$ converges in probability to $X(t)$ uniformly over any finite interval. Thus, for large $n$, $X^n(t)$ falls into $\Theta_X$, and Assumption A4 implies that 
 as $n \rightarrow \infty$, $R^n(X^n(t); \bU_n) = O_P(1)$, and  $n^{-1/2} R^n(X^n(t); \bU_n)| \rightarrow 0$.  Hence, 
 ODEs  (\ref{equ-3}) and (\ref{equ-4}) both converge to ODE (\ref{equ-2}). %and $X^n(t)$ converges to $X(t)$ in probability. 

 From Assumption A4 we have that $R^n(\theta; \bU_n)$  converges in distribution to $\bsigma(\theta) \bZ$ uniformly over $\theta \in \Theta_X$, and the generalization of Skorohod's representation theorem in Lemma \ref{Skorohod} below  shows that 
 %(and its construction proof in Billingsley (1999)), we have $R^n(\theta; \bU_n^\dagger) - \bsigma(\theta) \bZ_\dagger = o(1)$ uniformly over $\theta \in \Theta_X$, where $\bU_n^\dagger$ and $\bZ_\dagger$  are defined on some common probability spaces with $\bZ_\dagger \sim N_p(0,\bI_p)$ and $\bU_n^\dagger$ identically distributed as $\bU_n$. 
there exist $\bU_n^\dagger$ and $\bZ_\dagger$ defined on some common probability spaces with 
$\bZ_\dagger \sim N_p(0,\bI_p)$ and $\bU_n^\dagger$ identically distributed as $\bU_n$ such that as $n \rightarrow \infty$,
$R^n(\theta; \bU_n^\dagger) - \bsigma(\theta) \bZ_\dagger = o(1)$ uniformly over $\theta \in \Theta_X$. 
Thus we have that the solution $X^n(t)$ of equations (\ref{equ-3}) is identically distributed as the solution $X^n_\dagger(t)$ of 
\[  \ddot{X}^n_\dagger(t)+\frac{3}{t}\dot{X}^n_\dagger(t) +  \nabla \!g(X^n_\dagger(t)) + \frac{1}{\sqrt{n}} R^n(X^n_\dagger(t);\bU_n^\dagger) =0, \]
which in turn may be written as 

\begin{equation}\label{equ-5}
  \ddot{X}^n_\dagger(t)+\frac{3}{t}\dot{X}^n_\dagger(t) +  \nabla \!g(X^n_\dagger(t)) + \frac{1}{\sqrt{n}} \bsigma(X^n_\dagger(t)) \bZ_\dagger + o\left( n^{-1/2} \right)  =0. 
\end{equation}
In particular  (\ref{equ-5}) is equivalent to (\ref{equ-3-1}) up to the order of $n^{-1/2}$,  which implies that as $n \rightarrow \infty$, ODEs  (\ref{equ-3}), (\ref{equ-3-1}), and (\ref{equ-5}) all converge to ODE (\ref{equ-2}), and $X^n_\dagger(t)$ almost surely converges to $X(t)$. 
Since the solutions of equations (\ref{equ-3}), (\ref{equ-3-1})  and (\ref{equ-5}) are defined in the distribution sense, when there is no confusion, with a little abuse of notations we may drop index $\dagger$ %and $n$ 
and write equation (\ref{equ-5}) as
\begin{equation}\label{equ-6}
  \ddot{X}^n(t)+\frac{3}{t}\dot{X}^n(t) +  \nabla \!g(X^n(t)) + \frac{1}{\sqrt{n}} \bsigma(X^n(t)) \bZ + o\left( n^{-1/2} \right)  =0,
\end{equation}
where $\bZ$ is a Gaussian random vector with distribution $N_p(0, \bI_p)$, and initial conditions $X^n(0)=x_0$ and $\dot{X}^n(0)=0$.

The arguments for establishing Theorem 1 in Su et al. (2016) can be directly applied to establish the existence and uniqueness of the solution $X^n(t)$ to (\ref{equ-4}) for each $n$. We can employ the same arguments with $\nabla g(\cdot)$ replaced by $\boldsymbol{I\!\! H}\! g(X(t)) \,\Pi(t) + \bsigma(X(t))$ or  $\boldsymbol{I\!\! H}\! g(X(t)) \,V(t) + \bsigma(X(t)) \bZ$ 
to show that linear differential equations (\ref{limit-0}) and (\ref{limit-00})  have unique solutions. 

For the plain gradient descent case, Lemma \ref{lem-GD1} below shows that  $X^n(t)$ converges to $X(t)$ uniformly over any finite interval. Similarly we  
can establish that ODE (\ref{GD-c2}) is asymptotically equivalent to ODE (\ref{GD-c2-1}), and the standard ODE theory shows that they have unique solutions.

Now we give a generalization of Skorohod's representation theorem.
 Assumption A4 indicates $R^n(\theta; \bU_n)$ converges in distribution to $\bZ(\theta)$, and Skorohod's representation theorem  
 allows to realize $R^n(\theta; \bU_n)$ and $\bZ(\theta)$ on some common probability spaces with almost sure convergence. 
 The following lemma generalizes Skorohod's representation theorem to have a joint representation of $\bU_n$ and $R^n(\theta; \bU_n)$
along with almost sure convergence for $R^n(\theta; \bU_n)$ and $\bZ(\theta)$.
  \begin{lem} \label{Skorohod}
There exist $\bU_n^\dagger$ and $\bZ_\dagger(\theta)$ defined on some common probability spaces with 
$\bZ_\dagger(\theta)$ and $\bU_n^\dagger$ identically distributed as $\bZ(\theta)$ and $\bU_n$, respectively,  such that 
as $n \rightarrow \infty$, $R^n(\theta; \bU_n^\dagger) - \bZ_\dagger(\theta) = o(1)$ uniformly over $\theta \in \Theta_X$.  
 \end{lem}
 Proof. Our proof argument follows the construction proof of Skorohod's representation 
 theorem in  Billingsley (1999, Theorem 6.7), with some delicate modifications involving 
 joint distribution of $\bU_n$ and $R^n(\theta; \bU_n)$ as well as its associated conditional distributions. 
 
 Let $\Psi_\theta$ be the normal distribution of $\bZ(\theta)$.  Assume random variables $\bU_n$ are defined on 
 probability space $(\Omega, \cF, I\!\!P)$. Denote by $\Psi_{\theta,n}$ the joint distribution of $\bU_n$ 
 and $R^n(\theta; \bU_n)$, and by $\Psi_{\theta,n, U}$ and $\Psi_{\theta,n, R}$ the marginal distributions of $\bU_n$ and 
 $R^n(\theta; \bU_n)$, respectively. 
 Let $\Xi_0(\theta), \cdots, \Xi_k(\theta)$ be the partition of $I\!\!R^p$ (the support of normal distribution 
 $\Psi_\theta$) such that (i) $\Psi_\theta(\Xi_0(\theta)) < \epsilon$, (ii) the boundaries of $\Xi_0(\theta), \cdots, \Xi_k(\theta)$ have probability zero under $\Psi_\theta$, and (iii) the diameters of $\Xi_1(\theta), \cdots, \Xi_k(\theta)$ are bounded by $\epsilon$. Here we use notation $\Xi_i(\theta)$ to 
 indicate the possible dependence of the partitions on $\theta$. 
 For $r=1, 2, \cdots$,  we take $\epsilon_r = 2^{-r}$ and obtain partition $\Xi^r_0(\theta), \cdots, \Xi^r_{k_r}(\theta)$.  
 Assumption A4 indicates that  $R^n(\theta; \bU_n)$  converges in distribution to $\bZ(\theta)$ uniformly over  
 $\theta \in \Theta_X$, which implies that for each $r$ there exists an integer $n^*_r$ (free of $\theta$) such that for $n \geq n^*_r$, 
 \[   \Psi_{\theta, n, R}\left(\Xi^r_i(\theta) \right) \geq (1 - \epsilon_r) \Psi_\theta\left( \Xi^r_i(\theta) \right), \qquad i=1, \cdots, k_r, \; \theta \in \Theta_X. \]
 %Now for each $n^*_r \leq n < n^*_{r+1}$ and for each $\Xi^r_i(\theta)$ with non-zero probability under $P_\theta$, we take 
 %$\bU^\dagger_{n,i}$ to be independent random variable on $(\Omega_\dagger, \cF_\dagger, I\!\!P_\dagger)$ such that 
 %$R^n(\theta; \bU^\dagger_{n,i})$ follows distribution $P_{\theta,n}(\cdot | \Xi^r_i(\theta))$ (the restriction of $P_{\theta, n}$ 
 %on $\Xi^r_i(\theta)$), and 
 %$R^n(\theta; \cdot)$ as a function of the second argument may not be one to one, there are many choices for the values of $\bU^\dagger_{n,i}$. 
% With $\bU_n$ on probability space $(\Omega, I\!\!P)$, we can select $\bU_{n,i}^\dagger(\omega_\dagger)$ to take values in $\{\bU_n(\omega); R^n(\theta; \bU_n(\omega)) \in \Xi^r_i(\theta) \}$ so that 
 %$$ I\!\!P_\dagger\left[ \bU^\dagger_{n,i}(\omega_\dagger) \leq \bu, R^n\left(\theta; \bU^\dagger_{n,i}(\omega_\dagger) \right) \in \Xi^r_i(\theta)  \right] = I\!\!P\left[ \bU_n (\omega) \leq \bu, R^n\left(\theta; \bU_{n}(\omega) \right) \in \Xi^r_i(\theta)  \right],  $$
% where $\bU_n$ are defined on probability space $(\Omega, \cF, I\!\!P)$.   
 
 As in Billingsley (1999, Theorem 6.7), we can always find a probability space to support a random element with any given 
 distribution, and by passing to the appropriate large or infinite product space we can show that there exists a probability space  
 $(\Omega_\dagger, \cF_\dagger, I\!\!P_\dagger)$ to support random variables $\xi$, $\bZ_\dagger(\theta)$, $\check{\bU}_n$, and 
 $\bU^\dagger_{n,i}$, and $\bLambda_n$, $n, i \geq 1$, all independent of each other, with the following four properties. 
 
  (i) $\xi$  follows a uniform distribution on $[0,1]$. 
  
  (ii) $\bZ_\dagger(\theta)$ follows a normal distribution $\Psi_\theta$, and  $\check{\bU}_n$ has distribution $\Psi_{\theta,n,U}$. 
  
  (iii) For each $n^*_r \leq n < n^*_{r+1}$ and for each $\Xi^r_i(\theta)$ with non-zero probability under $\Psi_\theta$, we take 
 $\bU^\dagger_{n,i}$ to be independent random variable on $(\Omega_\dagger, \cF_\dagger, I\!\!P_\dagger)$ such that 
 $\bU^\dagger_{n,i}$ and $R^n(\theta; \bU^\dagger_{n,i})$ jointly follow distribution $\Psi_{\theta,n}(\cdot |  \Xi^r_i(\theta) )$, 
 which denotes the joint conditional distribution of $\bU_n(\omega)$ and $R^n\left(\theta; \bU_{n}(\omega) \right)$ given 
 $R^n\left(\theta; \bU_{n}(\omega) \right) \in   \Xi^r_i(\theta) $ (the restriction of the joint distribution $\Psi_{\theta, n}$ on 
 the set  $\{\bu, R^n\left(\theta; \bu \right) \in   \Xi^r_i(\theta)\}$). That is,  for any Borel sets $A _1 \subset  I\!\!R^m$ and $A_2 \subset I\!\!R^p$, 
\begin{eqnarray*}
&& I\!\!P_\dagger\left[ \bU^\dagger_{n,i}(\omega_\dagger) \in A_1, R^n\left(\theta; \bU^\dagger_{n,i}(\omega_\dagger) \right) \in A_2  \right] \\
&& =\left. I\!\!P_\dagger \left[ \check{\bU}_n (\omega_\dagger) \in A_1, R^n\left(\theta; \check{\bU}_{n}(\omega_\dagger) \right) \in A_2 
 \right|  R^n\left(\theta; \check{\bU}_{n}(\omega_\dagger) \right) \in \Xi^r_i(\theta)  \right] \\
&& = \left. I\!\!P \left[ \bU_n (\omega) \in A_1, R^n\left(\theta; \bU_{n}(\omega) \right) \in A_2 
 \right|  R^n\left(\theta; \bU_{n}(\omega) \right) \in \Xi^r_i(\theta)  \right] \\
&& = \Psi_{\theta,n}(A_1 \times A_2 | %R^n\left(\theta; \bU_{n}(\omega) \right) \in 
  \Xi^r_i(\theta) ). 
 \end{eqnarray*}
 Taking $A_1 = I\!\!R^m$ in above equality we obtain the marginal result for $R^n(\theta; \bU^\dagger_{n,i})$, 
 \begin{eqnarray*}
 && I\!\!P_\dagger\left[ R^n\left(\theta; \bU^\dagger_{n,i}(\omega_\dagger) \right) \in A_2  \right] = \left.  I\!\!P_\dagger\left[ R^n\left(\theta; \check{\bU}_{n}(\omega_\dagger) \right) \in A _2\right|  R^n\left(\theta; \check{\bU}_{n}(\omega_\dagger) \right) \in \Xi^r_i(\theta)  \right] \\
 && = \left.  I\!\!P \left[ R^n\left(\theta; \bU_{n}(\omega) \right) \in A_2 \right|  R^n\left(\theta; \bU_{n}(\omega) \right) \in \Xi^r_i(\theta)  \right]  = \Psi_{\theta, n, R} \left( A _2|  \Xi^r_i(\theta)  \right). 
 \end{eqnarray*}
(iv) For each $n^*_r \leq n < n^*_{r+1}$, the distribution of $\bLambda_n$ is given by  
\begin{eqnarray*}
\nu_n (A) = \epsilon_r^{-1} \sum_{i=0}^{k_r} \Psi_{\theta, n} (A \times I\!\!R^p  | \Xi^r_i(\theta)) \left[ \Psi_{\theta, n, R} (\Xi^r_i(\theta)) - 
    (1- \epsilon_r) \Psi_\theta(\Xi^r_i(\theta)) \right]. 
\end{eqnarray*}
Now we define $\bU_{n}^\dagger$ on $(\Omega_\dagger, \cF_\dagger, I\!\!P_\dagger)$ as follows. 
For $n<n^*_1$, take $\bU_{n,\dagger} = \check{\bU}_n$. For each $n^*_r \leq n< n^*_{r+1}$, define 
\[ \bU_{n}^\dagger = \sum_{i=0}^{k_r} \bU^\dagger_{n,i} 1\{ \xi \leq 1 - \epsilon_r, \bZ_\dagger(\theta) \in  \Xi^r_i(\theta)\} + \bLambda_n
1\{\xi >  1- \epsilon_r\}. \]
We derive the distribution of $\bU_{n}^\dagger$. For any Borel set $A_1 \subset I\!\!R^m$,
\begin{eqnarray*}
&& I\!\!P_\dagger( \bU_{n}^\dagger  \in A_1) = \sum_{i=0}^{k_r} I\!\!P_\dagger \left[ \bU^\dagger_{n,i} \in A_1, \bZ_\dagger(\theta) \in  \Xi^r_i(\theta), \xi \leq 1 - \epsilon_r \right]   \\
&& +  I\!\!P_\dagger( \bLambda_n \in A_1, \xi >  1- \epsilon_r) \\
\end{eqnarray*}
\begin{eqnarray*}
&& = (1- \epsilon_r) \sum_{i=0}^{k_r} I\!\!P_\dagger \left[ \bU^\dagger_{n,i} \in A_1 \right]   I\!\!P_\dagger \left[  \bZ_\dagger(\theta) \in  \Xi^r_i(\theta)\right] + \epsilon_r \nu_n(A_1) \\
&&=(1- \epsilon_r) \sum_{i=0}^{k_r} \Psi_{\theta,n}(A_1 \times I\!\!R^p |  \Xi^r_i(\theta) )  \Psi_\theta(  \Xi^r_i(\theta) )  + \epsilon_r \nu_n(A_1) \\
&& = \sum_{i=0}^{k_r} \Psi_{\theta, n} (A \times I\!\!R^p  | \Xi^r_i(\theta))  \Psi_{\theta, n, R} (\Xi^r_i(\theta)) \\
&&  = \sum_{i=0}^{k_r} I\!\!P \left[ \bU_{n} \in A_1, R^n\left(\theta; \bU_{n}(\omega) \right) \in \Xi^r_i(\theta)  \right]   \\
&& = I\!\!P [ \bU_{n} \in A_1] = \Psi_{\theta, n, U}(A_1),
\end{eqnarray*}
where the fifth equality is due to the definition of distribution $\nu_n$ (which is a reverse construction). 
That is, we have shown that $\bU_{n}^\dagger$ is identically distributed as $\bU_n$ for all $n$. 
Let $\Omega_{r,\dagger}=\{ \xi \leq 1 - \epsilon_r, \bZ_\dagger(\theta) \not \in \Xi^r_0(\theta) \}$ and $\Omega_{\dagger}^*= \liminf_{r \rightarrow \infty} \Omega_{r,\dagger}$. Then 
$I\!\!P_\dagger (\Omega_{r,\dagger}^*)  > 1 - 2 \epsilon_r$, and an application of the Borel-Cantelli lemma  leads to 
$I\!\!P_\dagger ( \Omega_\dagger^*)=1$. For $n^*_r \leq n < n^*_{r+1}$, on set $\Omega_{r,\dagger}$, 
$R^n(\theta; \bU^\dagger_{n}) $ and $\bZ_\dagger(\theta)$ fall into the same set $\Xi^r_i(\theta)$, whose diameter is less than $\epsilon_r$. Thus, on $\Omega_\dagger^*$, $R^n(\theta; \bU^\dagger_{n})$ a.s. converges to $\bZ_\dagger(\theta)$ uniformly over $\theta \in \Theta_X$.
%is reversely constructed so that $\bU_\dagger$ is identically distributed as $\bU_n$ as in Billingsley (1999).  
% Note that by iteratively defining random variables which are independent of everything previously defined, our resulting probability 
% space will be a large product space.
% This can be reverse reconstructed very easily, but requires more notation  than is ideal for this medium. 

\subsubsection{Weak convergence and tightness}
\label{section-tight}

To prove the weak convergence of $V_n(t)$ to $V(t)$, we need to establish the usual finite-dimensional convergence plus uniform 
tightness (or stochastic equicontinuity) (see Kim and Pollard (1990, Theorem 2.3), Pollard (1988) and Van der Vaart and Wellner (2000)).  
We establish finite-dimensional convergence below.  

For the accelerated case, taking a difference between ODEs (\ref{equ-2}) and (\ref{equ-5}), we have 
\begin{align*}
& [\ddot{X}^n_\dagger(t) - \ddot{X}(t)] + \frac{3}{t} [ \dot{X}^n_\dagger(t) - \dot{X}(t)] + \nabla \![g(X^n_\dagger(t)) - g(X(t))] + \frac{1}{\sqrt{n}} \bsigma(X^n_\dagger(t)) 
   \bZ_\dagger  = o\left( n^{-1/2} \right) . &
%& \ddot{V}^n(t) + \frac{3}{t}  \dot{V}^n(t) + \boldsymbol{I\!\! H}\! g(X(t)) V^n(t) + \bsigma(X^n(t)) \bZ = o(1), &
\end{align*}
Let $V^n_\dagger(t) = \sqrt{n} [ X^n_\dagger(t) - X(t) ]$. As $n\rightarrow \infty$, $X^n_\dagger(t)  \rightarrow_{a.s.}  X(t)$, 
\( \bsigma(X^n_\dagger(t)) = \bsigma(X(t)) + o(1)\),  and 
\(  \nabla \![g(X^n_\dagger(t)) - g(X(t))] = \nabla^2 \!g(X(t)) [ X^n_\dagger(t) - X(t)] + o(X^n_\dagger(t) - X(t)), \)
thus $V^n_\dagger(t)$ satisfies 

\begin{align*}
%& [\ddot{X}^n_\dagger(t) - \ddot{X}(t)] + \frac{3}{t} [ \dot{X}^n_\dagger(t) - \dot{X}(t)] + \nabla \![g(X^n_\dagger(t)) - g(X(t))] + \frac{1}{\sqrt{n}} \bsigma(X^n_\dagger(t)) \bZ  = o\left( n^{-1/2} \right) . &
& \ddot{V}^n_\dagger(t) + \frac{3}{t}  \dot{V}^n_\dagger(t) + \boldsymbol{I\!\! H}\! g(X(t)) V^n_\dagger(t) + \bsigma(X(t)) \bZ_\dagger= o(1). &
\end{align*}
As $n \rightarrow \infty$, $V^n_\dagger(t)$ almost surely converge to the unique solution $V_\dagger(t)$ of the following linear differential equation,
%\begin{equation} \label{equ-7}
\[ \ddot{V}_\dagger(t) + \frac{3}{t}  \dot{V}_\dagger(t) + [\boldsymbol{I\!\! H}\! g(X(t))] V_\dagger(t) + \bsigma(X(t)) \bZ_\dagger = 0, \]
%\end{equation}
where $X(t)$ is the solution of equation (\ref{equ-2}), random variable $\bZ_\dagger \sim N_p(0, \bI_p)$, and initial conditions $V_\dagger(0) = \dot{V}_\dagger(0)=0$. As $V(t)$ and $V_\dagger(t)$ are governed by the equations with the same form but identically distributed random  
coefficients $\bZ$ and $\bZ_\dagger$, we easily see that $V(t)$ and $V_\dagger(t)$ are identically distributed. 

The almost sure convergence of $V^n_\dagger(t)$ to $V_\dagger(t)$ implies the joint convergence of $(V^n_\dagger(t_1), \cdots, V^n_\dagger(t_k))$ to 
$(V_\dagger(t_1), \cdots, V_\dagger(t_k))$ for any integer $k$ and any $t_1, \cdots, t_k \in I\!\!R_+$. From the identical distributions of $X^n(t)$ with 
$X^n_\dagger(t)$, $V^n(t)$ with $V^n_\dagger(t)$, and $V(t)$ with $V_\dagger(t)$ we immediately conclude that 
%$(V^n_\dagger(t_1), \cdots, V^n_\dagger(t_k))$ converges in distribution to $(V_\dagger(t_1), \cdots, V_\dagger(t_k))$. 
$(V^n(t_1), \cdots, V^n(t_k))$ converges in distribution to $(V(t_1), \cdots, V(t_k))$. 
This establishes finite-dimensional distribution convergence of $V^n(t)$ to $V(t)$.

For the plain gradient descent case, an application of the similar argument to ODEs (\ref{GD-c1}) and (\ref{GD-c2-1}) can establish finite-dimensional 
convergence. 

Now we show tightness of $V_n(t)$. To establish the tightness of $V_n(t)$ on $[0, T]$,
we need to show that for any $\varepsilon>0$, and $\eta>0$, %and $T< \infty$, 
there exists a positive constant $\delta$ such that 
\begin{equation} \label{equ-7}
  \limsup_{n \rightarrow \infty} P\left[ \sup_{(t_1, t_2) \in \cT(T,\delta)}  | V_n(t_1) - V_n(t_2)| >\eta\right] < \varepsilon, 
\end{equation}
where 
$\cT(T,\delta)=\{(t_1, t_2), t_1, t_2 \in I\!\!R_+, \max(t_1, t_2) \leq T, |t_1- t_2| <\delta\}$.
The tightness of $V_n(t)$ on $I\!\!R_+$ requires above result for any $T<\infty$.
%the supremum runs over all pairs of $t_1, t_2 \in I\!\!R_+$ with $\max(t_1, t_2) \leq T$ and $|t_1- t_2| <\delta$.

Note that as (\ref{equ-7}) requires only some probability evaluation,  with the abuse of notations we have dropped index $\dagger$ and work on equation (\ref{equ-6}).

\subsubsection{Weak convergence proof for the plain gradient descent case}

%The same perturbation argument in Section \ref{section-tight} can be used to show finite-dimensional distribution convergence of $V^n(t)$ to $V(t)$ for simple 
%ODE (\ref{GD-c2}) in the plain gradient descent case. 
%We will establish tightness for $V^n(t)$. 
%and thus prove the weak convergence of  $V^n(t)$ to $V(t)$ in the plain gradient descent case. 

\begin{lem} \label{lem-GD1}
For any given $T>0$, we have 
\[ \max_{t \in [0, T]} |X^n(t) - X(t)| = O_P( n^{-1/2}).  \]
\end{lem}
Proof. From ODEs (\ref{GD-c1}) and (\ref{GD-c2}) we have 
\[ \dot{X}^n(t) - \dot{X}(t) = - [ \nabla g(X^n(t)) - \nabla g(X(t)) ] - n^{-1/2} R^n(X^n(t); \bU_n) , \]
and using assumption A1-A2 we obtain 
\begin{align*}
& | \nabla g(X^n(t)) - \nabla g(X(t)) | \leq L |X^n(t) - X(t)|, &\\
& n^{-1/2} |R^n(X^n(t); \bU_n) - R^n(X(t); \bU_n) | \leq \left( n^{-1} \sum_{i=1}^n h_1(U_i) + L\right) |X^n(t) - X(t)|. &
\end{align*}
Combining them together we arrive at  
\begin{align*}
|X^n(t) - X(t)| \leq n^{-1/2} \int^t_0 |R^n(X(s); \bU_n )|ds + \left( n^{-1} \sum_{i=1}^n h_1(U_i) + 2 L\right) \int_0^t |X^n(s) - X(s)| ds, 
\end{align*}
and an application of Gronwall's inequality leads to 
\begin{align*}
& |X^n(t) - X(t)| \leq n^{-1/2} \int^t_0 |R^n(X(s); \bU_n)| ds &\\ 
& + n^{-1/2}
\left( n^{-1} \sum_{i=1}^n h_1(U_i) + 2 L\right) \int_0^t  e^{ \left( n^{-1} \sum_{i=1}^n h_1(U_i) + 2 L\right) u   } du \int_0^u |R^n(X(s); \bU_n)|ds, 
\end{align*}
which implies 
\begin{align*}
& \max_{t \in [0, T]} |X^n(t) - X(t)| \leq n^{-1/2} \int^T_0 |R^n(X(s); \bU_n)|ds &\\ 
& + n^{-1/2}
\left( n^{-1} \sum_{i=1}^n h_1(U_i) + 2 L\right) \int_0^T  e^{ \left( n^{-1} \sum_{i=1}^n h_1(U_i) + 2 L\right) u  } du \int_0^u |R^n(X(s); \bU_n)|ds.
\end{align*}
Since assumptions A3 and A4 indicate $ \sup_t |R^n(X(t); \bU_n)| \sim \sup_t |\bsigma(X(t))  \bZ | = O_P(1)$, and 
$n^{-1} \sum_{i=1}^n h_1(U_i)$ converges in probability to $E[h_1(U)]<\infty$, the above inequality shows 
$\max_{t \in [0, T]} |X^n(t) - X(t)| = O_P( n^{-1/2})$.

\begin{lem} \label{lem-GD2}
For any given $T>0$, $V^n(t)$ is stochastically equicontinuous on [0,T].
\end{lem}
Proof. Lemma \ref{lem-GD1} has shown $\max_{t \in [0, T]} |V^n(t)| = O_P(1)$. 
From ODEs (\ref{GD-c1}) and (\ref{GD-c2}) we have 
\begin{align*}
& \dot{V}^n(t)= \sqrt{n} [ \dot{X}^n(t) - \dot{X}(t) ] = - \sqrt{n} [ \nabla g(X^n(t)) - \nabla g(X(t)) ] - R^n(X^n(t); \bU_n) , & \\
& | \dot{V}^n(t)| \leq \sqrt{n} | \nabla g(X^n(t)) - \nabla g(X(t)) | + | R^n(X^n(t); \bU_n)| &\\
& \leq L \sqrt{n} |X^n(t)) - X(t) | + | R^n(X^n(t); \bU_n)|. &
\end{align*}
Lemma \ref{lem-GD1} shows that $ \sqrt{n} |X^n(t)) - X(t) | = O_P(1)$, which indicates that for large $n$, $X^n(t)$ falls into $\Theta_X$ and 
assumption A4 in turn implies $| \sup_t R^n(X^n(t); \bU_n)| \sim \sup_t |\bsigma(X^n(t)) \bZ | = O_P(1)$. Substituting these into
the upper bound of $|\dot{V}^n(t)|$ we prove that $\max_{t \in [0,T]} |\dot{V}^n(t) | = O_P(1)$.  Combining this with 
$\max_{t \in [0,T]} |V^n(t) | = O_P(1)$ shown in Lemma \ref{lem-GD1}, we immediately establish the lemma. 

\textbf{Proof of Theorem \ref{thm-1} for the plain gradient descent case} The same perturbation argument in Section \ref{section-tight} can be used to show 
finite-dimensional distribution convergence of $V^n(t)$ to $V(t)$ for simple ODE (\ref{GD-c2}) in the plain gradient descent case. 
With the tightness of $V^n(t)$ shown in Lemma \ref{lem-GD2} together with the finite distribution convergence we immediately prove the weak convergence of  $V^n(t)$ to $V(t)$ in the plain gradient descent case. 

%\subsubsection{Convergence of $X^n(t)$ in the accelated case}
\subsubsection{Weak Convergence proof for the accelerated case}
\label{section-Xn}

We can use the same proof  in Su et al. (2016, Theorem 1) to show that 
%\begin{prop} \label{prop1}
ODE (\ref{equ-3}) 
%\[  \ddot{X}^n(t)+\frac{3}{t}\dot{X}^n(t) +  \nabla \!g(X^n(t)) + \frac{1}{\sqrt{n}} R^n(X^n(t);\bU_n) =0 \]
has a unique solution for each $n$ and $\bU_n$.
%\end{prop}
While the proof arguments in Su et al. (2016, Theorem 1) mainly require local ODE properties such as those near a neighbor of zero, our weak convergence analysis needs to investigate global behaviors of processes generated from SDEs and ODEs with random coefficients. We will first extend
and refine some local results for the global case and establish several preparatory lemmas for proving weak convergence in the theorem.
%for proving Proposition \ref{prop1}. 

Given an interval $\mathcal{I}=[s,t]$ and a process $Y(t)$, define for $a \in (0, 1]$, 
\begin{equation} \label{equ-Ma}
 M_a(s,t;Y)=M_a(\mathcal{I};Y) =  \sup_{u \in [s,t]} \left| \frac{\dot{Y}(u)-\dot{Y}(s)}{(u-s)^a} \right|. 
\end{equation}
In the proof of Theorem \ref{thm-1} we take $a=1$ and use $M_1(s,t; Y)$. %and write $M_1(s,t;Y)$ as $M(s,t;Y)$ for simplicity.
We will need $M_a(s,t; Y)$ with $a<1$ later in the proof of Theorems \ref{thm6}. 

\begin{lem} \label{lem1}  %\label{lem-Xn1}
For $X(t)$ and $X^n(t)$ we have the following inequalities, %for large enough $n$, 
\begin{align*}
& M_1(s,t; X) \leq \frac{1}{ 1 - L (t-s)^2/6} \left[ \left( \frac{3}{s}  + \frac{ L(t-s)}{2} \right) |\dot{X}(s)| + |\nabla g( X(s))|   \right], &\\
& M_1(s,t; X^n) \leq \frac{1}{ 1 - [\zeta(\bU_n) + 2 L] (t-s)^2/6} &\\ 
&\left[ \left( \frac{3}{s}  + \frac{ [\zeta(\bU_n) + 2 L] (t-s)}{2} \right) |\dot{X}^n(s)| +   |\nabla g( X^n(s))| +  n^{-1/2}  |R^n(X^n(s);\bU_n)|   \right] , & \\
& M_1(s,t;X^n - X) \leq \frac{1}{ 1 -  [\zeta(\bU_n) + 2 L] (t-s)^2/6}  \left\{   ( 3/s + (t-s) [\zeta(\bU_n) + 2 L]  )  |\dot{X}^n(s) - \dot{X}(s)|  \right. &\\ 
&  +  [2 \zeta(\bU_n) + 5 L] |X^n(s) - X(s)| + n^{-1/2} |R^n(X^n(s);\bU_n)| &\\
& \left. + n^{-1/2} \sup_{ u \in [s,t]} |R^n(X(u);\bU_n) - R^n(X(s);\bU_n)|  \right\} , &
\end{align*}
when $s>0$ and $t-s < \sqrt{6/[\zeta(\bU_n) + 2 L] }$, $\zeta(\bU_n) = \frac{1}{n} \sum_{i=1}^n h_1(U_i)$, and $h_1(\cdot)$ is given in assumption A1. 
In particular for $s=0$,
\[ M_1(0,t; X) \leq \frac{  |\nabla g(x_0)| }{ 1 - L t^2/6}, \;\;
M_1(0,t; X^n) \leq \frac{  |\nabla g(x_0)| +  n^{-1/2}  |R^n(x_0;\bU_n)|  
}{ 1 - [\zeta(\bU_n) + 2 L] t^2/6},  \]
\[ M_1(0,t;X^n - X) \leq \frac{n^{-1/2} }{ 1 - [\zeta(\bU_n) + 2 L] t^2/6}  \]
\[ \left[ |R^n(x_0;\bU_n)| + \sup_{ u \in [0,t]} |R^n(X(u);\bU_n) - R^n(x_0;\bU_n)| 
  \right]. \]
%If $s=0$, we replace  the coefficient $3/s$ by $1/t$ in  above inequality,  and $V^n(0) = \dot{V}^n(0) =0$, $X^n(0)=X(0)=x_0$.  Then 
%\[ M_1(0,t;X^n - X) \leq \frac{1}{ 1 - [\zeta(\bU_n) + 2 L] t^2/6} \left[ |R^n(x_0;\bU_n)| +  |R^n(X(t);\bU_n) - R^n(x_0;\bU_n)|   \right], \]
%which in particular implies that 
%\[ \sup_{ t \leq \sqrt{3/[\zeta(\bU_n) + 2L]}} \frac{ |\dot{X}^n(t) - \dot{X}(t)| }{ t } \leq 2 n^{-1/2} \left[ 2 |R^n(x_0;\bU_n)| +  |R^n(X(t);\bU_n)| 
%  \right]  \rightarrow 0,\]
%that is, $\dot{X}^n(t) \rightarrow \dot{X}(t)$ uniformly over $\left[0, \sqrt{3/[\zeta(\bU_n) + 2L]} \right]$. 
 \end{lem}
Proof. Because of similarity, we provide proof arguments only for $M_1(s,t; X^n-X)$. As $V^n(t) = \sqrt{n} [X^n(t) - X(t)]$, 
$M_1(s,t; V^n) = \sqrt{n} M_1(s,t; X^n-X)$, and we will establish the inequality for $M_1(s,t; V^n)$.
$V^n(t)$ satisfies the differential equation 
\begin{equation} \label{equ-l1}
 \ddot{V}^n(t) + \frac{3}{t}  \dot{V}^n(t) + \sqrt{n} \nabla \![g(X^n(t)) - g(X(t))] + R^n(X^n(t);\bU_n) =0. 
\end{equation}
Let 
 %\[  \ddot{X}^n(t)+\frac{3}{t}\dot{X}^n(t) +  \nabla \!g(X^n(t)) + \frac{1}{\sqrt{n}} T^n(X^n(t);\bU) =0. \]
\begin{align*}
%& [\ddot{X}^n_\dagger(t) - \ddot{X}(t)] + \frac{3}{t} [ \dot{X}^n_\dagger(t) - \dot{X}(t)] + \nabla \![g(X^n_\dagger(t)) - g(X(t))] + n^{-1/2} T^n(X^n(t);\bU)   =0. & \\
%& \ddot{V}^n(t) + \frac{3}{t}  \dot{V}^n(t) + \boldsymbol{I\!\! H}\! g(X(t)) V^n(t) + n^{-1/2} T^n(X^n(t);\bU) 
%  =0, & \\
%&  \ddot{V}^n(t) + \frac{3}{t}  \dot{V}^n(t) + \sqrt{n} \nabla \![g(X^n(t)) - g(X(t))] + T^n(X^n(t);\bU) =0, &\\
& H(t;V^n) = \sqrt{n} \nabla \![g(X^n(t)) - g(X(t))] + R^n(X^n(t);\bU_n), &
\end{align*}
and $J(s,t; H, V^n) = \int_s^t u^3 [ H(u;V^n) - H(s;V^n)] du$. Then
\begin{align*}
& | H(t;V^n) - H(s;V^n) | \leq \sqrt{n} |\nabla [g( X^n(t) ) - g(X^n(s)) - g(X(t)) + g(X(s))]| &\\
& + |R^n(X^n(t);\bU_n) - R^n(X^n(s);\bU_n)|. & 
\end{align*}
As in the proof of Lemma \ref{lem-GD1}, using assumptions A1-A2 we obtain 
\begin{align*}
&  \sqrt{n} \left| \nabla [g( X^n(t) ) - g(X^n(s)) - g(X(t)) + g(X(s))] \right| \\
& \leq L \sqrt{n} |X^n(t) - X(t) | + L \sqrt{n} | X^n(s) - X(s) |, & \\
& |R^n(X^n(t);\bU_n) - R^n(X^n(s);\bU_n)| \leq  |R^n(X^n(t);\bU_n) - R^n(X(t);\bU_n)| &\\
& +  |R^n(X^n(s);\bU_n) - R^n(X(s);\bU_n)| +  |R^n(X(t);\bU_n) - R^n(X(s);\bU_n)|, & \\
&  |R^n(X^n(u);\bU_n) - R^n(X(u);\bU_n)| \leq [ \zeta(\bU_n) + L ]  \sqrt{n} | X^n(u) - X(u) |, &\\
& \sqrt{n} [X^n(t) - X(t)] = V^n(t) %=\int_s^t \dot{V}^n(u) du + V^n(s) 
= \int_s^t [ \dot{V}^n(u) - \dot{V}^n(s)] du + V^n(s) + (t-s) \dot{V}^n(s). &
\end{align*}
Putting together these results we get 
\begin{align*}
& | H(t;V^n) - H(s;V^n) | \leq [\zeta(\bU_n) + 2 L] \left[  \int_s^t |\dot{V}^n(u) - \dot{V}^n(s)| du + 2 |V^n(s)| + (t-s) | \dot{V}^n(s)| \right] & \\ 
 &  + |R^n(X(t);\bU_n) - R^n(X(s);\bU_n)|. & 
 \end{align*}
 On the other hand, we have 
 
 \begin{align*}
&  \int_s^t |\dot{V}^n(u) - \dot{V}^n(s)| du \leq \int_s^t (u-s) \frac{|\dot{V}^n(u) - \dot{V}^n(s)|}{u-s} du \leq \int_s^t (u-s)M_1(s,t;V^n) du &\\
& = \frac{M_1(s,t;V^n) (t-s)^2}{2}, &\\
& \int_s^t  M_1(s,u;V^n) u^3 (u-s)^2 du /2 \leq  M_1(s,t;V^n) t^3 (t-s)^3/6. & 
\end{align*}
Substituting above inequalities into the upper bound for $|H(u; V^n) - H(s; V^n)| $ and the definition of $J(s, t; H, V^n)$ we conclude
\begin{align*}
&  |J(s,t;H, V^n)| \leq [\zeta(\bU_n) + 2 L] \left\{ M_1(s,t;V^n) t^3(t-s)^3/6 +  [ 2 |V^n(s)| +(t-s)| \dot{V}^n(s)|] t^3 (t-s) \right\} &\\
& +\,  t^3(t-s) \sup_{u \in [s,t]}  |R^n(X(u);\bU_n) - R^n(X(s);\bU_n)|. & 
\end{align*}
%\[ M_1(s,t) = \sup_{u \in [s, t]} \left|\frac{\dot{V}^n(u) - \dot{V}^n(s)}{u-s} \right|  \]

ODE (\ref{equ-l1}) is equivalent to 
\begin{align*} 
& \frac{ t^3 \dot{V}^n(t)}{dt} = - t^3 H(t;V^n), \mbox{which implies} & \\
& t^3 \dot{V}^n(t) -  s^3 \dot{V}^n(s) = -\int_s^t u^3 H(u;V^n) du =  - \frac{ t^4-s^4}{4} H(s;V^n)- J(s,t;H,V^n) ,&\\
& \frac{ \dot{V}^n(t) - \dot{V}^n(s)}{t-s} = -\frac{t^3-s^3}{t^3(t-s)} \dot{V}^n(s) - \frac{ t^4-s^4}{4 t^3(t-s)} H(s;V^n) - \frac{J(s,t;H,V^n)}{t^3(t-s)} , &
\end{align*}
and using  the upper bound of $|J(s,t;H, V^n)|$ and algebraic manipulation we get 
\begin{align*}
&  \frac{| \dot{V}^n(t) - \dot{V}^n(s)|}{t-s }\leq \frac{t^3-s^3}{t^3(t-s)} |\dot{V}^n(s)|  +  \frac{ t^4-s^4}{4 t^3(t-s)} |H(s;V^n)| + \frac{|J(s,t;H,V^n)|}{t^3 (t-s)} &\\
& \leq \frac{t^2 + st + s^2}{t^3} |\dot{V}^n(s)| + \frac{ (t^2+s^2)(t+s)}{4 t^3} |H(s;V^n)|&\\
& + [\zeta(\bU_n) + 2 L] \left[   M_1(s,t;V^n) \frac{(t-s)^2}{6} + 2 |V^n(s)| + (t-s) | \dot{V}^n(s)|  \right] &\\
&  +  \sup_{u \in [s, t]} |R^n(X(u);\bU_n) - R^n(X(s);\bU_n)|.  &
\end{align*}
As above inequality holds for any $t >s$, we may replace $t$ by $v$, take the maximum over $v \in [s,t]$,  and use the definition of 
$M_1(s,t; V^n)$ (which is increasing in $t$) to obtain 
\begin{align*}
&M_1(s,t;V^n) \leq \frac{3}{s} |\dot{V}^n(s)| + |H(s;V^n)| + [\zeta(\bU_n) + 2 L] M_1(s,t;V^n) \frac{(t-s)^2}{6} & \\
&  + [\zeta(\bU_n) + 2 L] [2 |V^n(s)| + (t-s)| \dot{V}^n(s)|]  + \sup_{ u \in [s,t]} |R^n(X(u);\bU_n) - R^n(X(s);\bU_n)|,  &\\
& \leq \frac{3}{s} |\dot{V}^n(s)| + L |V^n(s)| +   |R^n(X^n(s);\bU_n) |+ [\zeta(\bU_n) + 2 L] M_1(t,s;V^n) \frac{(t-s)^2}{6} & \\
&  + [\zeta(\bU_n) + 2 L] [2 |V^n(s)| + (t-s)| \dot{V}^n(s)|]  + \sup_{ u \in [s,t]}  |R^n(X(u);\bU_n) - R^n(X(s);\bU_n)|,  &
\end{align*}
and solving for  $M_1(s,t;  V^n)$ leads to 
\begin{align*}
& M_1(s,t;V^n) \leq \frac{1}{ 1 -  [\zeta(\bU_n) + 2 L] (t-s)^2/6}  \left\{   ( 3/s + (t-s) [\zeta(\bU_n) + 2 L]  )  |\dot{V}^n(s)|  \right. &\\ 
&\left.  +  [2 \zeta(\bU_n) + 5 L] |V^n(s)| + |R^n(X^n(s);\bU_n)| + \sup_{ u \in [s,t]} |R^n(X(u);\bU_n) - R^n(X(s);\bU_n)|  \right\} , &
\end{align*}
when $s>0$ and $t-s < \sqrt{6/[\zeta(\bU_n) + 2 L] }$.  If $s=0$, we replace  the coefficient $3/s$ by $1/t$ in  above inequality,  and $V^n(0) = \dot{V}^n(0) =0$, $X^n(0)=X(0)=x_0$.  Then 
\[ M_1(0,t;V^n) \leq \frac{1}{ 1 - [\zeta(\bU_n) + 2 L] t^2/6} \left[ |R^n(x_0;\bU_n)| + \sup_{ u \in [0,t]} |R^n(X(u);\bU_n) - R^n(x_0;\bU_n)| 
  \right], \]
%\[ \frac{1}{ 1 - L (t-s)^2/6} \left[ \bsigma(x_0) |\bZ |  +  (D(\bU) + L) \sup_{u \in [s,t]} | X(u) | \right]\]
which in particular implies that 
\[ \sup_{ t \leq \sqrt{3/[\zeta(\bU_n) + 2L]}} \frac{ |\dot{X}^n(t) - \dot{X}(t)| }{ t } \leq 2 n^{-1/2} \left[ 2 |R^n(x_0;\bU_n)| +  \sup_{ u \in [0,t]} |R^n(X(u);\bU_n)| 
  \right]  \rightarrow 0,\]
that is, $\dot{X}^n(t) \rightarrow \dot{X}(t)$ uniformly over $\left[0, \sqrt{3/[\zeta(\bU_n) + 2L]} \right]$. 
%Thus $X^n(t)$ weakly converges to $X(t)$ on $\mathcal{I}_1$.

\begin{lem} \label{lem2}
For any given $T>0$,  we have 
\[ \max_{t \in [0, T]} |X^n(t) - X(t)| = O_P(n^{-1/2}), \;\;\; \max_{t \in [0, T]} |V^n(t) | = O_P(1), \]
\[ \max_{t \in [0, T]} |\dot{X}^n(t) - \dot{X}(t)| = O_P(n^{-1/2}), \;\;\;  \max_{t \in [0, T]} |\dot{V}^n(t) | = O_P(1).  \]
\end{lem}
Proof. As $V^n(t) = \sqrt{n} [ X^n(t) - X(t)]$, we need to establish the results for $X^n(t) - X(t)$ only. 
Since as $n \rightarrow \infty$, $\zeta(\bU_n) = \frac{1}{n} \sum_{i=1}^n h_1(U_i) \rightarrow E(h_1(U))$. 
Divide the interval $[0,T]$ into $N=\left[T \sqrt{[E( h_1(U)) + 2 L]/3} \right]+1$ number of subintervals with length close to $\sqrt{3/[E(h_1(U))+2L]}$ (except for the last one), and denote them by $\mathcal{I}_i=[s_{i-1}, s_i]$, $i=1, \cdots, N$ (with $s_0=0$, $s_N=T$, $\mathcal{I}_1 = [0, s_1]$, %\sqrt{3/L}]$, 
$\mathcal{I}_N = [s_{N-1}, T]$).
First for $t \in \mathcal{I}_1$, from Lemma \ref{lem1} we have 
%\[  |\dot{X}^n(t)| \leq |\mathcal{I}_1| M_1(\mathcal{I}_1;X^n) \leq C \left[ |\nabla g(x_0)| +  n^{-1/2}  |R^n(x_0;\bU_n)|  \right],\]
%\[ |X^n(t)| \leq |X^n(0)| + \int_{\mathcal{I}_1} |\dot{X}^n(u) | du \leq x_0 +  C \left[  |\nabla g(x_0)| +   n^{-1/2}  |R^n(x_0;\bU_n)|    \right]. \]
\[  |\dot{X}^n(t) - \dot{X}(t)| \leq |\mathcal{I}_1| M_1(\mathcal{I}_1;X^n-X) \leq C n^{-1/2} \left[  |R^n(x_0;\bU_n)|  + |R^n(X(s_1);\bU_n)|  \right],\]
\[ |X^n(t) - X(t)| \leq \int_{\mathcal{I}_1} |\dot{X}^n(u) - \dot{X}(u) | du \leq  C n^{-1/2} \left[  |R^n(x_0;\bU_n)|  + |R^n(X(s_1);\bU_n)|   \right]. \]
Assumption A4 implies that $R^n(x_0;\bU_n) = O_P(1)$, and $R^n(X(s_1);\bU_n) = O_P(1)$, and thus 
 the upper bounds of $\dot{X}^n(t) - \dot{X}(t)$ and $X^n(t) - X(t)$ over $t \in \cI_1$ are $O_P(n^{-1/2})$. 
 %As a consequence, with $V^n(t) = \sqrt{n} [ X^n(t) - X(t)]$, we also conclude that the upper bounds of $\dot{V}^n(t)$ and $V^n(t)$ over $t \in \cI_1$ are $O_P(1)$, and hence $V^n(t)$ is stochastically  equicontinuous on $\cI_1$.
%  which implies that  $X^n(t)$ is stochastically equicontinuous and uniformly bounded over $\mathcal{I}_1$. 
%Since the arguments in Section \ref{section-Xconvergence} can show the finite-dimensional distribution convergence of $X^n(t)$ to $X(t)$, thus 
%$X^n(t)$ weakly converges to $X(t)$ over $t \in \cI_1$. With $X(t)$ deterministic, it implies that $X^n(t)$ converges in probability to $X(t)$. 

For $t \in \mathcal{I}_i$, $i =2, \cdots, N$, from Lemma \ref{lem1} we have 
\begin{eqnarray*}
&& |\dot{X}^n(t)- \dot{X}(t) - \dot{X}^n(s_{i-1})  + \dot{X}(s_{i-1})| \leq |\mathcal{I}_i| M_1(\mathcal{I}_i; X^n - X)  \\
&& \leq    C \left[ [ \zeta(\bU_n) + C_1 ] |\dot{X}^n(s_{i-1}) -\dot{X}(s_{i-1})| + [ \zeta(\bU_n) + C_2 ] |X^n(s_{i-1}) -X(s_{i-1})|  \right]  \\
&& + C n^{-1/2} \left\{    |R^n(X^n(s_{i-1});\bU_n)| +  2 \sup_{u \geq 0} |R^n(X(u);\bU_n) |  \right\} , 
\end{eqnarray*}
and
\begin{eqnarray*}
&& |X^n(t) - X(t)| \leq |X^n(s_{i-1}) - X(s_{i-1})| + |\mathcal{I}_i| | \dot{X}^n(s_{i-1}) - \dot{X}(s_{i-1}) | \\
&& + \int_{\mathcal{I}_i} |\dot{X}^n(u) -\dot{X}(u) - \dot{X}^n(s_{i-1}) + \dot{X}(s_{i-1}) | du \\
&& \leq     C \left[ [ \zeta(\bU_n) + C_1 ] |\dot{X}^n(s_{i-1}) -\dot{X}(s_{i-1})| + [ \zeta(\bU_n) + C_2 ] |X^n(s_{i-1}) -X(s_{i-1})|   \right]  \\
&& + C n^{-1/2} \left\{    |R^n(X^n(s_{i-1});\bU_n)| +  2 \sup_{u \geq 0} |R^n(X(u);\bU_n) |  \right\} .
\end{eqnarray*}
% $X^n(s_i)$ jointly converges to $X(s_i)$, $i=1,\cdots, N$. If we can show  that as $n \rightarrow \infty$, $\dot{X}^n(s_i) \rightarrow \dot{X}(s_i)$ in probability, 
We will use above two inequalities to prove by induction that the upper bounds of $X^n(t) - X(t)$ and $\dot{X}^n(t) - \dot{X}(t)$ on $[0, T]$ are $O_P(n^{-1/2})$,
and the upper bounds of $X^n(t) - X(t)$ and $\dot{X}^n(t) - \dot{X}(t)$ on $[0, T]$ are $O_P(n^{-1/2})$, 
Assume that the upper bounds of $X^n(t) - X(t)$ and $\dot{X}^n(t) - \dot{X}(t)$ on $\cup_{j=1}^{i-1} \mathcal{I}_j$ are $O_P(n^{-1/2})$. 
Note that $N$ is free of $n$, by induction we have shown that the upper bounds of $X^n(t) - X(t)$ and $\dot{X}^n(t) - \dot{X}(t)$ over $t \leq s_{i-1}$ are 
$O_P(n^{-1/2})$ in particular $X^n(s_{i-1}) \rightarrow X(s_{i-1}) $ in probability, and thus assumption A4 indicates that 
$R^n(X^n(s_{i-1});\bU_n) = O_P(1)$,  and $\sup_{u \geq 0} |R^n(X(u);\bU_n)| = O_P(1)$.
Above two inequalities immediately show that their upper bounds on $ \mathcal{I}_i$ are also $O_P(n^{-1/2})$. Hence, 
we establish that the bounds of $X^n(t) - X(t) $ and $\dot{X}^n(t) - \dot{X}(t)$ on $\cup_{j=1}^{N} \mathcal{I}_j=[0,T]$ are $O_P(n^{-1/2})$.
%Lemma \ref{lem2} together with the finite distribution convergence immediately show the following lemma. 

%\begin{lem} \label{lem3}
%As $n \rightarrow \infty$, $X^n(t)$ weakly converges to $X(t)$.
%\end{lem}

%Since $X(t)$ is deterministic, the weak convergence of $X^n(t)$ to $X(t)$ implies that $X^n(t)$ converges in probability to $X(t)$.

\begin{lem} \label{lem5}
$V^n(t) $ is stochastically equicontinuous on $[0, T]$.
\end{lem}
Proof. Lemma \ref{lem2} shows that $\max_{t \in [0, T]} |V^n(t)|=O_P(1)$ and $\max_{t \in [0, T]} |\dot{V}^n(t)| = O_P(1)$, which implies that $V^n(t)$ is stochastically equicontinuous on $[0, T]$. 

\textbf{Proof of Theorem \ref{thm-1} }
%Since $V^n(t) = \sqrt{n} [X^n(t) - X(t)]$, from Lemma \ref{lem2} we conclude that $\max_{t \in [0, T]} |V^n(t)|=O_P(1)$ and $\max_{t \in [0, T]} |\dot{V}^n(t)| = O_P(1)$, which implies that $V^n(t)$ is stochastically equicontinuous on $[0, T]$. 
Lemma \ref{lem5} together with the finite distribution convergence immediately lead to that 
%\begin{lem} \label{lem4}
as $n \rightarrow \infty$, $V^n(t)$ weakly converges to $V(t)$. 
%\end{lem}

%\subsubsection{Weak convergence of $V^n(t)$ in the accelerated case}
%\label{section-Vn}

%For deterministic solution curve $X(t)$ from ODE (\ref{equ-2}),  $\Theta_X$ contains a neighborhood of curve image $\{X(t): t \geq 0\}$ in $\Theta$, and for large 
%$n$, $X^n(t)$ will fall into $\Theta_X$. Throughout the rest of the proofs
%we often write $\sup_{\theta \in \Theta_X} $ as $\sup_\theta$ for simplicity, and inequalities for $M_a(s,t;X^n)$ and $M_a(s,t;V^n)$ hold for large $n$.

\subsection{Proof of Theorem \ref{thm-1-1}} 
\subsubsection{Proof for the plain gradient descent case} 
\begin{lem} \label{lem-GD3}
For the case of plain gradient descent algorithm, we have 
\[   \max_{ t \in [0, T]} | x^n_\delta(t) - X^n(t) | = O_P(\delta), \;\; \max_{ k \leq T \delta^{-1} } | x^n_k  - X^n( k \delta ) | = O_P(\delta), \] 
where $\{x^n_k\}$ is generated from algorithm (\ref{min-GD2}), with $x^n_\delta(t)$ its continuous-time step process, and $X^n(t)$ is the solution 
of ODE (\ref{GD-c2}). 
\end{lem}
Proof. Algorithm (\ref{min-GD2}) is the Euler scheme for solving ODE (\ref{GD-c2}), and we will apply the standard ODE theory to obtain the global 
approximation error for the Euler scheme. 
First by assumption A1 we have that $\nabla \cL^n(\theta; \bU_n)$ is Lipschtiz in $\theta$ with Lipschitz constant 
$\frac{1}{n} \sum_{i=1}^n h_1(U_i)$, which converges in probability to $E[h_1(U)]<\infty$. On the other hand, taking derivative on both sides of 
ODE (\ref{GD-c2}), we obtain 
\begin{align*}
& \ddot{X}^n(t) = - \boldsymbol{I\!\! H}\! \cL^n(X^n(t); \bU_n) \dot{X}^n(t) = \boldsymbol{I\!\! H}\! \cL^n(X^n(t); \bU_n) \nabla \cL^n(X^n(t); \bU_n). 
\end{align*}
Using Lemma \ref{lem-GD1}, we conclude that for large $n$, $X^n(t)$ falls into $\Theta_X$, and thus assumption A4 indicates that 
$\sup_t |\nabla^\kappa \cL^n(X^n(t); \bU_n)| \sim \sup_t |\nabla^\kappa g(X^n(t)) + n^{-1/2} \bsigma_k(X^n(t)) \bZ_\kappa | = O_P(1)$, where 
$\bZ_\kappa$ are standard normal random variables. Combining these results together we get $\sup_{t \in [0, T]} |\ddot{X}^n(t)| = O_P(1)$.
An application of the standard ODE theory for the global approximation error of the Euler scheme (Butcher (2008)) 
leads to 
\begin{align*}
&  \max_{ t \in [0, T]} | x^n_\delta(t) - X^n(t) | \leq \delta \left( \frac{2}{n} \sum_{i=1}^n h_1(U_i) \right)^{-1}
 \sup_{t \in [0, T]} |\ddot{X}^n(t)|  \left[ \exp\left( \frac{T}{n} \sum_{i=1}^n h_1(U_i) 
\right) -1 \right] &\\ 
& = O_P(\delta). 
\end{align*}

\textbf{Proof of Theorem \ref{thm-1-1}}. Lemma \ref{lem-GD3} establishes the first order result for $x^n_\delta(t)  - X^n(t)$, and the weak convergence result is 
the consequence of the order result and Theorem \ref{thm-1}. 

\subsubsection{Proof for the accelerated gradient descent case}
%\subsubsection{Difference between $x^n_\delta(t)$ and $X^n(t)$}

Note that $(x_k, y_k)$ and $(x^n_k, y^n_k)$ are generated from accelerated gradient descent algorithms (\ref{equ-Nest1}) and 
(\ref{min-Nest2}), respectively, and $X(t)$ and $X^n(t)$ are respective solutions of ODEs (\ref{equ-2}) and  (\ref{equ-3}). 

%and $X_\eta(t)$ and $X^n_\eta(t)$ the corresponding smoothed ODEs with $t$ replaced by $t \vee \eta$, 
%and $X_{k, \eta}$ and $X^n_{k, \eta}$ the 
%approximation sequences generated by the Euler scheme for the corresponding smoothed ODEs, where $\eta$ is a small positive constant. 

%\input{p1-add}
\begin{lem} \label{lem-add-1}
%Let $\{x_k, y_k\in \R^p\}_k$ be the sequence generated from the Nesterov's accelerated gradient descent algorithm when solving the optimization problem of a convex function $g$ defined on $I\!\!R^p$. $g$ has $L$-Lipschitz continuous gradients, that is $$|\nabla g(x)-\nabla g(y)| \leq L|x-y|$$
%for any $x,y\in I\!\!R^p$. The sequence starts with $x_0$ and $y_0=x_0$, iterates by 
%\begin{equation} \label{lm2-1}
%x_k=y_{k-1}-\delta\nabla g(y_{k-1}), y_k=x_k+\frac{k-1}{k+2}(x_k-x_{k-1}),
%\end{equation}
%where $\delta$ is the step size. Let $X\in C^1([0,\infty); \mR^p)\cap C^2((0,\infty); \mR^p)$ be the unique solution to the ODE
%\begin{equation} \label{lm2-2}
%\ddot{X}(t)+\frac{3}{t}\dot{X}(t)+\nabla g(X(t))=0
%\end{equation}
%with initial values $X(0)=x_0, \dot{X}(0)=0$. 
For fixed $T>0$, as $\delta\rightarrow 0$, we have 
\begin{equation} \label{lm2-3}
\max_{k\leq T\delta^{-1/2}}\left|x_k-X(k\delta^{1/2})\right|=O(\delta^{1/2}|\log\delta|), 
\end{equation}
\begin{equation} \label{lm2-3-0}
\max_{k\leq T\delta^{-1/2}}\left|z_k-\dot{X}(k\delta^{1/2})\right|=O(\delta^{1/2}|\log\delta|), 
\end{equation}
where sequence $x_k$ is generated from algorithm (\ref{equ-Nest1}), $X(t)$ is the solution of the corresponding ODE (\ref{equ-2}), and 
$z_k=(x_{k+1}-x_k)/\delta^\ot$ is given in (\ref{Nest-1-2}). 
\end{lem}
Proof. %As in (\ref{Nest-1-2}), we define $z_k=(x_{k+1}-x_k)/\delta^\ot$. 
We rewrite (\ref{equ-Nest1}) as 
$$x_{k+2}=y_{k+1}-\delta\nabla g(y_{k+1}), \;\; y_{k+1}=x_{k+1}+\frac{k}{k+3}(x_{k+1}-x_k)=x_k+\frac{2k+3}{k+3}\delta^\ot z_k,$$
and obtain 
\begin{equation} \label{lm2-3-1}
z_{k+1}=\left(1-\frac{3}{k+3}\right)z_k-\delta^\ot\nabla g\left(x_k+\frac{2k+3}{k+3}\delta^\ot z_k\right). 
\end{equation}
%Assume $y_k$ in the neighborhood of 
Denote by $y^*$ the critical point of $g(\cdot)$. Then
$$|\nabla g(y_k)|=|\nabla g(y_k)-\nabla g(y^*)|\leq L|y_k-y^*|\leq C_1,$$
where $C_1$ is some constant, and 
\begin{equation} \label{lm2-4}
|z_0|=|x_1-x_0|/\delta^\ot=\delta^\ot  |\nabla g(x_0)|\leq C_1\delta^\ot, 
\end{equation}
\begin{equation} \label{lm2-5}
|z_k|\leq\frac{k-1}{k+2}|z_{k-1}|+C_1\delta^\ot\leq (k+1)C_1\delta^\ot. 
\end{equation}
To compare $x_k$ and $X(k\delta^\ot)$ and derive their difference, we first need to find the relationship between $X(k\delta^\ot)$ and $X((k+1)\delta^\ot)$
and between $\dot{X}(k\delta^\ot)$ and $\dot{X}((k+1)\delta^\ot)$. As in (\ref{Nest-2-1}),  we let $Z=\dot{X}$, and ODE (\ref{equ-2}) is equivalent to 
$$\dot{X}=Z,\ \ \dot{Z}=-\frac{3}{t}Z-\nabla g(X).$$
Then with convention $t_k=k\delta^\ot$, we have for $k\geq 1$, %the integral form of the differential equation is
\begin{equation} \label{lm2-6}
X(t_{k+1})=X(t_k)+\int_{t_k}^{t_{k+1}}Z(u)du=X(t_k)+\delta^\ot Z(t_k)+\int_{t_k}^{t_{k+1}}[Z(u)-Z(t_k)]du, 
\end{equation}
\begin{eqnarray} \label{lm2-7}
Z(t_{k+1})&=&Z(t_k)-\int_{t_k}^{t_{k+1}}\frac{3}{u}Z(u)du-\int_{t_k}^{t_{k+1}}\nabla g(X(u))du \\ \nonumber
&=&\left(1-\frac{3}{k}\right)Z(t_k)-\int_{t_k}^{t_{k+1}}\left[\frac{3}{u}Z(u)-\frac{3}{t_k}Z(t_k)\right]du- \\ \nonumber
& &\delta^\ot\nabla g(X(t_k))-\int_{t_k}^{t_{k+1}}[\nabla g(X(u))-\nabla g(X(t_k))]du. 
\end{eqnarray}
Lemma \ref{lem1} shows that on $(0,T]$, $|\dot{X}(t)|/t$ is bounded,  $|Z(t)|\leq Ct$, and $|\dot{Z}(t)|=|\ddot{X}(t)|\leq C$ for some constant $C$. Then 
we easily derive bounds for the following integrals appeared on the right hand sides of (\ref{lm2-6}) and (\ref{lm2-7}), 
$$\left|\int_{t_k}^{t_{k+1}}[Z(u)-Z(t_k)]du\right|=O(\delta),$$
\begin{eqnarray*}
\left|\int_{t_k}^{t_{k+1}}\left[\frac{3}{u}Z(u)-\frac{3}{t_k}Z(t_k)\right]du\right|
&\leq&\int_{t_k}^{t_{k+1}}\left|\frac{3}{u}[Z(u)-Z(t_k)]\right|du   \\
&& +   \int_{t_k}^{t_{k+1}}\left|\left(\frac{3}{u}-\frac{3}{t_k}\right)Z(t_k)\right|du \\
&\leq&\frac{C\delta}{t_k}+\frac{3(t_{k+1}-t_k)^2}{t_kt_{k+1}}Ct_k 
= O(\delta^\ot k^{-1}),
\end{eqnarray*}
$$\left|\int_{t_k}^{t_{k+1}}[\nabla g(X(u))-\nabla g(X(t_k))]du\right|\leq L\int_{t_k}^{t_{k+1}}|X(u)-X(t_k)|du=O(\delta).$$
Plugging these integrals bounds into (\ref{lm2-6}) and (\ref{lm2-7}) we conclude
$$X(t_{k+1})=X(t_k)+\delta^\ot Z(t_k)+O(\delta),$$
$$Z(t_{k+1})=\left(1-\frac{3}{k}\right)Z(t_k)-\delta^\ot\nabla g(X(t_k))+O(\delta^\ot k^{-1})+O(\delta).$$
Let $a_k=|x_k-X(t_k)|, b_k=|z_k-Z(t_k)|$, and $S_k=b_0+b_1+...+b_k$. 
Using the definition of $z_k$ and (\ref{lm2-3-1})-(\ref{lm2-5}), we have
$$a_0=0, \;\; a_{k+1}\leq a_k+\delta^\ot b_k+O(\delta),$$
\begin{equation} \label{lm2-8}
a_k\leq \delta^\ot S_{k-1}+O(k\delta),
\end{equation}
$$b_0=|z_0|\leq C_1\delta^\ot, \;\; b_1=|z_1-Z(t_1)|=O(\delta^\ot),$$
\begin{eqnarray}
b_{k+1}&\leq&\left(1-\frac{3}{k+3}\right)b_k+\frac{9}{k(k+3)}|Z(t_k)|+ \nonumber \\
& &L\delta^\ot\left|x_k+\frac{2k+3}{k+3}\delta^\ot z_k-X(t_k)\right|+O(\delta^\ot k^{-1})+O(\delta) \nonumber \\
&\leq&b_k+O(\delta^\ot k^{-1})+L\delta^\ot a_k+2L\delta(k+1)C_1\delta^\ot+O(\delta^\ot k^{-1})+O(\delta)  \nonumber \\
&\leq&b_k+L\delta S_{k-1}+L\delta^\ot O(k\delta)+O(\delta)+O(\delta^\ot k^{-1})  \nonumber \\
&\leq&b_k+L\delta S_{k-1}+O(\delta^\ot (k+1)^{-1})   \label{lm2-9}
\end{eqnarray}
Since for $1\leq k\leq T\delta^{-\ot}$, $k\delta^\ot=O(1), O(\delta)=O(\delta^\ot k^{-1})$, $k^{-1}\leq 2(k+1)^{-1}$. Also with $b_1=O(\delta^\ot)$, we
can see that (\ref{lm2-9}) holds for $k=0$. Therefore, there exists some constant $C_2>0$ such that
$$b_{k+1}\leq b_k+L\delta S_{k-1}+C_2\delta^\ot (k+1)^{-1}.$$
Define a new sequence $b'_k$ from $b_k$ as follows. Let $b'_0=b_0$, $b'_{k+1}=b'_k+L\delta S'_{k-1}+C_2\delta^\ot (k+1)^{-1}$, where $S'_k=b'_0+b'_1+...+b'_k$. Then we can easily prove by induction that $b_k\leq b'_k$. Indeed, if $b_j\leq b'_j$ for $j=0,1,...,k$, then $S_{k-1}\leq S'_{k-1}$, 
$$b_{k+1}\leq b_k+L\delta S_{k-1}+C_2\delta^\ot (k+1)^{-1}\leq b'_k+L\delta S'_{k-1}+C_2\delta^\ot (k+1)^{-1}=b'_{k+1}.$$
On the other hand, as $L\delta S'_{k-1}+C_2\delta^\ot (k+1)^{-1}>0$, $\{b'_k\}$ is an increasing sequence. Thus,  $S'_{k-1}\leq kb'_k$, and 
$$b'_{k+1}\leq b'_k+L\delta kb'_k+C_2\delta^\ot (k+1)^{-1}.$$
Again define another sequence $b^*_k$ from $b'_k$ as follows. 
Let $b_0^*=b'_0$, $b_{k+1}^*=b_k^*+L\delta kb_k^*+C_2\delta^\ot (k+1)^{-1}$. The same induction argument can prove that $b'_k\leq b_k^*$.
The recursive definition of $b^*_k$ easily leads to the following expression, 
$$b_k^*=\delta^\ot\left(C_1\prod_{j=1}^{k-1}(1+L\delta j)+C_2\sum_{i=1}^{k}i^{-1}\prod_{j=i}^{k-1}(1+L\delta j)\right),  $$
and hence 
\begin{eqnarray*}
S_{\lfloor T\delta^{-\ot}\rfloor-1}&\leq&T\delta^{-\ot}b_{\lfloor T\delta^{-\ot}\rfloor}  \leq T\delta^{-\ot}b'_{\lfloor T\delta^{-\ot}\rfloor} \leq 
T\delta^{-\ot}b^*_{\lfloor T\delta^{-\ot}\rfloor} \\ 
&\leq&C\left(\prod_{j=1}^{\lfloor T\delta^{-\ot}\rfloor-1}(1+L\delta j)+\sum_{i=1}^{\lfloor T\delta^{-\ot}\rfloor}i^{-1}\prod_{j=i}^{\lfloor T\delta^{-\ot}\rfloor-1}(1+L\delta j)\right) \\ 
&\leq&C\left(\prod_{j=1}^{\lfloor T\delta^{-\ot}\rfloor-1}(1+L\delta T\delta^{-\ot})+\sum_{i=1}^{\lfloor T\delta^{-\ot}\rfloor}i^{-1}\prod_{j=i}^{\lfloor T\delta^{-\ot}\rfloor-1}(1+L\delta T\delta^{-\ot})\right) \\ 
\end{eqnarray*}
\begin{eqnarray*}
&\leq&Ce^{LT^2}\left(1+\sum_{i=1}^{\lfloor T\delta^{-\ot}\rfloor}i^{-1}\right) \leq C\log(T\delta^{-\ot}) =O(|\log\delta|). 
\end{eqnarray*}
Finally using above inequality and (\ref{lm2-8}) we arrive at 
$$\max_{k\leq T\delta^{-\ot}}\left|x_k-X(k\delta^\ot)\right|
\leq\delta^\ot S_{\lfloor T\delta^{-\ot}\rfloor-1}+O(T\delta^\ot)=O(\delta^\ot|\log\delta|),$$
which proves (\ref{lm2-3}). It is easy to see that the left hand side of (\ref{lm2-3-0}) is bounded by $b_{[T \delta^{-1/2}]}^*$, which is of order 
$\delta^\ot|\log\delta|$. 

%\newpage
\begin{lem} \label{lem-add-2}
%Under assumption A0-A4, as $\delta\rightarrow 0$ and $n\rightarrow\infty$, 
$$\max_{t\in[0,T]}\left|x_{\delta}^n(t)-X^n(t)\right|=O_p(\delta^\ot|\log\delta|), $$
where $x_{\delta}^n(t)$ is the continuous-time step processes for discrete sequence $x_k^n$ generated from algorithm (\ref{min-Nest2}), and $X^n(t)$ is the continuous-time solution of the corresponding ODE (\ref{equ-3}). 
\end{lem}
Proof. The objective function associated with (10) and (11) is $\nabla\cL^n(\theta; \bU_n)=\frac{1}{n}\sum_{i=1}^n\nabla\ell(\theta; U_i)$, which has Lipschitz constant $\frac{1}{n}\sum_{i=1}^n h_1(U_i)=O_p(1)$. Then for any $\epsilon>0$, there exists some constant $L_0>0$ such that for all $n$, 
$P\left(\frac{1}{n}\sum_{i=1}^n h_1(U_i) > L_0\right) < \epsilon$. For each $n$, on the event $\{\frac{1}{n}\sum_{i=1}^n h_1(U_i)\leq L_0\}$, Lemma \ref{lem-add-1} shows that there exists constant $M$ (which depends on $L_0$ only and is free of $n$) such that 
$$\max_{k\leq T\delta^{-\ot}}\left|x_k^n-X^n(k\delta^\ot)\right|
\leq M\delta^\ot|\log\delta|.$$
As a result we have that 
$$P\left(\max_{k\leq T\delta^{-\ot}}\left|x_k^n-X^n(k\delta^\ot)\right|
> M\delta^\ot|\log\delta|\right) \leq P\left(\frac{1}{n}\sum_{i=1}^n h_1(U_i)> L\right) < \epsilon$$
holds for each $n$, that is
$$\max_{k\leq T\delta^{-\ot}}\left|x_k^n-X^n(k\delta^\ot)\right|=O_p(\delta^\ot|\log\delta|).$$
Lemma \ref{lem1} indicates $\sup_{t\in[0, T]}|\dot{X}^n(t)|=O_p(1)$, and hence 
$$\sup_{s,t\in[0,T], t-s\leq\delta^\ot}|X^n(t)-X^n(s)|\leq \delta^\ot\sup_{t\in[0, T]}|\dot{X}^n(t)|=O_p(\delta^\ot).$$
Finally for any $t$ we can find  $k$ such that $t_k\leq t< t_{k+1}$,  and show 
$$\max_{t\in[0,T]}\left|x_{\delta}^n(t)-X^n(t)\right|
\leq \max_{t\in[0,T]}\left \{ |x_k^n-X^n(t_k)|+|X^n(t_k)-X^n(t)|\right \} =O_p(\delta^\ot|\log\delta|).$$

\textbf{Proof of Theorem \ref{thm-1-1}}. Lemma \ref{lem-add-2} establishes the order result for $x^n_\delta(t)  - X^n(t)$, and the weak convergence result is the consequence of
the order result and Theorem \ref{thm-1}. 
             
\subsection{Proof of Theorem \ref{thm-1-2}} 
%Then using the definitions of $\check{\theta}$ and $\hat{\theta}_n$ and Taylor expansion, we have 
%\[  0 = \nabla \cL^n(\hat{\theta}_n, \bU_n) = \nabla \cL^n(\check{\theta}, \bU_n) + \boldsymbol{I\!\! H}\! \cL^n(\check{\theta}, \bU_n) (\hat{\theta}_n - \check{\theta}) + \mbox{reminder},  \]
%and $\boldsymbol{I\!\! H}\! \cL^n(\check{\theta}, \bU_n)$ may approach to $\boldsymbol{I\!\! H}\! g(\check{\theta})$ for large $n$.
%On the other hand, 
%\[ \nabla \cL^n(\check{\theta}, \bU_n) \approx \nabla g(\check{\theta}) + n^{-1/2} \bsigma(\check{\theta}) \bZ = n^{-1/2} \bsigma(\check{\theta}) \bZ, \]
%where $\bZ$ stands for a standard normal random vector.  Thus, 
%\[   n^{1/2} (\hat{\theta}_n - \check{\theta}) \approx  [ \boldsymbol{I\!\! H}\! g(\check{\theta}) ]^{-1} 
%     \bsigma(\check{\theta}) \bZ, \]
%which leads to the asymptotic distribution of $n^{1/2} (\hat{\theta}_n - \check{\theta})$. The asymptotic derivation can be illustrated 
%from the optimization point of view. 
Using Assumption A4 and the standard empirical process argument (van der Vaart and Wellner (2000)) we can show that 
$\hat{\theta}_n$ is $\sqrt{n}$-consistent. Define $\vartheta = n^{1/2} (\theta - \check{\theta})$. We apply Taylor expansion to obtain 
\begin{align*}
&  \cL^n(\theta, \bU_n) = \cL^n(\check{\theta}, \bU_n) + \nabla \! \cL^n(\check{\theta}, \bU_n) (\theta - \check{\theta})
+ (\theta - \check{\theta})^\prime \boldsymbol{I\!\! H}\! \cL^n(\check{\theta}, \bU_n) (\theta - \check{\theta})/2 &\\
& + o_P(n^{-1/2}) &\\
& = \cL^n(\check{\theta}, \bU_n) + n^{-1/2} [\nabla \! g(\check{\theta}) + n^{-1/2} \bsigma(\check{\theta}) \bZ]  \vartheta 
+ n^{-1} \vartheta^\prime \boldsymbol{I\!\! H}\! g(\check{\theta}) \vartheta/2 + o_P(n^{-1}) &\\
& = \cL^n(\check{\theta}, \bU_n) +  n^{-1} \bsigma(\check{\theta}) \bZ  \vartheta + n^{-1} \vartheta^\prime \boldsymbol{I\!\! H}\! g(\check{\theta}) \vartheta/2 + o_P(n^{-1}), &
\end{align*}
where $\bZ$ stands for the standard normal random vector, the second equality is due to Assumptions 2 and 4, and 
the Skorokhod representation, and the law of large number, and the third equality is from $\nabla \! g(\check{\theta})=0$.
As $\hat{\theta}_n$ is the minimizer of $\cL^n(\theta, \bU_n)$, $\hat{\vartheta}_n = n^{1/2} (\hat{\theta}_n - \check{\theta})$ 
asymptotically minimizes $\bsigma(\check{\theta}) \bZ  \vartheta + \vartheta^\prime \boldsymbol{I\!\! H}\! g(\check{\theta}) \vartheta/2$ over $\vartheta$, 
%The minimizer $\hat{\theta}_n$ of $\cL^n(\theta, \bU_n )$ is immediately converted into the minimizer $\hat{\vartheta}_n = 
%n^{1/2} (\hat{\theta}_n- \check{\theta})$, which 
and thus has an asymptotic distribution $[ \boldsymbol{I\!\! H}\!  g(\check{\theta})]^{-1} 
\bsigma(\check{\theta} ) \bZ$. 
%Such asymptotic analysis is often rigorously carried out by empirical process arguments.
%Since gradient descent algorithms generate sequences corresponding to $X(t)$ and $X^n(t)$ are expected to approach the solutions of the two optimization problems  (\ref{min-0}) and (\ref{min-1}), respectively, $X(t)$ and $X^n(t)$ move towards $\check{\theta}$ and $\hat{\theta}_n$, respectively, and $V_n(t)$ and $V(t)$ are reaching their corresponding targets $n^{1/2} (\hat{\theta}_n- \check{\theta})$ and $[ \boldsymbol{I\!\! H}\! g( \check{\theta})]^{-1} \bsigma(\check{\theta} ) \bZ$. Below we will provide a framework to connect $(X(t), X^n(t))$ with 
%$(\check{\theta}, \hat{\theta}_n)$ and $V^n(t)$ with $n^{1/2}(\hat{\theta}_n - \check{\theta})$. 
%Solutions, $X(t)$, $X^n(t)$, $V(t)$, $V^n(t)$ from all ODEs (\ref{GD-c1}), (\ref{equ-2}), (\ref{GD-c2}),   (\ref{equ-3})-(\ref{limit-1}) 
%%(\ref{GD-c1}) \&(\ref{GD-c2}),  (\ref{equ-2}), (\ref{equ-3}), and (\ref{GD-limit-0})-(\ref{limit-1}) 
%live on $C(I\!\!R_+)$, and we can study their weak convergence on $I\!\!R_+$. Similarly we may adopt the Skorokhod space $D(I\!\!R_+)$ equipped with the Skorokhod metric for the weak convergence study of $x^n_\delta(t)$. 
Note that $C(I\!\!R_+)$ is a subspace of $D(I\!\!R_+)$, and because of the metrics used in $C(I\!\!R_+)$ and $D(I\!\!R_+)$, 
the weak convergence of these process on $D(I\!\!R_+ )$ is determined by their weak convergence on $D([0, T])$ for all integers $T$ 
only (see Billingsely (1999) and Jacod and Shiryaev (2002)).  Treating $X(t)$, $X^n(t)$, $V(t)$, $V^n(t)$, and $x^n_\delta(t)$
 as random elements in $D(I\!\!R_+)$, since the weak convergence results established in Theorems \ref{thm-1} and \ref{thm-1-1} hold 
 for $X^n(t)$ and $x^n_\delta(t)$ on $D([0, T])$ for any $T >0$, thus we may conclude from these established weak convergence results 
 that $V^n(t) = \sqrt{n} [X^n(t) - X(t)]$ and $\sqrt{n} [x^n_\delta(t) - X(t)]$ weakly converge to $V(t)$ on $D(I\!\!R_+)$. % $t \in [0, +\infty)$.

%Consider $V^n(t)$: it does not matter which first $n \rightarrow \infty$, $t \rightarrow \infty$ 

On the other hand, it is known that as $k \rightarrow \infty$, $x_k$ generated from algorithms (\ref{equ-GD1}) and (\ref{equ-Nest1}) converge 
to the solution $\check{\theta}$ of (\ref{min-0}) with speed of $(\delta k)^{-1}$ and $(\sqrt{\delta} k)^{-2}$, respectively, while as 
$t \rightarrow \infty$, their corresponding continuous curves $X(t)$ as the solutions of ODEs  (\ref{GD-c1}) and (\ref{equ-2}) approach $\check{\theta}$ with speed of $t^{-1}$ and $t^{-2}$, respectively (see Nesterov (1983) and Su et al. (2016)). 
%We may treat $\check{\theta}$ as the limits, $X(\infty)$, $x_\infty$, $x_\delta(\infty)$, of $x_k$, $X(t)$, and $x_\delta(t)$, as $k, t \rightarrow \infty$, respectively. 
Similarly for fixed $n$, as $k, t  \rightarrow \infty$, 
$x^n_k$ and $x^n_\delta(t)$ from algorithms (\ref{min-GD2}) and (\ref{min-Nest2}) and $X^n(t)$ from ODEs (\ref{GD-c2}) and (\ref{equ-3})
approach the solution $\hat{\theta}_n$ of (\ref{min-1}).
%and we may write their corresponding limit $\hat{\theta}_n$ as $x^n_\infty$, $x^n_\delta(\infty)$,
%and $X^n(\infty)$. Then $x_\infty = x_\delta(\infty) = X(\infty) = \check{\theta}$,  $x^n_\infty = x^n_\delta(\infty) = X^n(\infty) = \hat{\theta}_n$, 
%and $V^n(\infty)=\sqrt{n} [X^n(\infty)  - X(\infty)] = \sqrt{n} [x^n_\delta(\infty)  - X(\infty)] = \sqrt{n} ( \hat{\theta}_n - \check{\theta})$. 
For the weak limit $V(t)$ governed by (\ref{GD-limit-0}) or (\ref{GD-limit-00}),  as $t \rightarrow \infty$, 
both ODEs lead to $[\boldsymbol{I\!\! H}\! g(X(\infty)) ] V(\infty)  + \bsigma(X(\infty)) \bZ = 0$, or equivalently, $V(\infty)= [\boldsymbol{I\!\! H}\! g(X(\infty)) ]^{-1} \bsigma(X(\infty)) \bZ$. In fact, the solutions of (\ref{GD-limit-0}) and (\ref{GD-limit-00}) admit simple explicit expressions, for example,
\begin{align} \label{thm-1-2-equ1}
&  V(t) = %\exp\left[ - \int_0^t \boldsymbol{I\!\! H}\! g(X(s)) ds \right] 
  \int_0^t \exp\left[ - \int_s^t \boldsymbol{I\!\! H}\! g(X(u)) du \right] \bsigma(X(s)) ds \bZ, &\\
&  \mbox{ $\forall  \epsilon>0$, $\exists t_0>0$ such that $\forall s > t_0$}, 
\left |  [\boldsymbol{I\!\! H}\! g(X(s)) ]^{-1} \bsigma(X(s)) - [\boldsymbol{I\!\! H}\! g(X(\infty)) ]^{-1} \bsigma(X(\infty))   \right|  < \epsilon, \nonumber &\\
& \int_{t_0}^t \exp\left[  - \int_s^t \boldsymbol{I\!\! H}\! g(X(u)) du \right] \bsigma(X(s)) ds = \int_{t_0}^t \exp\left[ - \int_s^t \boldsymbol{I\!\! H}\! g(X(u)) du \right] \boldsymbol{I\!\! H}\! g(X(s))  \nonumber &\\
&    \left\{ [\boldsymbol{I\!\! H}\! g(X(s)) ]^{-1} \bsigma(X(s)) - [\boldsymbol{I\!\! H}\! g(X(\infty)) ]^{-1} \bsigma(X(\infty))   \right\}  ds  \nonumber &\\
&     + \int_{t_0}^t \exp\left[ - \int_s^t \boldsymbol{I\!\! H}\! g(X(u)) du \right] \boldsymbol{I\!\! H}\! g(X(s))  ds 
    [\boldsymbol{I\!\! H}\! g(X(\infty)) ]^{-1} \bsigma(X(\infty)).  \label{thm-1-2-equ2} &
\end{align}
Since the assumptions indicate that $\bsigma(X(s))$ and $\boldsymbol{I\!\! H}\! g(X(s))$ are bounded continuous on $[0, t_0]$, $\int_0^{t_0}  | \bsigma(X(s)) | ds$ is finite, 
and $\int_{t_0}^t \boldsymbol{I\!\! H}\! g(X(u)) du$ has finite eigenvalues. We immediately conclude that   
the eigenvalues of $\int_{t_0}^t \boldsymbol{I\!\! H}\! g(X(s)) ds$ are no less than the eigenvalues of $\int_{0}^t \boldsymbol{I\!\! H}\! g(X(s)) ds$ minus the maximum eigenvalue
of $\int^{t_0}_0 \boldsymbol{I\!\! H}\! g(X(s)) ds$, and thus diverge as $ t \rightarrow \infty$. Therefore, we can obtain 
\begin{align*}
& \left|  \int_0^{t_0} \exp\left[  - \int_s^t \boldsymbol{I\!\! H}\! g(X(u)) du \right] \bsigma(X(s)) ds \right | %&\\ &
\leq \left| \exp\left[  - \int_{t_0}^t \boldsymbol{I\!\! H}\! g(X(u)) du \right] \right|    \int_0^{t_0}  | \bsigma(X(s)) | ds  \rightarrow 0,  & \\ 
% \mbox{ as } t \rightarrow \infty ,  &
\end{align*}

\begin{align*}
& %\exp\left[ - \int_0^t \boldsymbol{I\!\! H}\! g(X(s)) ds \right] 
  \int_{t_0}^t \exp\left[ - \int_s^t \boldsymbol{I\!\! H}\! g(X(u)) du \right] \boldsymbol{I\!\! H}\! g(X(s))  ds = 1 - \exp\left[ - \int_{t_0}^t \boldsymbol{I\!\! H}\! g(X(s)) ds \right] \rightarrow 1, & \\
%&  \mbox{ $\forall  \epsilon>0$, $\exists t_0>0$ such that $\forall s > t_0$}, 
%\left |  [\boldsymbol{I\!\! H}\! g(X(s)) ]^{-1} \bsigma(X(s)) - [\boldsymbol{I\!\! H}\! g(X(\infty)) ]^{-1} \bsigma(X(\infty))   \right|  < \epsilon, &\\
& %\exp\left[ - \int_0^t \boldsymbol{I\!\! H}\! g(X(s)) ds \right]  
\int_{t_0}^t \left| \exp\left[ - \int_s^t \boldsymbol{I\!\! H}\! g(X(u)) du \right] \boldsymbol{I\!\! H}\! g(X(s)) \right|  %&\\&   
 \left|  [\boldsymbol{I\!\! H}\! g(X(s)) ]^{-1} \bsigma(X(s))  \right. &\\
 & \left.  - [\boldsymbol{I\!\! H}\! g(X(\infty)) ]^{-1} \bsigma(X(\infty))   \right|  ds & \\
& \leq \epsilon - \epsilon \left|  \exp\left[ - \int_{t_0}^t \boldsymbol{I\!\! H}\! g(X(s)) ds \right] \right| \leq \epsilon, & 
\end{align*}
which goes to zero, as we let $\epsilon \rightarrow 0$. Combining these results with (\ref{thm-1-2-equ1}) and (\ref{thm-1-2-equ2}) 
we conclude that as $t \rightarrow \infty$, 
\[ \int_0^t \exp\left[ - \int_s^t \boldsymbol{I\!\! H}\! g(X(u)) du \right] \bsigma(X(s)) ds \rightarrow [\boldsymbol{I\!\! H}\! g(X(\infty)) ]^{-1} \bsigma(X(\infty)), \]
and $V(t)$ converges in distribution to $[\boldsymbol{I\!\! H}\! g(X(\infty)) ]^{-1} \bsigma(X(\infty)) \bZ$. 

%we can easily show that if $\int_0^\infty \boldsymbol{I\!\! H}\! g(X(t)) dt = \infty$,  as $t \rightarrow \infty$, $V(t)$ approaches to $V(\infty)= [\boldsymbol{I\!\! H}\! g(X(\infty)) ]^{-1} \bsigma(X(\infty)) \bZ = [\boldsymbol{I\!\! H}\! g(\check{\theta}) ]^{-1} \bsigma(\check{\theta}) \bZ$, which is the exact weak limit  of  
%$\sqrt{n} ( \hat{\theta}_n - \check{\theta})=V^n(\infty)$. That is, $V^n(\infty)$ converges in distribution to $V(\infty)$ as $n \rightarrow \infty$.  
%As we have discussed in this section, the statistical arguments indicates that $\sqrt{n} ( \hat{\theta}_n - \check{\theta})=V^n(\infty)$ has an asymptotic distribution $[\boldsymbol{I\!\! H}\! g(\check{\theta})]^{-1} \bsigma(\check{\theta}) \bZ$. Since 
%$\sqrt{n} ( \hat{\theta}_n - \check{\theta})=V^n(\infty)$, $[\boldsymbol{I\!\! H}\! g(\check{\theta})]^{-1} \bsigma(\check{\theta}) \bZ= [\boldsymbol{I\!\! H}\! g(X(\infty)) ]^{-1} \bsigma(X(\infty)) \bZ$, which can be treated as the limit, denoted by $V(\infty)$, of $V(t)$ as $t \rightarrow \infty$. 
%With these treatments, we extend $[0, +\infty)$ further to $[0, +\infty]$ and consider $X(t)$, $x_\delta(t)$, 
%$X^n(t)$, $x^n_\delta(t)$, $V(t)$, and $V^n(t)$ on $t \in [0, \infty]$ and study the limits of $V^n(t)$ and $\sqrt{n} [x^n_\delta(t) - X(t)]$ on $[0, \infty]$.  With the augmentation of $t=\infty$, we establish  the following theorem.

%% file: p2-ML.tex
\subsection{Proofs of Theorems \ref{thm2}-\ref{thm4}}

Theorem \ref{thm2} is proved by Lemma \ref{lem-1}, with Theorems \ref{thm3} and \ref{thm4} shown in Lemma \ref{lem-7}, 
where both lemmas are established in this section below.
  
%%Denote by $\hat{Q}_n$ the empirical distribution of $U_1, \cdots, U_n$, and 
%For simplicity we will prove the case that mini-batches are sampled from the underlying distribution $Q$. Since mini-batch size $m$ is negligible in 
%comparison with data size $n$, and $\nabla\! \ell(\theta; U)$ has a moment generating function, the bootstrap sampling case can be handled 
%via strong approximation by converting the case into the proved scenario of sampling from the underlying distribution $Q$ 
%(Cs\"{o}rg\"{o} and Mason (1989), Cs\"{o}rg\"{o} et al. (1999), Massart (1989), Rio (1993a, b)). 
Denote by $\hat{Q}^*_{mk}$ the empirical distribution of mini-batch $U^*_{1k}, \cdots, U^*_{mk}$. Then 
%By KMT's strong approximation (Cs\"{o}rg\"{o} and Mason (1989) and Cs\"{o}rg\"{o} et al. (1999)) we have
%\[  \max_u |\hat{Q}_n(u) - Q(u) - n^{-1/2} B_n( Q(u)) |  = O(n^{-1} \log n), \]
%and for each $k$, 
%\[ P\left(  \max_u | \hat{Q}^*_{mk}(u) - \hat{Q}_n(u) - m^{-1/2} B^*_{mk}( Q(u)) | >  m^{-1/2} (x + c_2 \log m ) \right) \leq  e^{- c_1 x}, \]
%where $c_i$ are generic positive constants, and $B_n$ are Brownian bridges. 
%$B_n$ and $B_{mk}$ are Brownian bridges, and $B_{mk}$, $k=1, \cdots, N$, are independent. 
%Let $N = [T/\delta]+1$. Then $N = O( m^c)$ with $c$ given by Assumption 5.  Since $Q^*_{mk}$ are independent given $\bU_n$, 
%we have 
%\begin{align*}
%& P\left(  \max_{k \leq N } \max_u | \hat{Q}^*_{mk}(u) - \hat{Q}_n(u) - m^{-1/2} B^*_{mk}( Q(u)) | \leq  m^{-1/2} (x + c_2 \log m \right) & \\
%& \geq  \prod_{k=1}^{N} 
%   P\left(  \max_u | \hat{Q}^*_{mk}(u) - \hat{Q}_n(u) - m^{-1/2} B^*_{mk}( Q(u)) | \leq  m^{-1/2} (x + c_2 \log m \right)& \\
%& \geq \left[ 1 - e^{-c_1 x} \right]^{N} =  \left[ 1 - m^{-c_1 c_3 } \right]^{N} \sim 1 - N m^{-c_1 c_3 } = 1 - O(m^{c - c_1 c_3}), &
%\end{align*}
%where we take $x = c_3 \log m$ for some constant $c_3$ to satisfy $c - c_1 c_3> 1$ so that $\sum_m m^{c- c_1 c_3}$ converges. 
%Thus 
%\[  \max_{1 \leq k \leq T/\delta } \max_u | \hat{Q}^*_{mk}(u) - \hat{Q}_n(u) - m^{-1/2} B^*_{mk}( Q(u))| = O(m^{-1} \log m),  \]
%and 
\[ \nabla \!\hat{\cL}^m(\theta; \bU_{mk}^*) = \int \nabla \ell(\theta; u) \hat{Q}^*_{mk}(du), \;\; \]
 % \nabla \!\hat{\cL}^n(\theta; \bU_n) = \int \nabla \ell(\theta; u) \hat{Q}_{n}(du), \]
\[  \int \nabla \ell(\theta; u) Q(du) = E  [  \nabla \ell(\theta; U)] = \nabla g(\theta). \]
Let $R^m(\theta; \bU_m^*(t)) = (R^m_1(\theta;\bU_m^*(t)), \cdots, R^m_p(\theta;\bU_m^*(t)))^\prime$, where 
\begin{equation*}
R_{j}^{m}(\theta;\bU_m^*(t))=\sqrt{m}\left[ \frac{1}{m} \sum_{i=1}^m  \frac{\partial}{\partial \theta_j} \ell(\theta; U_{i}^*(t))-\frac{\partial}{\partial \theta_j}g(\theta)\right], \;\; j =1, \cdots, p.
\end{equation*}
We have 

\begin{align*}
& m^{-1/2} R^m(\theta; \bU_m^*(t))  = \int \nabla \ell(\theta; u) \hat{Q}^*_{mk}(du) - 
     \int \nabla \ell(\theta; u) Q(du) , &\\
%& = \int \nabla \ell(\theta; u, \hat{Q}_n) \hat{Q}^*_{mk}(du) - \int \nabla \ell(\theta; u) \hat{Q}_{n}(du)
%  +  \int \nabla \ell(\theta; u) \hat{Q}_{n}(du) & \\
%  & - \int \nabla \ell(\theta; u) Q(du), & \\
%%& = m^{-1/2} \int \nabla \ell(\theta; u, \hat{Q}_n) d B^*_{mk}( Q(u)) + n^{-1/2}  \int \nabla \ell(\theta; u) d B_n( Q(u))
%%    + O( m^{-1} \log m + n^{-1} \log n), & \\
& \nabla \!\hat{\cL}^m(x^m_{k-1}; \bU_{mk}^*) = \nabla g(x^m_{k-1}) + m^{-1/2} 
R^m(x^m_{k-1}; \bU^*_{mk}). &
%&= \nabla g(x^m_{k-1}) + \int \nabla \ell(x^m_{k-1}; u, \hat{Q}_n) d B^*_{mk}( Q(u)) + O( m^{-1} \log m) + O_P(n^{-1/2} ) , &
  \end{align*}          

It is easy to see that $R^m(x^m_{k-1}; \bU^*_{mk})$, $k=1, \cdots, T/\delta$, are martingale differences, and $H^m_\delta(t)$ is a martingale. 
We may use  martingale theory (He et al. (1992), Jacod and Shiryaev (2003)) to establish weak convergence of $H^m_\delta(t)$ to stochastic integral
$H(t)$. Below we will use a more direct approach to prove the weak convergence and obtain further convergence rate results. 

\begin{lem} \label{lem-1}
%Define partial sum process for $t \in [0, T]$, 
%\[  H^m_\delta(t) = [\delta/(m T)]^{1/2}  \sum_{k=1}^{[T/\delta] t} \left[  \int \nabla \ell(x^m_{k-1}; u, \hat{Q}_n) \hat{Q}^*_{mk}(du) - \int \nabla \ell(x^m_{k-1}; u) \hat{Q}_{n}(du) \right]. \]
As $\delta \rightarrow 0$ and $m \rightarrow \infty$, $H^m_\delta(t)$ weakly converges to $H(t)=\int_0^t \bsigma(X(u)) d\bB(u)$,
$t \in [0, T]$.
\end{lem}
Proof. Let  

\begin{align*}
&  \check{H}^m_\delta(t) = (m \delta)^{1/2}  \sum_{ t_k \leq t }  \left[ \nabla \!\hat{\cL}^m(X(t_{k-1}); \bU_{m}^*(t_k)) 
  - \nabla g (X(t_{k-1}) \right] & \\
&=  (m \delta)^{1/2}  \sum_{k=1}^{[t /\delta] } \left[  \int \nabla \ell(X((k-1)\delta); u) \hat{Q}^*_{mk}(du) \right. &\\
& \left. - \int \nabla \ell(X((k-1)\delta); u) Q(du) \right]. 
\end{align*}
%Denote by $E_*$ and $Var_*$ the expectation and variance with respect to bootstrap samples under $\hat{Q}_n$, respectively. 
Note that 
\[   E\left[ \int \nabla \ell(\theta; u) \hat{Q}^*_{mk}(du)\right] =  \int \nabla \ell(\theta; u) Q(du), \]
\begin{align*}
& \bsigma^2(\theta)= m Var\left[ \int \nabla \ell(\theta; u) \hat{Q}^*_{mk}(du)\right] =  \int [\nabla \ell(\theta; u) ]^2 Q(du) &\\
&  - \left[ \int \nabla \ell(\theta; u) Q(du)  \right]^2   , & 
\end{align*}
which are the mean and variance of $\nabla \ell(\theta; U)$, respectively.  Since %conditional on $\bU_n$,
$\bU^*_{mk}$, $k=1, 2,\cdots, [T/\delta]$, are independent, then % for fixed $m,n$, as $\delta \rightarrow 0$, 
$\check{H}^m_\delta (t)$ is a normalized partial sum process for independent random variables and weakly converges to 
$\int_0^t \bsigma(X(u)) d\bB(u)$. Indeed, its finite-dimensional distribution convergence can be 
easily established through Lyapunov Central Limit Theorem with assumptions A3-A4 and Lemma \ref{lem-3x} below, 
and its tightness can be shown by the fact that for $r \leq s \leq t$, 
\begin{equation} \label{tight-xx}
    E\left\{ \left |  \check{H}^m_\delta(t) - \check{H}^m_\delta(s) \right | ^2 \left | \check{H}^m_\delta(s) - \check{H}^m_\delta(r) \right |^2 \right \}  \leq 
     [ \Upsilon(t) - \Upsilon(r)]^2, 
\end{equation}
where $\Upsilon(\cdot)$ is a continuous non-decreasing function on $[0, T]$ (Billingsley (1999, equation (13.14)  \& theorem 13.5)). To establish (\ref{tight-xx}), we have 
that, because of independence, 
\begin{align*}
& E\left\{ \left |  \check{H}^m_\delta(t) - \check{H}^m_\delta(s) \right | ^2 \left | \check{H}^m_\delta(s) - \check{H}^m_\delta(r) \right | ^2 \right \} 
 = E\left\{ \left | \check{H}^m_\delta(t) - \check{H}^m_\delta(s) \right | ^2 \right\} E \left\{ \left | \check{H}^m_\delta(s) - \check{H}^m_\delta(r) \right |^2 \right \} 
& \\
& = \delta^2 \sum_{s< k \delta \leq t} tr[ \bsigma^2(X((k-1)\delta)) ]  \sum_{r< k \delta \leq s} tr[\bsigma^2(X((k-1)\delta))] & \\
& \sim \int_s^t  tr[\bsigma^2(X(u)) ] du \int_r^s  tr[ \bsigma^2(X(u))]  du.  &
 \end{align*}
Since $X(t)$ is a deterministic bounded  continuous curve, and $\bsigma^2(\theta)$ is a continuous positive definite matrix, 
\[  \int_s^t  tr[\bsigma^2(X(u)) ] du \int_r^s tr[ \bsigma^2(X(u)) ] du \leq  \left[ \int_r^t  tr[\bsigma^2(X(u))  ] du \right]^2 \equiv  [ \Upsilon(t) - \Upsilon(r)]^2. \]

%By Assumption A1, we have 
%\[  |\nabla \ell(X(t); U_i) | \leq  |\nabla \ell(x_0; U_i) | + h_1(U_i) \| X(t) - x_0\|, \]
%and the Marcinkiewicz-Zygmund theorem implies that 
%$\sup_{n} \sum_{i=1}^n |\nabla \ell(x_0; U_i) |^2 / n $ and $\sup_{n} \sum_{i=1}^n h^2_1(U_i) / n $ have finite moments.  
%As $n \rightarrow \infty$, $\bsigma_n(\theta) $ converges to $\bsigma(\theta)$, and $X(t)$ is a continuous curve from $x_0$ to $\check{\theta}$,
%using dominated convergence theorem 
%we have $\int_0^T |\bsigma_n(X(t)) - \bsigma(X(t))|^2 dt \rightarrow 0$, which implies by the quadratic variation convergence theorem 
%that $\int_0^t \sigma_n(X(t)) d\bB(u)$ weakly converge to $\int_0^t \sigma(X(t)) d\bB(u)$. 

We have shown that as $\delta \rightarrow 0$ and $m \rightarrow \infty$, %then $m,n \rightarrow \infty$, 
$\check{H}^m_\delta (t)$ weakly converges to $H(t)$. 
By Central Limit Theorem for stochastic processes (Jacod and Shiryaev 2003, Theorem 3.11 of Ch. VIII), %p.473
we obtain that quadratic variation $[ \check{H}^m_\delta, \check{H}^m_\delta]_t$ converges in probability to $[H, H]_t$ for $ t \in [0, T]$. 

The Lipschitz of $\nabla \ell(\theta; u, Q)$ in $\theta$ implies the Lipschitz of $\nabla \hat{\cL}^m(\theta; \bU^*_{mk}, Q)$ 
(which is proved at the beginning of the proof of Lemma \ref{lem-2x} below), and Lemma \ref{lem-2} below indicates that as $\delta \rightarrow 0$ and $m \rightarrow \infty$, $x^m_k - X( k \delta)$ converges to zero in probability (with order $\delta + m^{-1/2} \delta^{1/2}$) uniformly over $1 \leq k \leq T/\delta$. These two results together with the  Lipschitz of $\nabla g(\theta)$ immediately 
show that %$\max_{t \leq T} |\check{H}^m_\delta (t) - H^m_\delta(t) | = o_P(1)$ 
$$\max_{t \leq T} \left |[\check{H}^m_\delta, \check{H}^m_\delta]_t - [H^m_\delta, H^m_\delta]_t  \right| = 
  O_P\left( (m \delta) \; \delta^{-1}  [\delta + m^{-1/2} \delta^{1/2}] \right) =  o_P(1),$$
and hence quadratic variation $[ H^m_\delta, H^m_\delta]_t$ also converges in probability to $[H, H]_t$ for $t \in [0, T]$. 
An application of Central Limit Theorem for stochastic processes (Jacod and Shiryaev 2003, Theorem 3.11 of Ch. VIII) leads to that 
as $\delta \rightarrow 0$ and $m \rightarrow \infty$, $H^m_\delta (t)$ weakly converges to $H(t)$, 
that is, $\check{H}^m_\delta (t) $ and $H^m_\delta(t)$ share the same weak convergence limit $H(t)$.

\begin{lem} \label{lem-2x}
We have 
\[ %\max_{k \leq T/\delta}   |x^m_k | = O_P(1), \;\; 
\max_{k \leq T/\delta}   |x^m_k - x_k| = O_P(m^{-1/2}), \]%O_P(\delta +m^{-1/2}  \delta^{1/2} + n^{-1/2}), \]
where $x_k$ and $x^m_k$ are defined by (\ref{equ-GD1}) and (\ref{min-GD3}), respectively. 
%$x^m_k$ is  defined by (\ref{min-GD3}). 
\end{lem}
Proof. Let $\zeta(\bU^*_{mk}) = \frac{1}{m} \sum_{i=1}^m h_1(U^*_{ik})$, which converges in probability to $E[h_1(U)]$ as $m \rightarrow \infty$. 
Then 
\begin{align*}
& | \nabla \hat{\cL}^m(\theta; \bU^*_{mk}) - \nabla \hat{\cL}^m(\vartheta; \bU^*_{mk}) | \leq \zeta(\bU^*_{mk}) |\theta - \vartheta |,  \;\;
 | \boldsymbol{I\!\! H}\! \hat{\cL}^m(\theta; \bU^*_{mk}) | \leq \zeta(\bU^*_{mk}), & \\
&  |\check{\theta} - x^m_k| \leq |\check{\theta} - x^m_{k-1}|+  \delta | \nabla \hat{\cL}^m(x^m_{k-1}; \bU^*_{mk}) - \nabla \hat{\cL}^m(\check{\theta}; \bU^*_{mk})| + \delta |\nabla \hat{\cL}^m(\check{\theta}; \bU^*_{mk})|  & \\
&  \leq ( 1 + \delta \zeta(\bU^*_{mk}))  |\check{\theta} - x^m_{k-1}|  + \delta | \nabla \hat{\cL}^m(\check{\theta}; \bU^*_{mk})| &\\
& \leq \left ( 1 + \delta E[h_1(U)] + O_P(\delta m^{-1/2}) \right)^k &\\
& +  \left ( 1 + \delta E[h_1(U)] + O_P(\delta m^{-1/2}) \right)^k \delta \sum_{j=1}^k  [ |\nabla g(\check{\theta})| + m^{-1/2} |R^m(\check{\theta}; \bU^*_{mj})|] & \\
& \leq e^{ T E[h_1(U)] } [1 + |\nabla g(\check{\theta})| + O_P(m^{-1/2})] = e^{ T E[h_1(U)] } [1 + O_P(m^{-1/2})], &
\end{align*}
namely, $x^m_k$ are bounded uniformly over $k \leq T/\delta$. On the other hand, we have 
\begin{align*}
&  x^m_k - x_k = x^m_{k-1} - x_{k-1} - \delta [ \nabla \hat{\cL}^m(x^m_{k-1}; \bU^*_{mk}) - \nabla g(x_{k-1})] & \\
& = x^m_{k-1} - x_{k-1} - \delta [ \nabla \hat{\cL}^m(x^m_{k-1}; \bU^*_{mk}) - \nabla \hat{\cL}^m(x_{k-1}; \bU^*_{mk})] - \delta m^{-1/2} R^m(x_{k-1}; \bU^*_{mk})& \\
& = ( x^m_{k-1} - x_{k-1} ) [1 - \delta \boldsymbol{I\!\! H}\! \hat{\cL}^m(x^m_{\xi,k-1}; \bU^*_{mk})] - \delta m^{-1/2} R^m(x_{k-1}; \bU^*_{mk})& \\
& = - \delta m^{-1/2} \sum_{j=1}^k [1 - \delta \boldsymbol{I\!\! H}\! \hat{\cL}^m(x^m_{\xi,j-1}; \bU^*_{mj})]^j R^m(x_{j-1}; \bU^*_{mj}),
\end{align*}
where $x^m_{\xi, j-1}$ is between $x_{j-1}$ and $x^m_{j-1}$. Using $\zeta(\bU^*_{mj} ) \rightarrow E[h_1(U)]$ and assumption A4 we obtain for $j, k \leq T/\delta$,
\begin{align*}
&  |[1 - \delta \boldsymbol{I\!\! H}\! \hat{\cL}^m(x^m_{\xi,j-1}; \bU^*_{mj})]^j| \leq [1 + \delta \zeta(\bU^*_{mj})]^{T/\delta} \leq e^{T E[h_1(U)]} [1 + O_P(m^{-1/2})], &\\
&  R^m(x_{j-1}; \bU^*_{mj}) \sim \bsigma(x_{j-1}) \bZ = O_P(1), & \\
& | x^m_k - x_k | \leq \delta m^{-1/2} \sum_{j=1}^k |1 + \delta \zeta(\bU^*_{mj})]^{T/\delta} | R^m(x_{j-1}; \bU^*_{mj}) | = O_P(k \delta m^{-1/2}) 
   = O_P(m^{-1/2}) . &
\end{align*}

\begin{lem} \label{lem-2}
\[ \max_{k \leq T/\delta}   |X(k\delta) - x^m_k | = O_P(\delta +m^{-1/2}  \delta^{1/2} ) , \] %+ n^{-1/2}), \]
where $X(t)$ and $x^m_k$ are defined by (\ref{GD-c1}) and (\ref{min-GD3}), respectively. 
\end{lem}
Proof. For $k =1,\cdots, T/\delta$, %given $\hat{Q}_n$, 
\[ \int \nabla \ell(x^m_{k-1}; u) \hat{Q}^*_{mk}(du) - \int \nabla \ell(x^m_{k-1}; u) Q(du) \]
are martingale differences with conditional mean zero and conditional variance $\bsigma^2(x^m_{k-1})/m$. 
%$Var(\int [\nabla \ell(x^m_{k-1}; u)] ^2 d\hat{Q}_n(u))\leq C$,
Since $x_k$ in  (\ref{equ-GD1}) is the Euler approximation of solution $X(t)$ of ODE (\ref{GD-c1}), the standard ODE theory shows  
\begin{equation} \label{ODE-Euler}
  \max_{k \leq T/\delta}   |x_k - X(k \delta) | = O(\delta). 
\end{equation}
By Lemma \ref{lem-2x} we have that with probability tending to one, $x^m_{k-1}, k =1, \cdots, T/\delta, $ fall inside a neighborhood of solution curve of ODE (\ref{GD-c1}), and thus the maximum of $\bsigma^2(x^m_{k-1})$, $k=1, \cdots, T/\delta$, is bounded. Applying Burkholder's inequality (Chow and Teicher (1997), He et al. (1992), Jacod and Shiryaev (2003)) we obtain 
\[  \max_{1 \leq k \leq T/\delta} \left| \sqrt{m} \sum_{\ell=1}^k \left[
 \int \nabla \ell(x^m_{\ell-1}; u) \hat{Q}^*_{m\ell}(du) - \int \nabla \ell(x^m_{\ell-1}; u) Q(du) \right]
    \right| = O_P( \delta^{-1/2}), \]
that is, 
\[ \max_{k \leq T/\delta} \left | m^{-1/2} \sum_{\ell=1}^k R^m(x^m_{\ell-1}; \bU^*_{m\ell}) \right | = O_P( m^{-1/2} \delta^{-1/2}). \] % + n^{-1/2}). \]
Therefore, for $k =1, \cdots, T/\delta$,
\begin{align*}
& x^m_k = x_0 - \delta \sum_{\ell=1}^k  \nabla g(x^m_{\ell-1}) -  m^{-1/2} \delta   \sum_{\ell=1}^k R^m(x^m_{\ell-1}; \bU^*_{m\ell}) &\\
  %m^{-1/2} \delta  \sum_{\ell=1}^k  \int \nabla \ell(x^m_{\ell-1}; u, \hat{Q}_n) d B^*_{m \ell}( Q(u)) &\\
   % & +\, O_P(n^{-1/2}) +  O(m^{-1} \log m) & \\
&=x_0 -  \delta \sum_{\ell=1}^k  \nabla g(x^m_{\ell-1}) - O_P( m^{-1/2} \delta^{1/2} ). % + n^{-1/2}).  %k \delta n^{-1/2}). 
   & 
\end{align*}
%where the O terms are uniformly over $1 \leq k \leq T/\delta$. 
and with the same initial value $x_0$, comparing the expressions for $x_k$ and $x^m_k$ we obtain 
\begin{align*}
&  x^m_k - x_k %= x_{k-1} - x^m_{k-1} - \delta [ \nabla \hat{\cL}^m(x^m_{k-1}; \bU^*_{mk}) - \nabla g(x_{k-1})] & \\
  = x^m_{k-1} - x_{k-1} - \delta [ \nabla g(x^m_{k-1}) - \nabla g(x_{k-1})] - \delta m^{-1/2} R^m(x^m_{k-1}; \bU^*_{mk}) & \\
& = \delta \sum_{\ell=1}^k  [ \nabla g(x_{\ell-1}) - \nabla g(x^m_{\ell-1}) ] -  \delta m^{-1/2}  \sum_{\ell=1}^k R^m(x^m_{\ell-1}; \bU^*_{m\ell}).  &
\end{align*}
Using the L-Lipschitz assumption on $\nabla g(\cdot)$ we conclude for $k =1, \cdots, T/\delta$,
\begin{align*}
& |x^m_k - x_k| \leq   L \delta \sum_{\ell=1}^k  | x^m_{\ell-1} - x_{\ell-1}| + \delta m^{-1/2} \left |  \sum_{\ell=1}^k R^m(x^m_{\ell-1}; \bU^*_{m\ell}) \right| &\\
& \leq L T \max_{1 \leq \ell \leq k} | x^m_{\ell-1} - x_{\ell-1}| + \delta m^{-1/2} \left |  \sum_{\ell=1}^k R^m(x^m_{\ell-1}; \bU^*_{m\ell}) \right|.
\end{align*}
Finally we can easily show by induction that 
\[ \max_{k \leq T/\delta}   |x^m_k - x_k | = O_P(m^{-1/2}  \delta^{1/2} ). %+ n^{-1/2} ). 
\]
The lemma is a consequence of above result and (\ref{ODE-Euler}).

The following lemma refines the order regarding $m^{-1/2} \delta^{1/2}$ in Lemma \ref{lem-2}.
\begin{lem} \label{lem-3}
%Let $X^m_\delta(t)$ be the solution of the following stochastic differential equation 
%\begin{equation} \label{stoch-2}
%  dX^m_\delta(t) = - \nabla g(X^m_\delta (t)) dt - m^{-1/2} T^{1/2} \delta^{1/2} \bsigma(X(t)) d\bB(t). 
%\end{equation}
We have 
\[    \max_{k \leq T/\delta}  |x^m_k - X^m_\delta(k \delta) | = o_P( m^{-1/2}  \delta^{1/2}) + O_P(  \delta + \delta m^{-1/2} |\log \delta|^{1/2}), %+ n^{-1/2} 
\]
\[   \max_{t \leq T} |x^m_\delta(t) - X^m_\delta(t)| = o_P( m^{-1/2}  \delta^{1/2}) + O_P( \delta |\log \delta|^{1/2}),  % + n^{-1/2} 
\]
where $X^m_\delta(t)$ is given by (\ref{GD-stoch1}), and 
$x^m_k$ and $x^m_\delta(t)$ are defined by (\ref{min-GD3}) and (\ref{GD-xt}), respectively. 
\end{lem}
Proof. With weak convergence of $H^m_\delta(t)$ to $H(t)$ in Lemma \ref{lem-1}, 
by Skohorod's representation we may realize $H^m_\delta(t)$ and $H(t)$ 
on some common probability spaces such that as $\delta \rightarrow 0$ and $m \rightarrow \infty$, under the metric in $D([0, T])$, $H^m_\delta(t) - H(t)$ is $o_P(1)$. We may use arguments based on Lemma \ref{lm5} (in Section \ref{sec-thm7}) and stochastic equi-continuity to establish the convergence of $H^m_\delta(t) - H(t)$ under the maximum norm. Here we adopt a direct approach. 
Consider linear interpolation $\tilde{H}^m_\delta(t)$ between the values of $H^m_\delta( k \delta)$, $k=1, \cdots, T/\delta$,
which satisfies 
\begin{align*}
&  \max_{t \leq T} |\tilde{H}^m_\delta(t) - H^m_\delta(t) | \leq \delta^{1/2} \max_{k \leq T/\delta} | R^m(x^m_{k-1}; \bU^*_{mk}) | . & 
\end{align*}
By assumptions A1-A2, we have 
\begin{align*}
&  |[ \nabla \ell(x^m_{k-1}; U^*_{ik}) - \nabla g(x^m_{k-1}) ] - [ \nabla \ell(X((k-1)\delta); U^*_{ik}) - \nabla g(X((k-1)\delta)) ] |  &\\
& \leq [ h_1(U^*_{ik}) + L] | x^m_{k-1} - X((k-1)\delta)|, &
\end{align*}
%\[  |\nabla \ell(X(t); U^*_{ik})  | \leq  |\nabla \ell(u; U^*_{ik}) | + h_1(U^*_{ik}) \| X(t) - u\| , \]
and then 
\begin{align*}
&| R^m(x^m_{k-1}; \bU^*_{mk}) \leq  | R^m(X((k-1)\delta); \bU^*_{mk})| &\\
& +  m^{-1/2}  \sum_{i=1}^m [h_1(U^*_{i k}) +L] | x^m_{k-1} - X((k-1)\delta)|,  & \\
&  \max_{t \leq T} |\tilde{H}^m_\delta(t) - H^m_\delta(t) | \leq \delta^{1/2} \max_{k \leq T/\delta} | R^m(X((k-1)\delta); \bU^*_{mk})| &\\
& + \delta^{1/2}  \max_{k \leq T/\delta} \left\{ \frac{1}{m} \sum_{i=1}^m h_1(U^*_{i k}) + L \right\}  m^{1/2} \max_{k \leq T/\delta} | x^m_{k-1} - X((k-1)\delta)|. 
\end{align*}
Lemma \ref{lem-2} implies $ m^{1/2} \max_{k \leq T/\delta} | x^m_{k-1} - X((k-1)\delta)| =  m^{1/2} O_P( \delta + m^{-1/2} \delta^{1/2} ) %+ n^{-1/2}) 
= O_P( m^{1/2} \delta + \delta^{1/2}  %+  m^{1/2} n^{-1/2}
) = o_P(1)$, and by Lemma \ref{lem-3x} below we derive that  
$\max_{t \leq T} |\tilde{H}^m_\delta(t) - H^m_\delta(t) | =o_P(\delta^{1/4} |\log \delta| )$. Thus, 
$\tilde{H}^m_\delta(t)$ weakly converges to $H(t)$ in $D([0,T])$. As both $\tilde{H}^m_\delta(t)$ and $H(t)$ live in $C([0, T])$, the 
weak convergence of $\tilde{H}^m_\delta(t)$ to $H(t)$ holds in $C([0,T])$. 
Again by Skohorod's representation we may realize $\tilde{H}^m_\delta(t)$ and $H(t)$ 
on some common probability spaces such that as $\delta \rightarrow 0$ and $m \rightarrow \infty$, 
$ \max_{t \leq T} |\tilde{H}^m_\delta(t) - H(t) | = o_P(1)$, 
and hence $\max_{t \leq T} |H^m_\delta(t) - H(t) | = o_P(1)$.

%With weak convergence of $H^m_\delta(t)$ to $H(t)$, by Skohorod's representation we may realize $H^m_\delta(t)$ and $H(t)$ 
%on some common probability space such 
%that as $\delta \rightarrow 0$, $m,n \rightarrow \infty$, 
%\[   \max_{t \leq T} |H^m_\delta(t) - H(t) | = o(1). \]
Note that for $1 \leq k \leq T/\delta$,
\begin{align*}
& \delta \nabla \!\hat{\cL}^m(x^m_{k-1}; \bU_{mk}^*) = \delta \nabla g(x^m_{k-1}) + m^{-1/2}  \delta^{1/2} [H^m_\delta(k \delta) - H^m_\delta((k-1) \delta)] 
  %+ O_P(\delta n^{-1/2})
  , &
\end{align*}
\begin{align*}
&  x^m_k - x^m_{k-1}
= - \delta \nabla g(x^m(t_{k-1})) - m^{-1/2} \delta^{1/2} [H^m_\delta(k \delta) - H^m_\delta((k-1) \delta)] , %+ O_P(\delta n^{-1/2}), 
& \\
&  x^m_k = x_ 0  - \delta  \sum_{\ell=1}^k \nabla g(x^m_{\ell-1} )   - m^{-1/2}  \delta^{1/2} H^m_\delta(k \delta) %+   O_P(k \delta n^{-1/2} ) 
&\\
& = x_ 0  - \delta \sum_{\ell=1}^k \nabla g(x^m_{\ell-1} )   - m^{-1/2} \delta^{1/2} H(k \delta) + o_P(m^{-1/2} \delta^{1/2} ) %+O_P( n^{-1/2} ) 
. &
    %O_P( \delta^{1/2} + m^{-1/2} + \delta^{-1/2} n^{-1/2} + \delta^{-1/2} m^{-1} \log m)  & \\
%& = -\nabla g(x^m(t_{k-1}) )(t_k - t_{k-1})   - m^{-1/2} T^{1/2} \delta^{1/2} [H(t_k) - H(t_{k-1}) ] &\\
%& +   O(\delta m^{-1} \log m) + O_P(\delta n^{-1/2} ) + O_P( m^{-1/2}  \delta+ m^{-1} \delta^{1/2} + n^{-1/2} + m^{-1} \log m).  &     
\end{align*}
%where the last O and o terms are uniformly over $1 \leq k \leq T/\delta$.

%and $H^m_\delta(t)$ and $\check{H}_\delta(t)$ have the same weak limit wit rate of convergence  $\delta^{1/2} \log \delta$.

%\begin{align*}
%& \nabla \!\hat{L}^m(x^m_{k-1}; \bU_{mk}^*, \hat{Q}_n) = \nabla g(x^m_{k-1}) + m^{-1/2} T^m(x^m_{k-1}; \bU^*_{mk},\hat{Q}_n) &\\
%& = \nabla g(x^m_{k-1}) + m^{-1/2} \int \nabla \ell(x^m_{k-1}; u, \hat{Q}_n) d B^*_{mk}( Q(u)) + n^{-1/2}  \int \nabla \ell(x^m_{k-1}; u) d B_n( Q(u)) + O( m^{-1} \log m + n^{-1} \log n) & \\
%& = \nabla g(x^m_{k-1}) + m^{-1/2} T^{1/2} \delta^{-1/2} [H^m_\delta(k \delta) - H^m_\delta((k-1) \delta)] + 
%   O(m^{-1} \log m) + O_P(n^{-1/2}), &\\
%& \delta \nabla \!\hat{L}^m(x^m_{k-1}; \bU_{mk}^*, \hat{Q}_n) = \delta \nabla g(x^m_{k-1}) + m^{-1/2} T^{1/2} \delta^{1/2} [H^m_\delta(k \delta) - H^m_\delta((k-1) \delta)] + 
%   O(\delta m^{-1} \log m) + O_P(\delta n^{-1/2}), &\
%\end{align*}
%\begin{align*}
%&  x^m(t_k) - x^m(t_{k-1}) 
%= - \delta \nabla g(x^m(t_{k-1}) ) - m^{-1/2} T^{1/2} \delta^{1/2} [H^m_\delta(k \delta) - H^m_\delta((k-1) \delta)] + 
%   O(\delta m^{-1} \log n) + O_P(\delta n^{-1/2}) & \\
%& =  -\nabla g(x^m(t_{k-1}) (t_k - t_{k-1})   - m^{-1/2} T^{1/2} \delta^{1/2} [H(t_k) - H(t_{k-1}) ] &\\
%& +    O(\delta m^{-1} \log n) + O_P(\delta n^{-1/2} ) +O( m^{-1/2} T^{1/2} \delta \log \delta), & 
%\end{align*}
Define $\check{x}^m_0=x_0$, and 
\begin{equation} \label{GD-xcheck}
 \check{x}^m_k - \check{x}^m_{k-1} = -  \delta \nabla g(\check{x}^m_{k-1} )  - m^{-1/2} \delta^{1/2} [H(k \delta) - H((k-1)\delta) ]. 
 \end{equation}
Then the situation is the same as in the last proof part of Lemma \ref{lem-2},  and the same argument can be used to 
derive a recursive expression for $x^m_k - \check{x}^m_k$ and prove by induction that 
\[ \max_{k \leq T/\delta} |x^m_k - \check{x}^m_k| = o_P( m^{-1/2}  \delta^{1/2}) . %+ O_P(n^{-1/2}). 
\]
%O_P( m^{-1/2}  \delta+ m^{-1} \delta^{1/2} + n^{-1/2} + m^{-1} \log m), \]
%O(m^{-1} \log m) + O_P(n^{-1/2} ) + O( m^{-1/2} T^{1/2} \log \delta).  \]
The lemma is a consequence of above result and Lemma \ref{lem-4} below.

\begin{lem} \label{lem-3x}
$$\sup E\{ | R^m(X(t); \bU^*_{mk}) |^4:    t \in [0, T], k =1, \cdots, T/\delta\} < \infty, $$
\[ \delta^{1/2}  \max_{k \leq T/\delta} \left\{ \frac{1}{m} \sum_{i=1}^m h_1(U^*_{i k})  - E[h_1(U)]  \right\} = O_P(\delta^{1/4} |\log \delta|), \]
\[ \delta^{1/2} \max_{k \leq T/\delta} | R^m(X((k-1)\delta); \bU^*_{mk})|  = O_P( \delta^{1/4} |\log \delta| ).  \]
\end{lem}
Proof.  Direct calculations lead to 
\begin{align*}
&  P\left( \delta^{1/2} \max_{k \leq T/\delta} | R^m(X((k-1)\delta); \bU^*_{mk})| >  \delta^{1/4} |\log \delta|  \right) &\\
& = 1 -   \prod_{k \leq T/\delta}  P\left( \delta^{1/4} | R^m(X((k-1)\delta); \bU^*_{mk})| \leq  |\log \delta| \right) &\\    %\left |\bU_n \right) \right. & \\
\end{align*}
\begin{align*}
& \leq1 -   \prod_{k \leq T/\delta} \left [1 - \delta  E\left\{ | R^m(X((k-1)\delta); \bU^*_{mk}) |^4  \right \}   /|\log \delta|^4 \right] & \\
%& \leq 1 -   \exp\left[  - 2 \sum_{k \leq T/\delta}  \delta^{2} E\{ (R^m(X((k-1)\delta); \bU^*_{mk}))^4 | \bU_n \}\delta^{-1}/ |\log \delta|^4\right], & 
%\end{align*}
%and 
%\begin{align*}
%& P\left( \delta^{1/2} \max_{k \leq T/\delta} | R^m(X((k-1)\delta); \bU^*_{mk})| >  \delta^{1/4} |\log \delta| \right)  & \\
%& \leq  1 -   \exp  \left[  - 2 \sum_{k \leq T/\delta}  \delta^{2} E\left\{ (R^m(X((k-1)\delta); \bU^*_{mk}))^4 \right\} \delta^{-1}/ |\log \delta|^4  \right]  & \\
& \leq  1 - \exp\left [ - 2 T \tau / |\log \delta|^4 \right ] \sim 2 T \tau / |\log \delta|^4 \rightarrow 0, & 
\end{align*}
%as $\delta/b^4 \rightarrow \infty$ 
where we use Chebyshev's inequality, $\log (1 - u) \geq - 2 u$ for $0<u<0.75$, and 
$\tau=\sup_{t, k} E\{ | R^m(X(t); \bU^*_{mk}) |^4 \} \equiv \sup E\{ | R^m(X(t); \bU^*_{mk}) |^4:    t \in [0, T], k =1, \cdots, T/\delta\}
$ whose finiteness will be shown below.  Indeed,  it is enough to show that each component of
$R^m(X(t); \bU^*_{mk})$ has finite fourth moment uniformly over $t \in [0, T], k =1,\cdots, T/\delta$, and thus we need to prove it only in the one 
dimensional case with gradient equal to the partial derivative. With this simple set-up we have 
%$\leq C \max_t E\{ | R^m(X(t); \bU_{mk}) |^4\}$, and $E\{ | R^m(X(t); \bU_{mk}) |^4\} \leq  E [ \{\nabla \ell(X(t); U) - \nabla g(X(t))\}^4] /m + [Var( \nabla \ell(X(t)); U)) ]^2$ is bounded.
%\begin{align*}
%& [R^m(X(t); \bU_{mk})]^4 = m^{-2} \left[ \sum_{i=1}^m \{ \nabla \ell(X(t); U_{ik}) - \nabla g(X(t)) \} \right]^4 & \\
%& = m^{-2} \sum_{i \neq j} \{ \nabla \ell(X(t); U_{ik}) - \nabla g(X(t)) \}^2 \{ \nabla \ell(X(t); U_{jk}) - \nabla g(X(t)) \}^2 & \\
%& + m^{-2}  \sum_{i=1}^m \{ \nabla \ell(X(t); U_{ik}) - \nabla g(X(t)) \}^4 + \mbox{odd power terms}, & \\
%&  E\{ | R^m(X(t); \bU_{mk}) |^4\}  = m^{-2}  \sum_{i=1}^m E [\{ \nabla \ell(X(t); U_{ik}) - \nabla g(X(t)) \}^4]  & \\
%& + m^{-2} \sum_{i \neq j} E [\{ \nabla \ell(X(t); U_{ik}) - \nabla g(X(t)) \}^2] E [\{ \nabla \ell(X(t); U_{jk}) - \nabla g(X(t)) \}^2] & \\
%& \leq \{ E [\{ \nabla \ell(X(t); U_{1k}) - \nabla g(X(t)) \}^2]\}^2 + E[ \{ \nabla \ell(X(t); U_{1k}) - \nabla g(X(t)) \}^4]/m, &
%\end{align*}
%where we use the fact that all odd power terms have mean zero factor $\nabla \ell(X(t); U_{ik}) - \nabla g(X(t))$, 
\begin{align*}
& | R^m(X(t); \bU^*_{mk}) | ^4 = m^{-2} \left[ \sum_{i=1}^m \{ \nabla \ell(X(t); U^*_{ik}) - \nabla g(X(t)) \} \right]^4 & \\
& = m^{-2} \sum_{i \neq j} \{ \nabla \ell(X(t); U^*_{ik}) - \nabla g(X(t)) \}^2 \{ \nabla \ell(X(t); U^*_{jk}) - \nabla g(X(t)) \}^2 & \\
& + m^{-2}  \sum_{i=1}^m \{ \nabla \ell(X(t); U^*_{ik}) - \nabla g(X(t)) \}^4 + \mbox{odd power terms}, & \\
&  E\{ | R^m(X(t); \bU^*_{mk}) |^4\} = m^{-2}  \sum_{i=1}^m E [\{ \nabla \ell(X(t); U^*_{ik}) - \nabla g(X(t)) \}^4]  & \\
& + m^{-2} \sum_{i \neq j} E [\{ \nabla \ell(X(t); U^*_{ik}) - \nabla g(X(t)) \}^2] E [\{ \nabla \ell(X(t); U^*_{jk}) - \nabla g(X(t)) \}^2] & \\
& \leq \{ E [\{ \nabla \ell(X(t); U^*_{1k}) - \nabla g(X(t)) \}^2]\}^2 + E[ \{ \nabla \ell(X(t); U^*_{1k}) - \nabla g(X(t)) \}^4]/m & \\
& \leq \{ E [\{ \nabla \ell(X(t); U_{1k}) - \nabla g(X(t)) \}^2]\}^2 + E[ \{ \nabla \ell(X(t); U_{1k}) - \nabla g(X(t)) \}^4]/m,
\end{align*}
where we use the fact that all odd power terms have mean zero factor $\nabla \ell(X(t); U^*_{ik}) - \nabla g(X(t))$, and thus their expectations are equal to zero. %as well as the following bootstrap expectation properties, 
%\begin{align*}
%&  E [\{ \nabla \ell(X(t); U^*_{1k}, \hat{Q}_n) - \nabla g(X(t)) \}^2  | \bU_n] = \frac{1}{n} \sum_{i=1}^n  \{ \nabla \ell(X(t); U_{ik}) - \nabla g(X(t)) \}^2, &\\
%& E[ \{ \nabla \ell(X(t); U^*_{1k}, \hat{Q}_n) - \nabla g(X(t)) \}^4  |\bU_n ] =  \frac{1}{n} \sum_{i=1}^n  \{ \nabla \ell(X(t); U_{ik}) - \nabla g(X(t)) \}^4. &
%\end{align*}
By assumption A1, we have 
\begin{align*}
& \sup_{t, k} E [\{ \nabla \ell(X(t); U_{1k}) - \nabla g(X(t)) \}^2]  \leq 2 \sup_{t\geq 0} \left \{ | X(t) - x_0|^2 \right \}E [h^2_1(U)] &\\
& + 2 E [ \{\nabla \ell(x_0, U)\}^2 ]  + 2 \sup_{t \geq 0} \{ [\nabla g(X(t)) ]^2\}, & \\
\end{align*}
\begin{align*}
& \sup_{t, k} E[ \{ \nabla \ell(X(t); U_{1k}) - \nabla g(X(t)) \}^4] \leq 64 \sup_{t\geq 0} \left \{ | X(t) - x_0|^4 \right \}E [h^4_1(U)]   &\\
& + 64 E [ \{\nabla \ell(x_0, U)\}^4 ] + 8 \sup_{t \geq 0} \{ [\nabla g(X(t)) ]^4\}, & 
\end{align*}
which are finite because $X(t)$ is deterministic and bounded.  %a deterministic curve connecting $x_0$ and $\check{\theta}$.
Thus $\tau = \sup_{t,k} E\{ | R^m(X(t); \bU^*_{mk}) |^4\}$ is finite. 

Similarly as $h_1(U)$ has the fourth moment, we have %a moment generating function, 
\begin{align*}
&  E\left\{ \left| m^{-1/2}  \sum_{i=1}^m \{ h_1(U^*_{i k}) - E[ h_1(U^*_{ik}) ] \}  \right|^4 \right\} \leq [Var( h_1(U) )]^2
 %  \left(  E\left[  \{ h_1(U) - E[ h_1(U) ] \}^2 \right] \right)^2 
   + E\left[  \{ h_1(U) - E[ h_1(U) ] \}^4 \right] \equiv \tau_1, 
       \end{align*}
\begin{align*}
&   P \left( \delta^{1/2} \max_{k \leq T/\delta} \left| m^{-1}  \sum_{i=1}^m h_1(U^*_{i k}) - E[ h_1(U^*_{ik})] \right| > \delta^{1/4} | \log \delta | \right) &\\
& \leq  1 -   \prod_{k \leq T/\delta}  \left[ 1  - %\sum_{k \leq T/\delta}  
  \delta E\left\{ \left| m^{-1}  \sum_{i=1}^m h_1(U^*_{i k}) - E[ h_1(U^*_{ik})] \right|^4  \right\} / |\log \delta|^4  \right]  & \\
& \leq  1 - \exp \left [- 2 T \tau_1 / |\log \delta|^4 \right ]  \sim 2 T \tau_1 / |\log \delta|^4 \rightarrow 0, \mbox{ as $\delta \rightarrow 0$},& 
 %& \leq E \left\{  \left.  P \left( \max_{k \leq T/\delta} \left| m^{-1}  \sum_{i=1}^m h_1(U^*_{i k}) - E_*[ h_1(U^*_{ik})] \right| > 2 | \log \delta | \right | \hat{Q}_n \right)\right\} &\\
%& \leq 1 - \prod_{k=1}^{T/\delta} ( 1 -  e^{- 2 |\log \delta| + C} )  \leq C \delta \rightarrow 0, & 
\end{align*}
 which together with $E[ h_1(U^*_{ik})] = E[h_1(U)]$ imply $\delta^{1/2} \max_{k \leq T/\delta} \{  \sum_{i=1}^m h_1(U^*_{i k}) \}/m = \delta^{1/2} E[h_1(U)] +  O_P(\delta^{1/4} |\log \delta|)$.

\begin{lem} \label{lem-4}
\begin{align*}
&  \max_{t \in [0, T]}  |\check{x}^m_k - X^m_\delta (k \delta) | = O_P(\delta + \delta m^{-1/2} |\log \delta |^{1/2}), &\\
&  \max_{0 \leq t -s \leq \delta} |X^m_\delta(t) - X^m_\delta(s)| = O_P(\delta |\log \delta |^{1/2}), &
\end{align*}
where $\check{x}^m_k$ and $X^m_\delta(t)$ are defined by (\ref{GD-xcheck}) and (\ref{GD-stoch1}), respectively.
\end{lem}
Proof. By (\ref{GD-stoch1}) we have 
\begin{align*}
& |X^m_\delta(t) - X^m_\delta(s)| \leq  \int_{s}^t |\nabla g(X^m_\delta(u))| du + m^{-1/2}  \delta^{1/2} 
   \left| \int_{s}^t \bsigma(X(u)) d\bB(u) \right| & %\\
\end{align*}
\begin{align*}   
& = O_P(\delta + m^{-1/2} \delta |\log\delta|^{1/2}),    &
\end{align*}
where we use the fact that uniformly over $0\leq t -s \leq \delta$, 
\[ \int_s^t |\nabla g(X^m_\delta(u))| du = O_P(\delta), \;\;  \int_s^t \bsigma(X(u)) d\bB(u) = O_P( \delta^{1/2} |\log \delta|^{1/2}), \]  
and the order for the Brownian term is derived by law of the iterated logarithm for Brownian motion.

Note that $\check{x}^m_k$ are the Euler approximation of SDE (\ref{GD-stoch1}). The first 
result follows from the standard argument for the Euler approximation. Let 
$D(k) = |\check{x}^m_k - X^m_\delta(k \delta)|$. 
As $\check{x}^m_0=X^m_\delta(0)=x_0$, we have 
\begin{align*}
&   \check{x}^m_1 - X^m_\delta( \delta) = \int_0^\delta \nabla g(X^m_\delta(u)) du - \delta \nabla g(x_0),  &\\
& D(1)  = |\check{x}^m_1 - X^m_\delta( \delta) | =  |\int_0^\delta [ \nabla g(X^m_\delta(u))  - \nabla g(x_0)] du | &\\
& \leq C \delta \max_{0 \leq u \leq \delta}  | X^m_\delta(u) - x_0| = O_P(\delta^2 + m^{-1/2} \delta^2 |\log \delta|^{1/2}), &
\end{align*}
where we use the fact that for $u \in [0, \delta]$, 
\begin{align*}
& | X^m_\delta(u) - x_0 | \leq \int_0^u |\nabla g(X^m_\delta(v))| dv + m^{-1/2} \delta^{1/2} \left| \int_0^u \bsigma(X(v)) d\bB(v) \right| &\\
& = O_P(\delta + m^{-1/2} \delta |\log \delta|^{1/2}). &
\end{align*}
%and the order for the Brownian term is derived by law of the iterated iterated logarithm for Brownian motion. 
For the general $k$, we obtain
\begin{align*}
& D(k) = \left| \int_0^{k \delta} \nabla g(X^m_\delta(u)) du - \delta \sum_{\ell=1}^k \nabla g(\check{x}^m_{\ell-1}) \right| &\\
& \leq D(k-1) + \left| \int_{(k-1)\delta}^{k \delta} \nabla g(X^m_\delta(u)) du - \delta \nabla g(\check{x}^m_{k-1}) \right|, & 
\end{align*}
\begin{align*}
&  \int_{(k-1)\delta}^{k \delta} \nabla g(X^m_\delta(u)) du - \delta \nabla g(\check{x}^m_{k-1})  = 
   \int_{(k-1)\delta}^{k \delta} [ \nabla g(X^m_\delta(u)) - \nabla g(X^m_\delta((k-1)\delta)) ] du &\\
& + \delta[ \nabla g(X((k-1)\delta)) - \nabla g(\check{x}^m_{k-1})],  &\\   
& | \nabla g(X((k-1)\delta)) - \nabla g(\check{x}^m_{k-1}) | \leq C | X((k-1)\delta) - \check{x}^m_{k-1} | = C D(k-1), &\\
&  |  \nabla g(X^m_\delta(u)) - \nabla g(X^m_\delta((k-1)\delta)) | =|  \boldsymbol{I\!\! H}\! g(X^m_\delta(u_*) ) [ X^m_\delta(u) 
     - X^m_\delta((k-1)\delta) ]| & \\
& \leq C \int_{(k-1)\delta}^u |\nabla g(X^m_\delta(v))| dv + C m^{-1/2}  \delta^{1/2} 
   \left| \int_{(k-1)\delta}^u \bsigma(X(v)) d\bB(v) \right| &\\
 & = O_P(\delta + m^{-1/2} \delta |\log\delta|^{1/2}),    &
\end{align*}
%where again we use law of iterated logarithm for Brownian motion to show fact that for $u \in [(k-1)\delta, k \delta]$,
%$\int_{(k-1)\delta}^u \bsigma(X(v)) d\bB(v)$ is of order $\delta^{1/2} |\log \delta|^{1/2}$.
and thus 
\[ D(k) \leq D(k-1) + C \delta D(k-1) + O_P(\delta^2 + m^{-1/2} \delta^2 | \log \delta|^{1/2} ), \]
which %together with $D(1) = O_P(\delta^2 + m^{-1/2} \delta^2 | \log \delta|^{1/2} )$ 
shows  that for $k \leq T/\delta$,
\[ D(k) \leq (1+ C \delta )^{k-1} D(1) + O_P(k \delta^2 + k m^{-1/2} \delta^2 | \log \delta|^{1/2} ) 
         = O_P(\delta+  m^{-1/2} \delta | \log \delta|^{1/2} ). \]       
         
\begin{lem} \label{lem-5}
\[ \max_{t \leq T} |X^m_\delta (t) - X(t) | \leq C m^{-1/2} \delta^{1/2}  \max_{t \leq T} \left| \int_0^t \bsigma(X(u)) d\bB(u) \right|
        =O_P( m^{-1/2} \delta^{1/2} ), \]
where $X(t)$ and $X^m_\delta(t)$ are defined by (\ref{GD-c1}) and (\ref{GD-stoch1}), respectively. 
\end{lem}
Proof. With the same initial value for $X(t)$ and $X^m_\delta$, from (\ref{GD-c1}) and (\ref{GD-stoch1}) we have 
\begin{align*}
& |X^m_\delta (t) - X(t)| \leq \int_0^t | \nabla g(X^m_\delta(u)) - \nabla g(X(u))| du + m^{-1/2}  \delta^{1/2}
    \left| \int_0^t \bsigma(X(u)) d\bB(u) \right| &\\
& \leq C \int_0^t |X^m_\delta(u) - X(u) | du + m^{-1/2}  \delta^{1/2}
    \left| \int_0^t \bsigma(X(u)) d\bB(u) \right|. &
\end{align*}
Applying the Gronwall inequality we get 
\[  |X^m_\delta (t) - X(t)| \leq m^{-1/2} \delta^{1/2} \left[  \left| \int_0^t \bsigma(X(t)) d\bB(u) \right|  + 
     C \int_0^t e^{ C (t-s) } \left| \int_0^s \bsigma(X(u)) d\bB(u) \right|  ds \right] , \]
which implies 
\[ \max_{t \leq T} |X^m_\delta (t) - X(t) | \leq C m^{-1/2}  \delta^{1/2}  \max_{t \leq T} \left| \int_0^t \bsigma(X(u)) d\bB(u) \right|
        =O_P( m^{-1/2} \delta^{1/2} ), \]
where the last equality is due to Burkholder's inequality.

\begin{lem} \label{lem-6}
%the following stochastic differential equation
%\begin{equation} \label{GD-stoch1}
%  d\check{X}^m_\delta(t) = - \nabla g(\check{X}^m_\delta (t)) dt - m^{-1/2} T^{1/2} \delta^{1/2} \bsigma(\check{X}^m_\delta(t)) d\bB(t). 
%\end{equation}
%Then 
\[   \max_{t \leq T} | X^m_\delta(t) - \check{X}^m_\delta(t)| = O_P( m^{-1} \delta). \]
where $X^m_\delta(t)$ and $\check{X}^m_\delta(t)$ are the solutions of (\ref{GD-stoch1}) and (\ref{GD-stoch2}), respectively. 
\end{lem}
Proof. We have 
\begin{align*}
& |X^m_\delta (t) - \check{X}^m_\delta(t)| \leq \int_0^t | \nabla g(X^m_\delta(u)) - \nabla g(\check{X}^m_\delta(u))| du &\\
& + m^{-1/2} \delta^{1/2}   \left| \int_0^t [ \bsigma(X(u)) - \bsigma(\check{X}^m_\delta(u))]  d\bB(u) \right| 
&\\
& \leq C \int_0^t |X^m_\delta(u) - X(u) | du + m^{-1/2}  \delta^{1/2}
    \left| \int_0^t  [ \bsigma(X(u)) - \bsigma(\check{X}^m_\delta(u))]    
     d\bB(u) \right|. &\\
& E [  |X^m_\delta (t) - \check{X}^m_\delta(t)|^2] \leq C \int_0^t E [|X^m_\delta(u) - \check{X}^m_\delta(u) |^2] du &\\
& + 2 m^{-1}   \delta   E \left[ \left| \int_0^t  [ \bsigma(X(u)) - \bsigma(\check{X}^m_\delta(u))]    
     d\bB(u) \right|^2 \right ] &\\
\end{align*}
\begin{align*}
& \leq C \int_0^t E [|X^m_\delta(u) - \check{X}^m_\delta(u) |^2] du  + 2 m^{-1}   \delta    \int_0^t E  [ |\bsigma(X(u)) - \bsigma(\check{X}^m_\delta(u))|^2]    
     du   &\\
& \leq C \int_0^t E [|X^m_\delta(u) - \check{X}^m_\delta(u) |^2] du  + C_1 m^{-1}   \delta 
       \int_0^t E  [| X(u) - X^m_\delta(u) |^2] du &\\
&       + C_1 m^{-1}   \delta \int_0^t E[ | X^m_\delta(u) - \check{X}^m_\delta(u)|^2]    du,    &
\end{align*}
where the last inequality is due to 
\[ |\bsigma(X(u)) - \bsigma(\check{X}^m_\delta(u))| \leq C | X(u) - \check{X}^m_\delta(u) | \leq 
 C |X(u) - X^m_\delta(t)| + C | X^m_\delta(t) - \check{X}^m_\delta(t)|. \]
The Gronwall inequality leads to 
\[ E [  |X^m_\delta (t) - \check{X}^m_\delta(t)|^2] %\leq C m^{-1} T  \delta \max_{s \leq t} \int_0^s E  [| X(u) - X^m_\delta(u) |^2] du 
    \leq C m^{-1}  \delta \max_{s \leq t} E  [| X(s) - X^m_\delta(s) |^2] . \] 
Using Lemma \ref{lem-5} we have 
\begin{align*}
& \max_{s \leq t} E  [| X(s) - X^m_\delta(s) |^2] \leq C m^{-1} \delta E  \left[ \max_{s \leq t} \left| \int_0^s \bsigma(X(u)) d\bB(u) \right|^2 \right]
&\\
& \leq C m^{-1} \delta E \left[  \int_0^t [\bsigma(X(u))]^2 du \right], &
\end{align*}
where the last inequality is from Burkholder's inequality.
Hence 
\[  E [  |X^m_\delta (t) - \check{X}^m_\delta(t)|^2] \leq C m^{-2} \delta^2 E \left[ \int_0^t [\bsigma(X(u))]^2 du \right], \]
and we can adopt the same argument to establish it for $t$ as a bounded stopping time.
Finally we prove the lemma by applying Lenglart's inequality for semi-martingale with $\eta_i = D_i m^{-1} \delta $ for some positive constants $D_i$, 
\begin{align*}
& P\left( \max_{s \leq t} | X(s) - X^m_\delta(s) | > \eta_1 \right)  \leq \frac{ C m^{-2} \delta^2  \int_0^t [\bsigma(X(u))]^2 du  }{\eta_1^2} &\\
&  + P\left( C m^{-2} \delta^2  \int_0^t [\bsigma(X(u))]^2 du > \eta_2^2    \right)  \rightarrow 0, \mbox{ as } D_i \rightarrow \infty.  &
\end{align*}

 \begin{lem} \label{lem-7}
As $\delta \rightarrow 0$, and $m,n \rightarrow \infty$, we have $V^m_\delta(t)$ and $\check{V}^m_\delta(t)$ both 
%$=(m T/ \delta)^{1/2}  [ X^m_\delta (t) - X(t)] $ 
weakly converge to $V(t)$. 
Moreover, if $m (n \delta)^{1/2} \rightarrow 0$, and $m^{1/2} \delta |\log \delta|^{1/2} \rightarrow 0$, 
$(m /\delta)^{1/2} [x^m_\delta(t) - X(t)]$ weakly converges to $V(t)$.
%need to check as $\delta \rightarrow 0$ first and then  $m,n \rightarrow \infty$ may pose some issue for the limit for $x^m_\delta$.
\end{lem}        
Proof. As the solutions of (\ref{GD-stoch1}) and (\ref{GD-stoch2}) have difference of order $m^{-1}\delta$, they have the same asymptotic distribution, and we can easily establish the result for $\check{V}^m_\delta(t)$ by that for $V^m_\delta(t)$ and Lemma \ref{lem-3}. 

Let consider the easier one for $X^m_\delta(t)$. From (\ref{GD-stoch1}) and (\ref{GD-c1}), we have 
\[  d[X^m_\delta(t) - X(t)] = - [\nabla g(X^m_\delta (t) - \nabla g(X(t)) ] dt - m^{-1/2}  \delta^{1/2} \bsigma(X(t)) d\bB(t), \]
and for $ t \in [0, T]$,
\[ X^m_\delta(t) - X(t) = - \int_0^t [\boldsymbol{I\!\! H}\! g(X_\xi)] [ X^m_\delta (u) - X(u) ]  du - 
       m^{-1/2}  \delta^{1/2} \int_0^t \bsigma(X(u)) d\bB(u),\]
where $X_\xi$ is between $X(u)$ and $X^m_\delta(u)$ and thus Lemma \ref{lem-5} shows that uniformly over $[0, T]$, 
\[ |X_\xi - X(u)| \leq |X^m_\delta(u) - X(u) | = O_P(m^{-1/2}\delta). \] 
Then 
\begin{equation} \label{GD-v1-1}
  V^m_\delta(t) = - \int_0^t [\boldsymbol{I\!\! H}\! g(X_\xi)] V^m_\delta (u)  du - \int_0^t \bsigma(X(u)) d\bB(u). 
 \end{equation} 
First as  $\delta \rightarrow 0$, $m,n \rightarrow \infty$, %$V^m_\delta(t)$ converges to $V(t)$ governed by 
 equation (\ref{GD-v1-1}) converges to (\ref{GD-v0}).
%\[ V(t) = - \int_0^t [\boldsymbol{I\!\! H}\! g(X(t))] V(u)  du - \int_0^t \bsigma(X(u)) d\bB(u). \] 
 
We need to show stochastic equicontinuity for $V^m_\delta(t)$.
From (\ref{GD-v1-1}) we obtain 
\[  |V^m_\delta(t)|  \leq C \int_0^t |V^m_\delta(u) | du +  \left| \int_0^t \bsigma(X(u)) d\bB(u) \right|, \]
and by the Gronwall inequality we have 
\[  \max_{ t \leq T} |V^m_\delta(t)| \leq C \max_{t \leq T} \left| \int_0^t \bsigma(X(u)) d\bB(u) \right|, \]
that is $V^m_\delta(t)$ is bounded in probability uniformly over $[0, T]$. Again (\ref{GD-v1-1}) indicates that for any $s, t \in [0, T]$, and 
$t \in [s, s+\gamma]$, 
 \begin{align*}
& V^m_\delta(t) - V^m_\delta(s) = - \int_s^t [\boldsymbol{I\!\! H}\! g(X_\xi)] V^m_\delta (u)  du - \int_s^t \bsigma(X(u)) d\bB(u),& \\
& |V^m_\delta(t) - V^m_\delta(s)|  \leq C \int_s^t | V^m_\delta (u) |  du + \left|  \int_s^t \bsigma(X(u)) d\bB(u) \right| & \\
& \leq C \int_s^t | V^m_\delta (u) - V^m_\delta(s) |  du + C (t-s) |V^n_\delta(s)| + \left|  \int_s^t \bsigma(X(u)) d\bB(u) \right|. &
\end{align*}
Again applying the Gronwall inequality we obtain uniformly for $t \in [s, s+\gamma]$, 
\begin{align*}
& |V^m_\delta(t) - V^m_\delta(s)|  \leq  C \gamma |V^n_\delta(s)| + C \max_{s \leq t \leq s +\gamma} \left|  \int_s^t \bsigma(X(u)) d\bB(u) 
         \right| = O_P(\gamma + \gamma^{1/2} |\log\gamma |^{1/2}), &
\end{align*}
which proves stochastic equicontinuity for $V^n_\delta(t)$.

\subsection{Proof of Theorem \ref{thm5}}
Theorem \ref{thm5} can be proved by the same proof argument of Theorem \ref{thm2} except for changing step size from $\delta$ to $\delta^{1/2}$.

%% file: p4.tex
\subsection{Proof of Theorems \ref{thm6}}

\subsubsection{The unique solution of the second order SDEs}

In this section we will prove Lemma \ref{lem-s7} blow that 
%\begin{prop} \label{prop2}
the second order SDEs (\ref{Nest-stoch1}) (with fixed $(\delta, m)$) and  (\ref{Nest-limit-0}) have unique (weak) solutions in the distributional sense. 
%\end{prop}

Due to the similarity we provide proof arguments for (\ref{Nest-limit-0}) only. 
Consider the 2nd order SDE (\ref{Nest-limit-0}) with 
%\begin{equation}\label{equ-bb4}
%  \ddot{V}(t)+\frac{3}{t}\dot{V}(t) +  [\nabla \!g(X(t))]  V(t)+ \bsigma(X(t)) \dot{\bB}(t)  =0,
%\end{equation}
initial conditions $V(0)=c$ and $\dot{V}(0)=0$, where $\bB(t)$ is a standard Brownian motion, 
$\dot{V}(t)$ and $ \ddot{V}(t)$ are the first and second derivatives of $V(t)$, respectively, $\dot{\bB}(t) = \frac{d B(t)}{dt}$ is a white noise in a sense that 
for any smooth function $h(t)$ with compact support, 
\[ \int h(t) \dot{\bB}(t) dt = \int h(t) dB(t), \]
where the right hand side is It\^o integral.
%We can impose any smooth and growth condition on $\sigma(\cdot)$. 

%\begin{lem}
%Equation (\ref{limit-0}) has a unique solution. 
%\end{lem}
The second order SDE (\ref{Nest-limit-0}) is equivalent to 
\begin{equation} \label{equ-b1}
   Y(t) = V(t) + \frac{t}{2} \,\dot{V}(t), \;\;
 \dot{Y}(t) = -\frac{t}{2}\, [\nabla \! g(X(t))] V(t)- \frac{t}{2}\, \bsigma(X(t)) \dot{\bB}(t), 
 \end{equation}
 where $V(0)=c, \dot{V}(0)=0$, and $Y(0) = V(0)=c$. %$\dot{Y}(0) = \frac{3}{2} \,\dot{V}(0) +\frac{0}{2} \ddot{V}(0) = 0$.
%\begin{eqnarray*} 
% \dot{V}(t) &=& \frac{2}{t} \, Y(t) - \frac{2}{t} \, V(t) \\
% %\dot{Y}(t) &=&  -\frac{t}{2}\, \nabla \! g(X(t)) - \frac{t}{2}\, \ba(X(t)) \dot{\bB}(t) \\
%  dY(t) &=&  -\frac{t}{2}\, [\nabla \! g(X(t))] V(t) dt - \frac{t}{2}\, \bsigma(X(t)) d \bB(t)  \\
%   V(0) &=& Y(0) =x_0, \; \dot{V}(0)=\dot{Y}(0)=0 
% \end{eqnarray*}
 Denote by $V_\eta(t)$ the solution of the smoothed second order SDE 
 \begin{equation}  \label{equ-b2}
   \ddot{V}_\eta(t) + \frac{3}{t \vee \eta} \dot{V}_\eta(t) + [\nabla g(X(t))] V_\eta(t) + \bsigma(X(t)) \dot{\bB}(t)  = 0, 
 \end{equation}
 with initial conditions $V_\eta(0)=c$ and $\dot{V}_\eta(0)=0$. 
% Denote by $(V_\eta(t), Y_\eta(t))$ the solution of the smoothed stochastic equation system 
%\begin{equation} \label{equ-b2}
% \dot{V}(t) = \frac{2}{t \vee \eta} \, Y(t) - \frac{2}{t \vee \eta} \, V(t), \;\;
% \dot{Y}(t) =  -\frac{t \vee \eta}{2}\, [\boldsymbol{I\!\! H}\! g(X(t))] V(t) - \frac{t \vee \eta}{2}\, \bsigma(X(t)) \dot{\bB}(t), 
% \end{equation}
 %where $ V(0) = Y(0) =c, \dot{V}(0)=\dot{Y}(0)=0$. 

% Given an interval $\mathcal{I}=[s,t]$ and a process $Y(t)$, we have defined 
%\[ M_a(s,t;Y)=M_a(\mathcal{I};Y) =  \sup_{u \in [s,t]} \left| \frac{\dot{Y}(u)-\dot{Y}(s)}{(u-s)^{a}} \right|. \] 
We need notation $M_a(s,t;Y)$ defined in (\ref{equ-Ma}).
In the proofs of Theorems \ref{thm-1} and \ref{thm2}, we have employed $M_a(s,t;Y)$ with $a=1$, as curves and processes are solutions of ODE 
and thus differentiable. For this part of proofs we need to handle Brownian motion and SDEs,
and the related processes have less than $1/2$-derivatives, so we fix $a \in (0, 1/2)$ and consider $M_a(s,t;Y)$ with $a<1/2$.

\begin{lem}  \label{lem-s1}
\begin{align*}
& | \nabla g(X(t)) | %\leq | \nabla g(X(s)) | + L | X(t) - X(s)| 
  \leq  | \nabla g(X(s)) | + L (t-s) |\dot{X}(s)|  + L M_a(s, t; X) (t-s)^{1+a} /(1+a), &\\
& | \nabla g(X(t))  V_\eta(t) - \nabla g(X(s)) V_\eta(s) |  \leq L |V_\eta(s) | (t-s) |\dot{X}(s) | +  | \nabla g(X(t)) |  (t-s) | \dot{V}_\eta(s) | &\\
 & +  [ L|V_\eta(s) | M_a(s,t;X) +  | \nabla g(X(t)) |  M_a(s, t; V_\eta) ] (t-s)^{1+a} /(1+a). &
 \end{align*}
\end{lem}
Proof. We prove the lemma by the following direct calculation,
\begin{align*}
 & | \nabla g(X(t))  V_\eta(t) - \nabla g(X(s)) V_\eta(s) | \leq | \nabla g(X(t)) | |  V_\eta(t) - V_\eta(s) | &\\
 &  +  | \nabla g(X(t)) - \nabla g(X(s)) | | V_\eta(s) | &\\
&    \leq | \nabla g(X(t)) |  | V_\eta(t) - V_\eta(s)|  + L |V_\eta(s)| | X(t) - X(s) |  &\\
 & \leq L |V_\eta(s) |  | \int_s^t [\dot{X}(v) - \dot{X}(s)] dv + (t-s) \dot{X}(s) | &\\
 & + | \nabla g(X(t)) | | \int_s^t [\dot{V}_\eta(v) - \dot{V}_\eta(s)]  dv 
 + (t-s) \dot{V}_\eta(s) | &\\
 & \leq L |V_\eta(s) | (t-s) |\dot{X}(s) | +  | \nabla g(X(t)) |  (t-s) | \dot{V}_\eta(s) | &\\
  & +  L|V_\eta(s) | | \int_s^t  (v-s)^{a} \frac{\dot{X}(v)  -  \dot{X}(s)}{(v-s)^{a} } dv | + | \nabla g(X(t)) |
   | \int_s^t (v-s)^{a} \frac{\dot{V}_\eta(v)  - \dot{V}_\eta(s)}{(v-s)^{a} } dv |  &\\
 & \leq L |V_\eta(s) | (t-s) |\dot{X}(s) | +  | \nabla g(X(t)) |  (t-s) | \dot{V}_\eta(s) | &\\
 & +  [ L|V_\eta(s) | M_a(s,t;X) +  | \nabla g(X(t)) |  M_a(s, t; V_\eta)  (t-s)^{1+a} /(1+a), & \\
 & | \nabla g(X(t)) | \leq | \nabla g(X(s)) | + L | X(t) - X(s)| &\\
 &  \leq  | \nabla g(X(s)) | + L (t-s) |\dot{X}(s)|  +   L M_a(s, t; X) ] (t-s)^{1+a} /(1+a) . &
\end{align*}
\begin{lem}  \label{lem-s2}
There exists $\eta_0>0$ such that for $\eta \in (0, \eta_0]$, $1 - |\nabla g(X(0))|  \eta^{2}/[(1+a)(2+a)]   - L M_a(0, \eta; X) \eta^{3+a} /[(1+a)^2(3+2a) ] $ is bounded below from zero. Then we have for $\eta \in (0, \eta_0]$, 
\[ M_a(0,\eta; V_\eta) \leq \frac{1}{1 - |\nabla g(X(0))|  \eta^{2}/[(1+a)(2+a)]   - L M_a(0, \eta; X) \eta^{3+a} /[(1+a)^2(3+2a) ] } \]
\[ \left[ | \nabla g(X(0)) V_\eta(0)| \eta^{1-a} + 
  \frac{L |V_\eta(0)| M_a(0, \eta; X) \eta^{2}}{(1+a)(2+a)} 
   + \max_{t \in (0, \eta] } \left| 
 \frac{1}{t^a} e^{-3t/\eta} \int_0^t e^{3u/\eta} \bsigma(X(u)) d \bB(u)  \right|  \right]. \]
\end{lem}
 Proof.  Since $\nabla g(X(0))$ and $M_a(0, \eta; X)$ for each $\eta$ are deterministic and finite,  and $M_a(0, \eta; X)$ is continuous and increasing in $\eta$, we easily show that  $ |\nabla g(X(0))|  \eta^{2}/[(1+a)(2+a)]   + L M_a(0, \eta; X) \eta^{3+a} /[(1+a)^2(3+2a) ] $ approaches zero as $\eta \rightarrow 0$, which leads to the existence of $\eta_0$. 
 Lemma \ref{lem-s1} indicates 
 \begin{align*}
 & | \nabla g(X(u))  V_\eta(u) - \nabla g(X(0)) V_\eta(0) |  %\leq L | X(u) - X(0) | |  V_\eta(u) |  + | \nabla g(X(0)) |  | V_\eta(u) - V_\eta(0)|  
   &\\
 %& \leq L |  V_\eta(u) |   | \int_0^u \dot{X}(v) dv | +  | \nabla g(X(0)) |   | \int_0^u \dot{V}_\eta(v) dv | &\\
% & \leq L |  V_\eta(u) | | \int_0^u v^{a} \frac{\dot{X}(v)}{v^{a} } dv | + | \nabla g(X(0)) | | \int_0^u v^{a} \frac{\dot{V}_\eta(v)}{v^{a} } dv |  
&  \leq [ L  |  V_\eta(0) | M_a(0,u;X) + | \nabla g(X(u)) | M_a(0, u; V_\eta) ] u^{1+a} /(1+a), & \\
& | \nabla g(X(u)) | %\leq | \nabla g(X(0)) | + L | X(u) - X(0)| 
   \leq  | \nabla g(X(0)) | + L M_a(0, u; X)  u^{1+a} /(1+a). &
 \end{align*}
 %Take $\eta \in (0,1)$. 
 For $t \in (0, \eta]$, $V_\eta$ satisfies 
 \[ \ddot{V}_\eta(t) + \frac{3}{\eta} \dot{V}_\eta(t) + [\nabla g(X(t))] V_\eta(t) + \bsigma(X(t)) \dot{\bB}(t)  = 0, \]
 which is equivalent to 
 \[  \left[ \dot{V}_\eta(t) e^{3 t/\eta} \right]^\prime = -e^{3 t/\eta} [\nabla g(X(t)) ] V_\eta(t) - e^{3 t/\eta}\bsigma(X(t)) \dot{\bB}(t), \]
 \begin{align*}
 & \dot{V}_\eta(t) e^{3 t/\eta} = - \int_0^t e^{3 u/\eta} [\nabla g(X(u)) ] V_\eta(u)] du - \int_0^t e^{3u/\eta}\bsigma(X(u)) \dot{\bB}(u)  du &\\
 & = -\nabla g(X(0)) V_\eta(0) \int_0^t e^{3 u/\eta} du - 
   \int_0^t e^{3 u/\eta} [ \nabla g(X(u)) V_\eta(u) - \nabla g(X(0)) V_\eta(0)] du &\\
 & -\, \int_0^t e^{3u/\eta}\bsigma(X(u)) d \bB(u). &
  \end{align*}
 Thus for $t \in (0, \eta]$ we have 
 \begin{align*} 
 & \left| \frac{\dot{V}_\eta(t)}{t^a} \right| \leq \frac{1}{t^a} e^{-3t/\eta} | [\nabla g(X(0))] V_\eta(0) |  \int_0^t e^{3u/\eta}  du + 
    \frac{1}{t^a} e^{-3t/\eta} \left | \int_0^t e^{3u/\eta}\bsigma(X(u)) d \bB(u) \right|  &\\
 & +\,\frac{1}{(1+a)t^a} e^{-3t/\eta}
      \int_0^t  [ L  |  V_\eta(0) | M_a(0,u;X) + | \nabla g(X(u)) | M_a(0, u; V_\eta) ] u^{1+a} e^{3u/\eta}   du, & %\\
 %& + \,\frac{1}{t^a} e^{-3t/\eta} \left | \int_0^t e^{3u/\eta}\bsigma(X(u)) d \bB(u)  \right| & \\
  \end{align*}
 \begin{align*}
 & \leq t^{1-a} | \nabla g(X(0)) V_\eta(0)| + \frac{ [ L | V_\eta(0)| M_a(0, t; X)+ | \nabla g(X(0)) |   M_a(0,t, V_\eta)] \eta^{2}}{(1+a)(2+a)} & \\
 & + \frac{  L M_a(0, t; X)    M_a(0,t, V_\eta) \eta^{3+a}}{(1+a)^2(3 + 2a)} 
 + \frac{1}{t^a} e^{-3t/\eta} \left| \int_0^t e^{3u/\eta} \bsigma(X(u)) d \bB(u) \right|. & 
   \end{align*}
 Taking the maximum over $t \in (0, \eta]$ on both sides of above inequality, and using the definition of $M_a(0, t; \cdot)$ 
 (which is increasing in $t$), we can easily prove the lemma through the simple algebra manipulation (which is also employed in the proof of Lemma \ref{lem1}). 

\begin{lem} \label{lem-s3}
%For $t - \eta < [(1+a)(2+a)/L]^{1/2}$ and $\eta < t < 2 [(1+a)(2+a)/L]^{1/2}$, we have 
There exists $\eta_0>0$ such that for $\eta \in (0, \eta_0]$ and $\eta<t<\eta + \eta_0$, 
$1 -  \frac{(t-\eta)^{2}}{(1+a)(2+a)}  | \nabla g(X(\eta)) | - \frac{  L M_a(\eta, t; X) (t-\eta)^{3+a} }{(1+a)^2(3 + 2a) } $ is bounded below from zero. Then we have for $\eta \in (0, \eta_0]$ and $\eta<t<\eta + \eta_0$,
  \begin{align*}
 &  M_a(\eta, t; V_\eta) \left[ 1 -  \frac{(t-\eta)^{2}}{(1+a)(2+a)}  | \nabla g(X(\eta)) | - \frac{  L M_a(\eta, t; X) (t-\eta)^{3+a} }{(1+a)^2(3 + 2a) }   
   \right] & \\
 &  \leq C_1 M_a(0, \eta; V_\eta) +  C_2 | \nabla g(X(\eta)) V_\eta(\eta) | & \\
   &    + \frac{(t-\eta)^{2-a} } {2 } [  L (|V_\eta(\eta) | +1) |\dot{X}(\eta) | + | \nabla g(X(\eta) | ] +  \frac{(t-\eta)^3 }{(1+a)(3+a) }  
        L   | \dot{V}_\eta(\eta) |   M_a(\eta, t; X)   &\\
 & + \frac{(t-\eta)^{2}}{(1+a)(2+a)} L  |  V_\eta(\eta) |   M_a(\eta, t; X)   
   + \max_{t_0 \in (\eta, t]}  \left| \frac{1}{t_0^3 (t_0-\eta)^a}   \int_\eta^{t_0} u^3 \bsigma(X(u)) \dot{\bB}(u) du \right| .
 \end{align*} 
\end{lem}
Proof. Since $\nabla g(X(\eta))$ and $M_a(\eta, t; X)$ are deterministic and continuous in $\eta$, their maximum over $\eta$ in a 
neighborhood of $0$ is finite. As $t - \eta \rightarrow 0$, 
$\frac{(t-\eta)^{2}}{(1+a)(2+a)}  | \nabla g(X(\eta)) | + \frac{  L M_a(\eta, t; X) (t-\eta)^{3+a} }{(1+a)^2(3 + 2a) } $ approaches zero, and thus 
the existence of $\eta_0$ is obvious. For $t >\eta$,  $V_\eta$ satisfies 
 \[ \ddot{V}_\eta(t) + \frac{3}{t} \dot{V}_\eta(t) + [\nabla g(X(t))] V_\eta(t) + \bsigma(X(t)) \dot{\bB}(t)  = 0, \]
 which is equivalent to
\[  \left[ t^3 \dot{V}_\eta(t) \right]^\prime = - t^3  [\nabla g(X(t)) ] V_\eta(t) - t^3 \bsigma(X(t)) \dot{\bB}(t), \]
and 

\begin{align*}
& t^3 \dot{V}_\eta(t) = \eta^3 \dot{V}_\eta(\eta) - \int_\eta^t u^3  [\nabla g(X(u)) ] V_\eta(u) du  - \int_\eta^t u^3 \bsigma(X(u)) \dot{\bB}(u) du &\\
& = \eta^3 \dot{V}_\eta(\eta) - \int_\eta^t u^3  [\nabla g(X(u)) V_\eta(u)  - \nabla g(X(\eta)) V_\eta(\eta)]  du - 
 \int_\eta^t u^3  [\nabla g(X(\eta)) ] V_\eta(\eta) du  &\\
 & -\, \int_\eta^t u^3 \bsigma(X(u)) \dot{\bB}(u) du. &
\end{align*}
Thus, 
\begin{align*}
& \frac{|\dot{V}_\eta(t)  -\dot{V}_\eta(\eta)  |}{(t-\eta)^a} \leq \frac{(t^3 - \eta^3) \eta^a }{t^3 (t-\eta)^a} \frac{|\dot{V}_\eta(\eta)|}{\eta^a} + \frac{t^4 -\eta^4} {4 t^3 (t-\eta)^a}  | \nabla g(X(\eta)) V_\eta(\eta) |  &\\
&    + \frac{1}{ t^3 (t-\eta)^a } \int_\eta^t [ L |V_\eta(\eta) | (u-\eta) |\dot{X}(\eta) | +  | \nabla g(X(u) |  (u-\eta) | \dot{V}_\eta(\eta) | ]  u^3 du    &\\
  & + \, \frac{1}{(1+a) t^3 (t-\eta)^a } \int_\eta^t [ L  |  V_\eta(\eta) | M_a(\eta,u;X) + | \nabla g(X(u)) | M_a(\eta, u; V_\eta) ]  u^3 (u - \eta)^{1 + a} du &\\
  &  + \, \frac{1}{t^3 (t-\eta)^a }  \left | \int_\eta^t u^3 \bsigma(X(u)) \dot{\bB}(u) du \right| &\\
 & \leq \frac{(t^3 - \eta^3) \eta^a }{t^3 (t-\eta)^a}  M_a(0, \eta; V_\eta) + \frac{t^{4} - \eta^4}{4 t^3 (t-\eta)^a } | \nabla g(X(\eta)) V_\eta(\eta) | &\\
 &    + \frac{(t-\eta)^{2-a} } {2 } [  L (|V_\eta(\eta) | +1) |\dot{X}(\eta) | + | \nabla g(X(\eta) | ] +  \frac{(t-\eta)^3 }{(1+a)(3+a) }  
        L   | \dot{V}_\eta(\eta) |   M_a(\eta, t; X)   &\\
 & + \frac{(t-\eta)^2 }{(1+a)(2+a) } [ L  |  V_\eta(\eta) |
  M_a(\eta, t; X)+ | \nabla g(X(\eta)) |  M_a(\eta, t; V_\eta)] &\\
  & +\, \frac{  L M_a(\eta, t; X)    M_a(\eta,t, V_\eta) (t-\eta)^{3+a} }{(1+a)^2(3 + 2a) } + \frac{1}{t^3 (t-\eta)^{a}}  \left | \int_\eta^t u^3 \bsigma(X(u)) \dot{\bB}(u) du \right| . & 
% & \leq \frac{(t^3 - \eta^3) \eta^a }{t^3 (t-\eta)^a}  M_a(0, \eta; V_\eta) + \frac{t^{4} - \eta^4}{4 t^3 (t-\eta)^a } | \nabla g(X(0)) V_\eta(0) | &\\
% & + \frac{(t-\eta)^{2}}{(1+a)(2+a)} [L  |  V_\eta(0) |
 % M_a(0, t; X)+ | \nabla g(X(0)) |  M_a(0, t; V_\eta)] &\\
 %& +\, \frac{  L M_a(0, t; X)    M_a(0,t, V_\eta) (t-\eta)^{3+a} }{(1+a)^2(3 + 2a) } +  
 %   \frac{1}{t^3 (t-\eta)^{a}} \left | \int_\eta^t u^3 \bsigma(X(u)) \dot{\bB}(u) du \right|. &
   \end{align*}
As in the proof of Lemma \ref{lem-s2}, replacing $t$ by $u$ in above inequality, taking the maximum over $u \in (\eta, t]$ on both sides, and using the definition of $M_a(\eta, t;\cdot)$ (which is increasing in $t$), %the fact that the right hand side of above inequality is increasing in $t$, 
we conclude that 

 \begin{align*}
 &  M_a(\eta, t; V_\eta) \leq C_1 M_a(0, \eta; V_\eta) + % \frac{t^{1-a}}{4} 
   C_2 | \nabla g(X(\eta)) V_\eta(\eta) | & \\
   &    + \frac{(t-\eta)^{2-a} } {2 } [  L (|V_\eta(\eta) | +1) |\dot{X}(\eta) | + | \nabla g(X(\eta) | ] +  \frac{(t-\eta)^3 }{(1+a)(3+a) }  
        L   | \dot{V}_\eta(\eta) |   M_a(\eta, t; X)   &\\
 & + \frac{(t-\eta)^{2}}{(1+a)(2+a)} [L  |  V_\eta(\eta) |
  M_a(\eta, t; X)+ | \nabla g(X(\eta)) |  M_a(\eta, t; V_\eta)] & \\
  & +\, \frac{  L M_a(\eta, t; X)    M_a(\eta,t; V_\eta) (t-\eta)^{3+a} }{(1+a)^2(3 + 2a) }   
   + \max_{t_0 \in (\eta, t]}  \left| \frac{1}{t_0^3 (t_0-\eta)^a}   \int_\eta^{t_0} u^3 \bsigma(X(u)) \dot{\bB}(u) du \right|, 
 \end{align*}
 which leads to the lemma.

\begin{lem} \label{lem-s4}
%For $t - s < [(1+a)(2+a)/L]^{1/2}$ and $s < t < 2 [(1+a)(2+a)/L]^{1/2}$, we have 
%\begin{align*}
%& M_a(s,t; V_\eta) \leq \left[ 1 - \frac{ L (t-s)^2}{ (1+a)(2+a)} \right]^{-1}  \left[ 
%    \frac{3}{t^{1-a}} |\dot{V}_\eta(s)| + \frac{t^{1-a}}{4} | \nabla g(X(0)) V_\eta(0) | \right. & \\
 %   & \left. + \frac{(t-s)^{2}}{(1+a)(2+a)} M_a(s, t; X)  +  \frac{(t-s)^{2-a}}{2} [ L |V_\eta(s) | |\dot{X}(s) | +  | \nabla g(X(*)) |  | \dot{V}_\eta(s) |]      \right. &\\
%& \left.   + \max_{v \in (s, t]}  \left| \frac{1}{v^3(v-s)^a}   \int_s^{v} u^3 \bsigma(X(u)) \dot{\bB}(u) du \right| 
%   \right]. &
%\end{align*}
There exists $\eta_0>0$ such that for $\eta \in (0, \eta_0]$ and $\eta<s<t<\eta + s \leq T$, 
$1 -  \frac{(t-s)^{2}}{(1+a)(2+a)}  | \nabla g(X(s)) | - \frac{  L M_a(s, t; X) (t-s)^{3+a} }{(1+a)^2(3 + 2a) } $ is bounded below from zero. Then we have for $\eta \in (0, \eta_0]$ and $\eta<s<t<\eta + s$,
  \begin{align*}
 &  M_a(s, t; V_\eta) \left[ 1 -  \frac{(t-s)^{2}}{(1+a)(2+a)}  | \nabla g(X(s)) | - \frac{  L M_a(s, t; X) (t-s)^{3+a} }{(1+a)^2(3 + 2a) }   
   \right] & \\
 &  \leq C_1 M_a(0, s; V_\eta) +  C_2 | \nabla g(X(s)) V_\eta(s) | & \\
   &    + \frac{(t-s)^{2-a} } {2 } [  L (|V_\eta(s) | +1) |\dot{X}(s) | + | \nabla g(X(s) | ] +  \frac{(t-s)^3 }{(1+a)(3+a) }  
        L   | \dot{V}_\eta(s) |   M_a(s, t; X)   &\\
 & + \frac{(t-s)^{2}}{(1+a)(2+a)} L  |  V_\eta(s) |   M_a(s, t; X)   
   + \max_{t_0 \in (s, t]}  \left| \frac{1}{t_0^3 (t_0-s)^a}   \int_s^{t_0} u^3 \bsigma(X(u)) \dot{\bB}(u) du \right| .
 \end{align*} 
 \end{lem}
Proof. Note that for  $s, t >\eta$, $V_\eta$ satisfies 
\[  \left[ t^3 \dot{V}_\eta(t) \right]^\prime = - t^3  [\nabla g(X(t)) ] V_\eta(t) - t^3 \bsigma(X(t)) \dot{\bB}(t), \]
and 
\begin{align*}
& t^3 \dot{V}_\eta(t) = s^3 \dot{V}_\eta(s) - \int_s^t u^3  [\nabla g(X(u)) ] V_\eta(u) du  - \int_s^t u^3 \bsigma(X(u)) \dot{\bB}(u) du &\\
& = s^3 \dot{V}_\eta(s) - \int_s^t u^3  [\nabla g(X(u)) V_\eta(u)  - \nabla g(X(s)) V_\eta(s)]  du - 
 \int_s^t u^3  [\nabla g(X(s)) ] V_\eta(s) du  &\\
 & -\, \int_s^t u^3 \bsigma(X(u)) \dot{\bB}(u) du. &
\end{align*}
Then we work on  $\frac{|\dot{V}_\eta(t) - \dot{V}_\eta(s)|}{(t-s)^a} $. The rest of the proof argument is the same as in the proof of 
 Lemma \ref{lem-s3} with $\eta$ replaced by $s$.

\begin{lem}  \label{lem-s5}
We have  
\[ P \left( \max_{v \in (s, t]}  \left| \frac{1}{v^3 (v-s)^a}   \int_s^{v} u^3 \bsigma(X(u)) d \bB(u) \right| < \infty \mbox{ for all $0<s<t$}
\right) = 1. \]
\end{lem}
Proof. We need to show that Gaussian process $ \int_s^{v} u^3 \bsigma(X(u)) d \bB(u)$
has the $a$-th derivative. Indeed, 
%has mean zero and covariance 
%\[ Cov\left( \int_s^{v_1} u^3 \bsigma(X(u)) d \bB(u), \int_s^{v_2} u^3 \bsigma(X(u)) d \bB(u)  \right)  
 %   = \int_s^{v_1 \wedge v_2} u^6 \bsigma^2(X(u)) du.     \]
%Thus, it has variance 
%$\int_s^{v} u^6 \bsigma^2(X(u)) du \sim v - s $, and it has $a$ derivative. 
we have
\begin{align*}
& \frac{1}{v^3 (v-s)^a}   \int_s^{v} u^3 \bsigma(X(u)) d [\bB(u) - \bB(s)] = \bsigma(X(v)) \frac{ [\bB(v) - \bB(s)]}{ (v-s)^a}  &\\
& - \frac{1}{v^3 (v-s)^a}   \int_s^{v} \frac{d [ u^3 \bsigma(X(u))] }{du}    [\bB(u) - \bB(s)]  du &\\
& =\frac{ [\bB(v) - \bB(s)]}{ (v-s)^a} \bsigma(X(v))  
  -  \frac{ 1}{v^3}  \int_s^{v}  \frac{d [ u^3 \bsigma(X(u))] }{du}   \frac{(u-s)^a}{ (v-s)^a}   \frac{\bB(u) - \bB(s)}{(u-s)^a}   du, &
\end{align*}
which is a.s. finite, due to the facts that $0 < (u-s)^a/(v-s)^a \leq 1$, $X(\cdot)$ and $\bsigma(\cdot)$ are continuously differentiable and 
Lipschitz, and Brownian motion has a well-known property that for all $u > s >0$, $   \sup_{s<u} \frac{|\bB(u) - \bB(s)|}{(u-s)^a} $ is a.s. finite. 

\begin{lem} \label{lem-s6} %\label{lem8-bb}
For any given $T>0$,  
$V_\eta(t)$ is stochastically equicontinuous and stochastically bounded on $[0, T]$ uniformly over $\eta$.
\end{lem}

Proof. Take $\eta_*$ to be the smallest $\eta_0$ defined in Lemmas \ref{lem-s2} -\ref{lem-s4}. 
Divide the interval $[0,T]$ into $N = [T/\eta_* +1]$
%$N=[T \sqrt{L/(1+a)(2+a)}]+1$ 
number of subintervals with length almost equal to $\eta_*$ %$\sqrt{(1+a)(2+a)/L}$ 
(except for the last one), and denote by $\mathcal{I}_i=[s_{i-1}, s_i]$, $i=1, \cdots, N$ (with $s_0=0$, $s_N=T$, $\mathcal{I}_1 = [0, T/N]$,
$1/N < \eta_*/T$, %\sqrt{(1+a)(2+a)/L}/T$, 
$\mathcal{I}_N = [s_{N-1}, T]$).
First for $t \in \mathcal{I}_1$, we have 
\[  |\dot{V}_\eta(t)| \leq |\mathcal{I}_1|^a M_a(\mathcal{I}_1;V_\eta), \;\; |V_\eta(t)| \leq |V_\eta(0)| + \int_{\mathcal{I}_1} |\dot{V}_\eta(u) | du,  \]
and the upper bounds on $\dot{V}_\eta(t)$ and $V_\eta(t)$ over $\cI_1$
are a.s. finite uniformly over $\eta$, which implies that  $V_\eta(t)$ is stochastically equicontinuous and stochastically bounded over $\mathcal{I}_1$.

For $t \in \mathcal{I}_i$, $i =2, \cdots, N$, we have 
\begin{eqnarray*}
&& |\dot{V}_\eta(t)-\dot{V}_\eta(s_{i-1})| \leq |\mathcal{I}_i|^a M_a(\mathcal{I}_i; V_\eta), 
\end{eqnarray*}
and
\begin{eqnarray*}
&& |V_\eta(t)| \leq |V_\eta (s_{i-1})| + |\mathcal{I}_i| | \dot{V}_\eta(s_{i-1})| + \int_{\mathcal{I}_i} |\dot{V}_\eta(u) - \dot{V}_\eta(s_{i-1})| du. \\
\end{eqnarray*}
Note that  $N$ is free of $\eta$. 
We will use above two inequalities to prove by induction that the upper bounds of $V_\eta(t) $ and $\dot{V}_\eta(t)$ on $[0, T]$ are a.s. finite uniformly over $\eta$.  
Assume that the upper bounds of $V_\eta(t) $ and $\dot{V}_\eta(t)$ on $\cup_{j=1}^{i-1} \mathcal{I}_j$ are a.s. finite uniformly over $\eta$. 
Above two inequalities immediately show that their upper bounds on $ \mathcal{I}_i$ are also a.s. finite uniformly over $\eta$.
This implies that  the uniform finite bounds of $V_\eta(t) $ and $\dot{V}_\eta(t)$ on $\cup_{j=1}^{N} \mathcal{I}_j=[0,T]$, and thus 
$V_\eta(t)$ is stochastically equicontinuous and stochastically bounded on  $[0,T]$.

\begin{lem}  \label{lem-s7}
For fixed $(\delta, m)$, the second order SDEs (\ref{Nest-stoch1}) and  (\ref{Nest-limit-0}) have unique solutions in the distributional sense. 
\end{lem}
Proof. Due to the similarity we provide proof arguments for (\ref{Nest-limit-0}) only. 
Take a decreasing sequence of $\eta$ as follows: $\eta_k$, $k=1, 2, \cdots$, are decreasing, and as $k \rightarrow \infty$, $\eta_k \rightarrow 0$.
Lemma \ref{lem-s6} implies that $\{V_{\eta_k}(t), k=1,2, \dots, \}$ is tight and thus there exists a subsequence that has a weak limit process $V_\dagger(t)$. We will show that $V_\dagger(t)$ satisfies (\ref{Nest-limit-0}).  Without loss of generality, we may assume 
$V_{\eta_k}(t)$ weakly converges to $V_\dagger(t)$, and using SKorohod's representation theorem we may further assume that 
$V_{\eta_k}(t)$ converges to $V_\dagger(t)$ a.s.  
$V_{\eta_k}(t)$ obey the initial condition $V_{\eta_k}(0)= \dot{V}_{\eta_k}(0)=0$, thus 
$V_\dagger(0)=0$, and 
\[ \frac{ | V_\dagger(t) - V_\dagger(0)|}{t} = \lim_{k \rightarrow \infty} \frac{ | V_{\eta_k}(t) - V_{\eta_k}(0)|}{t} 
= \lim_{k \rightarrow \infty} | \dot{V}_{\eta_k}(\xi_k)| \leq \limsup_{k \rightarrow \infty} [t^a M_a(0,t, V_{\eta_k})].
\]
Since $M_a(0,t, V_{\eta_k})$ is a.s. finite uniformly over $\eta_k$, taking $t \rightarrow 0$ we obtain  $\dot{V}_\dagger(0) = 0$.
%Note that $(V_{\eta_k}(t), Y_{\eta_k}(t))$ satisfy 
For $t > \eta_k$, the second order SDE (\ref{equ-b2}) is equivalent to the following smoothed stochastic differential equation system 
\begin{eqnarray*} 
% \dot{V}_{\eta_k}(t) &=& \frac{2}{t \vee \eta_k} \, Y_{\eta_k}(t) - \frac{2}{t \vee \eta_k} \, V_{\eta_k}(t) \\
% \dot{Y}_{\eta_k}(t) &=&  -\frac{t \vee \eta_k }{2}\, [\nabla \! g(X(t))] V_{\eta_k}(t) - \frac{t \vee \eta_k}{2}\, \bsigma(X(t)) \dot{\bB}(t).  
\dot{V}_{\eta_k}(t) &=& \frac{2}{t} \, Y_{\eta_k}(t) - \frac{2}{t} \, V_{\eta_k}(t) \\
\dot{Y}_{\eta_k}(t) &=&  -\frac{t}{2}\, [\nabla \! g(X(t))] V_{\eta_k}(t) - \frac{t}{2}\, \bsigma(X(t)) \dot{\bB}(t).  
 \end{eqnarray*}
Its inherited  initial conditions are $ V_{\eta_k} (0) = Y_{\eta_k}(0) =c$ and $\dot{V}_{\eta_k}(0)=0$. %$\dot{Y}_{\eta_k}(0)=0$. 
The right hand side of the second equation in above system 
implies that as $k \rightarrow \infty$, $Y_{\eta_k}(t)$ converges to $Y(t)$ defined by  
\[ \dot{Y}(t) = -\frac{t}{2}\, [\nabla \! g(X(t))] V_\dagger(t) - \frac{t }{2}\, \bsigma(X(t)) \dot{\bB}(t), \;\; Y(0)=c, \]
which in turn shows that $\dot{V}_{\eta_k}(t)$ converges to $\dot{V}_*(t)$ given by 
\[ \dot{V}_*(t) = \frac{2}{t } \, Y(t) - \frac{2}{t } \, V_\dagger(t). \]
Since $V_{\eta_k}(t)$ converges to $V_\dagger(t)$, $\dot{V}_*(t)=\dot{V}_\dagger(t)$. Thus $V_\dagger(t)$ satisfies 
\[ \dot{V}_\dagger(t) = \frac{2}{t } \, Y(t) - \frac{2}{t } \, V_\dagger(t), \]
which impies $V_\dagger(t)$ obeys 
% \[ \ddot{V}_{\eta_k}(t) + \frac{3}{t \vee \eta_k } \dot{V}_{\eta_k}(t) + [\nabla g(X(t))] V_{\eta_k}(t) + \bsigma(X(t)) \dot{\bB}(t)  = 0, \]
 \[ \ddot{V}_\dagger(t) + \frac{3}{t } \dot{V}_{\dagger}(t) + [\nabla g(X(t))] V_{\dagger}(t) + \bsigma(X(t)) \dot{\bB}(t)  = 0. \] 
 
 Suppose that the equation has two solutions $(V(t), \bB(t))$ and $(V_*(t), \bB_*(t))$. Then we may 
 realize both solutions on some common probability space such that $\bB(t) = \bB_*(t)$. Then $U(t)=V(t) - V_*(t)$ obey 
 \[ \ddot{U}(t) + \frac{3}{t } \dot{U}(t) + [\nabla g(X(t))] U(t)   = 0,  \qquad U(0) = \dot{U}(0)=0,\]  
 which has a unique solution zero, as it is a second order ODE similar to   ODEs (\ref{equ-2}) and (\ref{limit-0}). 
 Thus $V(t) = V_*(t)$, that is, the two solutions have an identical distribution, which proves the 
 unique solution. 
 
\subsubsection{Weak convergence of $V^m_\delta(t)$}
%The basic proof idea is similar to that for $V^n(t)$ in Section \ref{section-Xn}.
\begin{lem} \label{lem-s8}
For $X(t)$, $X^m_\delta(t)$ and $V^m_\delta(t)$ we have the following inequalities,
\begin{align*}
& M_1(s,t; X) \leq \frac{1}{ 1 - L (t-s)^2/6} \left[ \left( \frac{3}{s}  + \frac{ L(t-s)}{2} \right) |\dot{X}(s)| + |\nabla g( X(s))|   \right], 
\mbox{if }\,  t -s < \sqrt{\frac{3}{L}}, \\
& M_a(s,t; X^m_\delta) \leq \frac{1}{ 1 - L (t-s)^2/[(a+1)(a+2)]}  \left[ (t-s)^{1-a} \left( \frac{3}{s}  + \frac{ L(t-s)}{2} \right) |\dot{X}^m_\delta(s)| 
 \right. & \\
& \left. + (t-s)^{1-a} |\nabla g( X^m_\delta(s))| +  \max_{v \in (s, t]} 
   \frac{\delta^{1/4} m^{-1/2}}{4 v^3 (v-s)^a}  \left|   \int_s^v  u^3 \bsigma(X(u)) d \bB(u)   \right | 
   \right],    & \\
%\end{align*}
%\begin{align*}
& M_a(s,t;V^m_\delta) \leq \frac{1}{ 1 - L (t-s)^2/[(a+1)(a+2)]} &\\
&  \left[ (t-s)^{1-a} \left\{ 
  2  L |V^m_\delta(s)| + [3/s + L (t-s) ] |\dot{V}^m_\delta(s)| \right\}   \right.   &\\ &\left.  %+(t-s)^{1-a}  \delta^{-1/4} (mT)^{1/2} |\nabla \![g(X^m_\delta(s)) - g(X(s)])|  
+ \max_{v \in (s, t]} \frac{1 }{v^3 (v-s)^a} \left|  \int_s^v u^3 \bsigma(X(u)) \dot{\bB}(u) du \right| 
   \right] , &
\end{align*}
when $s>0$ and $t-s < \sqrt{(a+1)(a+2)/(2 L)}$.   In particular, for $s=0$ we have 
\[ M_1(0,t; X) \leq \frac{  |\nabla g(x_0)| }{ 1 - L t^2/6}, \]
\[ M_a(0,t; X^m_\delta) \leq \frac{ t^{1-a} |\nabla g(x_0)| + \max_{v \in (s, t]} \frac{\delta^{1/4} (mT)^{-1/2}}{4 v^{3+a}}  \left|   \int_0^v  u^3 \bsigma(X(u)) d \bB(u)   \right |     }{ 1 - L t^2/[(a+1)(a+2)]}, 
\]
\[ M_a(0,t;V^m_\delta) \leq \frac{1}{ 1 - L t^2/[(a+1)(a+2)]} \max_{v \in (0, t]} \left[ \frac{ 1 }{v^{3+a}} \left|  \int_0^v u^3 \bsigma(X(u)) \dot{\bB}(u) du \right|   \right]. \]
 \end{lem}
%\subsubsection{Weak convergence of $V^m_\delta(t)$}
%The basic proof idea is similar to that for $V^n(t)$ in Section \ref{section-Vn}.
%\begin{lem}  \label{lem-s11}
%For $V^m_\delta(t)$ we have the following inequalities,
%\begin{align*}
%& M_a(s,t;V^m_\delta) \leq \frac{1}{ 1 - L (t-s)^2/[(a+1)(a+2)]} &\\
%&  \left[ (t-s)^{1-a} \left\{   2  L |V^m_\delta(s)| + [3/s + L (t-s) ] |\dot{V}^m_\delta(s)| \right\}   \right.   &\\ &\left.  %+(t-s)^{1-a}  \delta^{-1/4} (mT)^{1/2} |\nabla \![g(X^m_\delta(s)) - g(X(s)])|  + \max_{v \in (s, t]} \frac{1 }{v^3 (v-s)^a} \left|  \int_s^v u^3 \bsigma(X(u)) \dot{\bB}(u) du \right|    \right] , &
%\end{align*}
%when $s>0$ and $t-s < \sqrt{(a+1)(a+2)/(2 L)}$.  In particular for $s=0$, we have 
%\[ M_a(0,t;V^m_\delta) \leq \frac{1}{ 1 - L t^2/[(a+1)(a+2)]} \max_{v \in (0, t]} \left[ \frac{ 1 }{v^{3+a}} \left|  \int_0^v u^3 \bsigma(X(u)) \dot{\bB}(u) du \right|   \right]. \]
% \end{lem}
%$V^m_\delta(t)$ satisfies the differential equation 
%\begin{equation} \label{equ-l1}
% \ddot{V}^m_\delta(t) + \frac{3}{t}  \dot{V}^m_\delta(t) + \sqrt{n} \nabla \![g(X^m_\delta(t)) - g(X(t))] + T^n(X^n(t);\bU,Q) =0. 
%\end{equation}
Proof. 
Because of similarity, we provide proof arguments only for $M_1(s,t; V^m_\delta $. Let
 %\[  \ddot{X}^m_\delta(t)+\frac{3}{t}\dot{X}^m_\delta(t) +  \nabla \!g(X^m_\delta(t)) + \frac{1}{\sqrt{n}} T^n(X^m_\delta(t);\bU,Q) =0. \]
%\begin{align*}
%& [\ddot{X}^m_\delta_\dagger(t) - \ddot{X}(t)] + \frac{3}{t} [ \dot{X}^m_\delta_\dagger(t) - \dot{X}(t)] + \nabla \![g(X^m_\delta_\dagger(t)) - g(X(t))] + n^{-1/2} T^n(X^m_\delta(t);\bU,Q)   =0. & \\
%& \ddot{V}^m_\delta(t) + \frac{3}{t}  \dot{V}^m_\delta(t) + \boldsymbol{I\!\! H}\! g(X(t)) V^m_\delta(t) + n^{-1/2} T^n(X^m_\delta(t);\bU,Q) 
%  =0, & \\
%&  \ddot{V}^m_\delta(t) + \frac{3}{t}  \dot{V}^m_\delta(t) + \sqrt{n} \nabla \![g(X^m_\delta(t)) - g(X(t))] + T^n(X^m_\delta(t);\bU,Q) =0, &\\
%& H(t;V^m_\delta) = \delta^{-1/4} (m T)^{1/2} [ \nabla \! g(X^m_\delta(t)) - \nabla g(X(t))], %+ T^n(X^m_\delta(t);\bU,Q), 
%&
%\end{align*}
$H(t;V^m_\delta) = \delta^{-1/4} m^{1/2} [ \nabla \! g(X^m_\delta(t)) - \nabla g(X(t))]$, and $J(s,t; H, V^m_\delta) = \int_s^t u^3 [ H(u;V^m_\delta) - H(s;V^m_\delta)] du$. Then

\begin{align*}
& |H(t;V^m_\delta)| \leq  %|\boldsymbol{I\!\! H}\! g(X^m_{\delta, \xi})  
L \delta^{-1/4} m^{1/2} | X^m_\delta(t)  - X(t) | = L | V^m_\delta(t)|, &\\
& | H(t;V^m_\delta) - H(s;V^m_\delta) | = \delta^{-1/4} m^{1/2} |\nabla [g( X^m_\delta(t) ) - g(X^m_\delta(s)) - g(X(t)) + g(X(s))]|  &\\
& \leq L \delta^{-1/4} m^{1/2} | X^m_\delta(t) - X(t)| + L \delta^{-1/4} m^{1/2}  |X^m_\delta(s) - X(s) | =  L |V^n(t)| + L | V^m_\delta(s)|, & \\
%& \delta^{-1/4} (m T)^{1/2} [g( X^m_\delta(t) ) - g(X^m_\delta(s)) - g(X(t)) + g(X(s))] &\\
%& = \nabla g (X^m_{\delta,\xi}) \delta^{-1/4} (m T)^{1/2} [X^m_\delta(t) - X(t)]  - \,\nabla g (X^m_{\delta, \xi}) \delta^{-1/4} (m T)^{1/2}  [X^m_\delta(s) - X(s)] & \\
%& = \nabla g (X^m_{\delta,\xi}) \int_s^t \dot{V}^m_\delta(u) du + [\nabla g (X^m_{\delta,\xi}) - \nabla g (X^m_{\delta,\xi})] V^m_\delta(s), & \\
& %\delta^{-1/4} (m T)^{1/2}  [X^m_\delta(t) - X(t)] = 
V^m_\delta(t)=\int_s^t \dot{V}^m_\delta(u) du + V^m_\delta(s) = \int_s^t [ \dot{V}^m_\delta(u) - \dot{V}^n(s)] du + V^m_\delta(s) + (t-s) \dot{V}^n(s), & 
\end{align*} 
\begin{align*} 
& | H(t;V^m_\delta) - H(s;V^m_\delta) | \leq L \int_s^t |\dot{V}^m_\delta(u) - \dot{V}^m_\delta(s)| du + L [ 2 |V^m_\delta(s)| +|(t-s)  \dot{V}^m_\delta(s)|] , &\\ 
&  \int_s^t |\dot{V}^m_\delta(u) - \dot{V}^m_\delta(s)| du \leq \int_s^t (u-s)^a \frac{|\dot{V}^m_\delta(u) - \dot{V}^m_\delta(s)|}{(u-s)^a} du \leq \int_s^t (u-s)^a M_a(s,t;V^m_\delta) du &\\
& = \frac{M_a(s,t;V^m_\delta) (t-s)^{a+1} }{a+1}, &\\
& \frac{L}{a+1} \int_s^t M_a(s,u;V^m_\delta) u^3 (u-s)^{a+1} du  \leq \frac{ L M_a(s,t;V^m_\delta) t^3 (t-s)^{a+2}}{(a+1)(a+2)}, & \\
& |J(s,t;H,V^m_\delta)| \leq \frac{L t^3 (t-s)^{a+2}}{(a+1)(a+2)} M_a(s,t;V^m_\delta) + L [ 2 |V^m_\delta(s)| +(t-s)| \dot{V}^m_\delta(s)|] t^3(t-s).&
\end{align*}

The SDE (\ref{Nest-stoch1}) is equivalent to 
\begin{align*} 
& \frac{ t^3 \dot{V}^m_\delta(t)}{dt} = - t^3 H(t;V^m_\delta) -  t^3 \bsigma(X(t)) \dot{\bB}(t), \;\; \mbox{which implies} &\\
& t^3 \dot{V}^m_\delta(t) -  s^3 \dot{V}^m_\delta(s) = -\int_s^t u^3 H(u;V^m_\delta) du -  \int_s^t u^3 \bsigma(X(u)) \dot{\bB}(u) du &\\
& =  - \frac{ t^4-s^4}{4} H(s;V^m_\delta)- J(s,t;H,V^m_\delta) -  \int_s^t u^3 \bsigma(X(u)) \dot{\bB}(u) du ,&\\
& \frac{ \dot{V}^m_\delta(t) - \dot{V}^m_\delta(s)}{t-s} = -\frac{t^3-s^3}{t^3(t-s)} \dot{V}^m_\delta(s) - \frac{ t^4-s^4}{4 t^3(t-s)} H(s;V^m_\delta) - \frac{J(s,t;H,V^m_\delta)}{t^3(t-s)} &\\
& - \frac{ 1 }{t^3 (t-s)}  \int_s^t u^3 \bsigma(X(u)) \dot{\bB}(u) du,  &
\end{align*} 
and using the upper bounds of $H(s; V^m_\delta)$ and $J(s, t; H, V^m_\delta)$ and algebraic manipulations we get 
\begin{align*}   
&  \frac{| \dot{V}^m_\delta(t) - \dot{V}^m_\delta(s)|}{t-s }\leq \frac{t^3-s^3}{t^3(t-s)} |\dot{V}^m_\delta(s)|  +  \frac{ t^4-s^4}{4 t^3(t-s)} |H(s;V^m_\delta)| + \frac{|J(s,t;H,V^m_\delta)|}{t^3 (t-s)} &\\
& + \frac{ 1 }{t^3 (t-s)} \left|  \int_s^t u^3 \bsigma(X(u)) \dot{\bB}(u) du \right| &\\
& \leq \frac{t^2 + st + s^2}{t^3} |\dot{V}^m_\delta(s)| + \frac{ (t^2+s^2)(t+s)}{2 t^3}  L |V^m_\delta(s)| %|H(s;V^m_\delta)| + 
    +  M_a(s,t;V^m_\delta) \frac{L (t-s)^{a+1}}{(a+1)(a+2)}  & \\ 
& + L [ 2 |V^m_\delta(s)| + (t-s) | \dot{V}^m_\delta(s)|]     
  + \frac{ 1 }{t^3 (t-s)} \left|  \int_s^t u^3 \bsigma(X(u)) \dot{\bB}(u) du \right| . &
\end{align*} 
As above inequality holds for any $s<t$, an application of the definition of $M_a(s,t; V^m_\delta)$ leads to 
\begin{align*}   
&M_a(s,t;V^m_\delta) \leq (t-s)^{1-a} \left \{ \frac{3}{s} |\dot{V}^m_\delta(s)| + L [ 4 |V^m_\delta(s)| + (t-s)| \dot{V}^m_\delta(s)|]  \right\} &\\
& + M_a(t,s;V^m_\delta) \frac{ L(t-s)^{2}}{(a+1)(a+2)} +  \max_{v \in (s, t]} 
   \frac{ 1 }{v^3 (v-s)^a} \left|  \int_s^v u^3 \bsigma(X(u)) \dot{\bB}(u) du \right| ,  &
 \end{align*} 
and solving for  $M_a(s,t: V^m_\delta)$ to obtain 
\begin{align*}     
& M_a(s,t;V^m_\delta) \leq \frac{1}{ 1 - L (t-s)^2/[(a+1)(a+2)]} &\\
&   \left[ (t-s)^{1-a} \left\{ 
  4  L |V^m_\delta(s)| + [3/s + L (t-s) ] |\dot{V}^m_\delta(s)| \right\} \right. &\\
&\left.  %+(t-s)^{1-a}  \delta^{-1/4} (mT)^{1/2} |\nabla \![g(X^m_\delta(s)) - g(X(s)])|  
+ \max_{v \in (s, t]} \frac{1 }{v^3 (v-s)^a} \left|  \int_s^v u^3 \bsigma(X(u)) \dot{\bB}(u) du \right| 
   \right] , &
\end{align*}
when $s>0$ and $t-s < \sqrt{(a+1)(a+2)/(2 L)}$.  If $s=0$, we replace  the coefficient $3/s$ by $1/t$ in  above inequality,  and $V^m_\delta(0) = \dot{V}^m_\delta(0) =0$, $X^m_\delta(0)=X(0)=x_0$.  Then 
\[ M_a(0,t;V^m_\delta) \leq \frac{1}{ 1 - L t^2/[(a+1)(a+2)]} \max_{v \in (0, t]} \left[ \frac{ 1 }{v^{3+a}} \left|  \int_0^v u^3 \bsigma(X(u)) \dot{\bB}(u) du \right|   \right], \]
which proves the lemma.

%The bound on $M_a(0,t;V^m_\delta) $ implies %which in particular implies that 
%\[ \sup_{ t \leq \sqrt{(a+1)(a+2)/ (2 L)}} \frac{ |\dot{X}^m_\delta(t) - \dot{X}(t)| }{ t^a} \leq \max_{v \in (0, t]} 
%\frac{ 2 \delta^{1/4} (mT)^{-1/2} }{v^{3+a}} \left|  \int_0^v u^3 \bsigma(X(u)) \dot{\bB}(u) \right|      \rightarrow 0,\]
%that is, $\dot{X}^m_\delta(t) \rightarrow \dot{X}(t)$ uniformly over $\mathcal{I}_1=\left[0, \sqrt{(a+1)(a+2)/(2 L)}\right]$. 

\begin{lem} \label{lem-s12}
For any given $T>0$, we have 
\[  \max_{t \in [0, T]} | V^m_\delta(t) | = O_P(1), \;\;\;  \max_{t \in [0, T]} | X^m_\delta(t) - X(t) | = O_P( \delta^{1/4} m^{-1/2} ),  \]
\[ \max_{t \in [0, T]} | \dot{V}^m_\delta(t) | = O_P(1), \;\;\; \max_{t \in [0, T]} | \dot{X}^m_\delta(t) - \dot{X}(t) | = O_P( \delta^{1/4} m^{-1/2} ). \]
%As $\delta \rightarrow 0$, and $m,n \rightarrow \infty$, $V^m_\delta(t)$ weakly converges to $V(t)$.
%$V^m_\delta(t)$ is stochastically equicontinuous and stochastically bounded on $[0, T]$.
\end{lem}
Proof. As $V^m_\delta (t) = \delta^{-1/4} m^{1/2} [ X^m_\delta(t) - X(t)]$, we need to establish the results for $V^m_\delta(t) $ only. 
Divide interval $[0,T]$ into $N=\left[T \sqrt{2 L/ \{(a+1)(a+2)\}} \right]+1$ number of subintervals with length $\sqrt{(a+1)(a+2)/(2 L)}$ (except for the last one) and denote by $\mathcal{I}_i=[s_{i-1}, s_i]$, $i=1, \cdots, N$ (with $s_0=0$, $s_N=T$, $\mathcal{I}_1 = [0, \sqrt{3/L}]$, $\mathcal{I}_N = [s_{N-1}, T]$).
First for $t \in \mathcal{I}_1$, from Lemma \ref{lem-s8} we have 
\[  |\dot{V}^m_\delta(t)| \leq |\mathcal{I}_1|^a M_a(\mathcal{I}_1; V^m_\delta) \leq C \max_{v \in (0, s_1]} \left[ \frac{ 1 }{v^{3+a}} \left|  \int_0^v u^3 \bsigma(X(u)) \dot{\bB}(u) du \right|   \right], \]
\[ |V^m_\delta(t)| \leq |V^m_\delta(0)| + \int_{\mathcal{I}_1} |\dot{V}^m_\delta(u) | du \leq C \max_{v \in (0, s_1]} \left[ \frac{ 1 }{v^{3+a}} \left|  \int_0^v u^3 \bsigma(X(u)) \dot{\bB}(u) du \right|   \right]. \] 
The the upper bounds of  $V^m_\delta(t)$ and $ \dot{V}^m_\delta(t)$ on $\cI_1$ are a.s. finite uniformly over $(\delta, m)$. 
%and thus $V^m_\delta(t)$ is  stochastically equicontinuous and stochastically bounded over $\mathcal{I}_1$. 
%This together with finite distribution convergence show that $V^m_\delta(t)$ weakly converges to $V(t)$ on $\mathcal{I}_1$. 
%Moreover, 
%\[ \sup_{ t \leq \sqrt{3/L}} \frac{ |\dot{V}^m_\delta(t) | }{ t } \leq 6 \sup_\theta |T^n(\theta; \bU,Q)| \not \rightarrow 0,\]

For $t \in \mathcal{I}_i$, $i =2, \cdots, N$, from Lemma \ref{lem-s8} we have 
\begin{align*}
&  |\dot{V}^m_\delta(t)-\dot{V}^m_\delta(s_{i-1})| \leq |\mathcal{I}_i|^a M_a(\mathcal{I}_i, V^m_\delta) \leq C 
   \left[ 4 L |V^m_\delta(s_{i-1})| + (3/s_{1} + L s_1 ) |\dot{V}^m_\delta(s_{i-1})|  \right] &\\
&  +  C \max_{v \in (s_{i-1}, s_i]} 
\frac{1 }{v^3 (v-s_{i-1})^a} \left|  \int_{s_{i-1}}^v u^3 \bsigma(X(u)) \dot{\bB}(u) du \right|  ,  & \\
& |V^m_\delta(t)| \leq |V^m_\delta(s_{i-1})| + |\mathcal{I}_i| | \dot{V}^m_\delta(s_{i-1})| + \int_{\mathcal{I}_i} |\dot{V}^m_\delta(u) - \dot{V}^m_\delta(s_{i-1})| du &\\
& \leq |V^m_\delta(s_{i-1})| + \sqrt{3/L} | \dot{V}^m_\delta(s_{i-1})| + C 
    \left[ 4 L |V^m_\delta(s_{i-1})| + (3/s_{1} + L s_1 ) |\dot{V}^m_\delta(s_{i-1})| \right] &\\ 
    & + C \max_{v \in (s_{i-1}, s_i]} 
\frac{1 }{v^3 (v-s_{i-1})^a} \left|  \int_{s_{i-1}}^v u^3 \bsigma(X(u)) \dot{\bB}(u) du \right|  . 
 \end{align*}
 %As $N$ is finite free of $n$, $V^m_\delta(s_i)$ jointly converge to $V(s_i)$, $\dot{V}^m_\delta(s_i) \rightarrow \dot{V}(s_i)$,
%$X^m_\delta(s) \rightarrow X(s)$ uniformly. We need to prove $X^m_\delta(s)$ is equicontinuous. 
%Then $H(s; V^m_\delta) \rightarrow \boldsymbol{I\!\! H}\! g (X(s)) V(s) + \bsigma(X(s)) \bZ$.

We will use above two inequalities to prove by induction that the upper bounds of $V^m_\delta(t) $ and $\dot{V}^m_\delta(t)$ on $[0, T]$ are a.s. finite uniformly 
over $(m, \delta)$.  
Assume that the upper bounds of $V^m_\delta(t) $ and $\dot{V}^m_\delta(t)$ on $\cup_{j=1}^{i-1} \mathcal{I}_j$ are a.s. finite uniformly over $(m,\delta)$. 
Note that $\max_{v \in (s_{i-1}, s_i]} 
\frac{1 }{v^3 (v-s_{i-1})^a} \left|  \int_{s_{i-1}}^v u^3 \bsigma(X(u)) \dot{\bB}(u) du \right| $ is a.s. finite, and  $N$ is free of $(m, \delta)$. 
Above two inequalities %and the convergence of $X^m_\delta(t)$ to $X(t)$ 
 immediately show that the upper bounds of $V^m_\delta(t) $ and $\dot{V}^m_\delta(t)$ on $ \mathcal{I}_i$ are also a.s. finite uniformly over $(m,\delta)$.
This implies that  the uniform finite bounds of $V^m_\delta(t) $ and $\dot{V}^m_\delta(t)$ on $\cup_{j=1}^{N} \mathcal{I}_j=[0,T]$. 

%and thus $V^m_\delta(t)$ is stochastically equicontinuous and stochastically bounded on  $[0,T]$.

\begin{lem} \label{lem-s10}
For any given $T>0$, as $\delta \rightarrow 0$ and $m \rightarrow \infty$, $V^m_\delta(t)$  is stochastically equicontinuous on $[0, T]$. 
\end{lem}
Proof. Lemma \ref{lem-s12} proves that $\max_{t \in [0, T]} |V^m_\delta(t)|=O_P(1)$ and $\max_{t \in [0, T]} |\dot{V}^m_\delta(t)| = O_P(1)$, 
which implies that $V^m_\delta(t)$ is stochastically equicontinuous on $[0, T]$. 

\textbf{Proof  of Theorem \ref{thm6}}.
Lemma \ref{lem-s7} %Proposition \ref{prop2} 
shows the unique solutions of SDEs. \eqref{Nest-limit-0} is a linear SDE, and its constant term linearly depends on 
$\bB(t)$, thus its solution $V(t)$ is Gaussian. 
As in Section \ref{section-tight}, we can easily establish finite distribution convergence for $V^m_\delta(t)$. 
Lemma \ref{lem-s10} together with the finite distribution convergence immediately lead to that as $\delta \rightarrow 0$ 
and $m \rightarrow \infty$, $V^m_\delta(t)$ weakly converges to $V(t)$.

%\begin{lem} \label{lem12}
%As $\delta \rightarrow 0$, and $m,n \rightarrow \infty$, $V^m_\delta(t)$ weakly converges to $V(t)$.
%\end{lem}

%% file: p5.tex
\subsection{Proof of Theorem \ref{thm7} }
\label{sec-thm7}

\newcommand{\cx}{\check{x}}
\newcommand{\cy}{\check{y}}
\newcommand{\cz}{\check{z}}
\newcommand{\cd}{\check{d}}
\newcommand{\ccG}{\check{G}}
\newcommand{\tiF}{\tilde{F}}
\newcommand{\tiB}{\tilde{B}}
\newcommand{\tiG}{\tilde{G}}
\newcommand{\tiH}{\tilde{H}}
\newcommand{\tid}{\tilde{d}}
\newcommand{\tix}{\tilde{x}}
\newcommand{\tiy}{\tilde{y}}
\newcommand{\tiR}{\tilde{R}}
\newcommand{\tcx}{\tilde{\check{x}}}
\newcommand{\tcy}{\tilde{\check{y}}}
\newcommand{\tcd}{\tilde{\check{d}}}
\newcommand{\ve}{{\varepsilon}}

Recall that sequences $\{x_k, y_k\}$ and $\{x_{k}^{m}, y_{k}^{m}\}$ are, respectively, defined by algorithms (\ref{equ-Nest1}) and 
(\ref{min-Nest3}),  with initial values $x_{0}^{m}=y_{0}^{m}=x_{0}$, and $X_\delta^m(t)$ and $X(t)$ are the solutions of 
ODE (\ref{equ-2}) and SDE (\ref{Nest-stoch1}), respectively. 

%\[  \max_{k\leq T \delta^{-1/2} }\left|\cx_k^m-X_\delta^m(t_k)\right|  = O_p(\delta^{1/2}|\log\delta|),  \]
%where $\cx_k^m$ and $X_\delta^m$ are defined by (\ref{lm3-2}) and (\ref{Nest-stoch1}), 

We discretize SDE (\ref{Nest-stoch1-1}), which is equivalent to (\ref{Nest-stoch1}), to define a new sequence as follows. 
Let $\{\check{x}_{k}^{m},\check{y}_{k}^{m}\}$ be the sequence, with initial values $\check{x}_{0}^{m}=\check{y}_{0}^{m}=x_{0}$, generated by
\begin{equation} \label{lm3-2}
\check{x}_{k}^{m}=\check{y}_{k-1}^{m}-\delta\nabla g(\check{y}_{k-1}^{m})-m^{-1/2}\delta^{3/4} [ H(t_{k})-H(t_{k-1})],  \check{y}_{k}^{m}=\check{x}_{k}^{m}+\frac{k-1}{k+2}(\check{x}_{k}^{m}-\check{x}_{k-1}^{m}),
\end{equation}
where $H(t)=\displaystyle\int_{0}^{t}\bsigma(X(u))d\bB(u)$ and $t_{k}=k\delta^{1/2}$.

We rewrite algorithm (\ref{min-Nest3}) to generate $\{x_{k}^{m}, y_{k}^{m}\}$ as follows, 
\begin{equation} \label{lm3-1}
x_{k}^{m}=y_{k-1}^{m}-\delta\nabla g(y_{k-1}^{m})-m^{-1/2}\delta^{3/4} [ H_\delta^m(t_{k})-H_\delta^m(t_{k-1})],  
%\delta\nabla\hat{\mathcal{L}}^{m}(y_{k-1}^{m};\bU_{mk}^{*}),
y_{k}^{m}=x_{k}^{m}+\frac{k-1}{k+2}(x_{k}^{m}-x_{k-1}^{m}).
\end{equation}

%Algorithms (\ref{min-Nest3}) 
Note that (\ref{lm3-2}) and (\ref{lm3-1}) share the same recursive structure with only difference between $H_\delta^m(t)$ and $H(t)$. 
Our proof approach is to (i) Lemma \ref{lm3} below shows that $\{x_{k}^{m}, y_{k}^{m}\}$ and $\{\check{x}_{k}^{m},\check{y}_{k}^{m}\}$ can be realized on some probability spaces within small order distance; (ii) Lemma \ref{lm6} below derives an order bound for discretization error 
$\check{x}_{k}^{m} - X_\delta^m(t_k)$; (iii) the theorem is proved by combining two lemmas in (i) and (ii). 

%\begin{spacing}{1.5}

\begin{lem} \label{lm-1}
\[ \max_{k\leq T\delta^{-1/2}}|x_k-X(t_k)|=O(\delta^{1/2}|\log\delta|), \;\;
\max_{k\leq T\delta^{-1/2}}|z_k- \dot{X}(t_k)|=O(\delta^{1/2}|\log\delta|),  \]
\[\max_{k\leq T\delta^{-1/2}}|y_k-x_k| = O(\delta^{1/2}). \]
\end{lem}
Proof.   
As $X(t)$ is the solution of ODE (\ref{equ-2}), it has been shown that $X(t)$, $\dot{X}(t)$ and $\nabla g(X(t))$ are uniformly 
bounded on $[0,T]$, and Lemma \ref{lem1} further indicates that $\dot{X}(t)$ is Lipschitz. 
Let $Z(t)=\dot{X}(t)$. With deterministic sequence $\{x_k, y_k\}$ given by algorithm (\ref{equ-Nest1}), we define 
%\[x_k=y_{k-1}-\delta\nabla g(y_{k-1}), y_k=x_k+\frac{k-1}{k+2}(x_k-x_{k-1}),\  y_0=x_0 \]
$z_0=0$, $z_k=(x_k-x_{k-1})/\delta^{1/2}$. Lemma \ref{lem-add-1} has shown that 
%similar to the proof of the ASGD case, we have 
\[ \max_{k\leq T\delta^{-1/2}}|x_k-X(t_k)|=O(\delta^{1/2}|\log\delta|), \;\;
\max_{k\leq T\delta^{-1/2}}|z_k-Z(t_k)|=O(\delta^{1/2}|\log\delta|),  \]
and $ y_k - x_k = \frac{3k + 4}{k+3} \delta^{1/2} z_k$, which implies that 
\[\max_{k\leq T\delta^{-1/2}}|y_k-x_k| \leq 3 \delta^{1/2} \left( \max_{k\leq T\delta^{-1/2}}| z_k - Z(t_k)| + \max_{k\leq T\delta^{-1/2}} 
  |Z(t_k)| \right) = O(\delta^{1/2}). \]
%Recall ASGD
%\[x_k^m=y_{k-1}^m-\delta\nabla g(y_{k-1}^m)-\delta m^{-1/2}R^m(y_{k-1}^m;\bU_{mk}^*), y_k^m=x_k^m+\frac{k-1}{k+2}(x_k^m-x_{k-1}^m),\  y_0^m=x_0^m=x_0 \]
%where
%\[R^m(\theta;\bU_{mk}^*)=m^{1/2}(\nabla\hat\cL^m(\theta;\bU_{mk}^*)-\nabla g(\theta))\]

As in the proofs of Theorems \ref{thm2}-\ref{thm4}, we use  notations 
$R^m(\theta; \bU_m^*(t)) = (R^m_1(\theta;\bU_m^*(t)), \cdots, R^m_p(\theta;\bU_m^*(t)))^\prime$, where 
\begin{equation*}
R_{j}^{m}(\theta;\bU_m^*(t))=\sqrt{m}\left[ \frac{1}{m} \sum_{i=1}^m  \frac{\partial}{\partial \theta_j} \ell(\theta; U_{i}^*(t))-\frac{\partial}{\partial \theta_j}g(\theta)\right], \;\; j =1, \cdots, p.
\end{equation*}

\begin{lem} \label{lm-0}
\[  \max_{k \leq T \delta^{-1/2}} E[ |R^{m}(X(t_{k-1}); \bU_{m k}^{*})|^4] \leq C. \]
 \end{lem} 
 Proof. For simplicity we write $R_{i}=R^{m}(X(t_i); \bU_{m(i+1)}^{*})$, and $\mathrm{r}_{q} = \nabla\ell (X(t_i); U^*_{q(i+1)}) -\nabla g(X(t_i))$. 
 Then $R_{i}=m^{-1/2}\displaystyle \sum_{q=1}^{m}\mathrm{r}_{q}$. Since $X(t)$ is deterministic, and $U_{q k}^*$, $q=1, 2, \cdots,m$, are independent, we have that $\mathrm{r}_{1}, \mathrm{r}_{2}, \dots, \mathrm{r}_{m}$ are independent with mean $0$, and 
$$|\mathrm{r}_{q}|\leq h_{1}(U^*_{q(i+1)})|X(t_{i})-\theta_{0}|+|\nabla\ell(\theta_{0};U^*_{q(i+1)})|+|\nabla g(X(t_{i}))|,$$
$$E|\mathrm{r}_{q}|^{2}\leq 3\cdot(E[h_{1}^{2}(U)]|X(t_{i})-\theta_{0}|^{2}+E|\nabla\ell(\theta_{0};U)|^{2}+|\nabla g(X(t_{i}))|^{2}),
$$
$$E|\mathrm{r}_{q}|^{4}\leq 27\cdot(E[h_{1}^{4}(U)]|X(t_{i})-\theta_{0}|^{4}+E|\nabla\ell(\theta_{0};U)|^{4}+|\nabla g(X(t_{i}))|^{4}). $$
Note that $\displaystyle \sup_{t}|X(t)-\theta_{0}|$ and $\displaystyle \sup_{t}|\nabla g(X(t))|$ are bounded, and Assumption A1 implies that $E|\mathrm{r}_{q}|^{2}$ and $E|\mathrm{r}_{q}|^{4}$ are uniformly bounded. Therefore we obtain 
$$E|R_{i}|^{2}=m^{-1}E\left|\sum_{q=1}^{m}\mathrm{r}_{q}\right|^{2}=m^{-1}\sum_{q=1}^{m}E|\mathrm{r}_{q}|^{2}\leq C,$$
\begin{eqnarray*}
E|R_{i}|^{4}&=&m^{-2}E\left|\sum_{q=1}^{m}\mathrm{r}_{q}\right|^{4} \\
&=&m^{-2}E\left[\left(\sum_{q=1}^{m}\mathrm{r}_{q}'\right)\left(\sum_{q=1}^{m}\mathrm{r}_{q}\right)\left(\sum_{q=1}^{m}\mathrm{r}_{q}'\right)\left(\sum_{q=1}^{m}\mathrm{r}_{q}\right)\right] \\
&=&m^{-2}E\left[\sum_{q=1}^{m}|\mathrm{r}_{q}|^{4}+\sum_{p<q}(4(\mathrm{r}_{p}'\mathrm{r}_{q})^{2}+2|\mathrm{r}_{p}|^{2}|\mathrm{r}_{q}|^{2})\right] \\
&\leq& m^{-2}\left[\sum_{q=1}^{m}E|\mathrm{r}_{q}|^{4}+\sum_{p<q}6E|\mathrm{r}_{p}|^{2}E|\mathrm{r}_{q}|^{2}\right]  \\
%\ \ \ \ \ ((\mathrm{r}_{p}'\mathrm{r}_{q})^{2}\leq|\mathrm{r}_{p}|^{2}|\mathrm{r}_{q}|^{2}) \\
&\leq& m^{-2}(mC+3m^{2}C^{2})\leq C+3C^{2}, 
\end{eqnarray*}
where we have used the inequality $(\mathrm{r}_{p}'\mathrm{r}_{q})^{2}\leq|\mathrm{r}_{p}|^{2}|\mathrm{r}_{q}|^{2}$. 
That is, we have shown that $E(|R_{i}|^{2})$ and $E(|R_{i}|^{4})$ are uniformly bounded. 

\begin{lem} \label{lm-2}
\[\max_{t\in[0,T]}E[|G_\delta^m(t)-\ccG_\delta^m(t)|^2] \leq C \max_{k\leq k_T}E[|y_k^m-X(t_k)|^2], \]
where $G_\delta^m(t)$ and $\ccG_\delta^m(t)$ are defined in (\ref{lm3-6}) and (\ref{lm4-1}) below, respectively. 
\end{lem}
Proof. Define filtration $\cF_t=\sigma(y_\delta^m(s), \bU^*_m(s); s\leq t)$,  where $y_\delta^m(t)$ and $\bU_m^*(t)$ are given by (\ref{Nest-xyt}). 
%Define $y_\delta^m(t)=y_k^m$, $\bU_\delta^m(t)=\bU_{mk}^*$ when $t_k\leq t<t_{k+1}$. $\cF_t=\sigma(y_\delta^m(s), \bU_\delta^m(s); s\leq t)$, 
Then for $i>j$,
\[\begin{split}
&E\{(R^m(y_i^m;\bU_{m(i+1)}^*)-R^m(X(t_i);\bU_{m(i+1)}^*))'(R^m(y_j^m;\bU_{m(j+1)}^*)-R^m(X(t_j);\bU_{m(j+1)}^*)\}\\
=&E\{E[(R^m(y_i^m;\bU_{m(i+1)}^*)-R^m(X(t_i);\bU_{m(i+1)}^*))'(R^m(y_j^m;\bU_{m(j+1)}^*)-R^m(X(t_j);\bU_{m(j+1)}^*))|\cF_{t_j}]\}\\
=&E\{(R^m(y_i^m;\bU_{m(i+1)}^*)-R^m(X(t_i);\bU_{m(i+1)}^*))'E[R^m(y_j^m;\bU_{m(j+1)}^*)-R^m(X(t_j);\bU_{m(j+1)}^*)|\cF_{t_j}]\}\\
=&0.
\end{split}\]
Set $r_{qi}=\nabla\ell(y_i^m; U^*_{q(i+1)})-\nabla g(y_i^m)-(\nabla\ell(X(t_i); U^*_{q(i+1)})-\nabla g(X(t_i))$. We have  
\[D_i\overset{\triangle}{=}R^m(y_i^m;\bU_{m(i+1)}^*)-R^m(X(t_i);\bU_{m(i+1)}^*)=m^{-1/2}\sum_{q=1}^m r_{qi}, \]
and for $q\neq s$,
\[E(r_{qi}'r_{si})=E(E(r_{qi}'r_{si}|\cF_{t_i}))=E(E(r_{qi}'|\cF_{t_i})E(r_{si}|\cF_{t_i}))=0. \]
On the other hand, we get 
$|r_{qi}|\leq (h_1(U_{q(i+1)})+L)|y_i^m-X(t_i)|,$
\[E(|r_{qi}|^2)\leq E(E[(h_1(U_{q(i+1)})+L)^2|y_i^m-X(t_i)|^2|\cF_{t_i}])\leq C\cdot E|y_i^m-X(t_i)|^2, \]
and thus 
\[E|D_i|^2\leq C\cdot E|y_i^m-X(t_i)|^2.  \]
Direct calculations show that for $t_{k+1}\leq t<t_{k+2}$, 
\[\begin{split}
& E|G_\delta^m(t)-\ccG_\delta^m(t)|^2=\frac{\delta^{1/2}}{c_k^2}E\left|\sum_{i=1}^k c_iD_i\right|^2 
=\frac{\delta^{1/2}}{c_k^2}\sum_{i=1}^k c_i^2 E|D_i|^2  \\
&\leq k\delta^{1/2}C\cdot \max_{1\leq i\leq k}E|y_i^m-X(t_i)|^2 
\leq C\cdot \max_{1\leq i\leq k}E|y_i^m-X(t_i)|^2, 
\end{split}\]
and therefore
\[\max_{t\in[0,T]}E|G_\delta^m(t)-\ccG_\delta^m(t)|^2\leq C\cdot \max_{k\leq k_T}E|y_k^m-X(t_k)|^2. \]

\begin{lem} \label{lm-2_1}
\[\max_{k\leq T \delta^{-1/2}}E[|y_k^m-y_k|^4]=O(m^{-2}), \;\; 
    \max_{k\leq T \delta^{-1/2}}E[|y_k^m-X(t_k)|^4]=O(m^{-2} + \delta^{1/2} |\log \delta|). \]
\end{lem}
Proof. The second result can be easily established from the first one and Lemma \ref{lm-1}. 
%the relationship between $y_k$ and $X(t_k)$ (which is derived in the proof of Lemma \ref{lm-1} by Lemma \ref{lem-add-1}). 
We will prove the first result. 

Recall $d_0=0$, $d_k=x_k-x_{k-1}$, $d_0^m=0$, $d_k^m=x_k^m-x_{k-1}^m$, $a_k=|x_k-x_k^m|$, $b_k=|d_k-d_k^m|$. We have
$a_0=0$, $b_0=0$,
\[a_k\leq |x_{k-1}-x_{k-1}^m|+|d_k-d_k^m|=a_{k-1}+b_k\leq S_k, \]
where $S_k=b_0+b_1+\cdots+b_k$. Also we obtain 
\[d_{k+1}=\frac{k-1}{k+2}d_k-\delta\nabla g(y_k),\]
\[d_{k+1}^m=\frac{k-1}{k+2}d_k^m-\delta\nabla g(y_k^m)-\delta m^{-1/2}R^m(y_k^m;\bU_{m(k+1)}^*), \]
\[|y_k^m-y_k|\leq a_k+b_k. \]
Since $\max_{k\leq k_T}|y_k-X(t_k)|=O(\delta^{1/2}|\log\delta|)$, we have $\max_{k\leq k_T}|y_k|=O(1)$, and 
as in Lemma \ref{lm-0},  $ \max_{k\leq T\delta^{-1/2}} E[|R^m(y_k;\bU_{m(k+1)}^*)|^4]=O(1)$. For simplicity we let 
$R_k=|R^m(y_k;\bU_{m(k+1)}^*)|$. 
%\begin{equation} \label{4th-R}
%\[   \max_{k\leq T\delta^{-1/2}}ER_k^4=O(1), \]
%\end{equation}
Recall 
\[\begin{split}
|R^m(y_k^m;\bU_{m(k+1)}^*)|\leq&R_k+m^{-1/2}\sum_{i=1}^m[h_1(U_{i(k+1)}^*)+L]|y_k^m-y_k|\\
\leq&R_k+m^{-1/2}\left|\sum_{i=1}^m[h_1(U_{i(k+1)}^*)-E(h_1(U))]\right|\cdot|y_k^m-y_k|\\
&+m^{1/2}(E(h_1(U))+L)\cdot|y_k^m-y_k| .
\end{split}\]
Let $h_k=m^{-1/2}\left|\sum_{i=1}^m[h_1(U_{i(k+1)}^*)-E(h_1(U))]\right|$. Then
\[\max_{k\leq T\delta^{-1/2}}E[h_k^4]=O(1),\]
\[b_{k+1}\leq b_k+L\delta(a_k+b_k)+\delta m^{-1/2}R_k+(\delta m^{-1/2}h_k+C\delta)(a_k+b_k),\]
and using $a_k+b_k\leq 2S_k$, we obtain 
\[b_{k+1}\leq b_k+C\delta(1+m^{-1/2}h_k)S_k+\delta m^{-1/2}R_k. \]
Define sequence $b_k'$ satisfying $b_0'=0$,
\[b_{k+1}'=b_k'+C\delta(1+m^{-1/2}h_k)S_k'+\delta m^{-1/2}R_k.\]
Then $b_k\leq b_k'$, $b_k'$ is non-decreasing, and since $k\delta^{1/2}\leq T$, 
\[b_{k+1}'\leq b_k'+C\delta(1+m^{-1/2}h_k)kb_k'+\delta m^{-1/2}R_k\leq (1+C\delta^{1/2})(1+C\delta^{1/2}m^{-1/2}h_k)b_k'+\delta m^{-1/2}R_k. \]
Define another sequence $b_k^*$ satisfying $b_0^*=0$,
\[b_{k+1}^*=(1+C\delta^{1/2})(1+C\delta^{1/2}m^{-1/2}h_k)b_k^*+\delta m^{-1/2}R_k. \]
Then $b_k'\leq b_k^*$, and
\[\begin{split}
b_k^*&=\sum_{i=0}^{k-1}\left\{(1+C\delta^{1/2})^{k-i-1}\left[\prod_{j=i+1}^{k-1}(1+C\delta^{1/2}m^{-1/2}h_j)\right]\delta m^{-1/2}R_i\right\}\\
&\leq C\delta m^{-1/2}\sum_{i=0}^{k-1}\left\{\left[\prod_{j=i+1}^{k-1}(1+C\delta^{1/2}m^{-1/2}h_j)\right]R_i\right\}. 
\end{split}\]
Since for $r=C\delta^{1/2}m^{-1/2}<1$, 
\[ E(1+rh_j)^4\leq 1+4rEh_j+6rEh_j^2+4rEh_j^3+rEh_j^4\leq 1+Cr, \]
and $R_i, h_{i+1},\dots, h_{k-1}$ are independent, we obtain 
\[\begin{split}
\max_{k\leq k_T}Ea_k^4&\leq E(k_Tb_{k_T}^*)^4\\
&\leq C\delta^2m^{-2}k_T^3\sum_{i=0}^{k_T-1}E\left\{\left[\prod_{j=i+1}^{k_T-1}(1+C\delta^{1/2}m^{-1/2}h_j)\right]R_i\right\}^4\\
&\leq C\delta^{1/2}m^{-2}\sum_{i=0}^{k_T-1}\left\{\left[\prod_{j=i+1}^{k_T-1}E(1+C\delta^{1/2}m^{-1/2}h_j)^4\right]ER_i^4\right\}\\
&\leq C\delta^{1/2}m^{-2}k_T(C(1+C\delta^{1/2}m^{-1/2})^{k_T})\\
&=O(m^{-2}). 
\end{split}\]
Finally, we conclude 
\[\max_{k\leq k_T}E[|y_k^m-y_k|^4] \leq \max_{k\leq k_T} E[ (a_k+b_k)^4] \leq  C \max_{k\leq k_T} E[ a_k^4+b_k^4] =
O(m^{-2}). \]

 \begin{lem} \label{lm-0_1}
\[  \max_{k \leq T \delta^{-1/2}} E[ |R^{m}(y^m_{k-1}; \bU_{m k}^{*})|^4] \leq C. \]
 \end{lem} 
 Proof. % $y_\delta^m(t)=y_k^m$, $\bU_\delta^m(t)=\bU_{mk}^*$ when $t_k\leq t<t_{k+1}$. 
 Define filtration $\cF_t=\sigma(y_\delta^m(s), \bU^*_m(s); s\leq t)$,  where $y_\delta^m(t)$ and $\bU_m^*(t)$ are given by (\ref{Nest-xyt}). 
 For simplicity we write $R_k^m=R^m(y_k^m;\bU_{m(k+1)}^*)$, and $r_{qk}=\nabla\ell(y_k^m, U^*_{q(k+1)})-\nabla g(y_k^m)$. Then 
 given $\cF_{t_k}$, $r_{1k},\dots,r_{qk}$ are conditionally independent with conditional mean $0$, 
\[R_k^m=m^{-1/2}\sum_{q=1}^m r_{qk},\]
\[E|r_{qk}|^4\leq C \left (E[|h_1(U^*_{q(k+1)})|^4|y_k^m-\theta_0|^4]+E|y_k^m-\theta_0|^4+E|\nabla\ell(\theta_0, U^*_{q(k+1)})|^4+|\nabla g(\theta_0)|^4 \right), \]
which is bounded uniformly over $1 \leq k \leq k_T$, since $E[|y_k^m-\theta_0|^4] \leq E[|y_k^m - y_k|^4] + |y_k-\theta_0|^4] \leq C$
(implied by Lemmas \ref{lm-1} and  \ref{lm-2_1}),  and 
\[E[|h_1(U^*_{q(k+1)})|^4|y_k^m-\theta_0|^4]=E[E[|h_1(U^*_{q(k+1)})|^4|y_k^m-\theta_0|^4|\cF_{t_k}]]=E(|h_1(U)|^4)E[|y_k^m-\theta_0|^4] \leq C. \]
%we find $E|r_{qk}|^4$ is uniformly bounded. 
Finally, we conclude 
\[\begin{split}
E|R_k^m|^4&=m^{-2}E\left[E\left[\left.\left(\sum_{q=1}^m r_{qk}'\right)\left(\sum_{q=1}^m r_{qk}\right)\left(\sum_{q=1}^m r_{qk}'\right)\left(\sum_{q=1}^m r_{qk}\right)\right|\cF_{t_k}\right]\right] \\
&\leq m^{-2}\left(\sum_{q=1}^m E|r_{qk}|^4+6\sum_{p<q} (E|r_{pk}|^4\cdot E|r_{qk}|^4)^{\frac{1}{2}}\right)\\
&\leq C. 
\end{split}\]
%In the above calculation, we use the fact that, $r_{1k},\dots,r_{qk}$ are conditionally independent with mean $0$ conditional on $\cF_{t_k}$.

\begin{lem} \label{lm3}
%Let $\{\check{x}_{k}^{m},\check{y}_{k}^{m}\}$ be the sequence, with initial values $\check{x}_{0}^{m}=\check{y}_{0}^{m}=x_{0}$, generated by
%\begin{equation} \label{lm3-2}
%\check{x}_{k}^{m}=\check{y}_{k-1}^{m}-\delta\nabla g(\check{y}_{k-1}^{m})-m^{-1/2}\delta^{3/4}(H(t_{k})-H(t_{k-1})),  \check{y}_{k}^{m}=\check{x}_{k}^{m}+\frac{k-1}{k+2}(\check{x}_{k}^{m}-\check{x}_{k-1}^{m}),
%\end{equation}
%where $H(t)=\displaystyle\int_{0}^{t}\bsigma(X(u))d\bB(u)$ and $t_{k}=k\delta^{1/2}$.
%\begin{spacing}{1.5}
%Then for all ($\delta, m$) 
There exist simultaneous realizations  $\{\tilde{x}_k^m, \tilde{y}_k^m\}$, $\{\tilde{\cx}_k^m, \tilde{\cy}_k^m\}$, 
$\tilde{H}(t)=\displaystyle \int_{0}^{t}\bsigma(X(u))d\tilde{\bB}(u)$,  and standard Brownian motion $\tilde{B}$ on some common probability spaces, such that sequence $\displaystyle \{\tilde{x}_k^m, \tilde{y}_k^m, k\leq T/\delta^{1/2}\}$ have the same distribution as $\{x_k^m, y_k^m, k\leq T/\delta^{1/2}\}$,
sequence $\{\tilde{\cx}_k^m, \tilde{\cy}_k^m, k\leq T/\delta^{1/2}\}$ are generated from $\tilde{H}(t)$ the same way as $\{\cx_k^m, \cy_k^m, k\leq T/\delta^{1/2}\}$ generated from $H(t)$ according to (\ref{lm3-2}), 
%which (surely) have the same distribution as $\{\cx_k^m, \cy_k^m, k\leq T/\delta^{1/2}\}$ for all $\delta, m$, 
and as $\delta\rightarrow 0, m\rightarrow\infty$,
%\end{spacing}
\begin{equation} \label{lm3-3}
\max_{k\leq T/\delta^{1/2}}|\tilde{x}_{k}^{m}-\tilde{\cx}_{k}^{m}| =o_p(m^{-1/2}\delta^{1/4})
\end{equation}

\end{lem}

Proof. For $ k\geq 1$, let $\check{d}_{k}^m=\check{x}_{k+1}^{m}-\check{x}_{k}^{m},$ 
$$\check{d}_{0}^m=-\delta\nabla g(x_{0})-m^{-1/2}\delta^{3/4}(H(t_{1})-H(t_0)),$$
 and rewrite (\ref{lm3-2}) as 
$$
\check{x}_{k+1}^{m}=\check{x}_{k}^{m}+\frac{k-1}{k+2}(\check{x}_{k}^{m}-\check{x}_{k-1}^{m})-\delta\nabla g(\check{y}_{k}^{m})-m^{-1/2}\delta^{3/4}(H(t_{k+1})-H(t_k)). 
$$
Then 

\begin{eqnarray} \label{lm3-4} \nonumber 
\check{d}_{k}^m& =&  \frac{k-1}{k+2}\check{d}_{k-1}^m-\delta\nabla g(\check{y}_{k}^{m})-m^{-1/2}\delta^{3/4}(H(t_{k+1})-H(t_k)) \\ \nonumber
&=&\frac{k-1}{k+2}\left(\frac{k-2}{k+1}\check{d}_{k-2}^m-\delta\nabla g(\check{y}_{k-1}^{m})-m^{-1/2}\delta^{3/4}(H(t_{k})-H(t_{k-1}))\right) \\ \nonumber
&&-\delta\nabla g(\check{y}_{k}^{m})-m^{-1/2}\delta^{3/4}(H(t_{k+1})-H(t_{k})) \\ \nonumber
&=&-\sum_{i=1}^{k}\left(\frac{k-1}{k+2}\cdot\frac{k-2}{k+1}\cdot\cdots\cdot\frac{i}{i+3}\right)(\delta\nabla g(\check{y}_{i}^{m})+m^{-1/2}\delta^{3/4}(H(t_{i+1})-H(t_{i}))) \\ 
&=&-\sum_{i=1}^{k}\frac{(i+2)(i+1)i}{(k+2)(k+1)k}(\delta\nabla g(\check{y}_{i}^{m})+m^{-1/2}\delta^{3/4}(H(t_{i+1})-H(t_{i}))) . 
\end{eqnarray}

Similarly, let $d_{k}^m=x_{k+1}^{m}-x_{k}^{m},$
$$ 
d_{0}^m=-\delta\nabla\hat{\mathcal{L}}^{m}(x_{0};\bU_{m1}^{*})=-\delta\nabla g(x_{0})-\delta(\nabla\hat{\mathcal{L}}^{m}(x_{0};\bU_{m1}^{*})-\nabla g(x_{0})),
$$
and we have 
\begin{eqnarray} \label{lm3-5}
d_{k}^m&=&-\sum_{i=1}^{k}\frac{(i+2)(i+1)i}{(k+2)(k+1)k}\delta\nabla\hat{\mathcal{L}}^{m}(y_{i}^{m};\bU_{m(i+1)}^{*}) \\ \nonumber
&=& -\sum_{i=1}^{k}\frac{(i+2)(i+1)i}{(k+2)(k+1)k}\left(\delta\nabla g(y_{i}^{m})+\delta\left(\nabla\hat{\mathcal{L}}^{m}(y_{i}^{m};\bU_{m(i+1)}^{*})-\nabla g(y_{i}^{m})\right)\right) . 
\end{eqnarray}

Set $c_{i}=(i+2)(i+1)i$, 
$$
R^{m}(\theta;\bU_{mk}^{*})=m^{1/2}\left(\nabla\hat{\mathcal{L}}^{m}(\theta;\bU_{mk}^{*})-\nabla g(\theta)\right) , 
$$
and define c\`{a}dl\`{a}g processes $G_{\delta}^{m}(t)$ and $G_{\delta}(t)$ as follows,
\begin{equation} \label{lm3-6}
G_{\delta}^{m}(t)= \left\{
\begin{aligned}
0, & & & & 0\leq t<t_{1}\\
\delta^{1/4}R^{m}(x_{0};\bU_{m1}^{*}) , & & & & t_{1}\leq t<t_{2}\\
\delta^{1/4}\frac{1}{c_{k}}\sum_{i=1}^{k}c_{i}R^{m}(y_{i}^{m};\bU_{m(i+1)}^{*}) , & & & & t_{k+1}\leq t<t_{k+2}
\end{aligned}
\right.
\end{equation}
\begin{equation} \label{lm3-7}
G_{\delta}(t)=\left\{
\begin{aligned}
0 , & & & & 0\leq t<t_{1}\\
H(t_{1})-H(t_{0}), & & & & t_{1}\leq t<t_{2}\\
\frac{1}{c_{k}}\sum_{i=1}^{k}c_{i}(H(t_{i+1})-H(t_{i})), & & & & t_{k+1}\leq t<t_{k+2}  
\end{aligned}
\right.
\end{equation}
\begin{spacing}{1.5}
By Assumption A4, $R^{m}(\theta;\bU_{mk}^{*})$ weakly converges to $N(0,\bsigma^{2}(\theta))$ uniformly over $\theta$ as $m \rightarrow\infty$.
Note that $H(t_{i+1})-H(t_{i})$ follows $N(0, \displaystyle\int_{t_{i}}^{t_{i+1}}\bsigma^{2}(X(u))du)$, and Var$(\delta^{1/4}R^{m}(y_{i}^{m};\bU_{m(i+1)}^{*}))$ is approximately equal to $ %\approx\displaystyle 
\int_{t_{i}}^{t_{i+1}}\bsigma^{2}(X(u))du$.  
According to Lemma \ref{lm4} below, % for fix $T > 0$, 
there exist $\tilde{G}_{\delta}^m(t)$ and $\tilde{H}(t)=\displaystyle \int_{0}^{t}\bsigma(X(u))d\tilde{\bB}(u)$ on some common probability spaces 
such that $\tilde{G}_{\delta}^m(t)$ and $G_{\delta}^m(t)$ are identically distributed, $\tilde{G}_{\delta}(t)$ is generated by $\tilde{H}(t)$ the same way as $G_\delta(t)$ by $H(t)$ via scheme (\ref{lm3-7}), and as $\delta\rightarrow 0, m\rightarrow\infty$,
\end{spacing}
$$\max_{t\leq T}|\tilde{G}_{\delta}^{m}(t)-\tilde{G}_{\delta}(t)|=o_{p}(1).$$
%For any $\delta, m$, 
Using $\tilde{G}_{\delta}^m(t)$ we define associated sequences $\{\tilde{R}_k\}$ and $\{\tilde{x}_k^m, \tilde{y}_k^m\}$ as follows, 
$$\tilde{R}_0=\delta^{-1/4}\tilde{G}_{\delta}^m(t_1), \tilde{x}_0^m=\tilde{y}_0^m=x_0,$$
$$\tilde{x}_k^m=\tilde{y}_{k-1}^m-\delta\nabla g(\tilde{y}_{k-1}^m)-\delta m^{-1/2}\tilde{R}_{k-1},\ 
\tilde{y}_{k}^{m}=\tilde{x}_{k}^{m}+\frac{k-1}{k+2}(\tilde{x}_{k}^{m}-\tilde{x}_{k-1}^{m}),$$
$$\tilde{R}_k=\delta^{-1/4}\tilde{G}_\delta^m(t_{k+1})-\frac{1}{c_k}\sum_{i=1}^{k-1}c_i\tilde{R}_i.$$
Since $\tilde{G}_\delta^m(t_1),...,\tilde{G}_\delta^m(t_k)$ have the same distribution as $G_\delta^m(t_1),...G_\delta^m(t_k)$, we easily conclude 
that $\{\tilde{x}_k^m, \tilde{y}_k^m\}$ are identically distributed as $\{x_k^m,y_k^m\}$, and $\tid_k^m=\tix_{k+1}^m-\tix_k^m$ satisfies
$$\tid_{0}^m=-\delta\nabla g(x_{0})-m^{-1/2}\delta^{3/4}\tiG_\delta^m(t_1),\ \tid_{k}^m=-\sum_{i=1}^k\frac{c_{i}}{c_{k}}\delta\nabla g(\tiy_{i}^{m})-m^{-1/2}\delta^{3/4}\tiG_\delta^m(t_{k+1}).$$
Similarly, we define $\{\tcx_k^m, \tcy_k^m\}$ by $\tiG_\delta(t)$, and set $\tcd_k^m=\tcx_{k+1}^m-\tcx_k^m$, so that 

$$\tcd_{0}^m=-\delta\nabla g(x_{0})-m^{-1/2}\delta^{3/4}\tilde{G}_{\delta}(t_{1}),\tcd_{k}^m=-\sum_{i=1}^{k}\frac{c_{i}}{c_{k}}\delta\nabla g(\tcy_{i}^{m})-m^{-1/2}\delta^{3/4}\tilde{G}_{\delta}(t_{k+1}). $$

Let $a_{k}= |\tix_{k}^{m}-\tcx_{k}^{m}|, b_{k}= |\tid_{k}^m-\tcd_{k}^m|$, $S_k = b_0 + \cdots + b_k$, 
and $\cY=m^{-1/2}\displaystyle \delta^{3/4}\max_{t\leq T}|\tiG_{\delta}^{m}(t)-\tilde{G}_{\delta}(t)|$. Then we have $b_{0}\leq \cY$, 
$$a_{k}=|\tix_{k-1}^{m}+\tid_{k-1}^m-\tcx_{k-1}^{m}-\tcd_{k-1}^m|\leq a_{k-1}+b_{k-1}\leq b_{0}+b_{1}+\cdots+b_{k-1}=S_{k-1},$$
$$|\tiy_{k}^{m}-\tcy_{k}^{m}|=\left|\tix_{k}^{m}+\frac{k-1}{k+2}\tid_{k-1}-\tcx_{k}^{m}-\frac{k-1}{k+2}\tcd_{k-1}\right|\leq a_{k}+b_{k-1}\leq S_{k-1}+b_{k-1},$$
$$b_{k} \leq  L\delta\sum_{i=1}^k |\tiy_{i}^{m}-\tcy_{i}^{m}|+\cY\leq L\displaystyle \delta\sum_{i=1}^k (S_{i-1}+b_{i-1}) + \cY \leq 2L\delta kS_{k-1} + \cY \leq C\delta^{1/2}S_{k-1}+\cY.$$
Let $b_{0}^{*}=\cY, b_{k}^{*}=C\delta^{1/2}S_{k-1}^{*}+\cY$, where $S_{k}^{*}=b_{0}^{*}+\ldots+b_{k}^{*}$. Then by induction we easily conclude 
$$b_{0}\leq b_{0}^{*},b_{k}\leq C\delta^{1/2}S_{k-1}+\cY\leq C\delta^{1/2}S_{k-1}^{*}+\cY=b_{k}^{*}, S_k \leq S^*_k.$$
Since $b_{k+1}^{*}=C\delta^{1/2}S_{k}^{*}+\cY$ leads to $b_{k+1}^{*}-b_{k}^{*}=C\delta^{1/2}b_{k}^{*}$ for all $k \geq 0$, %which also holds for $k=0$, 
we immediately obtain geometric sequence $b_{k}^{*}=(1+C\delta^{1/2})^{k}\cY$ and find its sum $S^*_k$. Finally, 
$$\max_{k\leq T/\delta^{1/2}}|\tix_{k}^{m}-\tcx_{k}^{m}| \leq S_{\lfloor T/\delta^{1/2}\rfloor-1} \leq S^*_{\lfloor T/\delta^{1/2}\rfloor-1} 
\leq T/\delta^{1/2}(1+C\delta^{1/2})^{T/\delta^{1/2}}\cY$$
$$\leq C \cY/\delta^{1/2}=o_{p}(m^{-1/2}\delta^{1/4}).$$

\begin{lem} \label{lm4}
Given that $G_{\delta}^{m}(t)$ and $G_{\delta}(t)$ are defined by (\ref{lm3-6}) and (\ref{lm3-7}), respectively, we can show that there exist $\tilde{G}_{\delta}^m(t)$ and $\tilde{H}(t)=\displaystyle \int_{0}^{t}\bsigma(X(u))d\tilde{\bB}(u)$ on some common probability spaces, such that 
$\tilde{G}_{\delta}^m(t)$ and $G_{\delta}^m(t)$ are identically distributed, $\tilde{G}_{\delta}(t)$ are generated by $\tilde{H}(t)$ the same way as 
$G_\delta(t)$ by $H(t)$ via scheme (\ref{lm3-7}),  and as $\delta\rightarrow 0, m\rightarrow\infty$,
$$\max_{t\leq T}|\tilde{G}_{\delta}^{m}(t)-\tilde{G}_{\delta}(t)|=o_{p}(1).$$
\end{lem}
Proof. Define c\`{a}dl\`{a}g processes $\check{G}_{\delta}^{m}(t)$ 
\begin{equation} \label{lm4-1}
\ccG_{\delta}^{m}(t)= \left\{
\begin{aligned}
0, & & & & 0\leq t<t_{1}\\
\delta^{1/4}R^{m}(x_{0};\bU_{m1}^{*}) ,& & & & t_{1}\leq t<t_{2}\\
\delta^{1/4}\frac{1}{c_{k}}\sum_{i=1}^{k}c_{i}R^{m}(X(t_i);\bU_{m(i+1)}^{*}) ,& & & & t_{k+1}\leq t<t_{k+2}
\end{aligned}
\right.
\end{equation}
%\begin{spacing}{1.5}
Note that the only change in (\ref{lm4-1}) is to replace $y_{i}^{m}$ in (\ref{lm3-6}) by $X(t_{i})$. 
Define $G(t) = \displaystyle \frac{1}{t^{3}}\int_{0}^{t}u^{3}\bsigma(X(u))d\mathrm{B}(u)$. We will prove that $\check{G}_{\delta}^{m}(t)$ weakly converges to $G(t)$.
Set $\cC_{i}^{\delta} = \delta^{3/2}c_{i} = t_{i}t_{i+1}t_{i+2}$, and for any fixed $0 = \tau_{0} < \tau_{1} < \tau_{2} <\cdots < \tau_{l} \leq T,$ let $ k_{j}^{\delta}=\displaystyle \max (0,\lfloor\tau_{j}/\delta^{1/2}\rfloor-1)$. Then $\cC_{k_{j}^{\delta}}^{\delta} =\delta^{3/2}k_{j}^{\delta}(k_{j}^{\delta}+1)(k_{j}^{\delta}+2)\rightarrow\tau_{j}^{3}$, as $\delta\rightarrow 0.$
%\end{spacing}
Using the definition of $\check{G}_{\delta}^{m}(t)$ we have  
$$\check{G}_{\delta}^{m}(\tau_{j})=\delta^{1/4}\frac{1}{\cC_{k_{j}^{\delta}}^{\delta}}\sum_{i=1}^{k_{j}^{\delta}}\cC_{i}^{\delta}R^{m}(X(t_{i});\bU_{m(i+1)}^{*}),$$
$$\cC_{k_{j+1}^{\delta}}^{\delta} \check{G}_{\delta}^{m}(\tau_{j+1})-\cC_{k_{j}^{\delta}}^{\delta}\check{G}_{\delta}^{m}(\tau_{j})=\delta^{1/4}\sum_{i=k_{j}^{\delta}+1}^{k_{j+1}^{\delta}}\cC_{i}^{\delta}R^{m}(X(t_i); \bU_{m(i+1)}^{*}).$$
The right hand side of above equation is the sum of independent random variables, by Assumption A4, as $m\rightarrow \infty, \delta\rightarrow 0$,
$\cC_{k_{j+1}^{\delta}}^{\delta} \check{G}_{\delta}^{m}(\tau_{j+1})-\cC_{k_{j}^{\delta}}^{\delta}\check{G}_{\delta}^{m}(\tau_{j})$ converges in distribution to a normal distribution with mean $0$ and variance $\displaystyle \int_{\tau_{j}}^{\tau_{j+1}}u^{6}\sigma^{2}(X(u))du$. Because of independence between consecutive differences, we can easily arrive at that
$$\left(\cC_{k_{1}^{\delta}}^{\delta}\check{G}_{\delta}^{m}(\tau_{1}),\cC_{k_{2}^{\delta}}^{\delta}\check{G}_{\delta}^{m}(\tau_{2})-\cC_{k_{1}^{\delta}}^{\delta}\check{G}_{\delta}^{m}(\tau_{1}),\dots,\cC_{k_{l}^{\delta}}^{\delta}\check{G}_{\delta}^{m}(\tau_{l})-\cC_{k_{l-1}^{\delta}}^{\delta}\check{G}_{\delta}^{m}(\tau_{l-1})\right)$$
converges in distribution to
$$\left(\tau_{1}^{3}G(\tau_{1}),\tau_{2}^{3}G(\tau_{2})-\tau_{1}^{3}G(\tau_{1}),\dots,\tau_{l}^{3}G(\tau_{l})-\tau_{l-1}^{3}(\tau_{l-1})\right),$$
which immediately shows that 
$\left(\cC_{k_{1}^{\delta}}^{\delta}\check{G}_{\delta}^{m}(\tau_{1}),\cC_{k_{2}^{\delta}}^{\delta}\check{G}_{\delta}^{m}(\tau_{2}),\ldots,\cC_{k_{l}^{\delta}}^{\delta}\check{G}_{\delta}^{m}(\tau_{l})\right)$
converges in distribution to
$\left(\tau_{1}^{3}G(\tau_{1}),\tau_{2}^{3}G(\tau_{2}),\ldots,\tau_{l}^{3}G(\tau_{l})\right).$
Since $\cC_{k_{j}^{\delta}}^{\delta} \rightarrow \tau_{j}^{3}$ as $\delta \rightarrow 0$, we conclude that 
$\left(\check{G}_{\delta}^{m}(\tau_{1}),\check{G}_{\delta}^{m}(\tau_{2}),\ldots,\check{G}_{\delta}^{m}(\tau_{l})\right)$
converges in distribution to
$\left(G(\tau_{1}),G(\tau_{2}),\ldots,G(\tau_{l})\right).$ That is, we prove the finite-dimensional distribution convergence of 
$\check{G}_{\delta}^{m}(t)$. 

We will establish the tightness of $\check{G}_{\delta}^{m}(t)$ by proving that for any $0\leq r\leq s\leq t\leq T,$
\begin{equation} \label{lm4-2}
E\{|\check{G}_{\delta}^{m}(t)-\check{G}_{\delta}^{m}(s)|^{2}|\check{G}_{\delta}^{m}(s)-\check{G}_{\delta}^{m}(r)|^{2}\}\leq C(t-r)^{2}.
\end{equation}
To simplify some notation we let $R_{i}=R^{m}(X(t_i); \bU_{m(i+1)}^{*})$. First we will show that for any fixed $1 \leq j<k<l,$
\begin{equation} \label{lm4-3}
E \left\{\left|\frac{1}{c_{k}}\sum_{i=1}^{k}c_{i}R_{i}-\frac{1}{c_{j}}\sum_{i=1}^{j}c_{i}R_{i}\right|^{2}\left|\frac{1}{c_{l}}\sum_{i=1}^{l}c_{i}R_{i}-\frac{1}{c_{k}}\sum_{i=1}^{k}c_{i}R_{i}\right|^{2}\right\}\leq C(l-j)^{2}  ,
\end{equation}
where $C$ is a generic constant free of the choice of $(j, k, l)$. 
%Before dealing with this expectation, we need some preparation work. 

Lemma \ref{lm-0} implies that % \[\max_{k\leq T\delta^{-1/2}}E|R^m(X(t_{k-1});\bU_{mk}^*)|^4=O(1),\]
$E(|R_{i}|^{2})$ and $E(|R_{i}|^{4})$ are uniformly bounded  over $1 \leq i \leq T\delta^{-1/2}$. 
Since $R_0, R_1, R_2,\dots$ are independent with mean $0$, we have 
%direct calculations show 
$$E\left|\sum_{i=1}^{k}c_{i}R_{i}\right|^{2}=\sum_{i=1}^{k}c_{i}^{2}E|R_{i}|^{2}\leq Ckc_{k}^{2},$$
$$E\left|\sum_{i=1}^{k}c_{i}R_{i}\right|^{4}\leq\sum_{i=1}^{k}c_{i}^{4}E|R_{i}|^{4}+\sum_{i<j}6c_{i}^{2}c_{j}^{2}E|R_{i}|^{2}E|R_{j}|^{2}\leq Ck^{2}c_{k}^{4}, $$
where we remind the convention that $C$ denotes any generic constant free of $(\delta, m,n)$ and $(i,j,k,l)$, and its value may change from appearance to appearance. 

Let 
$$D_{1} = \sum_{i=1}^{j}c_{i}R_{i}, D_{2} = \sum_{i=j+1}^{k}c_{i}R_{i}, D_3= \sum_{i=k+1}^{l}c_{i}R_{i}. $$
Then $D_{1}, D_{2}, D_3$ are independent, and similarly we can show 
$$E|D_{2}|^{2}\leq C(k-j)c_{k}^{2}, E|D_{2}|^{4}\leq C(k-j)^{2}c_{k}^{4}, $$
$$E|D_{3}|^{2}\leq C(l-k)c_{l}^{2}, E|D_{3}|^{4}\leq C(l-k)^{2}c_{l}^{4},$$
$$\frac{(c_{k}-c_{j})^{2}}{c_{k}^{2}}\leq\frac{c_{k}-c_{j}}{c_{k}}=1-\frac{j(j+1)(j+2)}{k(k+1)(k+2)}\leq1-\frac{j^{3}}{k^{3}}\leq 3\left(\frac{k-j}{k}\right). $$
Therefore we establish (\ref{lm4-3}) as follows, 
\begin{eqnarray*}
&&E\left\{\left|\frac{1}{c_{k}}\sum_{i=1}^{k}c_{i}R_{i}-\frac{1}{c_{j}}\sum_{i=1}^{j}c_{i}R_{i}\right|^{2}\left|\frac{1}{c_{l}}\sum_{i=1}^{l}c_{i}R_{i}-\frac{1}{c_{k}}\sum_{i=1}^{k}c_{i}R_{i}\right|^{2}\right\} \\
&=&E\left\{\left|\frac{D_{1}+D_{2}}{c_{k}}-\frac{D_{1}}{c_{j}}\right|^{2}\left|\frac{D_{1}+D_{2}+D_{3}}{c_{l}}-\frac{D_{1}+D_{2}}{c_{k}}\right|^{2}\right\} \\
&=&E\left\{\left|\frac{D_{2}}{c_{k}}-\frac{c_{k}-c_{j}}{c_{k}c_{j}}D_{1}\right|^{2}\left|\frac{D_{3}}{c_{l}}-\frac{c_{l}-c_{k}}{c_{l}c_{k}}(D_{1}+D_{2})\right|^{2}\right\} \\
&\leq& C\cdot E\left\{\left(\frac{|D_{2}|^{2}}{c_{k}^{2}}+\frac{(c_{k}-c_{j})^{2}}{c_{k}^{2}c_{j}^{2}}|D_{1}|^{2}\right)\left(\frac{|D_{3}|^{2}}{c_{l}^{2}}+\frac{(c_{l}-c_{k})^{2}}{c_{l}^{2}c_{k}^{2}}(|D_{1}|^{2}+|D_{2}|^{2})\right)\right\} \\
&\leq& 9C\cdot E\left\{\left(\frac{|D_{2}|^{2}}{c_{k}^{2}}+\frac{k-j}{kc_{j}^{2}}|D_{1}|^{2}\right)\left(\frac{|D_{3}|^{2}}{c_{l}^{2}}+\frac{l-k}{lc_{k}^{2}}(|D_{1}|^{2}+|D_{2}|^{2})\right)\right\} \\
&\leq& C\left(\frac{(k-j)c_{k}^{2}(l-k)c_{l}^{2}}{c_{k}^{2}c_{l}^{2}}+\frac{(k-j)c_{k}^{2}(l-k)jc_{j}^{2}}{c_{k}^{2}lc_{k}^{2}}+\frac{(l-k)(k-j)^{2}c_{k}^{4}}{lc_{k}^{4}}\right) \\
&&+C\left(\frac{(k-j)jc_{j}^{2}(l-k)c_{l}^{2}}{kc_{j}^{2}c_{l}^{2}}+\frac{(k-j)(l-k)j^{2}c_{j}^{4}}{kc_{j}^{2}lc_{k}^{2}}+\frac{(k-j)jc_{j}^{2}(l-k)(k-j)c_{k}^{2}}{kc_{j}^{2}lc_{k}^{2}}\right) \\
&\leq& C(l-j)^{2}.
\end{eqnarray*}
Second we will prove 
\begin{equation} \label{lm4-4}
E \left\{\left|\frac{1}{c_{k}}\sum_{i=1}^{k}c_{i}R_{i}-\frac{1}{c_{j}}\sum_{i=1}^{j}c_{i}R_{i}\right|^{2}\left|\frac{1}{c_{j}}\sum_{i=1}^{j}c_{i}R_{i}\right|^{2}\right\} \leq Ck^{2} .
\end{equation}
Indeed, similar direct calculations lead to 

\begin{eqnarray*}
&&E\left\{\left|\frac{1}{c_{k}}\sum_{i=1}^{k}c_{i}R_{i}-\frac{1}{c_{j}}\sum_{i=1}^{j}c_{i}R_{i}\right|^{2}\left|\frac{1}{c_{j}}\sum_{i=1}^{j}c_{i}R_{i}\right|^{2}\right\} \\
&=&E\left\{\left|\frac{D_{1}+D_{2}}{c_{k}}-\frac{D_{1}}{c_{j}}\right|^{2}\left|\frac{D_{1}}{c_{j}}\right|^{2}\right\} \\
&\leq& C\cdot E\left\{\left(\frac{|D_{2}|^{2}}{c_{k}^{2}}+\frac{(c_{k}-c_{j})^{2}}{c_{k}^{2}c_{j}^{2}}|D_{1}|^{2}\right)\frac{|D_{1}|^{2}}{c_{j}^{2}}\right\} \\
&\leq& C\left(\frac{(k-j)c_{k}^{2}jc_{j}^{2}}{c_{k}^{2}c_{j}^{2}}+\frac{j^{2}c_{j}^{4}}{c_{j}^{4}}\right) \\
&\leq& Ck^{2}.
\end{eqnarray*}
Third, for any $0\leq r\leq s\leq t\leq T$, we may choose $j, k, l$ such that $t_{j+1} \leq r<t_{j+2}, t_{k+1} \leq s<t_{k+2}, t_{l+1} \leq t<t_{l+2}$.
If $j=k$ or $k=l$, then $r=s$ or $s=t$, and (\ref{lm4-2}) is obvious. Assume $j<k<l$, and we prove (\ref{lm4-2}) for each scenario.  
If $j=-1$ and $k=0$, then
\begin{eqnarray*}
&&E\left\{|\check{G}_{\delta}^{m}(t)-\check{G}_{\delta}^{m}(s)|^{2}|\check{G}_{\delta}^{m}(s)-\check{G}_{\delta}^{m}(r)|^{2}\right\} \\
&=&\delta E\left\{\left|\frac{1}{c_{l}}\sum_{i=1}^{l}c_{i}R_{i}-R_{0}\right|^{2}|R_{0}|^{2}\right\} \\
&\leq& C\delta l^{2}=C(t_{l+1}-t_{1})^{2}\leq C(t-r)^{2}.
\end{eqnarray*}
If $j=-1$ and $ k\geq 1$, then
\begin{eqnarray*}
&&E\left\{|\check{G}_{\delta}^{m}(t)-\check{G}_{\delta}^{m}(s)|^{2}|\check{G}_{\delta}^{m}(s)-\check{G}_{\delta}^{m}(r)|^{2}\right\} \\
&=&\delta E\left\{\left|\frac{1}{c_{l}}\sum_{i=1}^{l}c_{i}R_{i}-\frac{1}{c_{k}}\sum_{i=1}^{k}c_{i}R_{i}\right|^{2}\left|\frac{1}{c_{k}}\sum_{i=1}^{k}c_{i}R_{i}\right|^{2}\right\} \\
&\leq& C\delta l^{2}=C(t_{l+1}-t_{1})^{2}\leq C(t-r)^{2}.
\end{eqnarray*}
If $j=0$, then
\begin{eqnarray*}
&&E\left\{|\check{G}_{\delta}^{m}(t)-\check{G}_{\delta}^{m}(s)|^{2}|\check{G}_{\delta}^{m}(s)-\check{G}_{\delta}^{m}(r)|^{2}\right\} \\
&=&\delta E\left\{\left|\frac{1}{c_{l}}\sum_{i=1}^{l}c_{i}R_{i}-\frac{1}{c_{k}}\sum_{i=1}^{k}c_{i}R_{i}\right|^{2}\left|\frac{1}{c_{k}}\sum_{i=1}^{k}c_{i}R_{i}-R_0\right|^{2}\right\} \\
&\leq& C\delta l^{2}\leq 4C(t_{l+1}-t_{2})^{2}\leq 4C(t-r)^{2}.
\end{eqnarray*}
If $ j\geq 1$, then
\begin{eqnarray*}
&&E\left\{|\check{G}_{\delta}^{m}(t)-\check{G}_{\delta}^{m}(s)|^{2}|\check{G}_{\delta}^{m}(s)-\check{G}_{\delta}^{m}(r)|^{2}\right\} \\
&=&\delta E\left\{\left|\frac{1}{c_{l}}\sum_{i=1}^{l}c_{i}R_{i}-\frac{1}{c_{k}}\sum_{i=1}^{k}c_{i}R_{i}\right|^{2}\left|\frac{1}{c_{k}}\sum_{i=1}^{k}c_{i}R_{i}-\frac{1}{c_{j}}\sum_{i=1}^{j}c_{i}R_{i}\right|^{2}\right\} \\
&\leq& C\delta(l-j)^{2}\leq 4C(t_{l+1}-t_{j+2})^{2}\leq 4C(t-r)^{2}.
\end{eqnarray*}
%\begin{spacing}{1.5}
Now with the established finite-dimensional distribution convergence and tightness for $\ccG_{\delta}^{m}(t)$, we conclude that $\check{G}_{\delta}^{m}(t)$ weakly converges to $G(t)$. 

Note that the only difference between $\check{G}^m_\delta (t)$ and $G^m_\delta(t)$ is $y_i^m$ and $X(t_i)$ used in $R^m(\cdot; \bU^*_{mi})$. 
By Lemmas \ref{lm-2} and \ref{lm-2_1} we immediately show that the finite-dimensional distribution convergence of $G_{\delta}^{m}(t)$ to $G(t)$.

%Similar to Lemma \ref{lem-2} we can show that as $\delta\rightarrow 0$, $m\rightarrow \infty$, $y_k^m - X(t_k)$ converges to zero in probability uniformly over $1 \leq k \leq T/\delta^{1/2}$. Condition A1 implies that $\nabla \ell (\theta; u, Q)$ is Lipschitz in $\theta$, and thus 
%the difference between $\check{G}^m_\delta (t)$ and $G^m_\delta(t)$ is negligible. 
The same argument for deriving tightness of $\check{G}^m_\delta (t)$ can be used to establish tightness of $G^m_\delta(t)$ by proving that for any $0\leq r\leq s\leq t\leq T,$
\begin{equation} \label{lm4-2R}
 E\{| G_{\delta}^{m}(t)- G_{\delta}^{m}(s)|^{2}| G_{\delta}^{m}(s)-G_{\delta}^{m}(r)|^{2}\}\leq C(t-r)^{2}. 
\end{equation}
Again for simplicity we let $R_{k}=R^{m}(y^m_k; \bU_{m(k+1)}^{*})$, and we will show that for any fixed $1 \leq j<k<l,$
\begin{equation} \label{lm4-3R}
E \left\{\left|\frac{1}{c_{k}}\sum_{i=1}^{k}c_{i}R_{i}-\frac{1}{c_{j}}\sum_{i=1}^{j}c_{i}R_{i}\right|^{2}\left|\frac{1}{c_{l}}\sum_{i=1}^{l}c_{i}R_{i}-\frac{1}{c_{k}}\sum_{i=1}^{k}c_{i}R_{i}\right|^{2}\right\}\leq C(l-j)^{2} .
\end{equation}
%where $C$ is a generic constant free of the choice of $(j, k, l)$. 
Indeed, recall $c_i=i(i+1)(i+2)$, and define $S_k=\sum_{i=1}^k c_iR_i^m$. Then there exists a constant $C_1=\gamma^2 C$, $\gamma>1$, 
such that $E|S_k|^4\leq C_1k^2c_k^4$, which we will prove by induction. 
Lemma \ref{lm-0_1} implies it holds for $k=1$. Assume it holds for $k-1$, then using Lemma \ref{lm-0_1} we have 
\[\begin{split}
E|S_k|^4=&E\left[\left(\sum_{i=1}^k c_i(R_i^m)'\right)\left(\sum_{i=1}^k c_iR_i^m\right)\left(\sum_{i=1}^k c_i(R_i^m)'\right)\left(\sum_{i=1}^k c_iR_i^m\right)\right] \\
\leq&E|S_{k-1}|^4+c_k^4E|R_k^m|^4+4\sum_{i=1}^{k-1}c_ic_k^3E(|R_i^m|\cdot|R_k^m|^3)+6c_k^2E(|S_{k-1}|^2|R_k^m|^2)\\
&+4\sum_{i,j,l<k}c_ic_jc_lc_kE((R_i^m)'R_j^m(R_l^m)'R_k^m) \\
\leq&C_1(k-1)^2c_k^4+c_k^4C+4(k-1)c_k^4C+6c_k^2\sqrt{C_1}(k-1)c_k^2\sqrt{C} \\
\leq&c_k^4(C_1(k-1)^2+(4+6\gamma)Ck)\\
\leq&c_k^4(C_1(k-1)^2+\gamma^2Ck)\\
\leq&C_1k^2c_k^4, 
\end{split}\]
where we may take $\gamma=7$ so that $4+6\gamma<\gamma^2$, and  we have employed Cauchy-Schwarz inequality multiple times. 
Also in above derivation we use the facts that 
\[E(|R_i^m|\cdot|R_k^m|^3)\leq (E|R_i^m|^4)^{\frac{1}{4}}\cdot(E|R_k^m|^4)^{\frac{3}{4}}\leq C,\]
\[E(|S_{k-1}|^2|R_k^m|^2)\leq  \sqrt{E|S_{k-1}|^4\cdot E|R_k^m|^4}\leq \sqrt{C_1}(k-1)c_k^2\sqrt{C}\]
and the zero conditional mean for $i,j,l<k$,
\[E((R_i^m)'R_j^m(R_l^m)'R_k^m)=E[E[(R_i^m)'R_j^m(R_l^m)'R_k^m|\cF_{t_k}]]=E[(R_i^m)'R_j^m(R_l^m)'E[R_k^m|\cF_{t_k}]]=0. \]
Note that all we used in proving $E|S_k|^4\leq C_1k^2c_k^4$ are the above zero conditional mean and $E|R_k^m|^4\leq C$ implied by
Lemma \ref{lm-0_1}. Applying the argument to $S_k - S_j$ we obtain that 
%so the sum of $S_k$ does not necessarily start from $i=1$, i.e. for $j<k$, we also have 
$E|S_k-S_j|^4\leq C_1(k-j)^2c_k^4$. Since
\[\frac{(c_k-c_j)^2}{c_k^2}\leq\frac{c_k-c_j}{c_k}=1-\frac{j(j+1)(j+2)}{k(k+1)(k+2)}\leq1-\frac{j^3}{k^3}\leq 3\left(\frac{k-j}{k}\right),\]
direct calculations show 
\[\begin{split}
E\left|\frac{S_k}{c_k}-\frac{S_j}{c_j}\right|^4&=E\left|\frac{S_k-S_j}{c_k}-\frac{c_k-c_j}{c_jc_k}S_j\right|^4 \\
&\leq 8\left(\frac{E|S_k-S_j|^4}{c_k^4}+\frac{E|S_j|^4(c_k-c_j)^4}{c_j^4c_k^4}\right) \\
&\leq 8\left(C_1(k-j)^2+\frac{9C_1j^2(k-j)^2}{k^2}\right) \\
&\leq C_2(k-j)^2
\end{split}\]
Hence we conclude for $j<k<l$, 
\[E\left[\left|\frac{S_l}{c_l}-\frac{S_k}{c_k}\right|^2\left|\frac{S_k}{c_k}-\frac{S_j}{c_j}\right|^2\right]
\leq \left(E\left|\frac{S_l}{c_l}-\frac{S_k}{c_k}\right|^4\right)^{\frac{1}{2}}\left(E\left|\frac{S_k}{c_k}-\frac{S_j}{c_j}\right|^4\right)^{\frac{1}{2}}\leq C_2(l-j)^2, \]
which proves (\ref{lm4-3R}). The rest arguemnts for establishing (\ref{lm4-2R}) are easy and pretty much the same as for 
establishing (\ref{lm4-2}). 

The finite-dimensional distribution convergence and (\ref{lm4-2R}) show $G^m_\delta(t)$ has the same weak convergence limit $G(t)$ 
as $\check{G}^m_\delta (t)$.

Skorohod's representation theorem shows that there exist $\tiG^m_\delta(t)$ and $\tiG(t)$ on some common probability spaces, such that $\tiG^m_\delta(t)$ and $G^m_\delta(t)$ are identically distributed, $\tiG(t)$ and $G(t)$ are identically distributed, and as $\delta\rightarrow 0$, $m\rightarrow\infty$, under the metric $d$ in $D[0,T]$, 
%$\tiG^m_\delta(t)$ almost surely converges to $\tiG(t)$, which also converges in probability,
%\end{spacing}
$$d(\tiG_\delta^m(t), \tiG(t))=o_p(1). $$

By Lemma \ref{lm5} below we obtain that if we further prove the tightness of $\tiG(t)$, then above $o_p(1)$ result under the metric in $D[0,T]$ can be strengthen to %the metric in $C[0,T]$, that is, 
the maximum norm, that is, 
\begin{eqnarray} \label{lm4-5}
  \max_{t\leq T}|\tiG_\delta^m(t)-\tiG(t)|=o_p(1) . 
\end{eqnarray}
%and we complete the proof of the lemma. 
We need to establish the tightness of $\tiG(t)$. Because $G(t)$ and $\tiG(t)$ are identically distributed,  if we show that for any 
$0\leq r\leq s\leq t\leq T$, 
\begin{eqnarray} \label{lm4-6}
E\left\{|G(t)-G(s)|^2|G(s)-G(r)|^2\right\}\leq C(t-r)^2.
\end{eqnarray}
Then $\tiG$ also satisfies above inequality, and both $G(t)$ and $\tiG(t)$ are tight. 

We prove (\ref{lm4-6}). If $r>0$, let 
$$D_1=\int_0^r u^3\bsigma(X(u))d\bB(u), D_2=\int_r^s u^3\bsigma(X(u))d\bB(u), D_3=\int_s^t u^3\bsigma(X(u))d\bB(u).$$
By Assumption A3, we have 
$$\|\bsigma^{2}(X(t))\|\leq\|\bsigma^{2}(\theta_{0})\|+L|X(t)-\theta_{0}|\leq C.$$
$D_2$ follows a normal distribution with mean $0$ and variance $\displaystyle \Sigma=\int_r^s u^6\bsigma^2(X(u))du$, and 
$$\|\Sigma\|\leq\int_r^s u^6\|\bsigma^{2}(X(u))\|du\leq C(s-r)s^6. $$
Taking eigen-matrix decomposition $\Sigma=\Gamma'\Lambda\Gamma$, we obtain that $\Gamma D_2$ follows a normal distribution with mean $0$ and variance matrix $\Lambda$, and 
$$E|D_2|^2=E|\Gamma D_2|^2=tr(\Lambda)\leq C(s-r)s^6,$$
$$E|D_2|^4 \leq C E|\Gamma D_2|^4\leq Ctr(\Lambda^2)\leq C(s-r)^2s^{12}.$$
Similarly we have
$$E|D_1|^2\leq Cr^7, E|D_1|^4\leq Cr^{14},$$
$$E|D_3|^2\leq C(t-s)t^6, E|D_3|^4\leq C(t-s)^2t^{12}. $$
Putting them together we arrive at 

\begin{eqnarray*}
&&E\left\{|G(s)-G(r)|^2|G(t)-G(s)|^2\right\} \\
&=&E\left\{\left|\frac{D_1+D_2}{s^3}-\frac{D_1}{r^3}\right|^2\left|\frac{D_1+D_2+D_3}{t^3}-\frac{D_1+D_2}{s^3}\right|^2\right\} \\
&\leq& C\cdot E\left\{\left(\frac{|D_2|^2}{s^6}+\frac{(s^3-r^3)^2}{s^6r^6}|D_1|^2\right)\left(\frac{|D_3|^2}{t^6}+\frac{(t^3-s^3)^2}{t^6s^6}(|D_1|^2+|D_2|^2)\right)\right\} \\
&\leq& 9C\cdot E\left\{\left(\frac{|D_2|^2}{s^6}+\frac{s-r}{sr^6}|D_1|^2\right)\left(\frac{|D_3|^2}{t^6}+\frac{t-s}{ts^6}(|D_1|^2+|D_2|^2)\right)\right\} \\
&\leq& C\left(\frac{(s-r)s^6(t-s)t^6}{s^6t^6}+\frac{(s-r)s^6(t-s)r^7}{ts^{12}}+\frac{(t-s)(s-r)^2s^{12}}{ts^{12}}\right) \\
&& +C\left(\frac{(s-r)r^7(t-s)t^6}{sr^6t^6}+\frac{(s-r)(t-s)r^{14}}{sr^6ts^6}+\frac{(s-r)r^7(t-s)(s-r)s^6}{sr^6ts^6}\right) \\
&\leq& C(t-r)^2 . 
\end{eqnarray*}
Also similar arguments can show that for $0<r<s$,
\begin{eqnarray*}
&&E\left\{|G(s)-G(r)|^2|G(r)|^2\right\} \\
&=&E\left\{\left|\frac{D_1+D_2}{s^3}-\frac{D_1}{r^3}\right|^2\left|\frac{D_1}{r^3}\right|\right\} \\
&\leq& C\cdot E\left\{\left(\frac{|D_2|^2}{s^6}+\frac{s-r}{sr^6}|D_1|^2\right)\frac{|D_1|^2}{r^6}\right\} \\
&\leq& C\left(\frac{(s-r)s^6r^7}{s^6r^6}+\frac{(s-r)r^{14}}{sr^{12}}\right) \\
&\leq& Cs^2 .
\end{eqnarray*}

%Let $\displaystyle F(t)=t^3G(t)=\int_0^t u^3\bsigma(X(u))d\bB(u)$ is a local martingale, then $dF(t)=t^3\bsigma(X(t))d\bB(t)$,
%By Assumption A3, $\bsigma(X(t))$ is positive definite, so $d\bB(t)=t^{-3}[\bsigma(X(t))]^{-1}dF(t)$, 
%$$\bB(t)=\int_0^t u^{-3}[\bsigma(X(u))]^{-1}dF(u).$$
%Let $\tiF(t)=t^3\tiG(t)$, $\displaystyle \tilde{\bB}(t)=\int_0^t u^{-3}[\bsigma(X(u))]^{-1}d\tiF(u),$ then $F(t)$ and $\tiF(t)$, $\bB(t)$ and $\tilde{\bB}(t)$ are identically distributed, so $\tilde{\bB}(t)$ is also brownian process, and $\displaystyle \tiG(t)= \frac{1}{t^3}\int_0^t u^3\bsigma(X(u))d\tilde{\bB}(u)$.
With $\tiH(t)=\displaystyle \int_0^t\bsigma(X(u))d\tilde{\bB}(u)$, $\tiG_\delta(t)$ generated by $\tiH(t)$ via scheme (\ref{lm3-7}),  
$\displaystyle \tiG(t)= \frac{1}{t^3}\int_0^t u^3d\tiH(u)$. Lemma \ref{lm4_1} below indicates that 
%and using the same arguments for proving (\ref{lm4-5}) (but  it should be much easier), we can establish that
 as $\delta\rightarrow 0$, $$\max_{t\leq T}|\tiG_\delta(t)-\tiG(t)|=o_p(1).$$
Finally combining above result with (\ref{lm4-5}) we conclude 
$$\max_{t\leq T}|\tiG^m_\delta(t)-\tiG_{\delta}(t)|\leq \max_{t\leq T}|\tiG^m_\delta(t)-\tiG(t)|+
\max_{t\leq T}|\tiG_{\delta}(t)-\tiG(t)|=o_p(1).$$

\begin{lem} \label{lm4_1}
Given a Brownian motion $B(t)$ we define $G_\delta(t)$ by (\ref{lm3-7}) and $G(t) = t^{-3} \int_0^t u^3 \bsigma(X(u)) dB(u)$ as in the proof 
of Lemma \ref{lm4}. Then we have 
\[ \max_{ t |leq T }|G_\delta(t)-G(t)|=o_p(1). \]
\end{lem}
Proof. Denote by $\bSigma_k$ the variance of $G_\delta(t_k)-G(t_k)$. Then 
\[G_\delta(t_1)-G(t_1)=\frac{1}{t_1^3}\int_0^{t_1}(t_1^3-u^3)\bsigma(X(u))d\bB(u),\]
\[\bSigma_1=\frac{1}{t_1^6}\int_0^{t_1}(t_1^3-u^3)^2\bsigma^2(X(u))du,\]
\[\|\bSigma_1\|\leq \frac{Ct_1^7}{t_1^6}\leq C\delta^{1/2}. \]
Let $C_i^\delta=t_it_{i+1}t_{i+2}$. We have for $k\geq 1$,
\[G_\delta(t_{k+1})-G(t_{k+1})=\frac{1}{C_k^\delta t_{k+1}^3}\sum_{i=0}^k\int_{t_i}^{t_{i+1}}(C_i^\delta t_{k+1}^3-C_k^\delta u^3)\bsigma(X(u))d\bB(u),\]
\[\bSigma_{k+1}=\frac{1}{t_k^2t_{k+1}^2t_{k+2}^2 t_{k+1}^6}\sum_{i=0}^k\int_{t_i}^{t_{i+1}}(t_it_{i+1}t_{i+2} t_{k+1}^3-t_kt_{k+1}t_{k+2} u^3)^2\bsigma^2(X(u))du. \]
Since $|t_it_{i+1}t_{i+2} t_{k+1}^3-t_kt_{k+1}t_{k+2} u^3|\leq Ct_{k+1}^5\delta^{1/2}$, $t_kt_{k+2}\geq t_{k+1}^2/2$, we obtain 
\[\|\bSigma_{k+1}\|\leq \frac{C}{t_{k+1}^{12}}\sum_{i=0}^k\int_{t_i}^{t_{i+1}}t_{k+1}^{10}\delta du\leq \frac{C\delta}{t_{k+1}}\leq C\delta^{1/2}.\]
That is,  $\|\bSigma_{k}\|\leq C\delta^{1/2}$ uniformly over $k \leq T \delta^{-1/2}$.  As both $G_\delta(t)$ and $G(t)$ are normally distributed, 
we get 
\[E|G_\delta(t_k)-G(t_k)|^4\leq C\delta,\]
and hence for any $\eta>0$,
\[\begin{split}
&P(\max_{k\leq k_T}|G_\delta(t_k)-G(t_k)|>\eta) \leq \sum_{k=0}^{k_T}P(|G_\delta(t_k)-G(t_k)|>\eta)\\
&\leq \sum_{k=0}^{k_T}\frac{E|G_\delta(t_k)-G(t_k)|^4}{\eta^4} 
  \leq \sum_{k=0}^{k_T}\frac{C\delta}{\eta^4} 
 \leq \frac{CT\delta^{1/2}}{\eta^4} 
  \rightarrow 0. 
\end{split}\]
Finally, %we have prove $\max_{k\leq k_T}|G_\delta(t_k)-G(t_k)|=o_p(1)$, next by
the tightness of $G(t)$ implies that 
\[\max_{s,t\leq T, |t-s|\leq \delta^{1/2}}|G(t)-G(s)|=o_p(1),\]
and thus 
\[\max_{t\leq T}|G_\delta(t)-G(t)|\leq \max_{k\leq k_T}|G_\delta(t_k)-G(t_k)|+\max_{k\leq k_T, t_k\leq t<t_{k+1}}|G(t)-G(t_k)|=o_p(1).\]

The following lemma is a known result, but we state it explicitly in our context. 
\begin{lem}\label{lm5}
Let $D[0,T]$ be the space of all c\`{a}dl\`{a}g functions on $[0,T]$,  equipped with metric $d(X(t), Y(t))$ given by
\begin{eqnarray*}
& d(X(t), Y(t))=\inf\left\{\delta : \exists  \mbox{one to one map} \; \Gamma\ on\ [0,T]  \mbox{such that} \;  \sup_{t\leq T}|\Gamma(t)-t|\leq \delta, \right. & \\
& \left. \sup_{t\leq T}|X(\Gamma(t))-Y(t)|\leq \delta \right\}. &
\end{eqnarray*}
For processes $X_n(t)$ and $X(t)$ in $D[0,T]$, assume that $X(t)$ is tight, and as $n \rightarrow \infty$, 
$d(X_n(t), X(t))=o_p(1)$ under the metric in $D[0,T]$. Then we have $$\sup_{t\leq T}|X_n(t)-X(t)|=o_p(1).$$
\end{lem}
Proof.
For any $\ve>0, \eta>0$, by the tightness of $X(t)$, there exists $\delta<\eta/2$ such that
$$P\left(\sup_{s,t\leq T,|t-s|\leq \delta}|X(t)-X(s)|>\eta/2\right)<\ve/2.$$
Let 
$$A_n=\left\{\sup_{s,t\leq T,|t-s|\leq \delta}|X(t)-X(s)|\leq\eta/2\right\}\cap\left\{d(X_n(t), X(t))<\delta\right\},$$
$$B_n=\left\{\sup_{t\leq T}|X_n(t)-X(t)|\leq \eta\right\} . $$
Then $A_n\subseteq B_n$.
%\vspace{0.1in}
Indeed, if $d(X_n(t), X(t))<\delta$, then there exists a one-to-one map $\Gamma$ on $[0,T]$, such that $\sup_{t\leq T}|\Gamma(t)-t|\leq \delta$ and $\sup_{t\leq T}|X_n(t)-X(\Gamma(t))|\leq \delta$. If we also have $\displaystyle\sup_{|t-s|\leq \delta}|X(t)-X(s)|\leq\eta/2$, then
$$\sup_{t\leq T}|X(\Gamma(t))-X(t)|\leq \eta/2,$$
$$\sup_{t\leq T}|X_n(t)-X(t)|\leq \sup_{t\leq T}|X_n(t)-X(\Gamma(t))|+\sup_{t\leq T}|X(\Gamma(t))-X(t)|\leq \delta+\eta/2\leq \eta.$$

Hence we have $P(B_n^C)\leq P(A_n^C)$, and 
$$P\left(\sup_{t\leq T}|X_n(t)-X(t)|> \eta\right)\leq P\left(\sup_{s,t\leq T,|t-s|\leq \delta}|X(t)-X(s)|>\eta/2\right)
+P(d(X_n(t), X(t))\geq\delta).$$
Since $d(X_n(t), X(t))=o_p(1)$, $\exists N$, for $n>N$, $P(d(X_n(t), X(t))\geq\delta)< \ve/2$, then
$$P\left(\sup_{t\leq T}|X_n(t)-X(t)|> \eta\right)<\ve/2+\ve/2=\ve.$$
This completes the proof. 
%That is, 
%$$\sup_{t\leq T}|X_n(t)-X(t)|=o_p(1).$$

\begin{lem} \label{lm6}
\[  \max_{k\leq T \delta^{-1/2} }\left|\cx_k^m-X_\delta^m(t_k)\right|  = O_p(\delta^{1/2}|\log\delta|),  \]
where $\cx_k^m$ and $X_\delta^m$ are defined by (\ref{lm3-2}) and (\ref{Nest-stoch1}), 
respectively. 
\end{lem}
Proof. The same proof argument of Lemma \ref{lem-s5} can be easily used to show 
\begin{equation} \label{equHa}
\bPsi_a=\sup_{0\leq s<v\leq T}\left|\frac{1}{(v-s)^a}\int_s^v \bsigma(X(u))d\bB(u)\right| \; \mbox{is a.s. finite.}
\end{equation}
By Lemma \ref{lem-s12} we have 
$$\max_{t\in[0,T]}|\dot{X}_\delta^m(t)|\leq \max_{t\in[0,T]}|\dot{X}_\delta^m(t)-\dot{X}(t)|+\max_{t\in[0,T]}|\dot{X}(t)|=O_p(1),$$
$$\max_{t\in[0,T]}|X_\delta^m(t)|\leq \max_{t\in[0,T]}|X_\delta^m(t)-X(t)|+\max_{t\in[0,T]}|X(t)|=O_p(1),$$
$$\max_{t\in[0,T]}|\nabla g(X_\delta^m(t))|\leq |\nabla g(\theta_0)|+L\cdot\max_{t\in[0,T]}|X_\delta^m(t)-\theta_0|=O_p(1).$$
Let 
$$\Upsilon_\delta^m=\max\left\{\bPsi_a, \max_{t\in[0,T]}|\dot{X}_\delta^m(t)|, \max_{t\in[0,T]}|X_\delta^m(t)|, \max_{t\in[0,T]}|\nabla g(X_\delta^m(t))|\right\}.$$
Then $\Upsilon_\delta^m=O_p(1)$. For simplicity we will still use notation $\Upsilon_\delta^m$ to denote it after multiplying and adding 
some generic constant $C$ or adding random variable $\bPsi_a$ in (\ref{equHa}) so long as it is $O_p(1)$.
%we omit the subscript and superscript $\delta$ and $m$ for $R_\delta^m$ in the following derivation for simplicity.

For a fixed $a<1/2$,  set $\xi=\sqrt{(a+1)(a+2)/2L}$. By Lemma \ref{lem-s8} we have for $t<\xi$,
$$M_a(0,t;X_\delta^m)\leq 2(t^{1-a}|\nabla g(x_0)|+\delta^{1/4}m^{-1/2}\Upsilon_\delta^m),$$
$$|\dot{X}_\delta^m(t)|\leq C(t+t^a\delta^{1/4}m^{-1/2}\Upsilon_\delta^m) , $$
and for $s>0$ and $t-s<\xi$, 
$$M_a(s,t;X_\delta^m)\leq 2\left\{(t-s)^{1-a}\left(\frac{3}{s}+\frac{L(t-s)}{2}\right)|\dot{X}_\delta^m(s)|+(t-s)^{1-a}|\nabla g(X_\delta^m(s))|+\delta^{1/4}m^{-1/2}\Upsilon_\delta^m\right\}.$$
If further $t-s\leq s$, then for $s<\xi$, 
\begin{eqnarray*}
&&(t-s)^{1-a}\left(\frac{3}{s}+\frac{L(t-s)}{2}\right)|\dot{X}_\delta^m(s)| \\
&\leq&C(t-s)^{1-a}\left(\frac{3}{s}+\frac{L(t-s)}{2}\right)(s+s^a\delta^{1/4}m^{-1/2}\Upsilon_\delta^m) \\
&\leq&C(t-s)^{1-a}\left(3+\frac{L(t-s)s}{2}\right)+C(t-s)^{1-a}\left(3s^{a-1}+\frac{L(t-s)s^a}{2}\right)\delta^{1/4}m^{-1/2}\Upsilon_\delta^m \\
&\leq&C(t-s)^{1-a}+C\left[3\left(\frac{t-s}{s}\right)^{1-a}+\frac{L(t-s)^{2-a}s^a}{2}\right]\delta^{1/4}m^{-1/2}\Upsilon_\delta^m 
\end{eqnarray*}
\begin{eqnarray*}
&\leq&C(t-s)^{1-a}+C\delta^{1/4}m^{-1/2}\Upsilon_\delta^m,
\end{eqnarray*}
and for $s\geq\xi$,
$$(t-s)^{1-a}\left(\frac{3}{s}+\frac{L(t-s)}{2}\right)|\dot{X}_\delta^m(s)|\leq C(t-s)^{1-a}\Upsilon_\delta^m. $$
Putting them together we conclude 
\begin{eqnarray*}
&& |\dot{X}_\delta^m(t)-\dot{X}_\delta^m(s)|\leq C(t-s)(\Upsilon_\delta^m+1)+C(t-s)^a\delta^{1/4}m^{-1/2}\Upsilon_\delta^m \\
&& \leq C (\Upsilon_\delta^m+1) \left[(t-s)+(t-s)^a\delta^{1/4}m^{-1/2}\right] = \Upsilon_\delta^m  \left[(t-s)+(t-s)^a\delta^{1/4}m^{-1/2}\right],
\end{eqnarray*}
where we use the notation convention noted early to write $\Upsilon _\delta^m$ for $C(\Upsilon_\delta^m+1)$. 
%we will replace $R_1$ by $R$ in the following analysis.

%textbf{Remark 1:} We may also write all those uniformly bounded random elements as $O_p(1)$, but to deal with the confusion caused by different $O_p(1)$ in different occasions, and make sure there is no accumulative effects for the infinite indices in the sequence when $\delta\rightarrow 0$, I would prefer use a uniform upper bound for all indices which is $O_p(1)$, e.g., denote by R).

%Let's continue on the proof, 
The theorem assumption implies that $\delta^{a/2-1/4}m^{-1/2} < C_0$ for some generic constant $C_0$. For $\delta^{1/2}<\xi$, we have 
if $t-s\leq \delta^{1/2}$ and $t-s\leq s$,  
$$|\dot{X}_\delta^m(t)-\dot{X}_\delta^m(s)|\leq \left[\delta^{1/2}+\delta^{a/2}\delta^{1/4}m^{-1/2}\right]\Upsilon_\delta^m\leq \delta^{1/2}(1+C_0)\Upsilon_\delta^m 
   \leq \delta^{1/2} \Upsilon_\delta^m,$$
and if $t\leq \delta^{1/2}$, 
$$|\dot{X}_\delta^m(t)|\leq \left[\delta^{1/2}+\delta^{a/2}\delta^{1/4}m^{-1/2}\right]\Upsilon_\delta^m\leq \delta^{1/2}(1+C_0)\Upsilon_\delta^m \leq \delta^{1/2} \Upsilon_\delta^m. $$
%We wil update $R$ by $(1+C_0)R$, which is still $O_p(1)$.

Remind that $t_k=k\delta^{1/2}$ for any $k\geq 1$, and $t_{k+1}-t_k=\delta^{1/2}\leq t_k$. Then for any $t\in [t_k,t_{k+1}]$, 
$$|\dot{X}_\delta^m(t)-\dot{X}_\delta^m(t_k)|\leq \delta^{1/2}\Upsilon_\delta^m, \;\; |\dot{X}_\delta^m(t_{k+1})-\dot{X}_\delta^m(t)|\leq \delta^{1/2}\Upsilon_\delta^m,$$
and thus 
$$|\dot{X}_\delta^m(t_k)|\leq |\dot{X}_\delta^m(t_1)|+|\dot{X}_\delta^m(t_2)-\dot{X}_\delta^m(t_1)|+\cdots +|\dot{X}_\delta^m(t_k)-\dot{X}_\delta^m(t_{k-1})|\leq k\delta^{1/2}\Upsilon_\delta^m=t_k \Upsilon_\delta^m, $$
%in addition, we have
$$|X_\delta^m(t_{k+1})-X_\delta^m(t_k)|\leq \int_{t_k}^{t_{k+1}}|\dot{X}_\delta^m(t)|dt\leq \delta^{1/2}\Upsilon_\delta^m.$$
%Review the definition (\ref{lm3-2}),
%\begin{equation} 
%\check{x}_{k}^{m}=\check{y}_{k-1}^{m}-\delta\nabla g(\check{y}_{k-1}^{m})-m^{-1/2}\delta^{3/4}(H(t_{k})-H(t_{k-1})),  \check{y}_{k}^{m}=\check{x}_{k}^{m}+\frac{k-1}{k+2}(\check{x}_{k}^{m}-\check{x}_{k-1}^{m}),
%\end{equation}
Define $\cz_0^m=0$, $\cz_k^m=(\cx_k^m-\cx_{k-1}^m)/\delta^{1/2}.$ Using the definition of $\check{x}_{k}^{m}$ and $\check{y}_{k}^{m}$
in  (\ref{lm3-2}), we obtain 
$$\cz_1^m=(\cx_1^m-x_0)/\delta^{1/2}=-\delta^{1/2}\nabla g(x_0)-m^{-1/2}\delta^{1/4}(H(t_1)-H(t_0)).$$
Then using (\ref{equHa}) we get 
$$|\cz_1^m|\leq \delta^{1/2}|\nabla g(x_0)|+m^{-1/2}\delta^{1/4}\delta^{a/2}\bPsi_a\leq \delta^{1/2}(|\nabla g(x_0)|+C_0\bPsi_a). $$
Again we use the notation convention to write $\Upsilon_\delta^m$ for $\Upsilon_\delta^m + |\nabla g(x_0)|+C_0\bPsi_a$ (which is still $O_p(1)$). 
%(we will only update finite times, and we are almost done!)
Using above result, the notation convention, and the definition of $\check{x}_{k}^{m}$ and $\check{y}_{k}^{m}$ in  (\ref{lm3-2}), we get 
$$|\cx_1^m-x_0|\leq \delta^{1/2}\delta^{1/2}\Upsilon_\delta^m=\delta \Upsilon_\delta^m, \; \; |X_\delta^m(t_1)-x_0|\leq \int_0^{t_1}|\dot{X}_\delta^m(t)|dt\leq \delta \Upsilon_\delta^m . $$ 
Let $a_k=|\cx_k^m-X_\delta^m(t_k)|$. Then $a_0=0$, $a_1\leq 2\delta \Upsilon_\delta^m$. For $k\geq 2$, 
\begin{eqnarray*}
X_\delta^m(t_k)&=&X_\delta^m(t_{k-1})+\int_{t_{k-1}}^{t_k}\dot{X}_\delta^m(t)dt \\
&=&X_\delta^m(t_{k-1})+\delta^{1/2}\dot{X}_\delta^m(t_k)+\int_{t_{k-1}}^{t_k}(\dot{X}_\delta^m(t)-\dot{X}_\delta^m(t_k))dt . 
\end{eqnarray*}
Set $Z_\delta^m(t)=\dot{X}_\delta^m(t)$,  $b_k=|\cz_k^m-Z_\delta^m(t_k)|$. Then $b_0=0$, $b_1\leq 2\delta^{1/2}\Upsilon_\delta^m$. 
%update $R$ by $2R$, 
Combining above equality with the definition of $\cz_k^m$ (i.e. $\cx_k^m=\cx_{k-1}^m+\delta^{1/2}\cz_k^m$), we conclude 
\begin{eqnarray} \label{lm6-equ1}
a_k=|\cx_k^m-X_\delta^m(t_k)|&\leq& |\cx_{k-1}^m-X_\delta^m(t_{k-1})|+\delta^{1/2}|\cz_k^m-Z_\delta^m(t_k)|+\int_{t_{k-1}}^{t_k}|\dot{X}_\delta^m(t_k)-\dot{X}_\delta^m(t))|dt  \nonumber \\
&\leq& a_{k-1}+\delta^{1/2}b_k+\delta \Upsilon_\delta^m  \nonumber \\
&\leq& a_1+\delta^{1/2}(b_2+\dots+b_k)+(k-1)\delta \Upsilon_\delta^m  \nonumber \\
&\leq& \delta^{1/2}S_k+k\delta \Upsilon_\delta^m,
\end{eqnarray}
where $S_k=b_1+\dots+b_k$. 

Note that $Z_\delta^m(t)=\dot{X}_\delta^m(t)$ obeys 
$$dZ_\delta^m(t)=-\frac{3}{t}Z_\delta^m(t)dt-\nabla g(X_\delta^m(t))dt-m^{-1/2}\delta^{1/4}dH(t),$$
and thus 
\begin{eqnarray} \label{thm3-1}  \nonumber
Z_\delta^m(t_{k+1})&=&Z_\delta^m(t_k)-\int_{t_k}^{t_{k+1}}\frac{3}{t}Z_\delta^m(t)dt-\int_{t_k}^{t_{k+1}}\nabla g(X_\delta^m(t))dt \\ \nonumber
&&-m^{-1/2}\delta^{1/4}(H(t_{k+1})-H(t_k)) \\ \nonumber
&=&Z_\delta^m(t_k)-\frac{3\delta^{1/2}}{t_k}Z_\delta^m(t_k)-\int_{t_k}^{t_{k+1}}\left[\frac{3}{t}Z_\delta^m(t)dt-
\frac{3}{t_k}Z_\delta^m(t_k)\right]dt-\delta^{1/2}\nabla g(X_\delta^m(t_k)) \\ 
&&-\int_{t_k}^{t_{k+1}}[\nabla g(X_\delta^m(t))-\nabla g(X_\delta^m(t_k))]dt
-m^{-1/2}\delta^{1/4}(H(t_{k+1})-H(t_k)).
\end{eqnarray}
For $k\geq 1$,
\begin{eqnarray*}
\left|\int_{t_k}^{t_{k+1}}\left[\frac{3}{t}Z_\delta^m(t)-\frac{3}{t_k}Z_\delta^m(t_k)\right]dt\right|
&\leq&\int_{t_k}^{t_{k+1}}\left|\frac{3}{t}[Z_\delta^m(t)-Z_\delta^m(t_k)]\right|du+\int_{t_k}^{t_{k+1}}\left|\left(\frac{3}{t}-\frac{3}{t_k}\right)Z_\delta^m(t_k)\right|dt \\
&\leq&\frac{3\delta \Upsilon_\delta^m}{t_k}+\frac{3(t_{k+1}-t_k)^2}{t_kt_{k+1}}t_k \Upsilon_\delta^m \\
&\leq&6\delta^{1/2} k^{-1}\Upsilon_\delta^m,
\end{eqnarray*}
$$\left|\int_{t_k}^{t_{k+1}}[\nabla g(X_\delta^m(t))-\nabla g(X_\delta^m(t_k))]du\right|\leq L\int_{t_k}^{t_{k+1}}|X_\delta^m(t)-X_\delta^m(t_k)|du\leq L\delta \Upsilon_\delta^m.$$
Recall (\ref{lm3-4}) and note that $\cz_k^m=\cd_{k-1}^m/\delta^{1/2}$, 
$$\cz_{k+1}^m=\frac{k-1}{k+2}\cz_k^m-\delta^{1/2}\nabla g(\check{y}_{k}^{m})-m^{-1/2}\delta^{1/4}(H(t_{k+1})-H(t_k)). $$
Using above equality and (\ref{thm3-1}), we arrive at 
\begin{eqnarray*} 
b_{k+1}&=&|\cz_{k+1}^m-Z_\delta^m(t_{k+1})|
\leq\left(1-\frac{3}{k+2}\right)|\cz_k^m-Z_\delta^m(t_k)|+\frac{6}{k(k+2)}|Z_\delta^m(t_k)| \\
&&+\left|\int_{t_k}^{t_{k+1}}\left[\frac{3}{t}Z_\delta^m(t)dt-\frac{3}{t_k}Z_\delta^m(t_k)\right]dt\right|
+\delta^{1/2}|\nabla g(X_\delta^m(t_k))-\nabla g(\cy_k^m)| \\
&&\left|\int_{t_k}^{t_{k+1}}[\nabla g(X_\delta^m(t))-\nabla g(X_\delta^m(t_k))]dt\right| \\
&\leq&b_k+12\delta^{1/2}k^{-1}\Upsilon_\delta^m+L\delta^{1/2}\left|X_\delta^m(t_k)-\cx_k^m-\frac{k-1}{k+2}\delta^{1/2}\cz_k^m\right|
+L\delta \Upsilon_\delta^m 
\end{eqnarray*}
\begin{eqnarray*} 
&\leq&b_k+12\delta^{1/2}k^{-1}\Upsilon_\delta^m+L\delta^{1/2}\left(a_k+\delta^{1/2}(|Z_\delta^m(t_k)|+|\cz_k^m-Z_\delta^m(t_k)|)\right)
+L\delta \Upsilon_\delta^m\\
&\leq&b_k+12\delta^{1/2}k^{-1}\Upsilon_\delta^m+L\delta^{1/2}(\delta^{1/2}S_k+k\delta \Upsilon_\delta^m+\delta^{1/2}(\Upsilon_\delta^m+b_k))
+L\delta \Upsilon_\delta^m \\
&\leq&b_k+C\delta S_k+C\delta^{1/2}k^{-1}\Upsilon_\delta^m,
\end{eqnarray*}
where we use the fact $\delta\leq T\delta^{1/2}k^{-1}$. %above, $C$ is a constant that only depends on $L$ and $T$.

Let $b'_1=b_1$, $b'_{k+1}=b'_k+C\delta S'_k+C\delta^{1/2} k^{-1}\Upsilon_\delta^m$, where $S'_k=b'_1+b'_2+...+b'_k$. Then we can prove by induction that $b_k\leq b'_k$. Indeed, if $b_j\leq b'_j$ for $j=1,...,k$, then $S_k\leq S'_k$, 
$$b_{k+1}\leq b_k+C\delta S_k+C\delta^{1/2} k^{-1}\Upsilon_\delta^m\leq b'_k+C\delta S'_k+C\delta^{1/2} k^{-1}\Upsilon_\delta^m=b'_{k+1}.$$
Next, since $C\delta S'_k+C\delta^{1/2} k^{-1}\Upsilon_\delta^m\geq 0$, and $\{b'_k\}$ is non-decreasing, we have $S'_k\leq kb'_k$,
$$b'_{k+1}\leq b'_k+C\delta kb'_k+C\delta^{1/2} k^{-1}\Upsilon_\delta^m.$$
Similarly, let $b_1^*=b'_1$, $b_{k+1}^*=b_k^*+C\delta kb_k^*+C\delta^{1/2} k^{-1}\Upsilon_\delta^m$. The same argument leads to $b'_k\leq b_k^*$. 
It is easy to derive from the definition that 
\begin{eqnarray*}
b_{k+1}^*&=&(1+C\delta k)b_k^*+C\delta^{1/2} k^{-1}\Upsilon_\delta^m \\
&=&(1+C\delta k)((1+C\delta (k-1))b_{k-1}^*+C\delta^{1/2} (k-1)^{-1}\Upsilon_\delta^m)+C\delta^{1/2} k^{-1}\Upsilon_\delta^m \\
&=&\cdots\\
&\leq&(1+C\delta k)^k\left(b_1^*+C\delta^{1/2}\Upsilon_\delta^m\sum_{j=1}^k j^{-1}\right) \\
&\leq&(1+C\delta k)^k\left(\delta^{1/2}\Upsilon_\delta^m+C\delta^{1/2}\Upsilon_\delta^m (1+\log(k))\right).
\end{eqnarray*}
Let $k_T=\lfloor T/\delta^{1/2}\rfloor$. Then using (\ref{lm6-equ1}) and above bound result we conclude 
\begin{eqnarray*}
&& \max_{k\leq k_T}\left|\cx_k^m-X_\delta^m(t_k)\right| =\max_{k\leq k_T} a_k \leq \delta^{1/2}S_{k_T}^*+{k_T}\delta \Upsilon_\delta^m \\
&& \leq \delta^{1/2}k_Tb_{k_T}^*+T\delta^{1/2} \Upsilon_\delta^m \\
&& \leq (1+C\delta T/\delta^{1/2})^{T/\delta^{1/2}}C\delta^{1/2}\Upsilon_\delta^m(1+\log(T/\delta^{1/2}))+T\delta^{1/2} \Upsilon_\delta^m 
\end{eqnarray*}
\begin{eqnarray*}
&& \leq CTe^{CT^2}\delta^{1/2}\Upsilon_\delta^m(1+T+|\log\delta|/2)+T\delta^{1/2} \Upsilon_\delta^m \\
&& =O_p(\delta^{1/2}|\log\delta|). 
\end{eqnarray*}

\textbf{Proof of Theorem \ref{thm7}} 
As in Lemma \ref{lm3}, we may realize $x_{k}^{m}$, $B(t)$, $H(t)$, $\cx_k^m$, and $X_\delta^m(t)$ [defined by $B(t)$ via (\ref{Nest-stoch1})] on some common probability spaces and consider their versions 
$\tilde{x}_{k}^{m}$, $\tilde{B}(t)$, $\tilde{H}(t)$, $\tilde{\cx}_k^m$, and $\tilde{X}_\delta^m(t)$  on the probability spaces.  
Applying Lemma \ref{lm6}, we have 
\[ \max_{k \leq T \delta^{-1/2}} | \tilde{\cx}_k^m - \tilde{X}_\delta^m(t_k)| = O_P(\delta^{1/2}|\log\delta|). \]
Combining above result together with Lemmas \ref{lm3} we obtain 
$$\max_{k\leq T/\delta^{1/2}}|\tilde{x}_{k}^{m}- \tilde{X}_\delta^m(t_k) | =o_p(m^{-1/2}\delta^{1/4})  + O_P(\delta^{1/2}|\log\delta|). $$
For process $X_\delta^m(t)$, we have shown in the proof of Lemma \ref{lm6} 
$$\max_{t-s\leq \delta^{1/2}}| X_\delta^m(t)- X_\delta^m(s)|\leq \delta^{1/2}\Upsilon_\delta^m=O_p(\delta^{1/2}),$$
and thus the same result also holds for $\tilde{X}_\delta^m(t)$. 
Therefore we can conclude  
$$\max_{t\leq T}| \tilde{x}_\delta^m(t)-\tilde{X}_\delta^m(t)|=o_p(m^{-1/2}\delta^{1/4})+O_p(\delta^{1/2}|\log\delta|),$$
where we use the fact that $x_\delta^m(t)=x_k^m$ for $t_k\leq t<t_{k+1}$. With the theorem condition 
$m^{1/2}\delta^{1/4}|\log\delta|\rightarrow 0$ we immediately arrive at 
$$m^{1/2}\delta^{-1/4}\max_{t\leq T}|\tilde{x}_\delta^m(t)- \tilde{X}_\delta^m(t)|=o_p(1),$$
and Theorem \ref{thm6} indicates that $m^{1/2}\delta^{-1/4}[x_\delta^m(t)-X(t)]$ weakly converges to $V(t)$.

% for some $a<1/2$, we have $\delta^{a/2-1/4}m^{-1/2}<C_0$. These two conditions are not contradicted, e.g., for any $p>2$, let $\delta=m^{-p}$, then $$\delta^{a/2-1/4}m^{-1/2}\leq 1$$  holds when take any fixed $a>1/2-1/p$, and $$m^{1/2}\delta^{1/4}|\log\delta|=m^{1/2-p/4}p\log m\rightarrow 0.$$

%\textbf{Remark 2:} If there exists $a<1/2$ and $C_0$ such that $\delta^{a/2-1/4}m^{-1/2}<C_0$, then $\delta^{a-1/2}<C_0^2m$, $\delta>Cm^{-1/(1/2-a)}$, let $p=1/(1/2-a)$, then $\delta\geq Cm^{-p}$; On the other hand, if $\delta\geq Cm^{-p}$, then for any $a>1/2-1/p$, $\delta^{a/2-1/4}m^{-1/2}<C_1$. This condition mainly requires that $\delta$ should not decrease too fast, no faster than polynomial speed.

%\textbf{Remark 3}: The condition $m^{1/2}\delta^{1/4}|\log\delta|\rightarrow 0$ requires that $\delta$ should decrease faster than $m^{-2}$.

\begin{remark} 
% Under further condition $m \delta^{1/4} / |\log\delta|^2  \rightarrow \infty$ and 
%With lengthy arguments
The proof arguments in fact also establish 
$$\max_{t\leq T}| y_\delta^m(t)- X(t)|=O_p( m^{-1/2}\delta^{1/4} + \delta^{1/2}|\log\delta|). $$
%We can prove $y_k^m-X(t_k) $ converges in probability to zero uniformly over $k \leq T \delta^{-1/2}$, and derive the order 
%for $ \max_{k\leq T \delta^{-1/2} }|y_k^m-X(t_k)| $ as $m \rightarrow \infty$ and $ \delta \rightarrow 0$.
\end{remark}

%% file: p6.tex
\subsection{Proof of Theorem \ref{thm8}}
Part (i) can be proved by using the same argument for showing Theorem \ref{thm-1-2}.  First we will show parts (ii) and (iii) in one dimension. 
%For simplicity we will show parts (ii) and (iii) in one dimension only, since the multivariate case 
%can be shown by following matrix arguments in Gardiner (2009, chapter 6) and Da Prato and Zabczyk (1996, chapter 9). 
From solution (\ref{GD-v1}) of SDE (\ref{GD-v0}) we find that $V(t)$ follows a normal 
distribution with mean zero and variance 
\[  \Gamma(t)=\int_0^t \exp \left[    - 2 \int_u^t \boldsymbol{I\!\! H}\! g(X(v)) dv \right] \bsigma^2(X(u)) du. \]
%\[  \Gamma(t)=\int_0^t \exp \left[    - \int_u^t \boldsymbol{I\!\! H}\! g(X(v)) dv \right] \bsigma(X(u)) [ \bsigma(X(u))]^\prime  \exp \left[    - \int_u^t \boldsymbol{I\!\! H}\! g(X(v)) dv \right] du. \]
It is easy to check that $\Gamma(t)$ satisfies 
ODE \[   \dot{\Gamma}(t)  + 2 [\boldsymbol{I\!\! H}\! g(X(t))] \Gamma(t) - \bsigma^2(X(t)) =0, %[\bsigma(X(t))]^\prime  = 0, 
\]
and show that the limit $\Gamma(\infty)$ of $\Gamma(t)$ as $t \rightarrow \infty$ is equal to 
\[ %2 [\boldsymbol{I\!\! H}\! g(X(\infty))] \Sigma(\infty) + \bsigma(X(\infty)) [\bsigma(X(\infty))]^\prime  = 0, \;\;\;
     \Gamma(\infty) =  \bsigma^2(X(\infty))  [2 \boldsymbol{I\!\! H}\! g(X(\infty))]^{-1}.  \]
Thus %we may show under some conditions that 
as $t \rightarrow \infty$, $V(t)$ converges in distribution to $V(\infty)=[\Gamma(\infty)]^{1/2} \bZ$, where $\bZ$ is a standard normal 
random variable. 

Denote by $P(\theta;t)$ the probability distribution of $X^m_\delta(t)$ at time $t$. Then from the Fokker-Planck equation we have 
\[ \frac{\partial P(\theta;t)}{\partial t} = \nabla \left[ - \nabla g(\theta) P(\theta; t) - \frac{\delta}{2 m} \bsigma^2(X(t)) %[\bsigma(X(t))]^\prime 
   \, \nabla P(\theta;t) \right], \]
and its stationary distribution $P(\theta)$ satisfies 
\[ 0 = \nabla \left[ - \nabla g(\theta) P(\theta) - \frac{\delta}{2 m} \bsigma^2(X(\infty)) %[\bsigma(X(\infty))]^\prime
  \, \nabla P(\theta) \right], \]
which has solution 
\[ P(\theta) \propto  \exp\left\{ - \frac{2 m}{\delta \bsigma^2(\check{\theta})} g(\theta) \right\}. \]
The corresponding stationary distribution $P_0(v)$ for $V^m_\delta(\infty) = (m/\delta)^{1/2} (X^m_\delta(\infty)  - \check{\theta})$ takes the form 

\begin{align*}
& P_0(v) \propto \exp\left\{  - \frac{m}{\delta \bsigma^2(\check{\theta})} g\left(\check{\theta} + \sqrt{\delta/m } \, v \right) \right\}  
    \sim   \exp\left\{ - \frac{2 m}{\delta \bsigma^2(\check{\theta})} \left[  g(\check{\theta}) + \frac{ \delta \boldsymbol{I\!\! H}\! g(\check{\theta})  }{2 m} v^2 
     \right] \right\} & \\
& \propto \exp\left\{ - \frac{ \boldsymbol{I\!\! H}\! g(\check{\theta})}{  \bsigma^2(\check{\theta})}  v^2 \right\}, &
%\sim N\left(0,  \frac{ \bsigma^2(\check{\theta}) }{ 2  \boldsymbol{I\!\! H}\!(\check{\theta}) } \right), &
\end{align*}
where %$v$ is the variable, 
we use the fact that $\nabla g(\check{\theta})=0$, and the asymptotics are based on taking $\delta \rightarrow 0$, $m \rightarrow \infty$.
Therefore, $P_0$ converges to $N\left(0,  \frac{ \bsigma^2(\check{\theta}) }{ 2  \boldsymbol{I\!\! H}\!(\check{\theta}) } \right)$, and we conclude that $V^m_\delta(\infty)$ has a limiting normal distribution 
with mean zero and variance $\bsigma^2(\check{\theta}) [2  \boldsymbol{I\!\! H}\! g(\check{\theta})]^{-1}=\Gamma(\infty)$. 

Similarly we can show parts (ii) and (iii) in the multivariate case by following matrix arguments in Gardiner (2009, chapters 4 \& 6) and Da Prato and Zabczyk (1996, chapter 9) as follows. Using the explicit solution (\ref{GD-v1}) of SDE (\ref{GD-v0}) we find that $V(t)$ follows a normal 
distribution with mean zero and variance matrix 
\begin{align} \label{balance1}
&  \Gamma(t)=\int_0^t \exp \left[    - \int_u^t \boldsymbol{I\!\! H}\! g(X(v)) dv \right] \bsigma(X(u)) [ \bsigma(X(u))]^\prime  \exp \left[    - \int_u^t \boldsymbol{I\!\! H}\! g(X(v)) dv \right] du \nonumber &\\
& = \int_0^t \exp \left[    - \int_u^t \boldsymbol{I\!\! H}\! g(X(v)) dv \right] \bsigma(X(\infty)) [ \bsigma(X(\infty))]^\prime  \exp \left[    - \int_u^t \boldsymbol{I\!\! H}\! g(X(v)) dv \right] du + 
\zeta_t, & 
\end{align}
where 
\begin{align*}
& \zeta_t = \int_0^t \exp \left[    - \int_u^t \boldsymbol{I\!\! H}\! g(X(v)) dv \right] \left\{ \bsigma(X(u)) [ \bsigma(X(u))]^\prime - \bsigma(X(\infty)) [ \bsigma(X(\infty))]^\prime 
\right\} &\\
&  \exp \left[    - \int_u^t \boldsymbol{I\!\! H}\! g(X(v)) dv \right] du. &
\end{align*}
Similar to the proof for Part 3 of Theorem \ref{thm-1-2}, we can show that as $t \rightarrow \infty$, $| \zeta_t | \rightarrow 0$. Indeed, for any $\epsilon>0$, there exists $t_0>0$ such that for any $u > t_0$, 
\begin{align*}
& \left|  \bsigma(X(u)) [ \bsigma(X(u))]^\prime - \bsigma(X(\infty)) [ \bsigma(X(\infty))]^\prime \right| < \epsilon, \;\;
  \left |  \boldsymbol{I\!\! H}\! g(X(u)) [ \boldsymbol{I\!\! H}\! g(X(\infty))]^{-1} \right| > 1 - \epsilon, &
  \end{align*}
   \begin{align*}
  & \left |  \int_0^{t_0} \exp \left[    - \int_u^t \boldsymbol{I\!\! H}\! g(X(v)) dv \right] \left\{ \bsigma(X(u)) [ \bsigma(X(u))]^\prime - \bsigma(X(\infty)) [ \bsigma(X(\infty))]^\prime \right\} \right. &\\
& \left.  \exp \left[    - \int_u^t \boldsymbol{I\!\! H}\! g(X(v)) dv \right] du \right | &\\ 
& \leq  \left|  \exp \left[ -2  \int_{t_0}^t \boldsymbol{I\!\! H}\! g(X(v)) dv \right] \right |
   \int_0^{t_0} \left |  \bsigma(X(u)) [ \bsigma(X(u))]^\prime - \bsigma(X(\infty)) [ \bsigma(X(\infty))]^\prime \right | du & \\
& \leq C  \left|  \exp \left[ -2  \int_{t_0}^t \boldsymbol{I\!\! H}\! g(X(v)) dv \right] \right | \rightarrow 0, & 
\end{align*}
\begin{align*}
& \left |  \int_{t_0}^t \exp \left[    - \int_u^t \boldsymbol{I\!\! H}\! g(X(v)) dv \right] \left\{ \bsigma(X(u)) [ \bsigma(X(u))]^\prime - \bsigma(X(\infty)) [ \bsigma(X(\infty))]^\prime \right\} \right. &\\
& \left.  \exp \left[    - \int_u^t \boldsymbol{I\!\! H}\! g(X(v)) dv \right] du \right | &\\ 
& \leq \frac{\epsilon}{1-\epsilon}  \int_{t_0}^t  \left | \exp \left[    - 2 \int_u^t \boldsymbol{I\!\! H}\! g(X(v)) dv \right]    \boldsymbol{I\!\! H}\! g(X(u))   \right| d u |  \boldsymbol{I\!\! H}\! g(X(\infty)) |^{-1} & \\
&  \leq \frac{\epsilon}{2(1-\epsilon)}   \left | 1 -  \exp \left[    - 2 \int_{t_0}^t \boldsymbol{I\!\! H}\! g(X(v)) dv \right]    \right|  |  \boldsymbol{I\!\! H}\! g(X(\infty)) |^{-1} & \\
& \leq \frac{\epsilon}{2(1-\epsilon)}   |  \boldsymbol{I\!\! H}\! g(X(\infty)) |^{-1}  \rightarrow 0, \mbox{ as we let $\epsilon \rightarrow 0$, } & 
\end{align*}
and these results implies that the integral in $\zeta_t$ can be divided into two parts over $[0, t_0]$ and $[t_0, t]$, both of which go to zero as $ t \rightarrow \infty$. 

Now we will verify the detailed balance condition using (\ref{balance1}) and $\zeta_t \rightarrow 0$.
Direct algebraic manipulations show 
\begin{align*}
& \boldsymbol{I\!\! H}\! g(X(t)) \Gamma(t)  +  \Gamma(t) \boldsymbol{I\!\! H}\! g(X(t)) & \\
& = \int_0^t  \boldsymbol{I\!\! H}\! g(X(t)) \exp \left[    - \int_u^t \boldsymbol{I\!\! H}\! g(X(v)) dv \right] \bsigma(X(u)) [ \bsigma(X(u))]^\prime  \exp \left[    - \int_u^t \boldsymbol{I\!\! H}\! g(X(v)) dv \right] du &\\
& + \int_0^t  \exp \left[    - \int_u^t \boldsymbol{I\!\! H}\! g(X(v)) dv \right] \bsigma(X(u)) [ \bsigma(X(u))]^\prime  \exp \left[    - \int_u^t \boldsymbol{I\!\! H}\! g(X(v)) dv \right] \boldsymbol{I\!\! H}\! g(X(t)) du & 
\end{align*}
\begin{align*}
& = \int_0^t  \boldsymbol{I\!\! H}\! g(X(t)) \exp \left[    - \int_u^t \boldsymbol{I\!\! H}\! g(X(v)) dv \right] \bsigma(X(\infty)) [ \bsigma(X(\infty))]^\prime  \exp \left[    - \int_u^t \boldsymbol{I\!\! H}\! g(X(v)) dv \right] du &\\
& + \int_0^t  \exp \left[    - \int_u^t \boldsymbol{I\!\! H}\! g(X(v)) dv \right] \bsigma(X(\infty)) [ \bsigma(X(\infty))]^\prime  \exp \left[    - \int_u^t \boldsymbol{I\!\! H}\! g(X(v)) dv \right] \boldsymbol{I\!\! H}\! g(X(t)) du &\\
& +  \boldsymbol{I\!\! H}\! g(X(t)) \zeta_t  +  \zeta_t \boldsymbol{I\!\! H}\! g(X(t)) & \\
%\end{align*}
%\begin{align*}
&=\int_0^t \frac{d}{ du} \left\{ \exp \left[ -\int_u^t \boldsymbol{I\!\! H}\! g(X(v)) dv \right] \bsigma(X(\infty)) [ \bsigma(X(\infty))]^\prime  \exp \left[ -\int_u^t \boldsymbol{I\!\! H}\! g(X(v)) dv \right] \right\} du  &\\
& +  \boldsymbol{I\!\! H}\! g(X(t)) \zeta_t  +  \zeta_t \boldsymbol{I\!\! H}\! g(X(t)) & \\
& = \bsigma(X(\infty)) [ \bsigma(X(\infty))]^\prime  & \\
& - \exp \left[ -\int_0^t \boldsymbol{I\!\! H}\! g(X(v)) dv \right] \bsigma(X(\infty)) [ \bsigma(X(\infty))]^\prime  \exp \left[ -\int_0^t \boldsymbol{I\!\! H}\! g(X(v)) dv \right]  &\\
& +  \boldsymbol{I\!\! H}\! g(X(t)) \zeta_t  +  \zeta_t \boldsymbol{I\!\! H}\! g(X(t)), & 
\end{align*}
where by the assumption we have that as $t \rightarrow \infty$, $\int_0^t \boldsymbol{I\!\! H}\! g(X(v)) dv \rightarrow \infty$, which together with $\zeta_t \rightarrow 0$ 
indicate that the last three terms on the right hand size of above expression go to zero. Hence we have shown that
as $t \rightarrow \infty$, $ \boldsymbol{I\!\! H}\! g(X(t)) \Gamma(t)  +  \Gamma(t) \boldsymbol{I\!\! H}\! g(X(t)) \rightarrow \bsigma(X(\infty)) [ \bsigma(X(\infty))]^\prime$, that is, 
their limits obey the following detailed balance condition, 
\begin{equation} \label{balance} 
\boldsymbol{I\!\! H}\! g(X(\infty)) \Gamma(\infty) + \Gamma(\infty) \boldsymbol{I\!\! H}\! g(X(\infty)) = \bsigma(X(\infty)) [\bsigma(X(\infty))]^\prime. 
 \end{equation}
With the limit $\Gamma(\infty)$ of $\Gamma(t)$ as $t \rightarrow \infty$, we  conclude that $V(t)$ converges in distribution to $V(\infty)=[\Gamma(\infty)]^{1/2} \bZ$, where $\bZ$ is a standard normal 
random vector. 

Denote by $P(\theta;t)$ the probability distribution of $X^m_\delta(t)$ at time $t$. Then from the Fokker-Planck equation we have 
\[ \frac{\partial P(\theta;t)}{\partial t} = \nabla \left[ - \nabla g(\theta) P(\theta; t) - \frac{\delta}{2 m} \bsigma(X(t)) [\bsigma(X(t))]^\prime 
   \, \nabla P(\theta;t) \right], \]
and under the detailed balance condition (\ref{balance}) %(\ref{balance1})-(\ref{balance2}), 
its stationary distribution $P(\theta)$ satisfies 
\[ 0 = \nabla \left[ - \nabla g(\theta) P(\theta) - \frac{\delta}{2 m} \bsigma(X(\infty)) [\bsigma(X(\infty))]^\prime
  \, \nabla P(\theta) \right], \]
which corresponds to a normal stationary distribution $N(0, \Gamma(\infty))$ for $V^m_\delta(\infty) = (m/\delta)^{1/2} (X^m_\delta(\infty)  - \check{\theta})$.
Thus, we conclude that $V^m_\delta(\infty)$ has a limiting normal distribution 
with mean zero and variance $\Gamma(\infty)$.

\subsection{Proof of Theorem \ref{thm9}}
As  $\nabla \! g(\check{\theta})=0$, by Taylor expansion we have 
\begin{align*}
& g(X^m_\delta(t)) = g(X(t)) + (\delta/ m)^{-1/2} \nabla g(X(t)) V^m_\delta(t) + \frac{\delta}{2 m} [V^m_\delta(t)]^\prime \boldsymbol{I\!\! H}\! g(X(t)) V^m_\delta(t)  + o_P(\delta/m) , & \\
& \nabla g(X^m_\delta(t)) = \nabla g(X(t)) + (\delta/ m)^{-1/2}  \boldsymbol{I\!\! H}\! g(X(t)) V^m_\delta(t)  + o_P( (\delta/m)^{1/2}) , & \\
& g(X(t)) \sim g(\check{\theta}) + \nabla g(\check{\theta}) [X(t) - \check{\theta}] + \frac{1}{2} [X(t) - \check{\theta}]^\prime \boldsymbol{I\!\! H}\! g(\check{\theta}) [X(t) - \check{\theta}] &\\
& = g(\check{\theta}) + \frac{1}{2} [X(t) - \check{\theta}]^\prime \boldsymbol{I\!\! H}\! g(\check{\theta}) [X(t) - \check{\theta}], & \\
& \nabla g(X(t)) \sim \boldsymbol{I\!\! H}\! g(\check{\theta}) [X(t) - \check{\theta}], \;\; \boldsymbol{I\!\! H}\! g(X(t)) \sim \boldsymbol{I\!\! H}\! g(\check{\theta}) , &\\
& g(X^m_\delta(t)) \sim g(\check{\theta}) +  \frac{1}{2} [X^m_\delta(t) - \check{\theta}]^\prime 
\boldsymbol{I\!\! H}\! g(\check{\theta}) [X^m_\delta(t) - \check{\theta}], &\\
%& =  g(\check{\theta}) +  \frac{1}{2} \boldsymbol{I\!\! H}\! g(\check{\theta}) \left\{   [X(t) - \check{\theta}]^2 + 2 [X(t) - \check{\theta}] [X^m_\delta(t) - X(t)] + [X^m_\delta(t) - X(t)]^2  \right\} & \\
%& =  g(\check{\theta}) +  \frac{1}{2} \boldsymbol{I\!\! H}\! g(\check{\theta}) \left\{   [X(t) - \check{\theta}]^2 + 2 \eta  [X(t) - \check{\theta}] V(t) + \eta^2 [V(t)]^2  \right\}. 
& X^m_\delta(t) - \check{\theta} =  X(t) - \check{\theta}+  (\delta/ m)^{-1/2}  V^m_\delta(t) .
 \end{align*}
%min when $X(t) -\check{\theta} \sim \eta V(t)$ i.e. $X(t) -  \eta V(t) \sim \check{\theta}$
Thus by Theorems \ref{thm3} we have that $g(X^m_\delta(t))$ and $\nabla \!g(X^m_\delta(t))$ behave like, respectively, 
\[ g(X(t)) + (\delta /m)^{1/2} \nabla g(X(t)) V(t) + \frac{\delta}{2 m} [V(t)]^\prime \boldsymbol{I\!\! H}\! g(X(t)) V(t), \]
and 
\[ \nabla \!g(X(t)) + (\delta /m)^{1/2}  \boldsymbol{I\!\! H}\! g(X(t)) V(t). \]
Similar to the stationary distribution part of the proof for Theorem \ref{thm8}, when $\boldsymbol{I\!\! H}\! g(\check{\theta})$ is positive definite,
we can derive the stationary distribution of $V(t)$ to be a normal distribution with mean zero and variance 
$\Gamma(\infty)$ defined by (\ref{stationary-xx}). Thus, we have  
\begin{eqnarray*}
&& E  [V(t)] =0, \qquad 
E \{  [V(t)]^\prime \boldsymbol{I\!\! H}\! g(X(t)) V(t) \} = tr  [ \Gamma(\infty)  \boldsymbol{I\!\! H}\! g(X(t)) ] , \\
&& Var \{ \boldsymbol{I\!\! H}\! g(X(t)) V(t) \} = tr[ \Gamma(\infty)  \{ \boldsymbol{I\!\! H}\! g(X(t)) \}^2 ] .
\end{eqnarray*}
where the expectation is taken under the stationary distribution. 
Taking the trace on both sides of (\ref{stationary-xx}) we obtain 
\[  tr  [ \Gamma(\infty)  \boldsymbol{I\!\! H}\! g(X(\infty)) ] =  tr  [ \boldsymbol{I\!\! H}\! g(X(\infty))  \Gamma(\infty)  ]
= tr[ \bsigma^2( X(\infty))]/2,  \]
and multiplying  $\boldsymbol{I\!\! H}\! g(X(\infty))$ on both sides of  (\ref{stationary-xx})  and then taking the trace operation we arrive at 
\[ tr  [ \Gamma(\infty)  \{ \boldsymbol{I\!\! H}\! g(X(\infty)) \}^2 ] = tr  [  \boldsymbol{I\!\! H}\! g(X(\infty)) \Gamma(\infty)  \boldsymbol{I\!\! H}\! g(X(\infty))  ] =  tr[ \bsigma^2( X(\infty))   \boldsymbol{I\!\! H}\! g(X(\infty))  ]/2.   \]
Putting these results together and using $\boldsymbol{I\!\! H}\! g(X(t))  \rightarrow  \boldsymbol{I\!\! H}\! g(X(\infty))$
%$tr  [ \Gamma(\infty)  \boldsymbol{I\!\! H}\! g(X(t)) ] \rightarrow tr  [ \Gamma(\infty)  \boldsymbol{I\!\! H}\! g(X(\infty))$ 
as $t \rightarrow \infty$, we prove \eqref{optimization-x3} and  \eqref{optimization-x4}. 

For the saddle point case, for simplicity we assume that 
$\boldsymbol{I\!\! H}\! g(\check{\theta})$ is diagonal with eigenvalues $\lambda_i$. Then $V(t)$ has covariance function (Gardiner (2009)) 
\[ [Cov(V(t), V(s))]_{ii} = \frac{ \bsigma_{ii}(X(t)) }{2 \lambda_i} \left[  e^{ - \lambda_i |t + s| } - e^{ -\lambda_i |t-s| } \right], \]
which, for negative $\lambda_i$, diverge as $t, s \rightarrow \infty$. Thus, $V(t)$ does not have any limiting stationary distribution.